\pdfoutput=1
\documentclass{article} 
\usepackage{conference,times}

\usepackage{url}

\usepackage{times}
\usepackage{epsfig}
\usepackage{graphicx}
\usepackage{amsmath}
\usepackage{amssymb}
\usepackage{color}
\usepackage{xcolor,colortbl}
\usepackage{multicol}
\usepackage{comment}
\usepackage{booktabs}
\usepackage{comment}
\usepackage{multirow}

\title{Image GANs meet Differentiable Rendering for Inverse Graphics and Interpretable 3D Neural Rendering}

\iclrfinalcopy 

\author{\quad \quad \quad \quad \quad \quad Yuxuan Zhang$^{1,4}$\thanks{indicates equal contribution.} \quad Wenzheng Chen$^{1,2,3*}$  \quad Huan Ling$^{1,2,3}$  \\
	\And
	\quad \quad \quad \quad Jun Gao$^{1,2,3}$  \quad Yinan Zhang$^{5}$  \quad Antonio Torralba$^{6}$  \quad Sanja Fidler$^{1,2,3}$ \\
	\\
	\quad \ \ \ \ \ \ \ \ \ \ \ \ \ \ \small{NVIDIA\textsuperscript{1} \quad University of Toronto\textsuperscript{2} \quad Vector Institute\textsuperscript{3} \quad University of Waterloo \textsuperscript{4} }
	\\ \ \ \ \ \ \ \ \ \ \ \ \ \ \ \ \ \ \ \ \ \ \ \ \ \ \ \ \ \ \ \ \ \ \ \ \ \ \ \ \ \ \  \small{\quad Stanford University\textsuperscript{5} \quad MIT CSAIL\textsuperscript{6} } \vspace{3pt}\\
	\quad \quad \texttt{\tiny \{alezhang, wenzchen, huling, jung, sfidler\}@nvidia.com, yinanzy@stanford.edu, torralba@mit.edu} \\
}

\newcommand{\AZ}[1]{{\color{purple}{[Alex: #1]}}}

\newcommand{\JUN}[1]{{\color{blue}{[Jun: #1]}}}
\newcommand{\JG}[1]{{\color{blue}{[Jun: #1]}}}

\newcommand{\SF}[1]{{\color{magenta}{[Sanja: #1]}}}
\newcommand{\ours}{StyleGAN-R}
\newcommand{\wz}[1]{{\color{red}{[WZ: #1]}}}

\iclrfinalcopy 
\begin{document}

\maketitle

\begin{abstract}
Differentiable rendering has paved the way to training neural networks to perform ``inverse graphics" tasks such as predicting 3D geometry from monocular photographs. To train high performing models, most of the current approaches rely on multi-view imagery which are not readily available in practice.  Recent Generative Adversarial Networks (GANs) that synthesize images, in contrast, seem to acquire 3D knowledge implicitly during training: object viewpoints can be manipulated by simply manipulating the latent codes. However, these latent codes often lack further physical interpretation and thus GANs cannot easily be inverted to perform explicit 3D reasoning. In this paper, we aim to extract and disentangle 3D knowledge learned by generative models by utilizing differentiable renderers. Key to our approach is to exploit GANs as a multi-view data generator to train an inverse graphics network using an off-the-shelf differentiable renderer, and the trained inverse graphics network as a teacher to disentangle  the GAN's latent code into interpretable 3D properties. The entire architecture is trained iteratively using cycle consistency losses. We show that our approach significantly outperforms state-of-the-art inverse graphics networks trained on existing datasets,both quantitatively and via user studies. We further showcase  the disentangled GAN as a controllable 3D ``neural renderer", complementing traditional graphics renderers.

\end{abstract}

\vspace{-2mm}
\section{Introduction}
\vspace{-1mm}

The ability to infer 3D properties such as geometry, texture, material, and light from photographs is key in many domains such as AR/VR, robotics, architecture, and computer vision. Interest in this problem has been explosive, particularly in the past few years, as evidenced by a large body of published works and several released 3D libraries (TensorflowGraphics by~\cite{TensorflowGraphicsIO2019}, Kaolin by~\cite{kaolin2019arxiv}, PyTorch3D by~\cite{ravi2020pytorch3d}). 

The process of going from images to 3D is often called ``inverse graphics", since the problem is inverse to the process of rendering in graphics in which a 3D scene is projected onto an image by taking into account the geometry and material properties of objects, and light sources present in the scene. 
Most work on inverse graphics assumes that 3D labels are available during training~\citep{pixel2mesh,occnet,groueix2018,DISN,3dr2n2}, and trains a neural network to predict these labels. To ensure high quality 3D ground-truth, synthetic datasets such as ShapeNet~\citep{shapenet} are typically used. 
However, models trained on synthetic datasets often struggle on real photographs due to the domain gap with synthetic imagery. 

To circumvent these issues, recent work has explored an alternative way to train inverse graphics networks that sidesteps the need for 3D ground-truth during training. The main idea is to make graphics renderers differentiable which allows one to infer 3D properties directly from images using gradient based optimization,~\cite{NMR,softras,li2018differentiable,dib-r}. These methods employ a neural network to predict geometry, texture and light from images, by minimizing the difference between the input image with the image  rendered from these properties. While impressive results have been obtained in~\cite{softras,sitzmann2019srns,liuadvgeo2018,henderson2018learning,dib-r,yao20183d,kanazawa2018learning}, most of these works  still require some form of implicit 3D supervision such as multi-view images of the same object with known cameras. 
Thus, most results 
have been reported on the synthetic ShapeNet dataset, or the large-scale CUB~\citep{cub} bird dataset annotated with keypoints from which cameras can be accurately computed using structure-from-motion techniques.

On the other hand, generative models of images appear to learn 3D information implicitly, where several works have shown that manipulating the latent code can produce images of the same scene from a different viewpoint~\citep{stylegan}. 
However, the learned latent space typically lacks physical interpretation and is usually not disentangled, where properties such as the 3D shape and color of the object often cannot be manipulated independently. 

In this paper, we aim to extract and disentangle 3D knowledge learned by generative models by utilizing differentiable graphics renderers.  We exploit a GAN, specifically StyleGAN~\citep{stylegan}, as a generator of multi-view imagery  to train an inverse graphics neural network using a differentiable renderer. In turn, we use the inverse graphics network to inform StyleGAN about the image formation process through the knowledge from graphics, effectively disentangling the GAN's latent space. We connect StyleGAN and the inverse graphics network into a single architecture which we iteratively train using cycle-consistency losses. We demonstrate our approach to significantly outperform inverse graphics networks on existing datasets, and showcase controllable 3D generation and manipulation of imagery using the disentangled generative model. 

\vspace{-2mm}
\section{Related Work}
\vspace{-2mm}

\begin{figure*}[t!]
\vspace{-10mm}
	\centering
	
	\setlength{\tabcolsep}{1pt}
	\setlength{\fboxrule}{0pt}
	\includegraphics[width=0.87\linewidth]{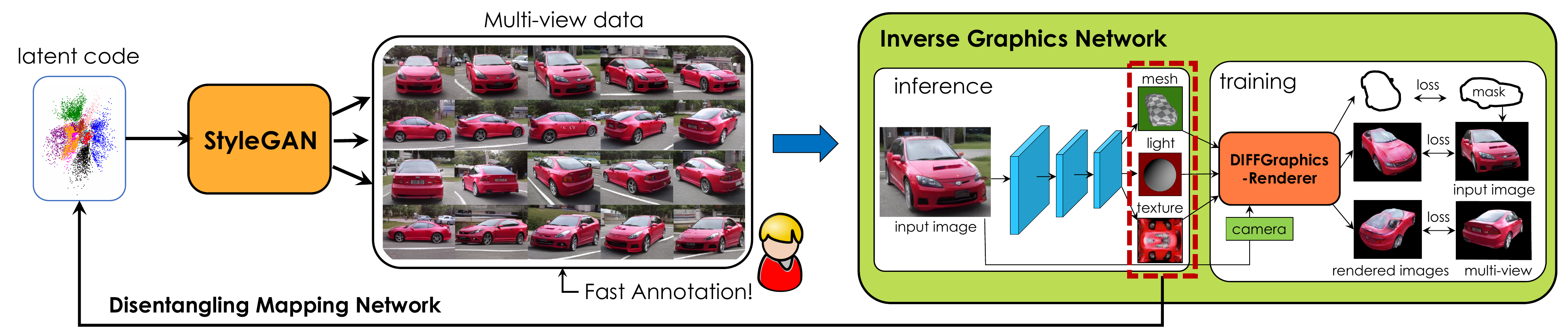}
	\vspace{-3.5mm}
		\caption{\footnotesize We employ two ``renderers": a GAN (StyleGAN in our work), and a differentiable graphics renderer (DIB-R in our work). We exploit StyleGAN as a synthetic data generator, and we label  this data extremely efficiently. This ``dataset" is used to train an inverse graphics network that predicts 3D properties from images. 
		We use this network to disentangle StyleGAN's latent code through a carefully designed mapping network.
		}		
		\label{fig:pip}
		\vspace{-3mm}
\end{figure*}

\vspace{-1mm}
\paragraph{3D from 2D:}
Reconstructing 3D objects from 2D images is one of the mainstream problems in 3D computer vision. We here focus our review to single-image 3D reconstruction which is the domain of our work. 
Most of the existing approaches train neural networks to predict 3D shapes from images by utilizing 3D labels during training,~\cite{pixel2mesh,occnet,3dr2n2,Park_2019_CVPR}. However, the need for 3D training data limits these methods to the use of synthetic datasets. When tested on real imagery there is a noticeable performance gap.

Newer works propose to differentiate through the traditional rendering process in the training loop of neural networks,~\cite{OpenDR,NMR,softras,dib-r,pix2vec,deftet}. Differentiable renderers allow one to infer 3D  from 2D images without requiring 3D ground-truth. However, in order to make these methods work in practice, several additional losses are utilized in learning, such as the multi-view consistency loss whereby the cameras are assumed known. Impressive reconstruction results have been obtained on the synthetic ShapeNet dataset. 
 While CMR by~\cite{kanazawa2018learning} and DIB-R by~\cite{dib-r} show real-image 3D reconstructions on CUB and Pascal3D~\citep{xiang_wacv14} datasets, they rely on manually annotated keypoints, while still failing to produce accurate results. 

A few recent works,~\cite{Wu_2020_CVPR,umr2020,ucmrGoel20,kato2019selfsupervised}, explore 3D reconstruction from 2D images in a completely unsupervised fashion. They recover both 3D shapes and camera viewpoints from 2D images by minimizing the difference between original and re-projected images with additional unsupervised constraints, e.g., semantic information (\cite{umr2020}), symmetry (\cite{Wu_2020_CVPR}), GAN loss (\cite{kato2019selfsupervised}) or viewpoint distribution (\cite{ucmrGoel20}). Their reconstruction is typically limited to 2.5D (\cite{Wu_2020_CVPR}), and produces lower quality results than when additional supervision is used (\cite{ucmrGoel20,umr2020,kato2019selfsupervised}).
In contrast, we utilize GANs to generate multi-view realistic datasets that can be annotated \emph{extremely efficiently}, which leads to accurate 3D results. Furthermore, our model achieves disentanglement in GANs and turns them into interpretable 3D neural renderers. 

\vspace{-3mm}
\paragraph{Neural Rendering with GANs:} 
GANs~\citep{NIPS2014_5423,stylegan} can be regarded as neural renderers, as they take a latent code as input and ``render" an image. However, the latent code is sampled from a predefined prior 
and lacks interpretability. 
Several works generate images with conditions: a semantic mask~\citep{cyclegan}, scene layout~\cite{karacan2016learning}, or a caption~\citep{reed2016generative}, and manipulate the generated images by modifying the input condition. Despite tremendous progress in this direction, there is little work on generating images through an interpretable 3D physics process. \cite{dosovitskiy2016learning} synthesizes images conditioned on object style, viewpoint, and color. 
Most relevant work to ours is~\cite{zhu2018visual}, which utilizes a learnt 3D geometry prior and generates images with a given viewpoint and texture code. We differ in three important ways. First, we do not require a 3D dataset to train the 3D prior. Second, the texture in our model has 3D physical meaning, while~\cite{zhu2018visual} still samples from a prior. We further control background while~\cite{zhu2018visual} synthesizes objects onto white background. 


 \vspace{-3mm}
\paragraph{Disentangling GANs:}
Learning disentangled representations has been widely explored,~\cite{lee2020high,lin2019infogan,perarnau2016invertible}. Representative work is InfoGAN~\cite{infogan}, which tries to maximize the mutual information between the prior and the generated image distribution. However, the disentangled code often still lacks physical interpretability. 
~\cite{tewari2020stylerig} transfers face rigging information from an existing model to control face attribute disentanglement in the StyleGAN latent space.~\cite{shen2020interfacegan} aims to find the latent space vectors that correspond to meaningful edits, while~\cite{harkonen2020ganspace} exploits PCA to disentangle the latent space. 
Parallel to our work, ~\cite{zhang21,segGAN21} attempt to interpret the semantic meaning of StyleGAN latent space. In our work, we disentangle the latent space with knowledge from graphics. 


\begin{figure*}[t!]
	{
		\vspace*{-10mm}
		\begin{center}
			\setlength{\tabcolsep}{1pt}
			\setlength{\fboxrule}{0pt}
			\hspace*{0pt}
			\begin{tabular}{c}
				\begin{tabular}{cccccc}
					
					
					\hspace{-1mm}\includegraphics[width=.161\textwidth,trim=0 0 0 30,clip]{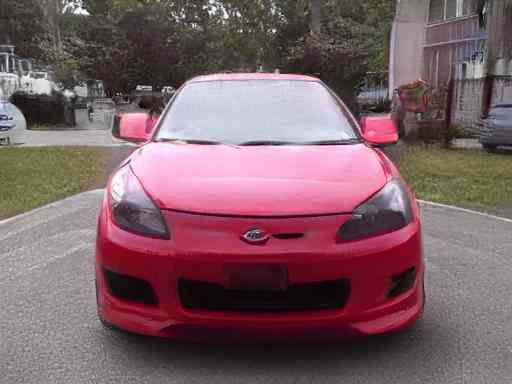}&
					\includegraphics[width=.161\textwidth,trim=0 0 0 30,clip]{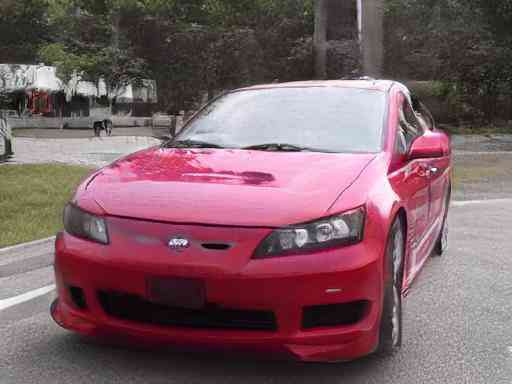}&
					\includegraphics[width=.161\textwidth,trim=0 0 0 30,clip]{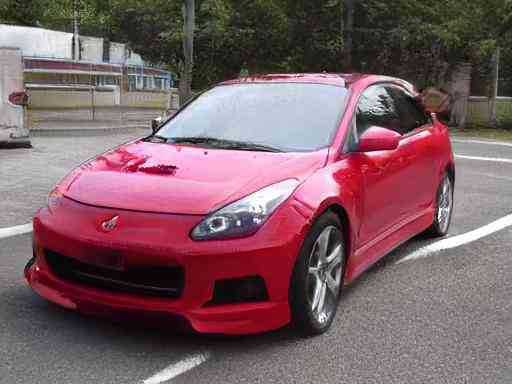}&
					\includegraphics[width=.161\textwidth,trim=0 0 0 30,clip]{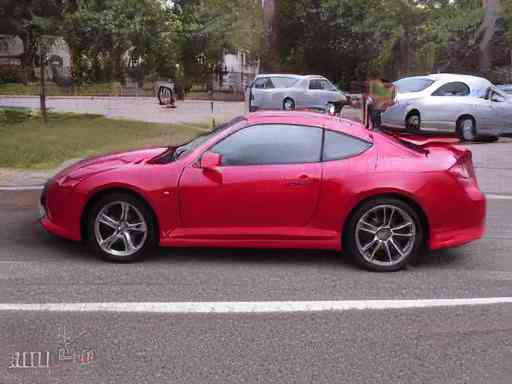}&
					\includegraphics[width=.161\textwidth,trim=0 0 0 30,clip]{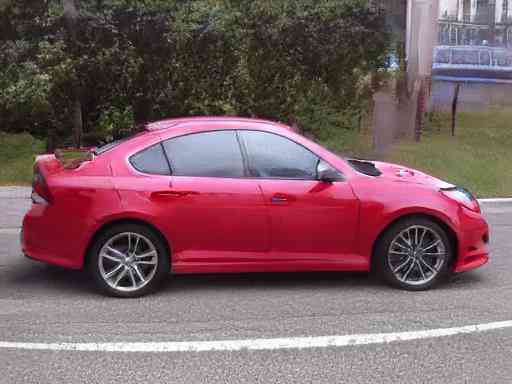}
					&
					\includegraphics[width=.161\textwidth,trim=0 0 0 30,clip]{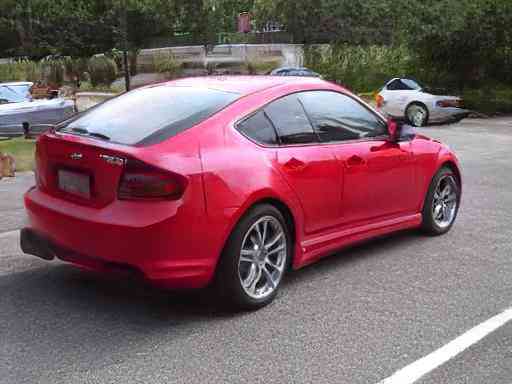}
					\\[-0.7mm]
					\hspace{-1mm}\includegraphics[width=.161\textwidth,trim=0 20 0 40,clip]{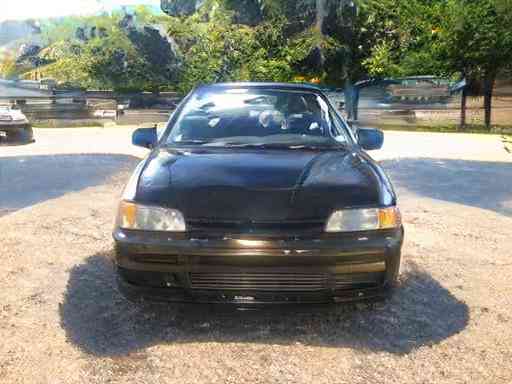}&
					\includegraphics[width=.161\textwidth,trim=0 20 0 40,clip]{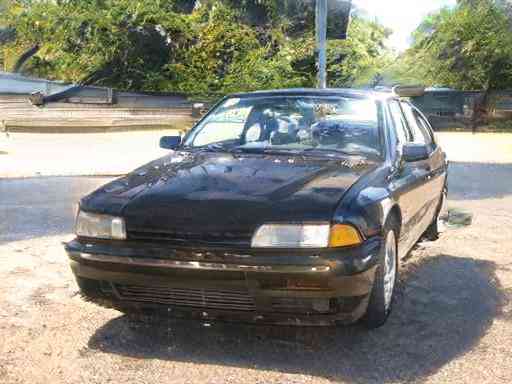}&
					\includegraphics[width=.161\textwidth,trim=0 20 0 40,clip]{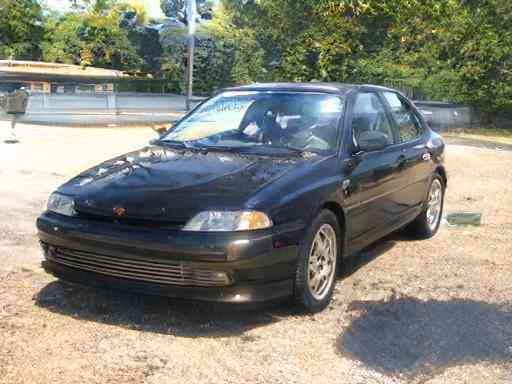}&
					\includegraphics[width=.161\textwidth,trim=0 40 0 20,clip]{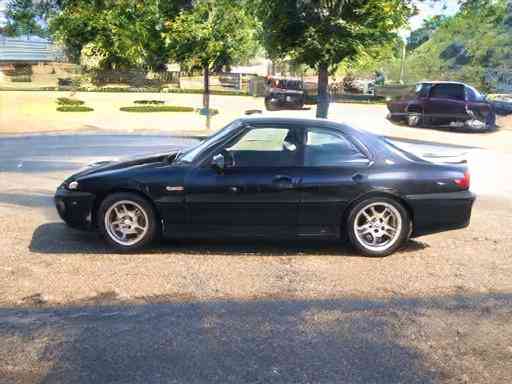}&
					\includegraphics[width=.161\textwidth,trim=0 40 0 20,clip]{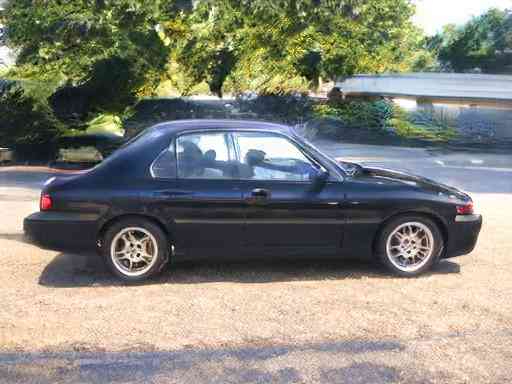}&
					\includegraphics[width=.161\textwidth,trim=0 40 0 20,clip]{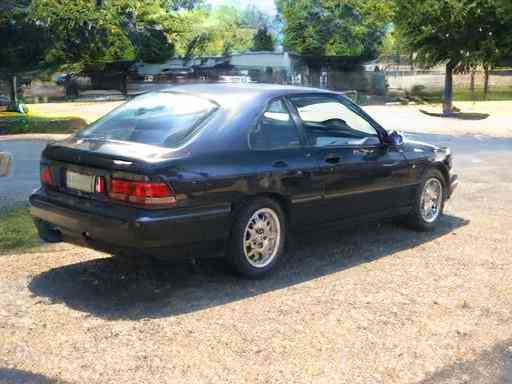}
					
					\\[-0.8mm]
					\end{tabular}
					\end{tabular}
					\begin{tabular}{ccccccccccccc}
					\includegraphics[width=.077\textwidth]{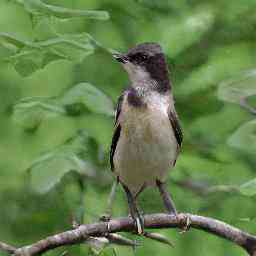}&
					\includegraphics[width=.077\textwidth]{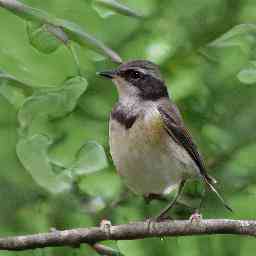}&
					\includegraphics[width=.077\textwidth]{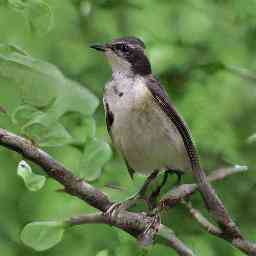}&
					\includegraphics[width=.077\textwidth]{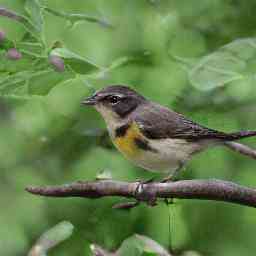}&
					\includegraphics[width=.077\textwidth]{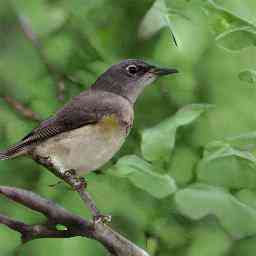}
					&
					\includegraphics[width=.077\textwidth]{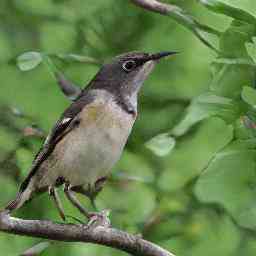}						
					&\hspace{0.8mm}
					\includegraphics[width=.077\textwidth]{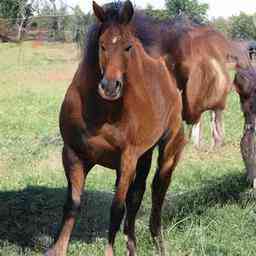}&
					\includegraphics[width=.077\textwidth]{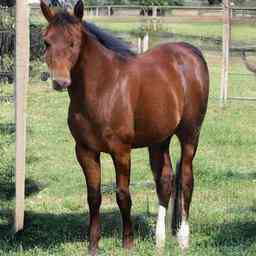}&
					\includegraphics[width=.077\textwidth]{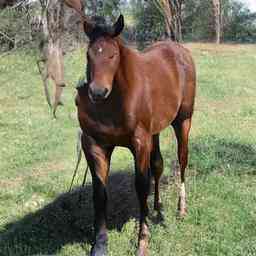}&
					\includegraphics[width=.077\textwidth]{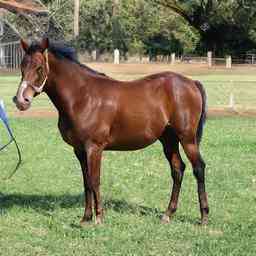}&
					\includegraphics[width=.077\textwidth]{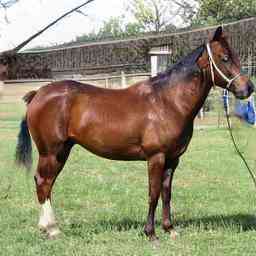}
					&
					\includegraphics[width=.077\textwidth]{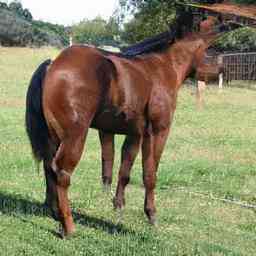}
					\\
					\end{tabular}
		\end{center}
		\vspace*{-6mm}
	}

	\caption{\label{fig:viewpoints}\footnotesize We show examples of cars (first two rows) synthesized in chosen viewpoints (columns). To get these, we fix the latent code $w_v^*$ that controls the viewpoint (one code per column) and randomly sample the remaining dimensions of (Style)GAN's latent code (to get rows). Notice how well aligned the two cars are in each column. In the third row we show the same approach applied to horse and bird StyleGAN. }
	\vspace{-5mm}
\end{figure*}

\vspace{-3mm}
\section{Our Approach}
\label{sec:approach}
\vspace{-2mm}

We start by providing an overview of our approach (Fig.~\ref{fig:pip}), and describe the individual components in more detail in the following sections. Our approach marries two types of renderers:  a GAN-based neural ``renderer'' and a differentiable graphics renderer. 
Specifically, we leverage the fact that the recent state-of-the-art GAN architecture StyleGAN by~\cite{stylegan,stylegan2} learns to produce highly realistic images of objects, and allows for a reliable control over the camera. We manually select a few camera views with a rough viewpoint annotation, and use StyleGAN to generate a large number of examples per view, which we explain in Sec.~\ref{sec:genedata}. 
In Sec.~\ref{sec:dibr}, we exploit this dataset to train an inverse graphics network utilizing the state-of-the-art differentiable renderer, DIB-R by~\cite{dib-r} in our work,  with a small modification that allows it to deal with noisy cameras during training. In Sec.~\ref{sec:disentangle}, we employ the trained inverse graphics network to disentangle StyleGAN's latent code and turn StyleGAN into a 3D neural renderer, allowing for control over explicit 3D properties. 
We fine-tune the entire architecture, leading to significantly improved results. 



\vspace{-2mm}
\subsection{StyleGAN as Synthetic Data Generator}
\label{sec:genedata}
\vspace{-1mm}
We first aim to utilize  StyleGAN to generate multi-view imagery. 
StyleGAN is a 16 layers neural network that maps a latent code $z \in Z$ drawn from a normal distribution into a realistic image. The code $z$   is first mapped to an intermediate latent code $w \in W$ which is transformed to ${w}^*=(w_1^*, w_2^*, ..., w_{16}^*) \in W^*$ through 16 learned affine transformations. We call $W^*$ the transformed latent space to differentiate it from the intermediate latent space $W$. Transformed latent codes $w^*$  are then injected as the style information to the StyleGAN Synthesis network. 

Different layers control different image attributes. As observed in~\cite{stylegan}, styles in early layers adjust the camera viewpoint while styles in the intermediate and higher  layers influence shape, texture and background. We provide a careful analysis of all layers in Appendix. 
We empirically find that the latent code $w_v^*:=(w_1^*, w_2^*, w_3^*, w_4^*)$ in the first 4 layers controls camera viewpoints. That is, if we sample a new  code $w_v^*$ but keep the remaining dimensions of $w^*$ fixed (which we call the content code), we generate images of the same object depicted in a different viewpoint. Examples are shown in Fig.~\ref{fig:viewpoints}.

We further observe that a sampled code $w_v^*$ in fact represents a fixed camera viewpoint. That is, if we keep $w_v^*$ fixed but sample the remaining dimensions of $w^*$, StyleGAN produces imagery of different objects in the same camera viewpoint. This is shown in columns in Fig.~\ref{fig:viewpoints}. Notice how aligned the objects are in each of the viewpoints. 
This makes StyleGAN a \emph{multi-view}  data generator! 

{\bf ``StyleGAN" multi-view dataset:} We manually select several views, which cover all the common viewpoints of an object ranging from 0-360 in azimuth and roughly 0-30 in elevation. We pay attention to choosing viewpoints in which the objects look most consistent.  Since inverse graphics works require camera pose information, we annotate the chosen viewpoint codes with a rough absolute camera pose. To be specific, we classify each viewpoint code into one of 12 azimuth angles, uniformly sampled along $360\deg$. We assign each code a fixed elevation (0$^\circ$) and camera distance.  These camera poses provide a very coarse annotation of the actual pose -- the annotation serves as the initialization of the camera which we will optimize during training. This allows us to annotate all views (and thus the entire dataset) in {\bf only 1 minute} -- making annotation effort neglible. 
For each viewpoint, we sample a large number of content codes to synthesize different objects in these views. Fig.~\ref{fig:viewpoints} shows 2 cars, and a horse and a bird. Appendix provides more examples.

Since DIB-R also utilizes segmentation masks during training, we further apply MaskRCNN by~\cite{maskrcnn} to get instance segmentation in our generated dataset. As StyleGAN sometimes generates unrealistic images or images with multiple objects, we filter out ``bad'' images which have more than one instance, or small masks (less than 10\% of the whole image area). 


\vspace{-2mm}
\subsection{Training an Inverse Graphics Neural Network}
\label{sec:dibr}
\vspace{-2mm}

Following CMR by~\cite{kanazawa2018learning}, and DIB-R by~\cite{dib-r}, we aim to train a 3D prediction network $f$, parameterized by $\theta$,	 to infer 3D shapes (represented as meshes) along with textures from  images. 
Let $I_V$ denote an image in viewpoint $V$ from our StyleGAN dataset, and $M$ its corresponding object mask. 
The inverse graphics network makes a prediction as follows: 
$\{S, T\} = f_\theta(I_V),$
where $S$ denotes the predicted shape, and $T$ a texture map. Shape $S$ is deformed from a sphere as in~\cite{dib-r}. While DIB-R also supports prediction of lighting, we empirically found its performance is weak for realistic imagery and we thus omit lighting estimation in our work.

To train the network, we adopt DIB-R as the differentiable graphics renderer that takes $\{S, T\}$ and $V$ as input and produces a rendered image $I'_{V} = r(S, T, V)$ along with a rendered mask $M'$. 
Following DIB-R, the loss function then takes the following form:
\begin{equation}
\begin{aligned}
L(I,S,T,V;\theta) = & \lambda_{\text{col}}L_{\text{col}}(I, I') + \lambda_{\text{percpt}}L_{\text{pecept}}(I, I') +  L_{\text{IOU}}(M, M') 
\\&+ \lambda_{\text{sm}}L_{\text{sm}}(S)  + \lambda_{\text{lap}}L_{\text{lap}}(S) + \lambda_{\text{mov}}L_{\text{mov}}(S)
\end{aligned}
\end{equation}
Here, $L_{\text{col}}$ is the standard $L_1$ image reconstruction loss defined in the RGB color space while $L_{\text{percpt}}$ is the perceptual loss that helps the predicted texture look more realistic. Note that rendered images do not have background, so $L_{\text{col}}$ and $L_{\text{percept}}$ are calculated by utilizing the mask. $L_{\text{IOU}}$ computes the intersection-over-union between the ground-truth mask and the rendered mask. 
Regularization losses such as the Laplacian loss $L_{\text{lap}}$ and  flatten loss $L_{\text{sm}}$ are commonly used to ensure that the shape is well behaved. Finally, $L_{\text{mov}}$ regularizes the shape deformation to be uniform and small.



Since we also have access to multi-view images for each object we also include a multi-view consistency loss. 
In particular, our  loss per object $k$ is:
\begin{equation}
\begin{aligned}
\mathcal L_k(\theta)=  \sum_{i,j,i\neq j}\big( & L(I_{V_i^k}, S_k, T_k, V_i^k;\theta) + L(I_{V_j^k}, S_k, T_k, V_j^k;\theta)\big) \\
& \text{where} \ \  \{S_k, T_k, L_k \} = f_\theta(I_{V_i^k})
\end{aligned}
\end{equation}
While more views provide more constraints, empirically, two views have been proven sufficient. We randomly sample view pairs $(i, j)$ for efficiency. 


We use the above loss functions to jointly train the neural network $f$ and optimize viewpoint cameras $V$ (which were fixed in~\cite{dib-r}). We assume that different images generated from the same $w_v^*$ correspond to the same viewpoint $V$. Optimizing the camera jointly with the weights of the network allows us to effectively deal with noisy initial camera annotations. 

\vspace{-2mm}
\subsection{Disentangling StyleGAN with the Inverse Graphics Model} 
\label{sec:disentangle}

The inverse graphics model allows us to infer a 3D mesh and texture from a given image. We now utilize these 3D properties to disentangle StyleGAN's latent space, and turn StyleGAN into a fully controllable 3D neural renderer, which we refer to as StyleGAN-R. Note that StyleGAN in fact synthesizes more than just an object, it also produces the background, i.e., the entire scene. Ideally we want control over the background as well, allowing the neural renderer to render 3D objects into desired scenes. To get the background from a given image, we simply mask out the object. 

We propose to learn a mapping network to map the viewpoint, shape (mesh), texture and background into the StyleGAN's latent code. 
Since StyleGAN may not be completely disentangled, we further fine-tune the entire StyleGAN model while keeping the inverse graphics network fixed. 

\begin{figure*}[t!]
	\vspace{-10mm}
	\begin{minipage}{0.615\linewidth}
	\centering
	\includegraphics[width=1\linewidth]{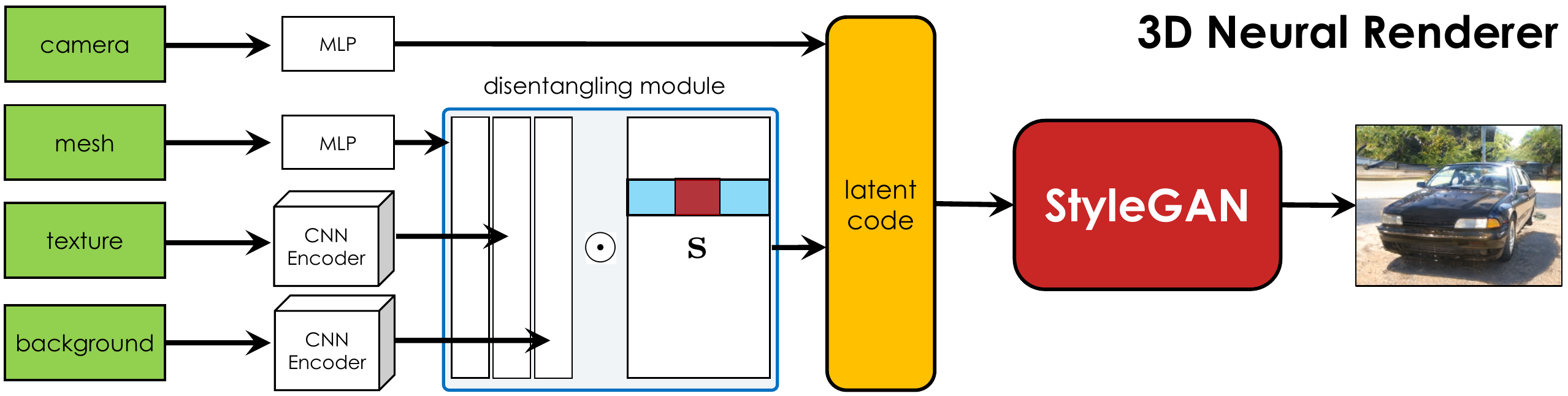}
	\end{minipage}
	\hspace{3.5mm}
	\begin{minipage}{0.35\linewidth}
	\vspace{-4mm}
		\caption{\footnotesize A mapping network maps camera, shape, texture and background into a disentangled code that is passed to StyleGAN for ``rendering". We refer to this network as {\ours}.}	
		\label{fig:mapping}
		\end{minipage}
		\vspace{-5mm}
\end{figure*}

\vspace{-2mm}
\paragraph{Mapping Network:} 
Our mapping network, visualized in Figure~\ref{fig:mapping}, maps the viewpoints to first 4 layers and maps the shape, texture and background to the last 12 layers of $W^{*}$. For simplicity, we denote the first 4 layers as $W_V^{*}$ and the last 12 layers as $W_{STB}^{*}$, where $W_V^{*} \in \mathbb R^{2048}$ and $W_{STB}^{*} \in \mathbb R^{3008}$. 
Specifically, the mapping network $g_v$ for viewpoint $V$ and $g_s$ for shape $S$ are separate MLPs while $g_t$ for texture $T$ and $g_b$ for background $B$ are CNN layers:
\begin{equation}
\begin{aligned}
& \mathbf{z}^{\mathrm{view}}=g_v(V;\theta_v), \ \mathbf{z}^{\mathrm{shape}}=g_s(S;\theta_s), \mathbf{z}^{\mathrm{txt}}=g_t(T;\theta_t),\ \mathbf{z}^{\mathrm{bck}}=g_b(B;\theta_b),
\end{aligned}
\end{equation}
where $\mathbf{z}^{\mathrm{view}}\in \mathbb R^{2048}, \mathbf{z}^{\mathrm{shape}},\mathbf{z}^{\mathrm{txt}},\mathbf{z}^{\mathrm{bck}}\in \mathbb R^{3008}  \text{ and }\ \theta_v,\theta_s,\theta_t,\theta_b$ are network parameters.
We softly combine the shape, texture and background codes into the final latent code as follows:
\begin{eqnarray}
	\tilde w^{mtb}&=&\mathbf{s}^{\mathrm{m}} \odot \mathbf{z}^{\mathrm{shape}} + \mathbf{s}^{\mathrm{t}} \odot \mathbf{z}^{\mathrm{txt}} + \mathbf{s}^{\mathrm{b}} \odot \mathbf{z}^{\mathrm{bck}},
\end{eqnarray}
where $\odot$ denotes element-wise product, and $\mathbf{s}^{\mathrm{m}}, \mathbf{s}^{\mathrm{t}}, \mathbf{s}^{\mathrm{b}} \in \mathbb R^{3008}$ are shared across all the samples. To achieve disentanglement,  we want each dimension of the final code to be explained by only one property (shape, texture or background). We thus normalize each dimension of $\mathbf{s}$ using softmax. 

In practice, we found that mapping $V$ to a high dimensional code is challenging since our dataset only contains a limited number of views, and $V$ is limited to azimuth, elevation and scale. We thus map $V$ to the subset of $W_V^{*}$, where we empirically choose 144 of the 2048 dimensions with the highest correlation with the annotated viewpoints. Thus, $\mathbf{z}^{\mathrm{view}}\in \mathbb R^{144}$ in our case. 

\vspace{-3mm}
\paragraph{Training Scheme:}
We train the mapping network and fine-tune StyleGAN in two separate stages. We first freeze StyleGAN's weights and train the mapping network only. This warms up the mapping network to output reasonable latent codes for StyleGAN. We then fine-tune both StyleGAN and the mapping network to better disentangle different attributes.  We provide details next.

In the warm up stage, we sample viewpoint codes $w_v^*$ among the chosen viewpoints, and sample the remaining dimensions of $w^* \in W^*$. 
We try to minimize the $L_2$ difference between the mapped code $\tilde w$ and StyleGAN's code $w^*$. To encourage the disentanglement in the latent space, we  penalize the entropy of each dimension $i$ of $\mathbf{s}$. Our overall loss function for our mapping network is:
\begin{align}
L_{\text{mapnet}}(\theta_v,\theta_s,\theta_t,\theta_v) = ||\tilde{w} -  w^*||_2 - \sum_i \sum_{k\in \{m,t,b\}} \mathbf{s}_{i}^k \log(\mathbf{s}_{i}^k).
\end{align}
By training the mapping network, we find that view, shape and texture can be disentangled in the original StyleGAN model but the background remains entangled. We thus fine-tune the model to get a better disentanglement.
To fine-tune the StyleGAN network we incorporate a cycle consistency loss. 
In particular, by feeding a sampled shape, texture and background to StyleGAN we obtain a synthesized image.
We encourage  consistency between the original sampled properties and the shape, texture and background predicted from the StyleGAN-synthesized image via the inverse graphics network.  We further feed the same background $B$ with two different $\{S, T\}$ pairs to generate two images $I_1$ and $I_2$. We then encourage the re-synthesized backgrounds $\bar{B_1}$ and $\bar{B_2}$ to be similar. This loss tries to disentangle the background from the foreground object. During training, we find that imposing the consistency loss on $B$ in image space results in blurry images, thus we constrain it in the code space. 
Our fine-tuning loss takes the following form:
\begin{equation}
\begin{aligned}
L_{\text{stylegan}}(\theta_{\text{gan}}) =  ||S - \bar{S}||_2 +  ||T - \bar{T}||_2 +  ||g_b(B) - g_b(\bar{B})||_2  + ||g_b(\bar{B_1}) - g_b(\bar{B_2})||_2
\end{aligned}
\end{equation}

\vspace{-4mm}
\section{Experiments}
\vspace{-3mm}
In this section, we showcase our approach on inverse graphics tasks (3D image reconstruction), as well as on the task of 3D neural rendering and 3D image manipulation. 

\paragraph{\textbf{Image Datasets for training StyleGAN:}}
We use three category-specific StyleGAN models, one representing a rigid object class, and two representing articulated (and thus more challenging) classes. We use the official car and horse model from StyleGAN2~\citep{stylegan2} repo which are trained on LSUN Car and LSUN Horse with 5.7M and 2M images. We also train a bird model on NABirds~\citep{7298658} dataset, which contains 48k images.

\paragraph{\textbf{Our ``StyleGAN" Dataset:}} We first randomly sample 6000 cars, 1000 horse and 1000 birds with diverse shapes, textures, and backgrounds from StyleGAN. 
After filtering out images with bad masks as described in Sec.~\ref{sec:approach}, 55429 cars, 16392 horses and 7948 birds images remain in our dataset which is significant larger than the Pascal3D car dataset~\citep{xiang_wacv14} (4175 car images).  
Note that nothing prevents us from synthesizing a significantly larger amount of data, but in practice, this amount turned out to be sufficient to train good models. 
We provide  more examples in Appendix.


\begin{figure*}[t!]
	{
		\vspace*{-4mm}
		\begin{center}
			\setlength{\tabcolsep}{1pt}
			\setlength{\fboxrule}{0pt}
			\hspace*{-0.15cm}
			\begin{tabular}{c}
				\begin{tabular}{cccccccccc}
					\includegraphics[height=0.086\linewidth]{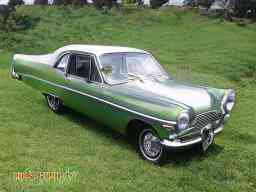}&
					\includegraphics[height=0.086\linewidth]{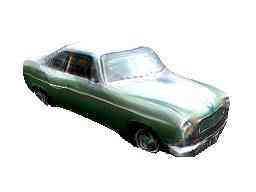}&
					\multicolumn{3}{c}{\includegraphics[height=0.086\linewidth]{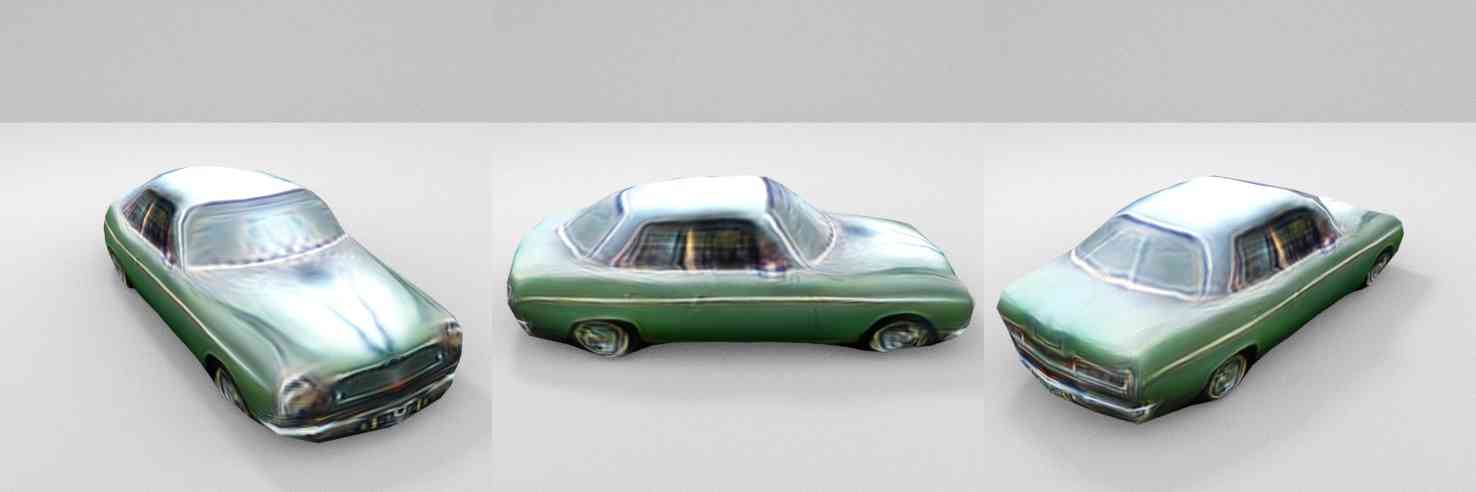}} &
					\includegraphics[height=0.086\linewidth]{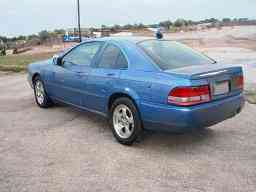}&
					\includegraphics[height=0.086\linewidth]{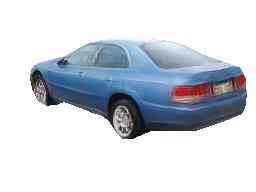}&
					\multicolumn{3}{c}{\includegraphics[height=0.086\linewidth]{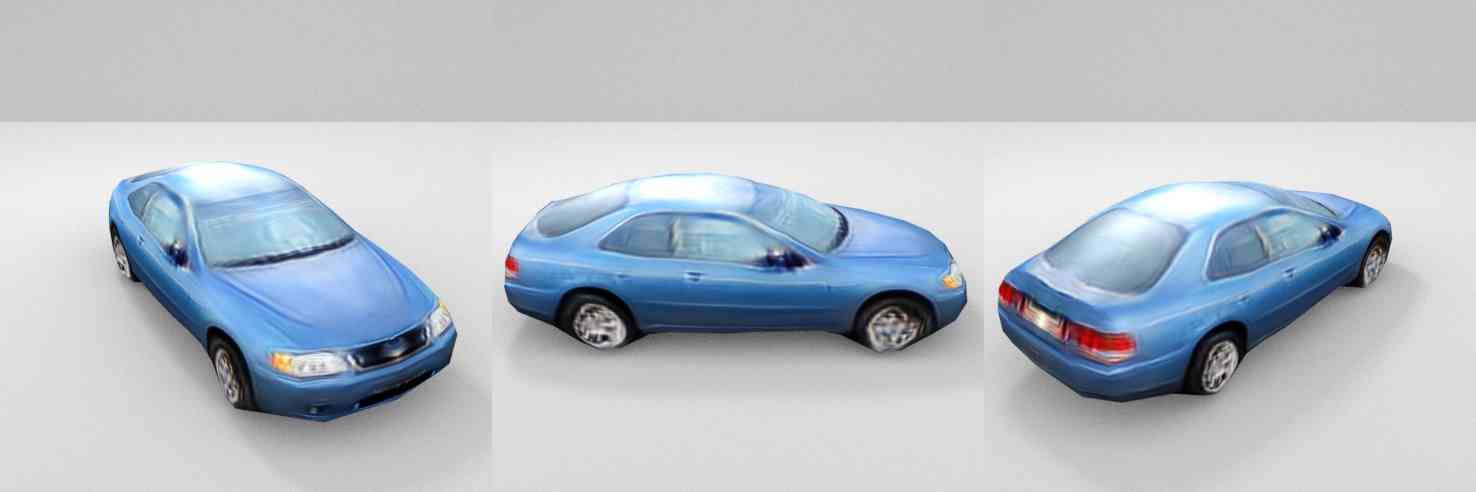}}
					\\[-0.5mm]
					\includegraphics[height=0.086\linewidth]{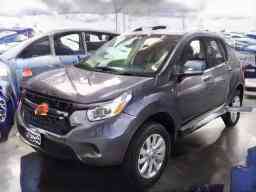}&
					\includegraphics[height=0.086\linewidth]{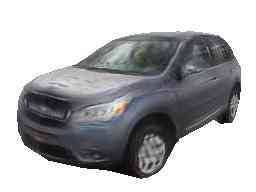}&
					\multicolumn{3}{c}{\includegraphics[height=0.086\linewidth]{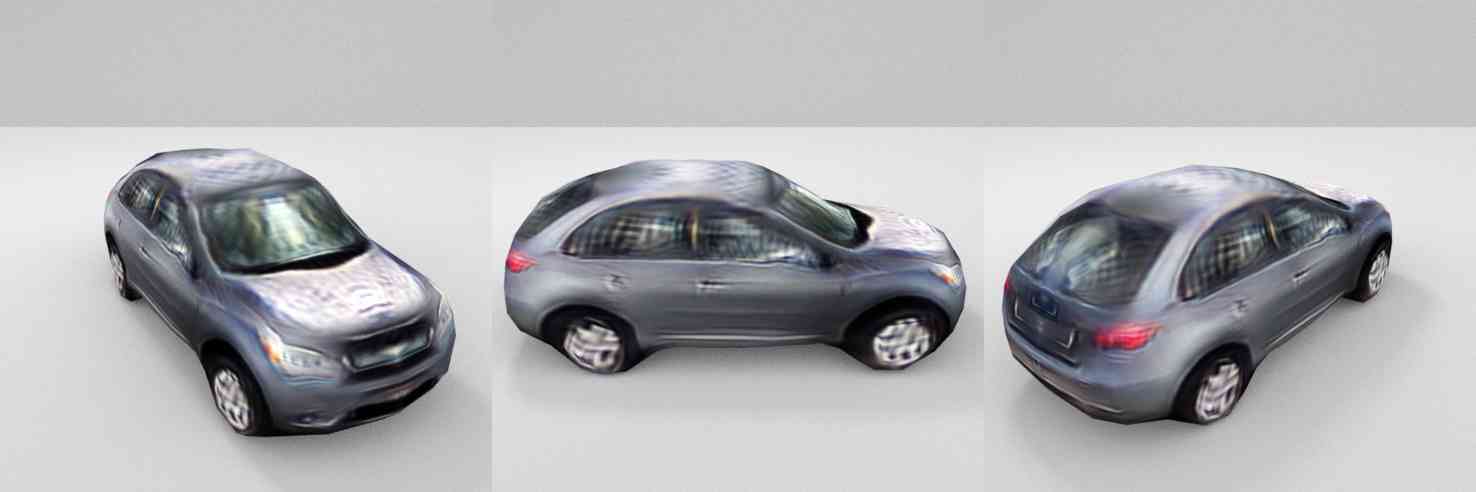}} &
					\includegraphics[height=0.086\linewidth]{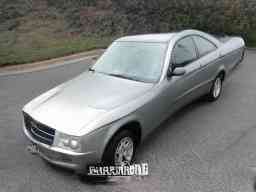}&
					\includegraphics[height=0.086\linewidth]{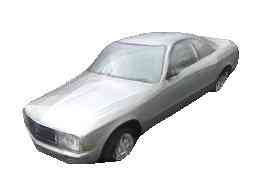}&
					\multicolumn{3}{c}{\includegraphics[height=0.086\linewidth]{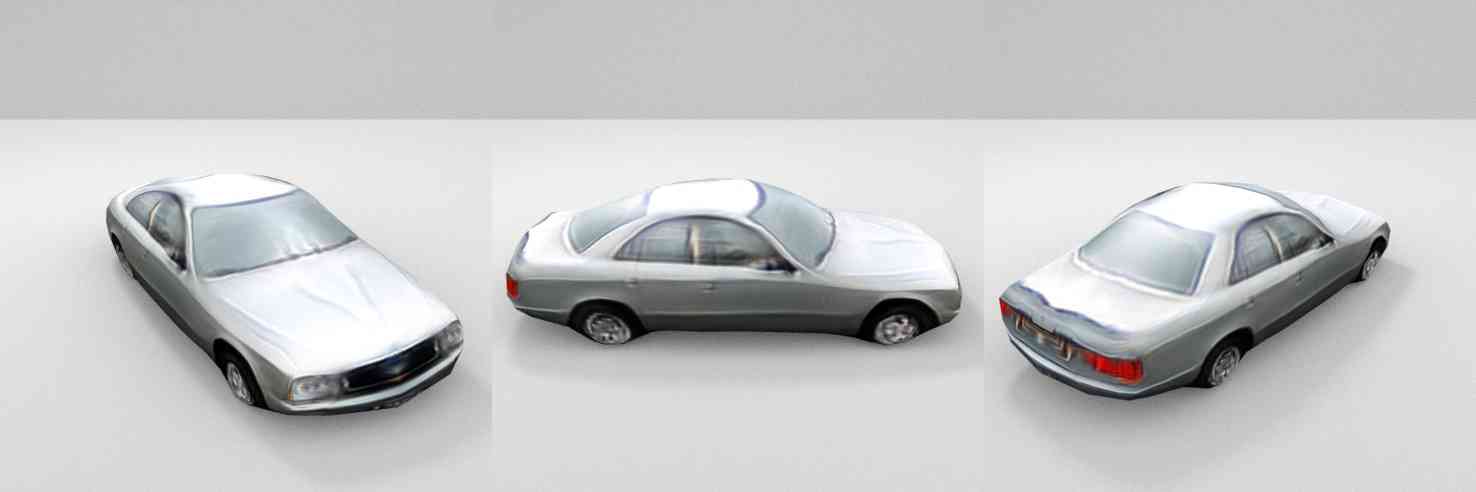}}
					\\[-0.5mm]	
					
					\includegraphics[height=0.086\linewidth]{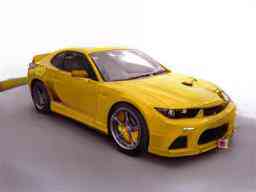}&
					\includegraphics[height=0.086\linewidth]{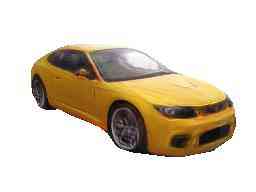}&
					\multicolumn{3}{c}{\includegraphics[height=0.086\linewidth]{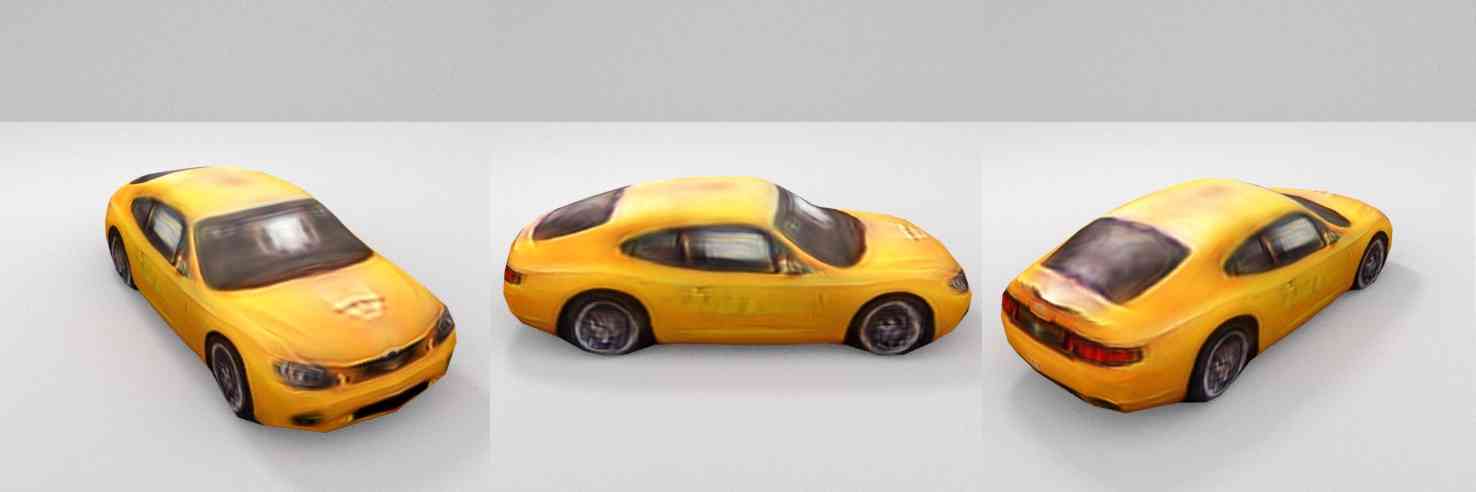}} &
					\includegraphics[height=0.086\linewidth]{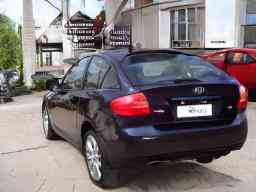}&
					\includegraphics[height=0.086\linewidth]{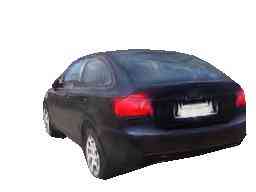}&
					\multicolumn{3}{c}{\includegraphics[height=0.086\linewidth]{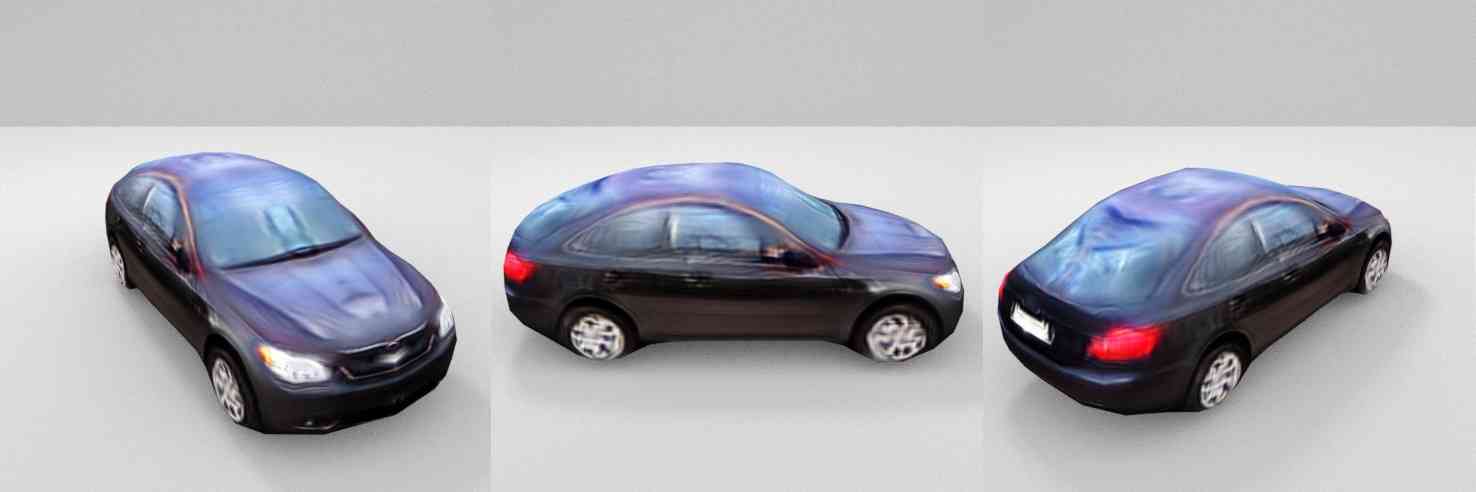}}
					\\[-0.5mm]

					\includegraphics[height=0.086\linewidth]{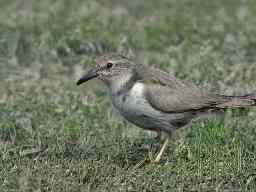}&
					\includegraphics[height=0.086\linewidth]{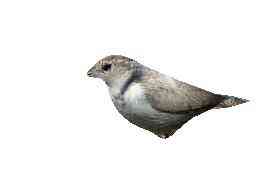}&
					\multicolumn{3}{c}{\includegraphics[height=0.086\linewidth]{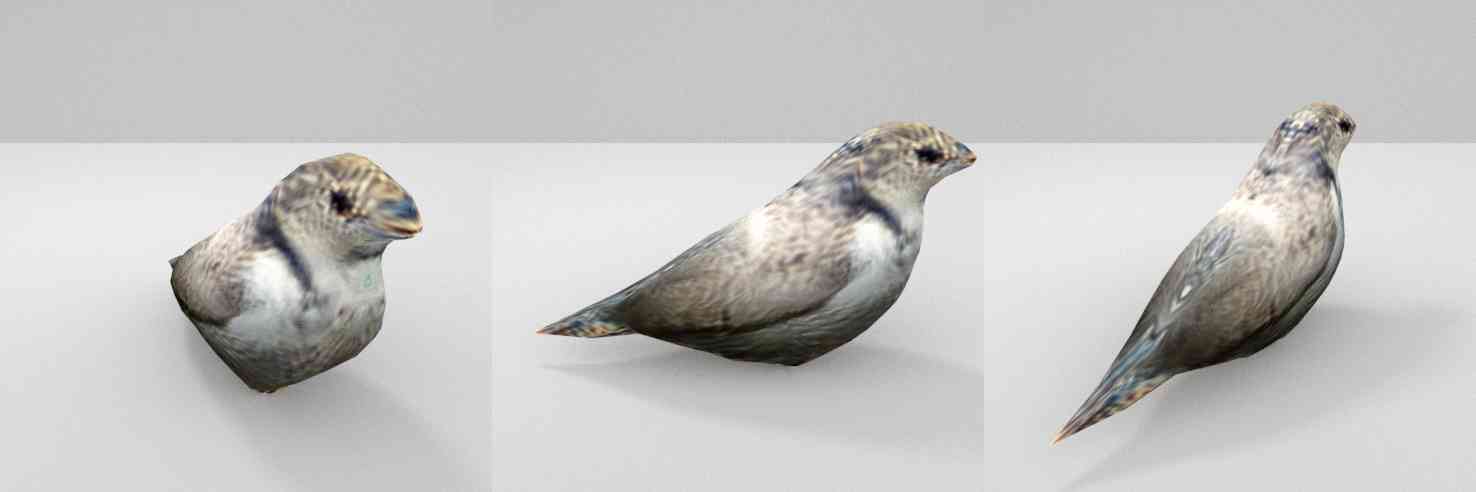}} &
					\includegraphics[height=0.086\linewidth]{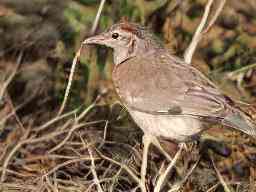}&
					\includegraphics[height=0.086\linewidth]{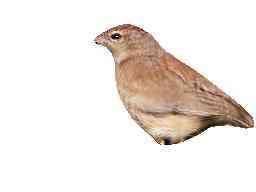}&
					\multicolumn{3}{c}{\includegraphics[height=0.086\linewidth]{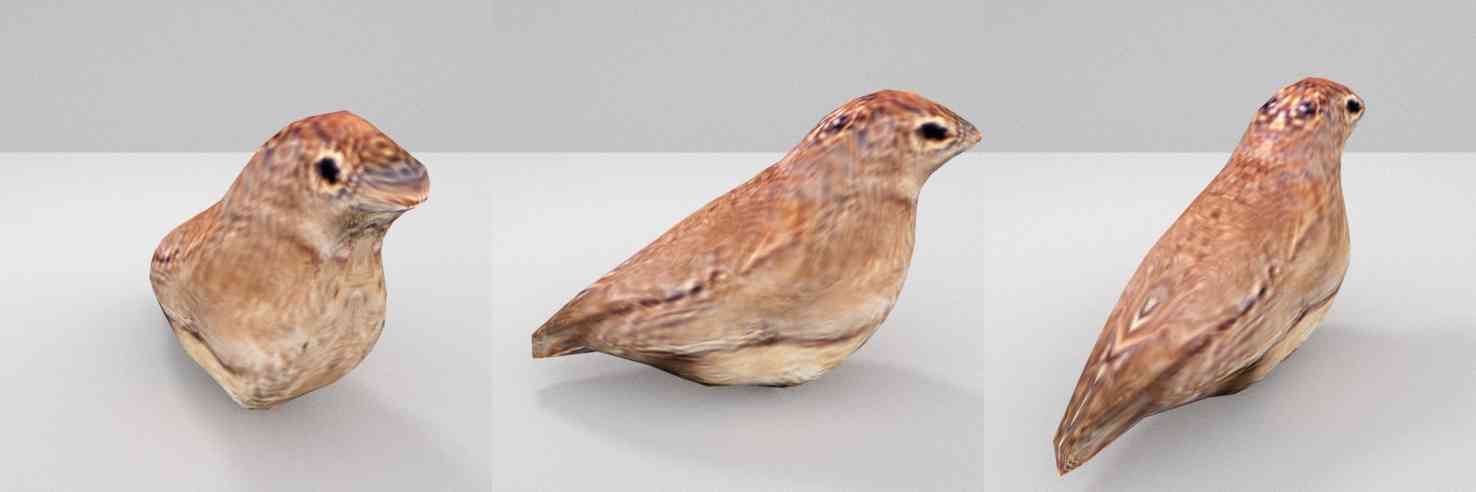}}
					\\[-0.5mm]						
					\includegraphics[height=0.086\linewidth]{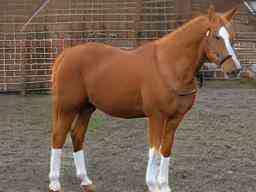}&
					\includegraphics[height=0.086\linewidth]{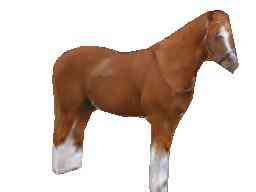}&
					\multicolumn{3}{c}{\includegraphics[height=0.086\linewidth]{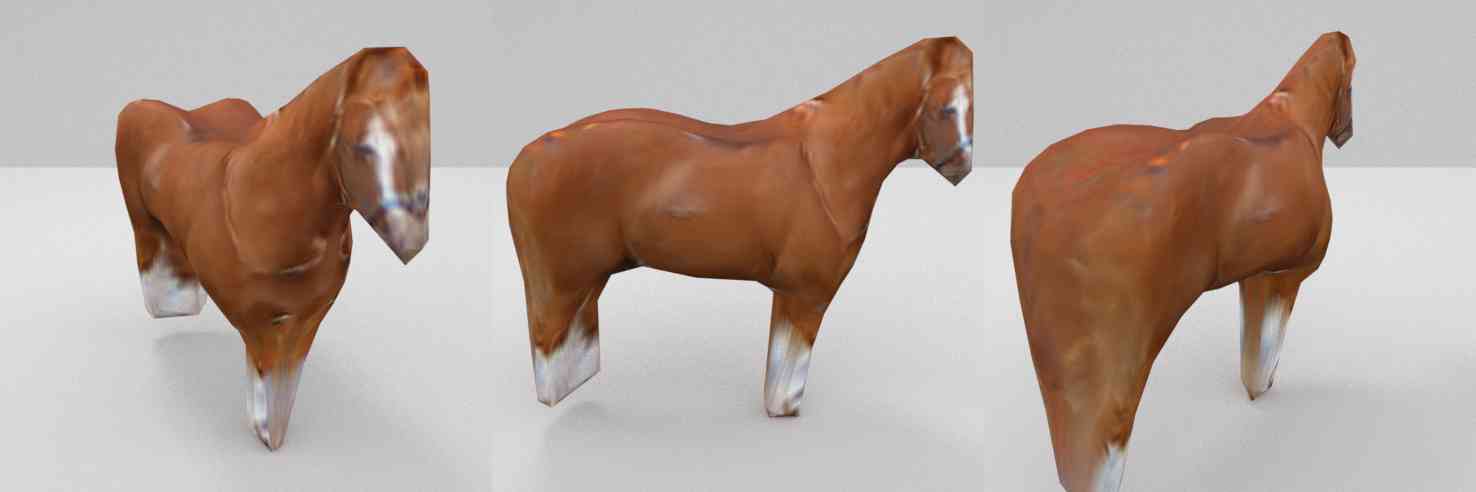}} &
					\includegraphics[height=0.086\linewidth]{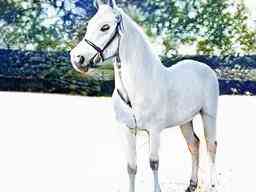}&
					\includegraphics[height=0.086\linewidth]{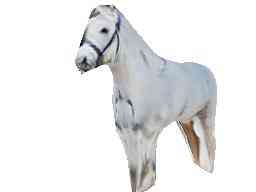}&
					\multicolumn{3}{c}{\includegraphics[height=0.086\linewidth]{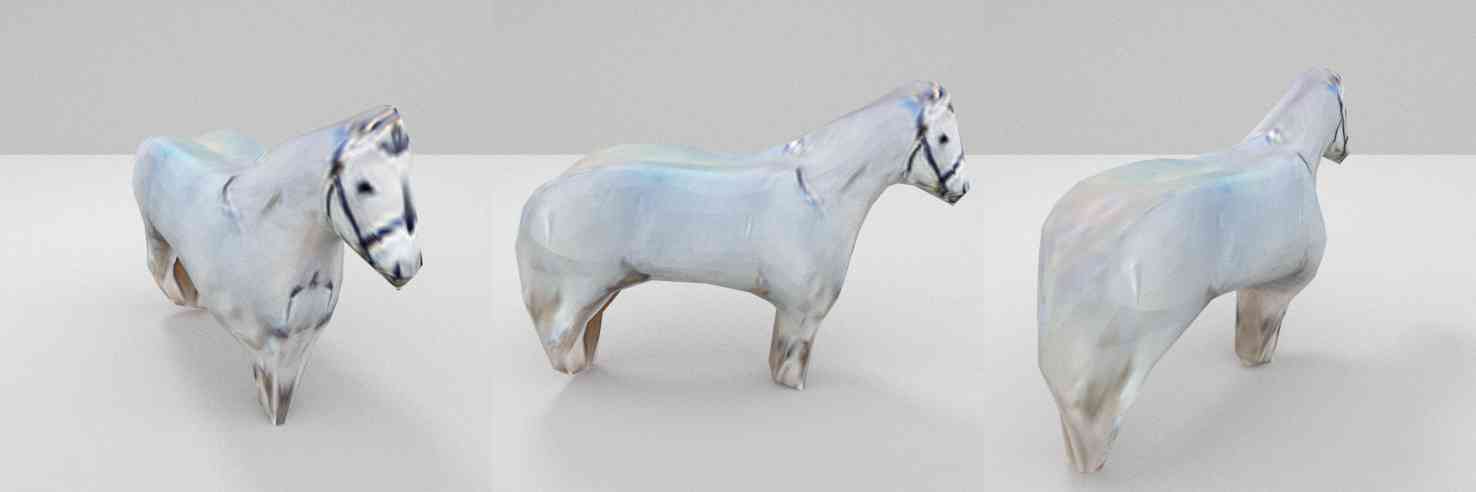}}
					\\[-1.5mm]						
								
					{\scriptsize Input} & {\scriptsize Prediction}  & \multicolumn{3}{c}{{\scriptsize Multiple Views}}  & {\scriptsize Input} & {\scriptsize Prediction}  & \multicolumn{3}{c}{{\scriptsize Multiple Views}} 
				\end{tabular}
			\end{tabular}
		\end{center}
		\vspace*{-3mm}
	}\vspace*{-2mm}
	\caption{\label{fig:3d} \footnotesize \textbf{3D Reconstruction Results:} Given input images (1st column), we predict 3D shape, texture, and render them into the same viewpoint (2nd column). We also show renderings in 3 other views in remaining columns to showcase 3D quality. Our model is able to reconstruct cars with various shapes, textures and viewpoints. We also show the same approach on harder (articulated) objects, i.e., bird and horse.}
	\vspace*{-1mm}
\end{figure*}

\vspace{-2mm}
\subsection{3D Reconstruction Results}

\vspace{-2mm}
\paragraph{Training Details:}
Our DIB-R based inverse graphics model was trained with Adam (\cite{adam}), with a learning rate of 1e-4. We set $\lambda_{\text{IOU}}$, $\lambda_{\text{col}}$,  $\lambda_{\text{lap}}$, $\lambda_{\text{sm}}$ and $\lambda_{\text{mov}}$ to 3, 20, 5, 5, and 2.5, respectively. We first train the model with $L_{\text{col}}$ loss for 3K iterations, and then fine-tune the model by adding $L_{\text{pecept}}$ to make the texture more realistic. We set $\lambda_{\text{percept}}$ to 0.5. The model converges in 200K iterations with batch size 16. Training takes around 120 hours on four V100 GPUs. 

\begin{figure*}[t!]
	{
	\begin{minipage}{0.75\linewidth}
	
		\vspace*{-4mm}
			\setlength{\tabcolsep}{1pt}
			\setlength{\fboxrule}{0pt}
			\hspace*{-0.3cm}
			\begin{tabular}{c}
				\begin{tabular}{ccccc}
					\rotatebox{90}{\,\,\, {\color{black}{\scriptsize Pascal3D }}}&
					\includegraphics[height=0.137\linewidth]{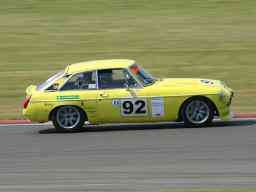}&
					\includegraphics[height=0.137\linewidth]{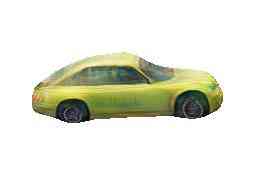}&
					\includegraphics[height=0.137\linewidth]{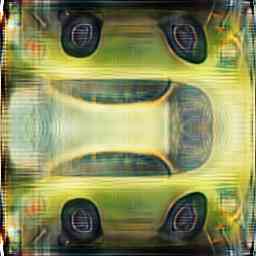}&
					\includegraphics[height=0.137\linewidth]{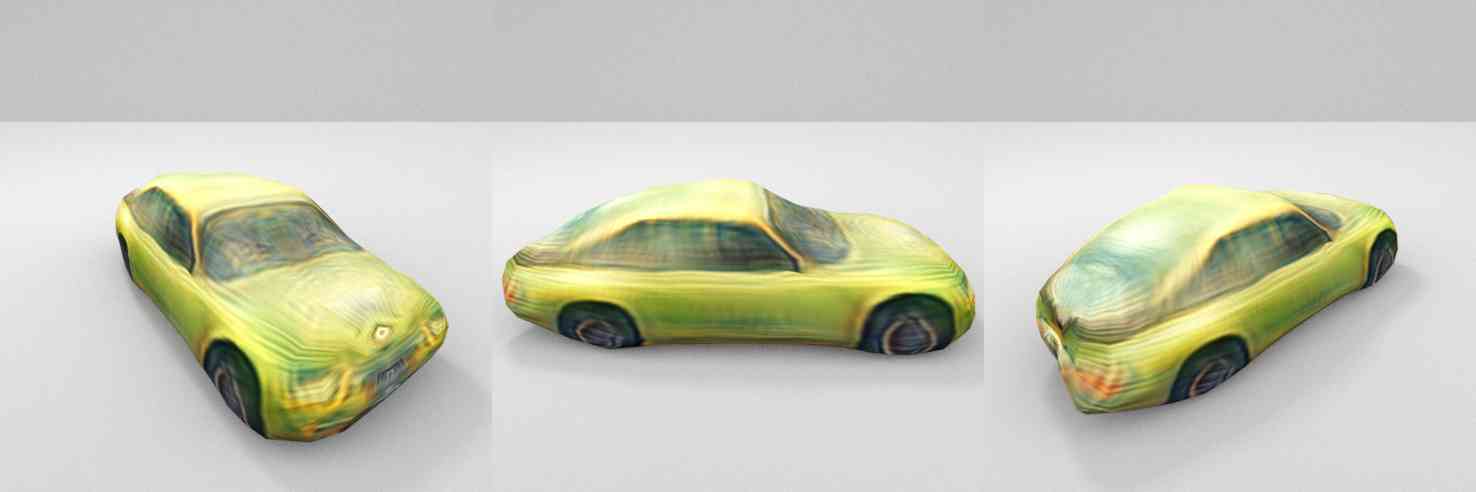}
					\\[-0.5mm]
					\rotatebox{90}{\,\,\,\,\,\,\,{\color{black}{\scriptsize Ours}}}&
					\includegraphics[height=0.137\linewidth]{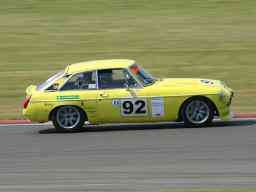}&
					\includegraphics[height=0.137\linewidth]{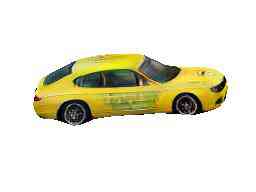}&
					\includegraphics[height=0.137\linewidth]{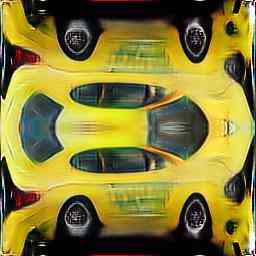}&
					\includegraphics[height=0.137\linewidth]{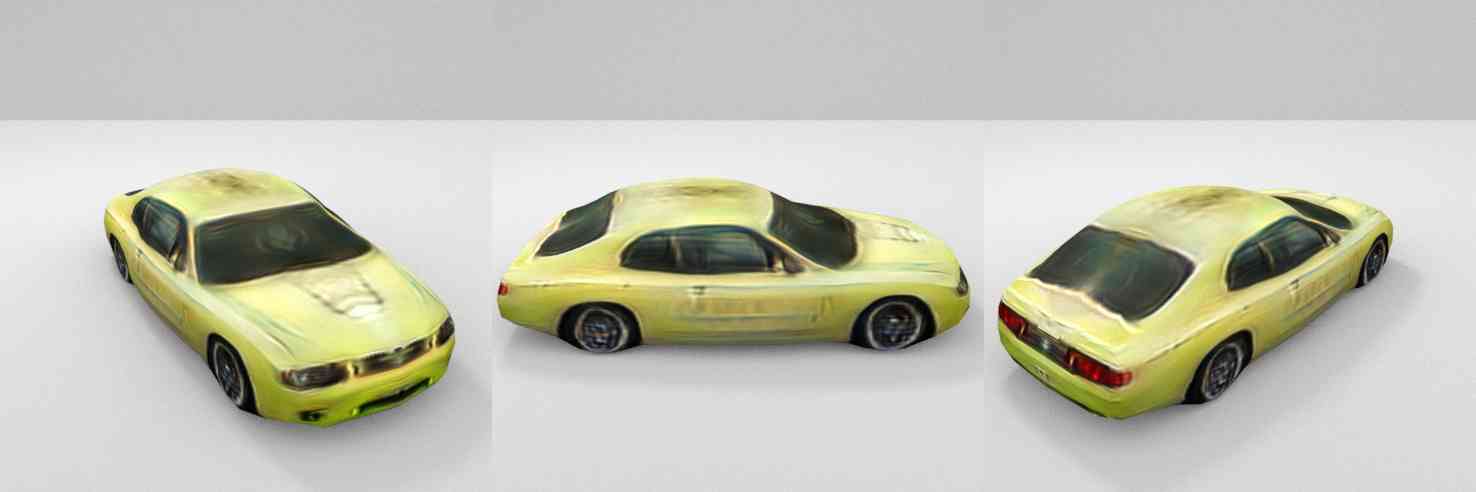}
					\\[-1.1mm]
					& {\scriptsize Input }& {\scriptsize Prediction} & {\scriptsize Texture} & {\scriptsize Multiple Rendered Views of Prediction}
					\hspace*{0pt}
				\end{tabular}
			\end{tabular}
		\vspace*{-0.5cm}
	
	\end{minipage}
	\begin{minipage}{0.245\linewidth}
	\vspace{-5mm}
	\caption{\label{fig:comppas}\footnotesize \textbf{Comparison on Pascal3D test set:} We compare inverse graphics networks trained on Pascal3D and  our StyleGAN dataset. Notice considerably higher quality of prediction when training on the StyleGAN dataset.}
	\end{minipage}
	}
	\vspace*{-0.2cm}
\end{figure*}

\vspace{-2mm}
\paragraph{Results:}
We show 3D reconstruction results in Fig.~\ref{fig:3d}. Notice the  quality of the predicted shapes and textures, and the diversity of the 3D car shapes we obtain. Our method also works well on more challenging (articulated) classes, e.g. horse and bird. We provide additional examples in Appendix. 

\vspace*{-2mm}
\paragraph{Qualitative Comparison:}
To showcase our approach, we compare our inverse graphics network trained on our StyleGAN dataset with exactly the same model but which we train on the Pascal3D car dataset. 
Pascal3D dataset has annotated keypoints, which we utilize to train the baseline model, termed as as Pascal3D-model. We show qualitative comparison on Pascal3D test set in Fig.~\ref{fig:comppas}. Note that the images from Pascal3D dataset are different from those our StyleGAN-model was trained on. Although the Pascal3D-model's prediction is visually good in the input image view, rendered predictions in other views are of noticeably lower quality than ours, which demonstrates that we recover 3D geometry and texture better than the baseline. 


\begin{table*} [htb!]  
	\vspace{-3mm}
	\begin{center}
		\addtolength{\tabcolsep}{-1.8pt}
		\scriptsize
		\centering
		
			\begin{tabular}{ccc}
				
		\begin{tabular}{lcc}
			\toprule
			Dataset & Size & Annotation \\                                   
			\midrule
			Pascal3D& 4K  & 200-350h \\
			StyleGAN   & \textbf{50K}  & $\sim$\textbf{1min}
			\\
			\bottomrule
		\end{tabular}
	&
		\begin{tabular}{lcc}
		\toprule
		 Model & Pascal3D test & StyleGAN test \\                                      
		\midrule
		Pascal3D&  \textbf{0.80} & 0.81\\
		Ours   & 0.76 & \textbf{0.95}\\ 
		\bottomrule
	\end{tabular}
&
\begin{tabular}{lccc}
	\toprule
	 & Overall & Shape & Texture \\
	\midrule
	Ours  & \textbf{57.5\%}  & \textbf{61.6\%}  & \textbf{56.3\%}  \\
	Pascal3D-model & 25.9\% &26.4\% & 32.8\% \\
	No Preference & 16.6\% & 11.9\% & 10.8\%  \\
	\bottomrule
\end{tabular}
\\
(a) Dataset Comparison & (b) 2D IOU Evaluation & (c) User Study	
\end{tabular}	

	\end{center}
	\vspace{-4mm}
\caption{\footnotesize \textbf{(a):} We compare dataset size and annotation time of Pascal3D with our StyleGAN dataset. 
	\textbf{(b):} We evaluate re-projected 2D IOU score of our StyleGAN-model vs the baseline Pascal3D-model on the two datasets. 
	\textbf{(c):} We conduct a user study to judge the quality of 3D estimation. 
	}
	\vspace{-5pt}
	\label{tab:pascal}
\end{table*}

\vspace*{-2mm}
\paragraph{Quantitative Comparison:}
We evaluate the two networks in Table~\ref{tab:pascal} for the car class. We report the estimated annotation time in Table.~\ref{tab:pascal} (a) to showcase efficiency behind our StyleGAN dataset. It takes 3-5 minutes to annotate keypoints for one object, which we empirically verify. Thus, labeling Pascal3D required around 200-350 hours while ours takes only 1 minute to annotate a 10 times larger dataset. In Table~\ref{tab:pascal} (b), we evaluate shape prediction quality by the re-projected 2D IOU score. Our model outperforms the Pascal3D-model on the SyleGAN test set while Pascal3D-model is better on the Pascal test set. This is not surprising since there is a domain gap between two datasets and thus each one performs best on their own test set. Note that this metric only evaluates quality of the prediction in input view and thus not reflect the actual quality of the predicted 3D shape/texture. 

To analyze the quality of 3D prediction, we conduct an AMT user study on the \emph{Pascal3D test set} which contains 220 images. We provide users with the input image and predictions rendered in 6 views (shown in Fig.~\ref{fig:comppas}, right) for both models. We ask them to choose the model with a more realistic shape and texture prediction that matches the input object. We provide details of the study in the Appendix. 
We report results in  Table.~\ref{tab:pascal} (c). Users show significant preference of our results versus the baseline, which confirms that the quality of our 3D estimation. 

\begin{figure*}[t]
	{
		\vspace*{-0pt}
		\begin{center}
			\setlength{\tabcolsep}{1pt}
			\setlength{\fboxrule}{0pt}
			\hspace*{-0.25cm}
				\begin{tabular}{ccccc}
					\rotatebox{90}{\,\,\,\,\,\, {\color{black}{\tiny Full}}}&
					\includegraphics[width=0.13\textwidth]{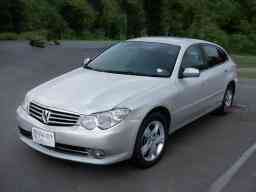}&
					\includegraphics[width=0.13\textwidth]{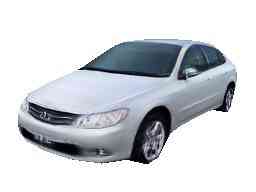}&
					\includegraphics[width=0.1\textwidth]{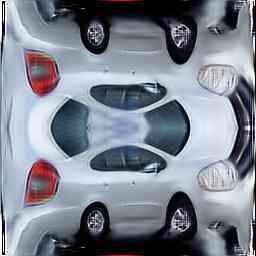}&
					\includegraphics[width=0.6\textwidth]{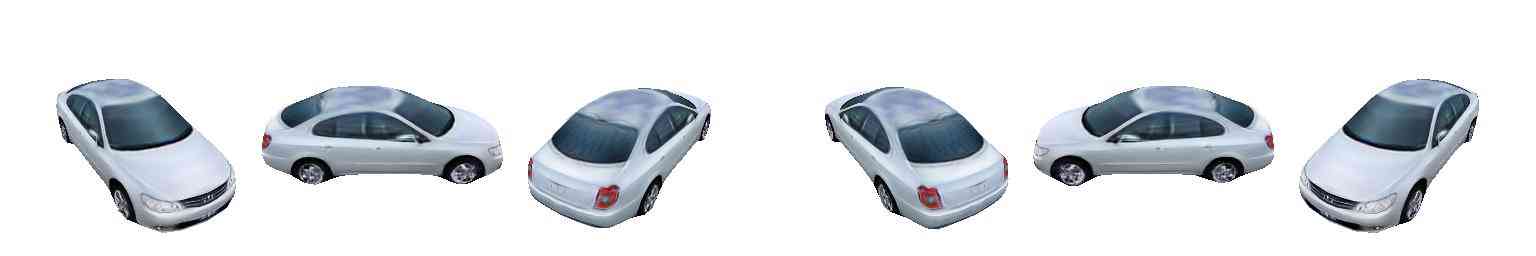}
					\\
					\rotatebox{90}{\,\,\, {\color{black}{\tiny w.o M. V.}}}&
					\includegraphics[width=0.13\textwidth]{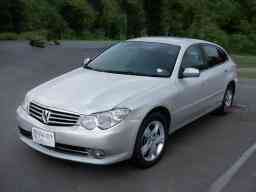}&
					\includegraphics[width=0.13\textwidth]{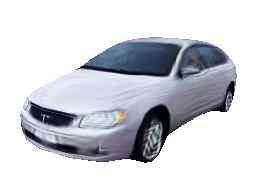}&
					\includegraphics[width=0.1\textwidth]{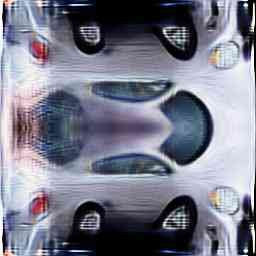}&
					\includegraphics[width=0.6\textwidth]{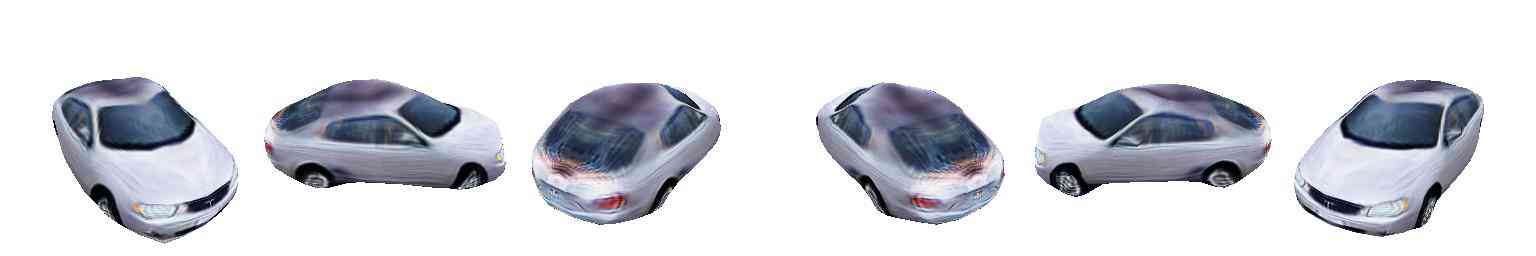}
					\\
					& {\scriptsize Input } & {\scriptsize Prediction } & {\scriptsize Texture} & {\scriptsize Prediction rendered in Multiple Views}
					\hspace*{0pt}
				\end{tabular}
		\end{center}
	}
	\vspace{-4mm}
	\caption{\footnotesize \textbf{Ablation Study:} We ablate the use of multi-view consistency loss. Both texture are shape are worse without this loss, especially in the invisible parts (rows 2, 5, denoted by ``w.o M. V.'' -- no multi-view consistency used during training), showcasing the importance of our StyleGAN-multivew dataset.}
	\label{fig:ablation}
	\vspace*{-2mm}
\end{figure*}

\vspace{-2mm}
\paragraph{Ablation study:} In Fig~\ref{fig:ablation} we ablate the importance of using multiple views in our dataset, i.e., by encouraging multi-view consistency loss during training. We compare predictions from inverse graphics networks trained with and without this loss, with significant differences in quality.

\begin{figure*}[t!]
	{
		\vspace*{-5mm}
		\begin{center}
			\setlength{\tabcolsep}{0.8pt}
			\setlength{\fboxrule}{0pt}
			\hspace*{-0.3cm}
			\begin{tabular}{c}	
				\begin{tabular}{cccccc}
					\scriptsize Input & \scriptsize DIBR-R  & \scriptsize Stylegan-R & \scriptsize Input & \scriptsize DIBR-R  & \scriptsize Stylegan-R 
					\\
					\includegraphics[width=.16\textwidth,height=.13\textwidth]{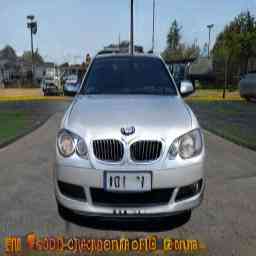}&
					\includegraphics[width=.16\textwidth,height=.13\textwidth]{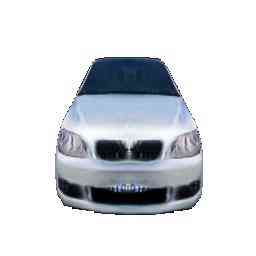}&
					\includegraphics[width=.16\textwidth,height=.13\textwidth]{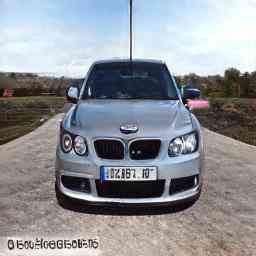}&
					\includegraphics[width=.16\textwidth,height=.13\textwidth]{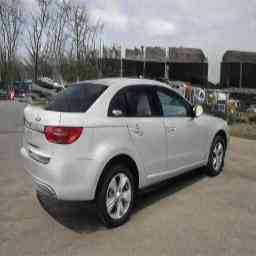}&
					\includegraphics[width=.16\textwidth,height=.13\textwidth]{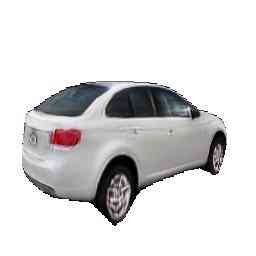}&
					\includegraphics[width=.16\textwidth,height=.13\textwidth]{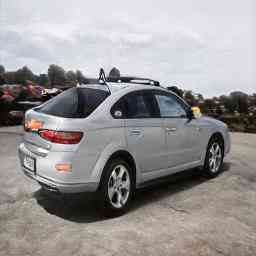} 
					\\
				\end{tabular}
			\end{tabular}
		\end{center}
		\vspace*{-6mm}
	}
	\caption{\footnotesize \textbf{Dual Renderer: } Given input images (1st column), we first predict mesh and texture, and render them with the graphics renderer (2nd column), and our {\ours} (3rd column). 
	}	
	\label{fig:dualrender} 
	\vspace*{-1.5mm}
\end{figure*}

\vspace{-2mm}
\subsection{Dual Renderers}
\vspace{-1mm}
\paragraph{Training Details:} We train {\ours} using Adam with learning rate of 1e-5 and batch size 16. Warmup stage takes 700 iterations, and we perform joint fine-tuning for another 2500 iterations. 

With the provided input image, we first predict mesh and texture using the trained inverse graphics model, and then feed these 3D properties into {\ours} to generate a new image. For comparison, we feed the same 3D properties to the DIB-R graphics renderer (which is the OpenGL renderer). Results are provided in Fig.~\ref{fig:dualrender}. Note that DIB-R can only render the predicted object, while {\ours} also has the ability to render the object into a desired background. We find that {\ours} produces relatively consistent images compared to the input image. Shape and texture are well preserved, while only the background has a slight content shift. 

\begin{figure*}[t]
	\vspace*{-3mm}
	{
		\begin{center}
			\setlength{\tabcolsep}{1pt}
			\setlength{\fboxrule}{0pt}
			\hspace*{-0.15cm}
			\begin{tabular}{c}
				\begin{tabular}{ccccccc}
					\includegraphics[width=.138\textwidth]{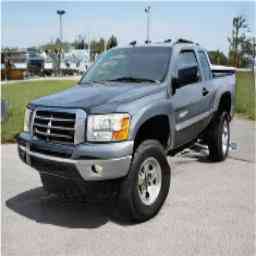}&
					\includegraphics[width=.138\textwidth]{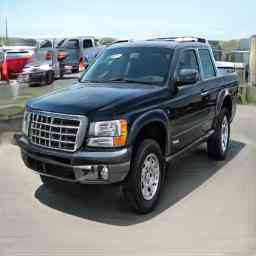}&
					\includegraphics[width=.138\textwidth]{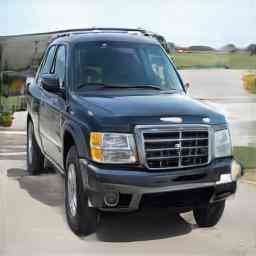}&
					\includegraphics[width=.138\textwidth]{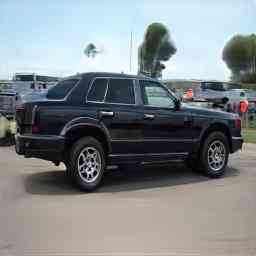}&
					\includegraphics[width=.138\textwidth]{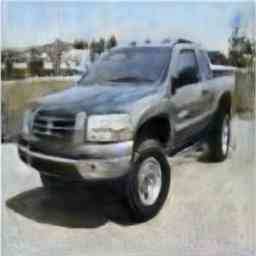}&
					\includegraphics[width=.138\textwidth]{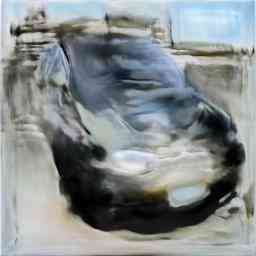}&
					\includegraphics[width=.138\textwidth]{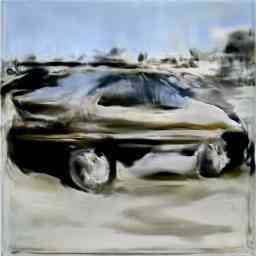}
					\\[-1.5mm]
					{\footnotesize Input} & \multicolumn{3}{c}{{\scriptsize Mapping from 3D properties}} & \multicolumn{3}{c}{{\scriptsize latent code optimization}} 
					\hspace*{0pt}
				\end{tabular}
			\end{tabular}
		\end{center}
		\vspace*{-4.5mm}
	}
	\caption{\footnotesize\textbf{Latent code manipulation:} Given an input image (col 1), we predict 3D properties and synthesize a new image with {\ours}, by manipulating the viewpoint (col 2, 3, 4). Alternatively, we directly optimize the (original) StyleGAN latent code w.r.t. image, however this leads to a blurry reconstruction (col 5). Moreover, when we try to adjust the style for the optimized code, we get low quality results (col 6, 7).}
	\label{fig:anothercode}
	\vspace*{-3mm}	
\end{figure*}

\begin{figure*}[t!]
	{
		\vspace{-3mm}
		\hspace{-1mm}
		\begin{minipage}{0.70\linewidth}
			\setlength{\tabcolsep}{1pt}
			\setlength{\fboxrule}{0pt}
			\begin{tabular}{c}
				\begin{tabular}{ccccccc}
					\\
					\rotatebox{90}{\,\,\,\,{\color{black}{\scriptsize Scale}}}&
					\includegraphics[width=.143\textwidth]{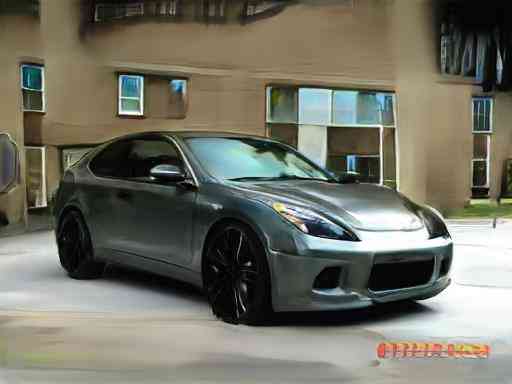}&
					\includegraphics[width=.143\textwidth]{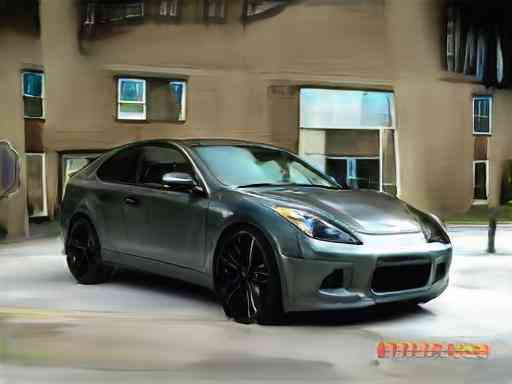}&
					\includegraphics[width=.143\textwidth]{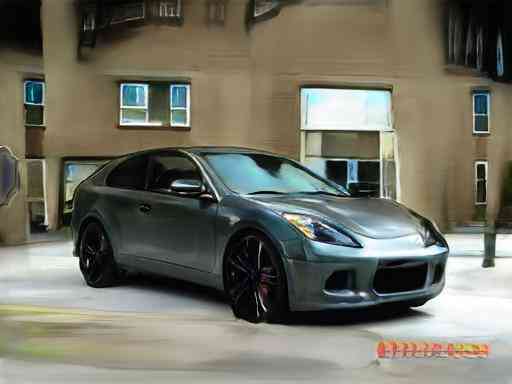}&
					\includegraphics[width=.143\textwidth]{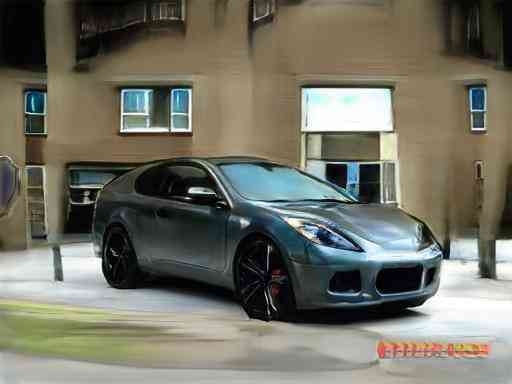}&
					\includegraphics[width=.143\textwidth]{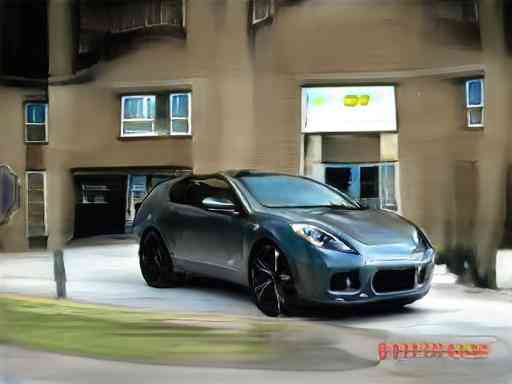}&
					\includegraphics[width=.143\textwidth]{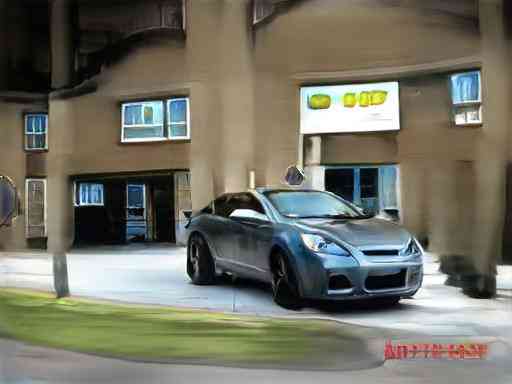}
					\\[-0.9mm]
					\rotatebox{90}{\,{\color{black}{\scriptsize Azimuth}}}&
					\includegraphics[width=.143\textwidth]{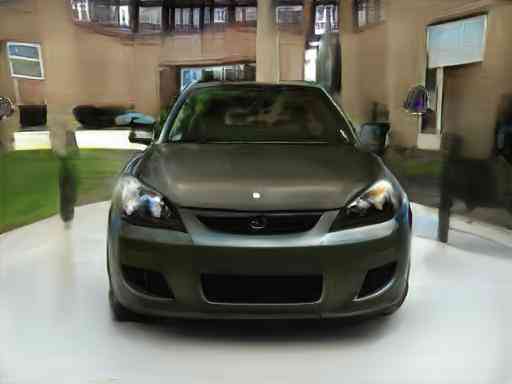}&
					\includegraphics[width=.143\textwidth]{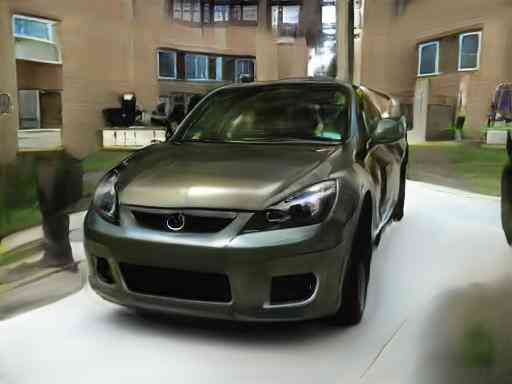}&
					\includegraphics[width=.143\textwidth]{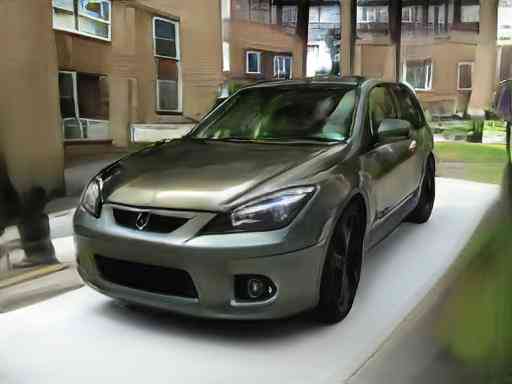}&
					\includegraphics[width=.143\textwidth]{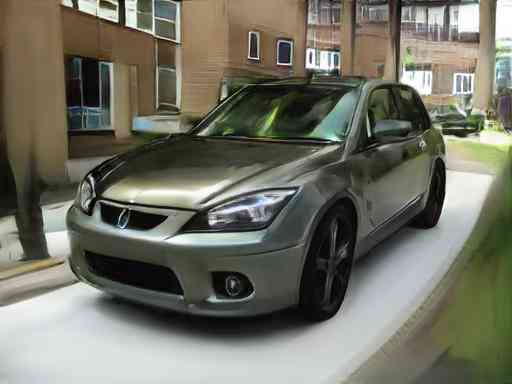}&
					\includegraphics[width=.143\textwidth]{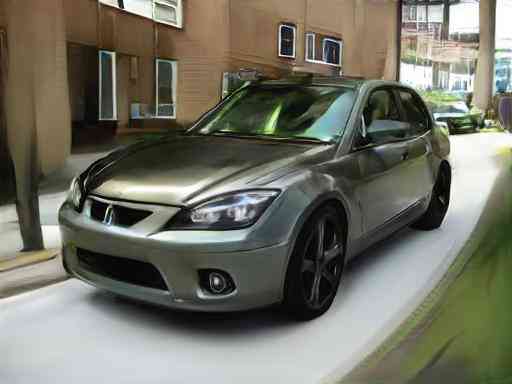}&
					\includegraphics[width=.143\textwidth]{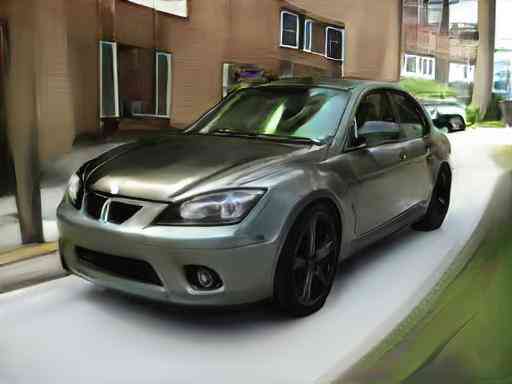}
					\\[-0.9mm]
					\rotatebox{90}{\,{\color{black}{\scriptsize Elevation}}}&
					\includegraphics[width=.143\textwidth]{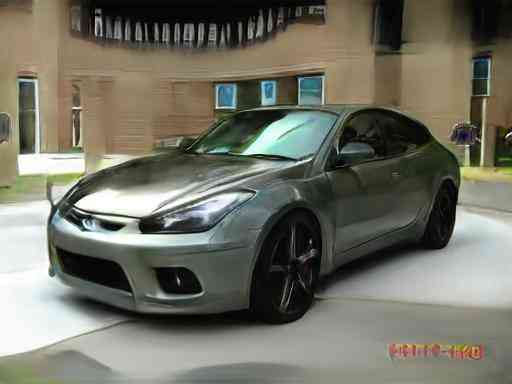}&
					\includegraphics[width=.143\textwidth]{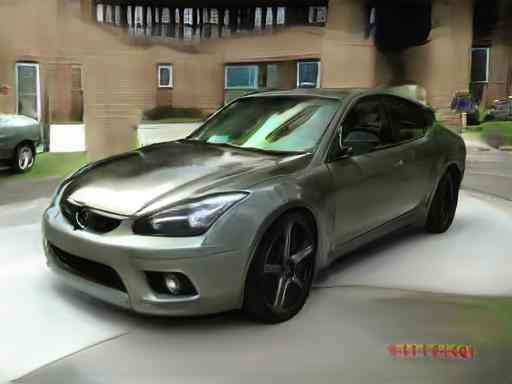}&
					\includegraphics[width=.143\textwidth]{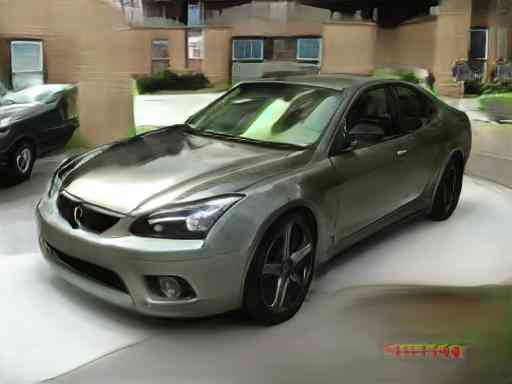}&
					\includegraphics[width=.143\textwidth]{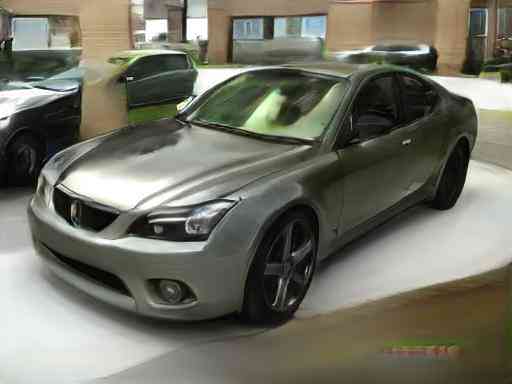}&
					\includegraphics[width=.143\textwidth]{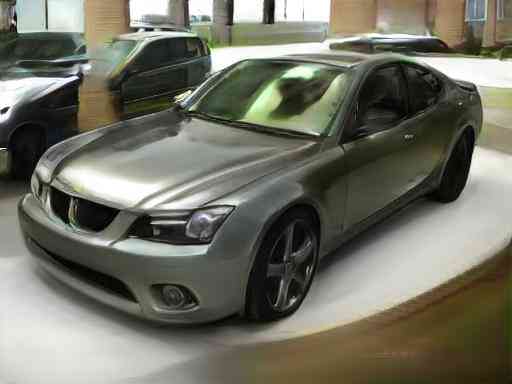}&
					\includegraphics[width=.143\textwidth]{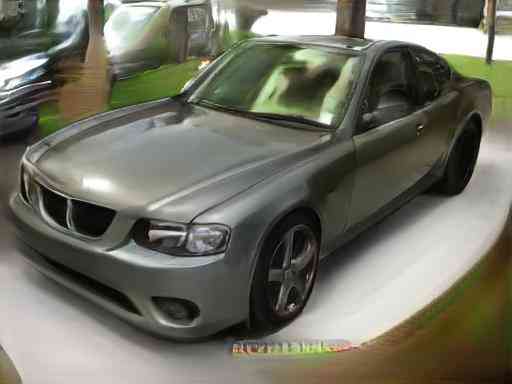}\\[-1mm]
				\end{tabular}
			\end{tabular}
		\end{minipage}
	\hspace{-4mm}
	\begin{minipage}{0.3\linewidth}
		\caption{\label{fig:camera_controlll}\footnotesize \textbf{Camera Controller:} We manipulate azimuth, scale, elevation parameters with {\ours} to synthesize images in new viewpoints while keeping content code fixed.}
	\end{minipage}
	}
	\vspace*{-1mm}
\end{figure*}

\begin{figure*}[t!]
	{
		\vspace*{-1mm}
		\begin{minipage}{0.69\linewidth}
			\setlength{\tabcolsep}{0.0pt}
			\setlength{\fboxrule}{0pt}
			\hspace*{-0.25cm}
			\begin{tabular}{c}
				\begin{tabular}{ccccc}
					& {\scriptsize Sampled Cars} & \cellcolor{red!40}{\scriptsize Shape Swap} & {\cellcolor{lime!50}{\scriptsize Texture Swap}} & {\cellcolor{cyan!40}{\scriptsize Background Swap}}
					\\
					\rotatebox{90}{\,\,\,\,\,{\color{black}{\scriptsize Car 1}}}&
					{\setlength{\fboxrule}{0mm}\fcolorbox{red}{red}{\includegraphics[width=.224\textwidth]{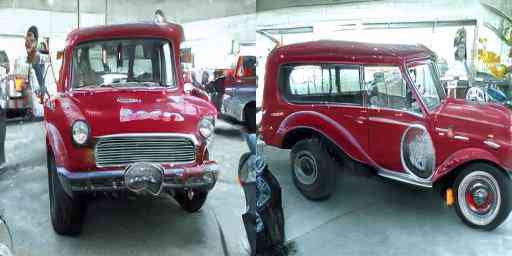}}}&
					{\setlength{\fboxrule}{0mm}\fcolorbox{white}{red}{\includegraphics[width=.224\textwidth]{figures/5_cars_example_2/change_shape/car_000.jpg}}}&
					\includegraphics[width=.224\textwidth]{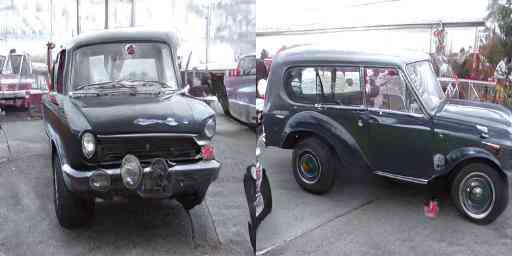}&
					\includegraphics[width=.224\textwidth]{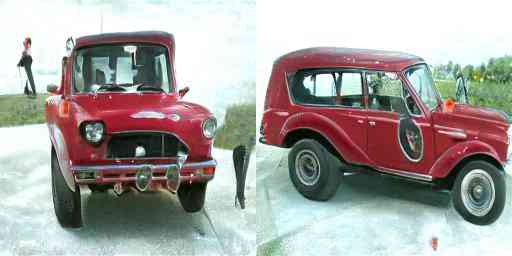}
					\\
					\rotatebox{90}{\,\,\,\,\, {\color{black}{\scriptsize Car 2}}}&
					{\setlength{\fboxrule}{0mm}\fcolorbox{lime}{lime}{\includegraphics[width=.224\textwidth]{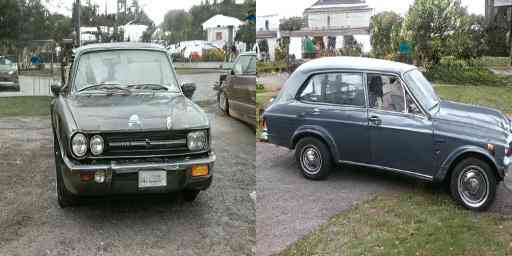}}}&
					\includegraphics[width=.224\textwidth]{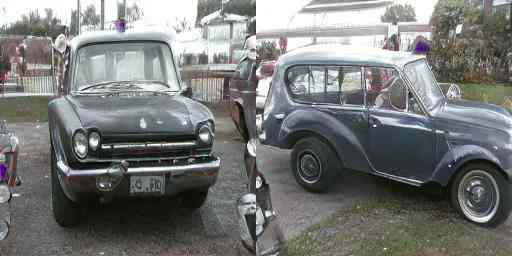}&
					\fcolorbox{white}{lime}{\includegraphics[width=.224\textwidth]{figures/5_cars_example_2/change_texture/car_003.jpg}}&
					\includegraphics[width=.224\textwidth]{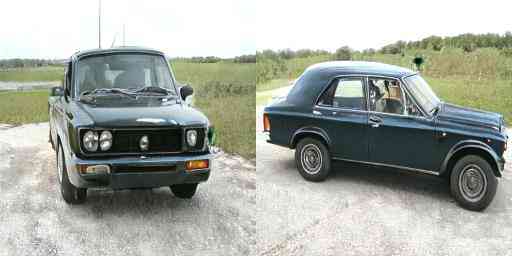}
					\\
					\rotatebox{90}{\,\,\,\,\,{\color{black}{\scriptsize Car 3}}}&
					{\setlength{\fboxrule}{0mm}\fcolorbox{white}{cyan}{\includegraphics[width=.224\textwidth]{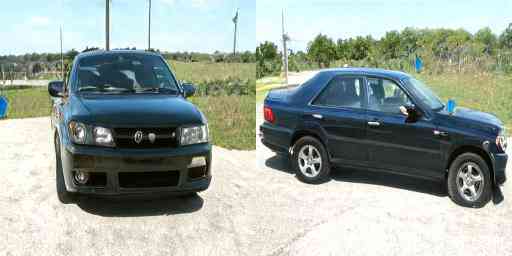}}}&
					\includegraphics[width=.224\textwidth]{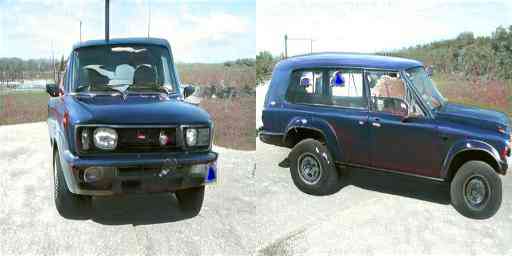}&
					\includegraphics[width=.224\textwidth]{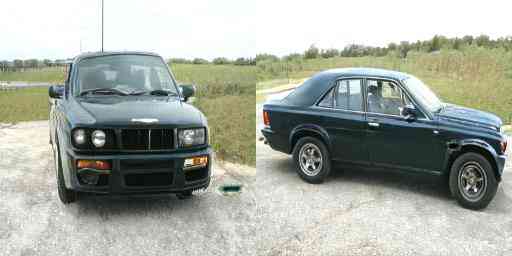}&
					{\setlength{\fboxrule}{0mm}\fcolorbox{white}{cyan}{\includegraphics[width=.224\textwidth]{figures/5_cars_example_2/change_background/car_001.jpg}}}
					\hspace*{0pt}
				\end{tabular}
			\end{tabular}
		\end{minipage}
		\hspace{-1mm}
		\begin{minipage}{0.295\linewidth}
		\vspace*{-3mm}
	\caption{\footnotesize\textbf{3D Manipulation:} We  sample 3 cars in column 1. We replace the shape of all cars with the shape of Car 1 ({\color{red}red} box) in 2nd column. We transfer texture of Car 2 ({\color{green}green} box)  to other cars (3rd col). In last column, we paste background of Car 3 ({\color{cyan}cyan} box) to the other cars. Examples indicated with boxes are unchanged. Zoom in to  see details.}	
	\label{fig:styleganmanip}
	\end{minipage}
		}
	\vspace*{-2mm}
\end{figure*}

\begin{figure*}[t!]
	{
		\vspace*{0pt}
		\begin{center}
			\setlength{\tabcolsep}{1pt}
			\setlength{\fboxrule}{0pt}
			\hspace*{-0.3cm}
			\begin{tabular}{c}
				\begin{tabular}{cccccc}
					\rotatebox{90}{\,\,\,\,\,\,{\color{black}{\scriptsize Car 1}}}&
					\includegraphics[width=.13\textwidth,height=.09\textwidth]{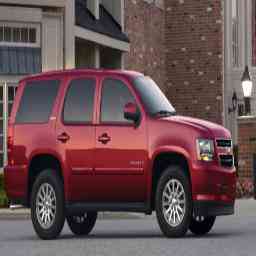}&
					\includegraphics[width=.21\textwidth,height=.09\textwidth]{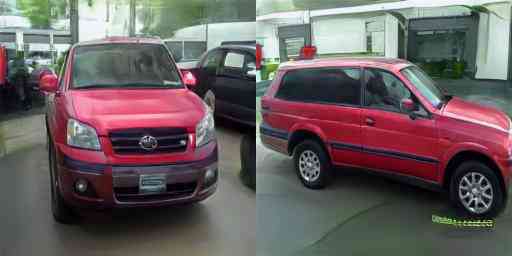}&
					\includegraphics[width=.21\textwidth,height=.09\textwidth]{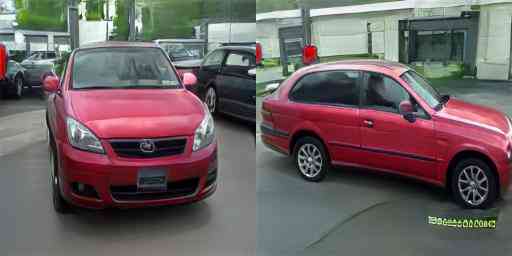}&
					\includegraphics[width=.21\textwidth,height=.09\textwidth]{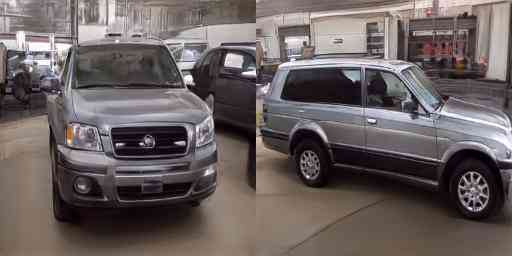}&
					\includegraphics[width=.21\textwidth,height=.09\textwidth]{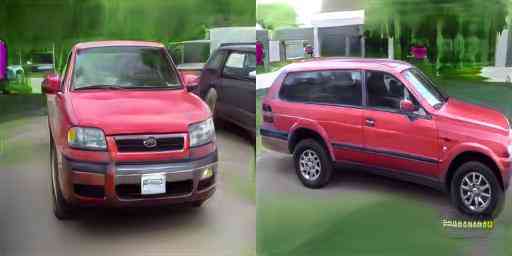}
					\\[-0.5mm]
					\rotatebox{90}{\,\,\,\, {\color{black}{\scriptsize Car 2}}}&
					\includegraphics[width=.13\textwidth,height=.09\textwidth]{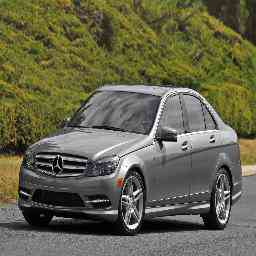}&
					\includegraphics[width=.21\textwidth,height=.09\textwidth]{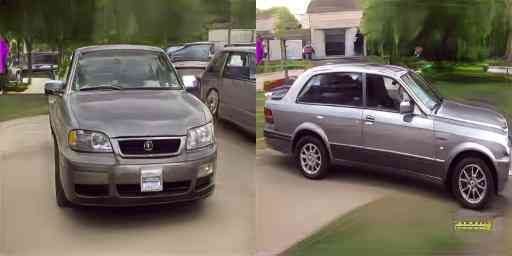}&
					\includegraphics[width=.21\textwidth,height=.09\textwidth]{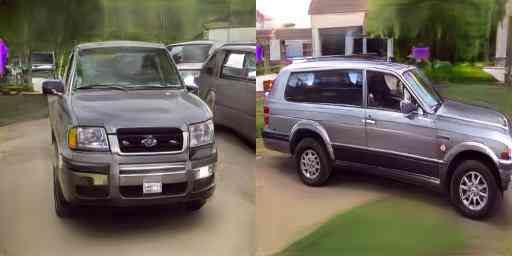}&
					\includegraphics[width=.21\textwidth,height=.09\textwidth]{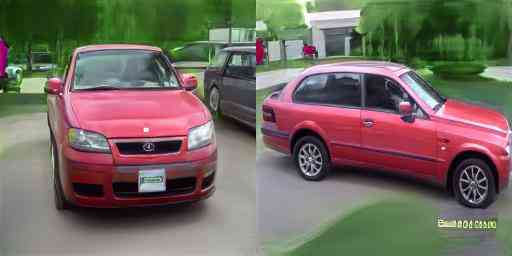}&
					\includegraphics[width=.21\textwidth,height=.09\textwidth]{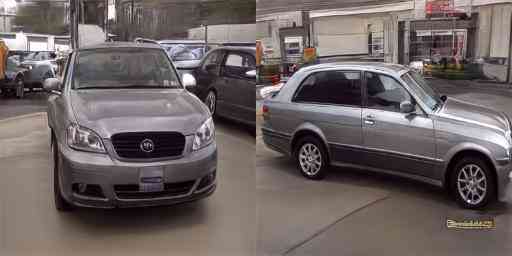}
					\\[-1mm]
					& {\scriptsize Input} & {\scriptsize Neural R.} & {\scriptsize  Shape Swap} & {\scriptsize  Texture Swap} & {\scriptsize Bck. Swap}
					\hspace*{0pt}
				\end{tabular}
			\end{tabular}
		\end{center}
		\vspace*{-5mm}
	}
	\caption{\footnotesize \textbf{Real Image Manipulation:} Given input images (1st col), we predict 3D properties and use our {\ours} to render them back (2nd col). We swap out shape, texture \& background in cols 3-5.}
	\label{fig:real} 
	\vspace*{-3.5mm}	
\end{figure*}

\vspace{-2mm}
\subsection{3D Image Manipulation with {\ours}}
\vspace{-1mm}
We test our approach in manipulating StyleGAN-synthesized images from our test set and real images. Specifically, given an input image, we predict 3D properties using the inverse graphics network, and extract background by masking out the object with Mask-RCNN. We then manipulate and feed these properties to {\ours} to synthesize new views.
\vspace{-2mm}
\paragraph{Controlling Viewpoints:} We first freeze shape, texture and background, and change the camera viewpoint. Example is shown in Fig.~\ref{fig:camera_controlll}. We obtain meaningful results, particularly for shape and texture.  
For comparison, an alternative way that has been explored in literature is to directly optimize the GAN's latent code (in our case the original StyleGAN's code) via an L2 image reconstruction loss. Results are shown in the last three columns in Fig.~\ref{fig:anothercode}. As also observed in~\cite{wonka19}, this approach fails to generate plausible images, showcasing the importance of the mapping network and fine-tuning the entire architecture with 3D inverse graphics network in the loop. 
\vspace{-2mm}
\paragraph{Controlling Shape, Texture and Background:} We further aim to manipulate 3D properties, while keeping the camera viewpoint fixed. In the second column of Fig~\ref{fig:styleganmanip}, we replace the shapes of all cars to one chosen shape (red box) and perform neural rendering using {\ours}. We successfully  swap the shape of the car while maintaining other properties. We are able to modify tiny parts of the car, such as trunk and headlights. We do the same experiment but swapping texture and background in the third and forth column of Fig~\ref{fig:styleganmanip}. We notice that swapping textures also slightly modifies the background, pointing that further improvements are possible in disentangling the two. 
\vspace{-2mm}
\paragraph{Real Image Editing:} 
As shown in Fig.~\ref{fig:real}, our framework also works well when provided with real images, since StyleGAN's images, which we use in training, are quite realistic. 



\vspace{-3mm}
\subsection{Limitations}
\label{limit}
\vspace{-3mm}
While recovering faithful 3D gemetry and texture, our model fails to predict correct lighting. Real images and StyleGAN-generated images contain advanced lighting effects such as reflection, transparency and shadows, and our spherical harmonic lighting model is incapable in dealing with it successfully.  We also only partly succeed at disentangling the background, which one can see by noticing slight changes in background in Fig.~\ref{fig:dualrender}, Fig.~\ref{fig:styleganmanip} and Fig.~\ref{fig:real}. Predicting faithful shapes for out-of-distribution objects as discussed in Appendix is also a significant challenge. We leave improvements to future work. 


\vspace{-3mm}
\section{Conclusion}
\vspace{-3mm}
In this paper, we introduced a new powerful architecture that links two renderers: a state-of-the-art image synthesis network and a differentiable graphics renderer. The image synthesis network generates training data for an inverse graphics network. In turn, the inverse graphics network teaches the synthesis network about the physical 3D controls. We showcased our approach to obtain significantly higher quality 3D reconstruction results while requiring 10,000$\times$ less annotation effort than standard datasets. We also provided 3D neural rendering and image manipulation results demonstrating the effectiveness of our approach.

\begin{small}
\bibliography{egbib}

\begin{thebibliography}{50}
\providecommand{\natexlab}[1]{#1}
\providecommand{\url}[1]{\texttt{#1}}
\expandafter\ifx\csname urlstyle\endcsname\relax
  \providecommand{\doi}[1]{doi: #1}\else
  \providecommand{\doi}{doi: \begingroup \urlstyle{rm}\Url}\fi

\bibitem[Abdal et~al.(2019)Abdal, Qin, and Wonka]{wonka19}
Rameen Abdal, Yipeng Qin, and Peter Wonka.
\newblock Image2stylegan: How to embed images into the stylegan latent space?
\newblock \emph{CoRR}, abs/1904.03189, 2019.
\newblock URL \url{http://arxiv.org/abs/1904.03189}.

\bibitem[Chang et~al.(2015)Chang, Funkhouser, Guibas, Hanrahan, Huang, Li,
  Savarese, Savva, Song, Su, et~al.]{shapenet}
Angel~X Chang, Thomas Funkhouser, Leonidas Guibas, Pat Hanrahan, Qixing Huang,
  Zimo Li, Silvio Savarese, Manolis Savva, Shuran Song, Hao Su, et~al.
\newblock Shapenet: An information-rich 3d model repository.
\newblock \emph{arXiv preprint arXiv:1512.03012}, 2015.

\bibitem[Chen et~al.(2019)Chen, Gao, Ling, Smith, Lehtinen, Jacobson, and
  Fidler]{dib-r}
Wenzheng Chen, Jun Gao, Huan Ling, Edward Smith, Jaakko Lehtinen, Alec
  Jacobson, and Sanja Fidler.
\newblock Learning to predict 3d objects with an interpolation-based
  differentiable renderer.
\newblock In \emph{Advances In Neural Information Processing Systems}, 2019.

\bibitem[Chen et~al.(2016)Chen, Duan, Houthooft, Schulman, Sutskever, and
  Abbeel]{infogan}
Xi~Chen, Yan Duan, Rein Houthooft, John Schulman, Ilya Sutskever, and Pieter
  Abbeel.
\newblock Infogan: Interpretable representation learning by information
  maximizing generative adversarial nets.
\newblock In \emph{Advances in neural information processing systems}, pp.\
  2172--2180, 2016.

\bibitem[Choy et~al.(2016)Choy, Xu, Gwak, Chen, and Savarese]{3dr2n2}
Christopher~B Choy, Danfei Xu, JunYoung Gwak, Kevin Chen, and Silvio Savarese.
\newblock 3d-r2n2: A unified approach for single and multi-view 3d object
  reconstruction.
\newblock In \emph{ECCV}, 2016.

\bibitem[Dosovitskiy et~al.(2016)Dosovitskiy, Springenberg, Tatarchenko, and
  Brox]{dosovitskiy2016learning}
Alexey Dosovitskiy, Jost~Tobias Springenberg, Maxim Tatarchenko, and Thomas
  Brox.
\newblock Learning to generate chairs, tables and cars with convolutional
  networks.
\newblock \emph{IEEE transactions on pattern analysis and machine
  intelligence}, 39\penalty0 (4):\penalty0 692--705, 2016.

\bibitem[Gao et~al.(2020)Gao, Chen, Xiang, Tsang, Jacobson, McGuire, and
  Fidler]{deftet}
Jun Gao, Wenzheng Chen, Tommy Xiang, Clement~Fuji Tsang, Alec Jacobson, Morgan
  McGuire, and Sanja Fidler.
\newblock Learning deformable tetrahedral meshes for 3d reconstruction.
\newblock In \emph{Advances In Neural Information Processing Systems}, 2020.

\bibitem[Goel et~al.(2020)Goel, Kanazawa, , and Malik]{ucmrGoel20}
Shubham Goel, Angjoo Kanazawa, , and Jitendra Malik.
\newblock Shape and viewpoints without keypoints.
\newblock In \emph{ECCV}, 2020.

\bibitem[Goodfellow et~al.(2014)Goodfellow, Pouget-Abadie, Mirza, Xu,
  Warde-Farley, Ozair, Courville, and Bengio]{NIPS2014_5423}
Ian Goodfellow, Jean Pouget-Abadie, Mehdi Mirza, Bing Xu, David Warde-Farley,
  Sherjil Ozair, Aaron Courville, and Yoshua Bengio.
\newblock Generative adversarial nets.
\newblock In Z.~Ghahramani, M.~Welling, C.~Cortes, N.~D. Lawrence, and K.~Q.
  Weinberger (eds.), \emph{Advances in Neural Information Processing Systems
  27}, pp.\  2672--2680. Curran Associates, Inc., 2014.
\newblock URL
  \url{http://papers.nips.cc/paper/5423-generative-adversarial-nets.pdf}.

\bibitem[Groueix et~al.(2018)Groueix, Fisher, Kim, Russell, and
  Aubry]{groueix2018}
Thibault Groueix, Matthew Fisher, Vladimir~G. Kim, Bryan Russell, and Mathieu
  Aubry.
\newblock {AtlasNet: A Papier-M\^ach\'e Approach to Learning 3D Surface
  Generation}.
\newblock In \emph{Proceedings IEEE Conf. on Computer Vision and Pattern
  Recognition (CVPR)}, 2018.

\bibitem[H{\"a}rk{\"o}nen et~al.(2020)H{\"a}rk{\"o}nen, Hertzmann, Lehtinen,
  and Paris]{harkonen2020ganspace}
Erik H{\"a}rk{\"o}nen, Aaron Hertzmann, Jaakko Lehtinen, and Sylvain Paris.
\newblock Ganspace: Discovering interpretable gan controls.
\newblock \emph{arXiv preprint arXiv:2004.02546}, 2020.

\bibitem[He et~al.(2017)He, Gkioxari, Doll{\'{a}}r, and Girshick]{maskrcnn}
Kaiming He, Georgia Gkioxari, Piotr Doll{\'{a}}r, and Ross~B. Girshick.
\newblock Mask {R-CNN}.
\newblock \emph{CoRR}, abs/1703.06870, 2017.
\newblock URL \url{http://arxiv.org/abs/1703.06870}.

\bibitem[Henderson \& Ferrari(2018)Henderson and
  Ferrari]{henderson2018learning}
Paul Henderson and Vittorio Ferrari.
\newblock Learning to generate and reconstruct 3d meshes with only 2d
  supervision.
\newblock \emph{arXiv preprint arXiv:1807.09259}, 2018.

\bibitem[J. et~al.(2019)J., Smith, Lafleche, {Fuji Tsang}, Rozantsev, Chen,
  Xiang, Lebaredian, and Fidler]{kaolin2019arxiv}
{Krishna Murthy} J., Edward Smith, Jean-Francois Lafleche, Clement {Fuji
  Tsang}, Artem Rozantsev, Wenzheng Chen, Tommy Xiang, Rev Lebaredian, and
  Sanja Fidler.
\newblock Kaolin: A pytorch library for accelerating 3d deep learning research.
\newblock \emph{arXiv:1911.05063}, 2019.

\bibitem[Kanazawa et~al.(2018)Kanazawa, Tulsiani, Efros, and
  Malik]{kanazawa2018learning}
Angjoo Kanazawa, Shubham Tulsiani, Alexei~A Efros, and Jitendra Malik.
\newblock Learning category-specific mesh reconstruction from image
  collections.
\newblock In \emph{ECCV}, pp.\  371--386, 2018.

\bibitem[Karacan et~al.(2016)Karacan, Akata, Erdem, and
  Erdem]{karacan2016learning}
Levent Karacan, Zeynep Akata, Aykut Erdem, and Erkut Erdem.
\newblock Learning to generate images of outdoor scenes from attributes and
  semantic layouts.
\newblock \emph{arXiv preprint arXiv:1612.00215}, 2016.

\bibitem[Karras et~al.(2019{\natexlab{a}})Karras, Laine, and Aila]{stylegan}
Tero Karras, Samuli Laine, and Timo Aila.
\newblock A style-based generator architecture for generative adversarial
  networks.
\newblock In \emph{CVPR}, 2019{\natexlab{a}}.

\bibitem[Karras et~al.(2019{\natexlab{b}})Karras, Laine, Aittala, Hellsten,
  Lehtinen, and Aila]{stylegan2}
Tero Karras, Samuli Laine, Miika Aittala, Janne Hellsten, Jaakko Lehtinen, and
  Timo Aila.
\newblock Analyzing and improving the image quality of {StyleGAN}.
\newblock \emph{CoRR}, abs/1912.04958, 2019{\natexlab{b}}.

\bibitem[Kato \& Harada(2019)Kato and Harada]{kato2019selfsupervised}
Hiroharu Kato and Tatsuya Harada.
\newblock Self-supervised learning of 3d objects from natural images, 2019.

\bibitem[Kato et~al.(2018)Kato, Ushiku, and Harada]{NMR}
Hiroharu Kato, Yoshitaka Ushiku, and Tatsuya Harada.
\newblock Neural 3d mesh renderer.
\newblock In \emph{CVPR}, 2018.

\bibitem[Kingma \& Ba(2015)Kingma and Ba]{adam}
Diederik~P. Kingma and Jimmy Ba.
\newblock Adam: {A} method for stochastic optimization.
\newblock In Yoshua Bengio and Yann LeCun (eds.), \emph{3rd International
  Conference on Learning Representations, {ICLR} 2015, San Diego, CA, USA, May
  7-9, 2015, Conference Track Proceedings}, 2015.
\newblock URL \url{http://arxiv.org/abs/1412.6980}.

\bibitem[Lee et~al.(2020)Lee, Kim, Hong, and Lee]{lee2020high}
Wonkwang Lee, Donggyun Kim, Seunghoon Hong, and Honglak Lee.
\newblock High-fidelity synthesis with disentangled representation.
\newblock \emph{arXiv preprint arXiv:2001.04296}, 2020.

\bibitem[Li et~al.(2021)Li, Yang, Kreis, Torralba, and Fidler]{segGAN21}
Daiqing Li, Junlin Yang, Karsten Kreis, Antonio Torralba, and Sanja Fidler.
\newblock Semantic segmentation with generative models: Semi-supervised
  learning and strong out-of-domain generalization.
\newblock In \emph{CVPR}, 2021.

\bibitem[Li et~al.(2018)Li, Aittala, Durand, and
  Lehtinen]{li2018differentiable}
Tzu-Mao Li, Miika Aittala, Fr{\'e}do Durand, and Jaakko Lehtinen.
\newblock Differentiable monte carlo ray tracing through edge sampling.
\newblock In \emph{SIGGRAPH Asia 2018 Technical Papers}, pp.\  222. ACM, 2018.

\bibitem[Li et~al.(2020)Li, Liu, Kim, De~Mello, Jampani, Yang, and
  Kautz]{umr2020}
Xueting Li, Sifei Liu, Kihwan Kim, Shalini De~Mello, Varun Jampani, Ming-Hsuan
  Yang, and Jan Kautz.
\newblock Self-supervised single-view 3d reconstruction via semantic
  consistency.
\newblock In \emph{ECCV}, 2020.

\bibitem[Lin et~al.(2019)Lin, Thekumparampil, Fanti, and Oh]{lin2019infogan}
Zinan Lin, Kiran~Koshy Thekumparampil, Giulia Fanti, and Sewoong Oh.
\newblock Infogan-cr: Disentangling generative adversarial networks with
  contrastive regularizers.
\newblock \emph{arXiv preprint arXiv:1906.06034}, 2019.

\bibitem[Liu et~al.(2019{\natexlab{a}})Liu, Tao, Li, Nowrouzezahrai, and
  Jacobson]{liuadvgeo2018}
Hsueh{-}Ti~Derek Liu, Michael Tao, Chun{-}Liang Li, Derek Nowrouzezahrai, and
  Alec Jacobson.
\newblock Beyond pixel norm-balls: Parametric adversaries using an analytically
  differentiable renderer.
\newblock In \emph{ICLR}, 2019{\natexlab{a}}.

\bibitem[Liu et~al.(2019{\natexlab{b}})Liu, Li, Chen, and Li]{softras}
Shichen Liu, Tianye Li, Weikai Chen, and Hao Li.
\newblock Soft rasterizer: A differentiable renderer for image-based 3d
  reasoning.
\newblock \emph{ICCV}, 2019{\natexlab{b}}.

\bibitem[Loper \& Black(2014)Loper and Black]{OpenDR}
Matthew~M. Loper and Michael~J. Black.
\newblock Opendr: An approximate differentiable renderer.
\newblock In David~J. Fleet, Tom{\'{a}}s Pajdla, Bernt Schiele, and Tinne
  Tuytelaars (eds.), \emph{Computer Vision - {ECCV} 2014 - 13th European
  Conference, Zurich, Switzerland, September 6-12, 2014, Proceedings, Part
  {VII}}, volume 8695 of \emph{Lecture Notes in Computer Science}, pp.\
  154--169. Springer, 2014.
\newblock \doi{10.1007/978-3-319-10584-0\_11}.
\newblock URL \url{https://doi.org/10.1007/978-3-319-10584-0\_11}.

\bibitem[Mescheder et~al.(2019)Mescheder, Oechsle, Niemeyer, Nowozin, and
  Geiger]{occnet}
Lars Mescheder, Michael Oechsle, Michael Niemeyer, Sebastian Nowozin, and
  Andreas Geiger.
\newblock Occupancy networks: Learning 3d reconstruction in function space.
\newblock In \emph{Proceedings of the IEEE Conference on Computer Vision and
  Pattern Recognition}, pp.\  4460--4470, 2019.

\bibitem[Park et~al.(2019)Park, Florence, Straub, Newcombe, and
  Lovegrove]{Park_2019_CVPR}
Jeong~Joon Park, Peter Florence, Julian Straub, Richard Newcombe, and Steven
  Lovegrove.
\newblock Deepsdf: Learning continuous signed distance functions for shape
  representation.
\newblock In \emph{The IEEE Conference on Computer Vision and Pattern
  Recognition (CVPR)}, June 2019.

\bibitem[Perarnau et~al.(2016)Perarnau, Van De~Weijer, Raducanu, and
  {\'A}lvarez]{perarnau2016invertible}
Guim Perarnau, Joost Van De~Weijer, Bogdan Raducanu, and Jose~M {\'A}lvarez.
\newblock Invertible conditional gans for image editing.
\newblock \emph{arXiv preprint arXiv:1611.06355}, 2016.

\bibitem[Petersen et~al.(2019)Petersen, Bermano, Deussen, and
  Cohen{-}Or]{pix2vec}
Felix Petersen, Amit~H. Bermano, Oliver Deussen, and Daniel Cohen{-}Or.
\newblock Pix2vex: Image-to-geometry reconstruction using a smooth
  differentiable renderer.
\newblock \emph{CoRR}, abs/1903.11149, 2019.
\newblock URL \url{http://arxiv.org/abs/1903.11149}.

\bibitem[Ravi et~al.(2020)Ravi, Reizenstein, Novotny, Gordon, Lo, Johnson, and
  Gkioxari]{ravi2020pytorch3d}
Nikhila Ravi, Jeremy Reizenstein, David Novotny, Taylor Gordon, Wan-Yen Lo,
  Justin Johnson, and Georgia Gkioxari.
\newblock Pytorch3d.
\newblock \url{https://github.com/facebookresearch/pytorch3d}, 2020.

\bibitem[Reed et~al.(2016)Reed, Akata, Yan, Logeswaran, Schiele, and
  Lee]{reed2016generative}
Scott Reed, Zeynep Akata, Xinchen Yan, Lajanugen Logeswaran, Bernt Schiele, and
  Honglak Lee.
\newblock Generative adversarial text to image synthesis.
\newblock \emph{arXiv preprint arXiv:1605.05396}, 2016.

\bibitem[Shen et~al.(2020)Shen, Yang, Tang, and Zhou]{shen2020interfacegan}
Yujun Shen, Ceyuan Yang, Xiaoou Tang, and Bolei Zhou.
\newblock Interfacegan: Interpreting the disentangled face representation
  learned by gans.
\newblock \emph{arXiv preprint arXiv:2005.09635}, 2020.

\bibitem[Sitzmann et~al.(2019)Sitzmann, Zollh{\"o}fer, and
  Wetzstein]{sitzmann2019srns}
Vincent Sitzmann, Michael Zollh{\"o}fer, and Gordon Wetzstein.
\newblock Scene representation networks: Continuous 3d-structure-aware neural
  scene representations.
\newblock In \emph{Advances in Neural Information Processing Systems}, 2019.

\bibitem[Tewari et~al.(2020)Tewari, Elgharib, Bharaj, Bernard, Seidel,
  P{\'e}rez, Z{\"o}llhofer, and Theobalt]{tewari2020stylerig}
Ayush Tewari, Mohamed Elgharib, Gaurav Bharaj, Florian Bernard, Hans-Peter
  Seidel, Patrick P{\'e}rez, Michael Z{\"o}llhofer, and Christian Theobalt.
\newblock Stylerig: Rigging stylegan for 3d control over portrait images, cvpr
  2020.
\newblock In \emph{{IEEE} Conference on Computer Vision and Pattern Recognition
  (CVPR)}. {IEEE}, june 2020.

\bibitem[Torralba et~al.(2010)Torralba, Russell, and Yuen]{labelme}
Antonio Torralba, Bryan~C Russell, and Jenny Yuen.
\newblock Labelme: Online image annotation and applications.
\newblock \emph{Proceedings of the IEEE}, 98\penalty0 (8):\penalty0
  1467–1484, 2010.
\newblock \doi{10.1109/jproc.2010.2050290}.

\bibitem[Valentin et~al.(2019)Valentin, Keskin, Pidlypenskyi, Makadia, Sud, and
  Bouaziz]{TensorflowGraphicsIO2019}
Julien Valentin, Cem Keskin, Pavel Pidlypenskyi, Ameesh Makadia, Avneesh Sud,
  and Sofien Bouaziz.
\newblock Tensorflow graphics: Computer graphics meets deep learning.
\newblock 2019.

\bibitem[{Van Horn} et~al.(2015){Van Horn}, {Branson}, {Farrell}, {Haber},
  {Barry}, {Ipeirotis}, {Perona}, and {Belongie}]{7298658}
G.~{Van Horn}, S.~{Branson}, R.~{Farrell}, S.~{Haber}, J.~{Barry},
  P.~{Ipeirotis}, P.~{Perona}, and S.~{Belongie}.
\newblock Building a bird recognition app and large scale dataset with citizen
  scientists: The fine print in fine-grained dataset collection.
\newblock In \emph{2015 IEEE Conference on Computer Vision and Pattern
  Recognition (CVPR)}, pp.\  595--604, 2015.

\bibitem[Wang et~al.(2018)Wang, Zhang, Li, Fu, Liu, and Jiang]{pixel2mesh}
Nanyang Wang, Yinda Zhang, Zhuwen Li, Yanwei Fu, Wei Liu, and Yu-Gang Jiang.
\newblock Pixel2mesh: Generating 3d mesh models from single rgb images.
\newblock In \emph{ECCV}, 2018.

\bibitem[Wang et~al.(2019)Wang, Qiangeng, Ceylan, Mech, and Neumann]{DISN}
Weiyue Wang, Xu~Qiangeng, Duygu Ceylan, Radomir Mech, and Ulrich Neumann.
\newblock Disn: Deep implicit surface network for high-quality single-view 3d
  reconstruction.
\newblock \emph{arXiv preprint arXiv:1905.10711}, 2019.

\bibitem[Welinder et~al.(2010)Welinder, Branson, Mita, Wah, Schroff, Belongie,
  and Perona]{cub}
P.~Welinder, S.~Branson, T.~Mita, C.~Wah, F.~Schroff, S.~Belongie, and
  P.~Perona.
\newblock {Caltech-UCSD Birds 200}.
\newblock Technical Report CNS-TR-2010-001, California Institute of Technology,
  2010.

\bibitem[Wu et~al.(2020)Wu, Rupprecht, and Vedaldi]{Wu_2020_CVPR}
Shangzhe Wu, Christian Rupprecht, and Andrea Vedaldi.
\newblock Unsupervised learning of probably symmetric deformable 3d objects
  from images in the wild.
\newblock In \emph{Proceedings of the IEEE/CVF Conference on Computer Vision
  and Pattern Recognition (CVPR)}, June 2020.

\bibitem[Xiang et~al.(2014)Xiang, Mottaghi, and Savarese]{xiang_wacv14}
Yu~Xiang, Roozbeh Mottaghi, and Silvio Savarese.
\newblock Beyond pascal: A benchmark for 3d object detection in the wild.
\newblock In \emph{IEEE Winter Conference on Applications of Computer Vision
  (WACV)}, 2014.

\bibitem[Yao et~al.(2018)Yao, Hsu, Zhu, Wu, Torralba, Freeman, and
  Tenenbaum]{yao20183d}
Shunyu Yao, Tzu~Ming Hsu, Jun-Yan Zhu, Jiajun Wu, Antonio Torralba, Bill
  Freeman, and Josh Tenenbaum.
\newblock 3d-aware scene manipulation via inverse graphics.
\newblock In \emph{Advances in neural information processing systems}, pp.\
  1887--1898, 2018.

\bibitem[Zhang et~al.(2021)Zhang, Ling, Gao, Yin, Lafleche, Barriuso, Torralba,
  and Fidler]{zhang21}
Yuxuan Zhang, Huan Ling, Jun Gao, Kangxue Yin, Jean-Francois Lafleche, Adela
  Barriuso, Antonio Torralba, and Sanja Fidler.
\newblock Datasetgan: Efficient labeled data factory with minimal human effort.
\newblock In \emph{CVPR}, 2021.

\bibitem[Zhu et~al.(2017)Zhu, Park, Isola, and Efros]{cyclegan}
Jun-Yan Zhu, Taesung Park, Phillip Isola, and Alexei~A Efros.
\newblock Unpaired image-to-image translation using cycle-consistent
  adversarial networks.
\newblock In \emph{Proceedings of the IEEE international conference on computer
  vision}, pp.\  2223--2232, 2017.

\bibitem[Zhu et~al.(2018)Zhu, Zhang, Zhang, Wu, Torralba, Tenenbaum, and
  Freeman]{zhu2018visual}
Jun-Yan Zhu, Zhoutong Zhang, Chengkai Zhang, Jiajun Wu, Antonio Torralba, Josh
  Tenenbaum, and Bill Freeman.
\newblock Visual object networks: Image generation with disentangled 3d
  representations.
\newblock In \emph{Advances in neural information processing systems}, pp.\
  118--129, 2018.

\end{thebibliography}
\bibliographystyle{iclr2021_conference}
\end{small}
\clearpage


\setcounter{section}{0}
\renewcommand{\thesection}{\Alph{section}}
\renewcommand\thefigure{\Alph{figure}}   
\renewcommand\thetable{\Alph{table}}     
\setcounter{figure}{0} 
\setcounter{table}{0} 
\section*{\LARGE Appendix}

\section{Overview}
In the Appendix, we first show  feature visualization of StyleGAN layers in Sec.~\ref{sec:stylegan_layer}. We then provide a detailed explanation of our StyleGAN dataset creation in Sec.~\ref{sec:dataset}, including examples of the generated images and selected viewpoints. Next, we do a systematic analysis of our camera initialization method  in Sec.~\ref{sec:camera_init}. Finally, we show additional results on the 3D inverse graphics task in Sec.~\ref{sec:3d_inference}, additional details of the user study  in Sec.~\ref{sec:user_study}, futher examples of StyleGAN disentanglement in Sec.~\ref{sec:stylegan_disentangle}, with ablation studies and a discussion of limitations in Sec.~\ref{sec:ablation} and  Sec.~\ref{sec:supp_limit}, respectively.

\section{StyleGAN Layers Visualization}
\label{sec:stylegan_layer}

The official StyleGAN code repository provides models of different object categories at different resolutions. Here we take the $512 \times 384$ car model as the example. This model contains 16 layers, where every two consecutive layers form a block. Each block has a different number of channels. In the last block, the model produces a 32-channel feature map at a $512 \times 384$ resolution. Finally, a learned RGB transformation function is applied to convert the feature map into an RGB image.

We visualize the feature map for each block via the learned RGB transformation function. Specifically, for the feature map in each block with the size of $h\times w \times c$, we first sum along the feature dimension, forming a $h\times w \times 1$ tensor. We then repeat the feature 32 times and generate a $h\times w \times 32$ new feature map. This allows us to keep the information of all the channels and directly apply the RGB transformation function in the last block to convert it to the RGB image. 

\begin{figure*}[htb!]
	{
		\vspace*{-10pt}
		\begin{center}
			\setlength{\tabcolsep}{1pt}
			\setlength{\fboxrule}{0pt}
			\hspace*{-0.3cm}
			\begin{tabular}{c}
				\begin{tabular}{ccccc}
					Block 1 & Block 2 & Block 3 & Block 4 &  
					\\
					\includegraphics[width=.19\textwidth,height=.15\textwidth]{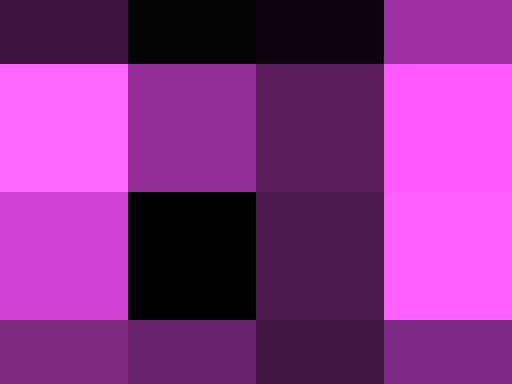}&
					\includegraphics[width=.19\textwidth,height=.15\textwidth]{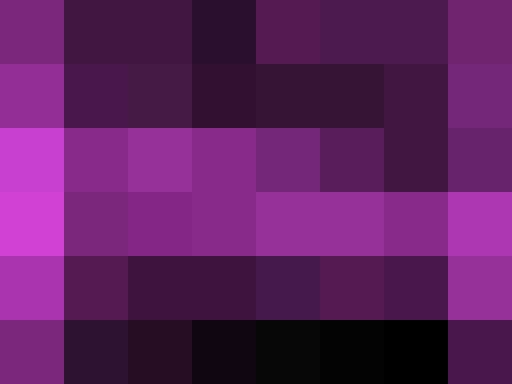}&
					\includegraphics[width=.19\textwidth,height=.15\textwidth]{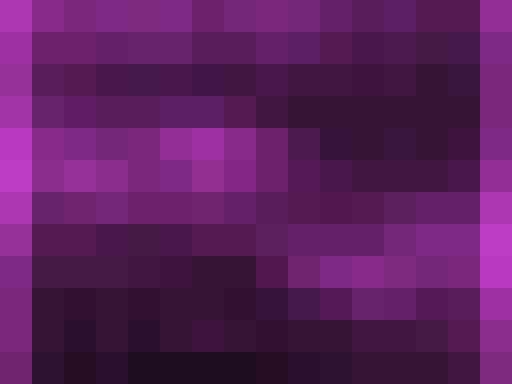}&
					\includegraphics[width=.19\textwidth,height=.15\textwidth]{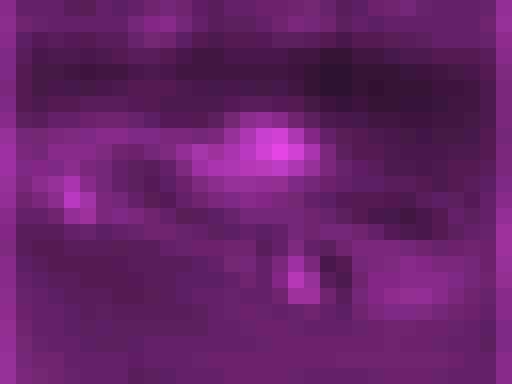}&
					\\
					Block 5 & Block 6 & Block 7 & Block 8 & Generated Image
					\\
					\includegraphics[width=.19\textwidth,height=.15\textwidth]{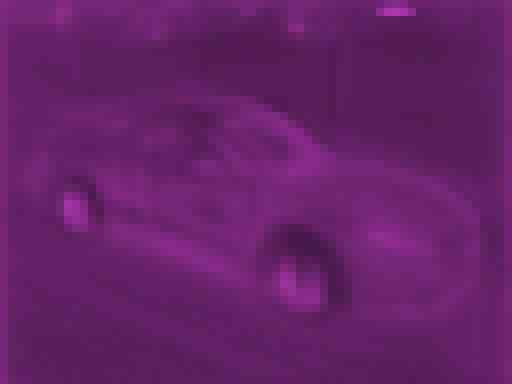}&
					\includegraphics[width=.19\textwidth,height=.15\textwidth]{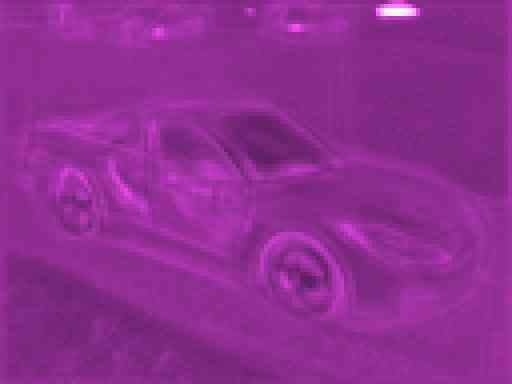}&
					\includegraphics[width=.19\textwidth,height=.15\textwidth]{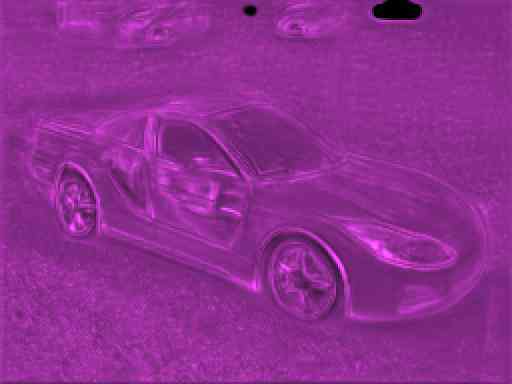}&
					\includegraphics[width=.19\textwidth,height=.15\textwidth]{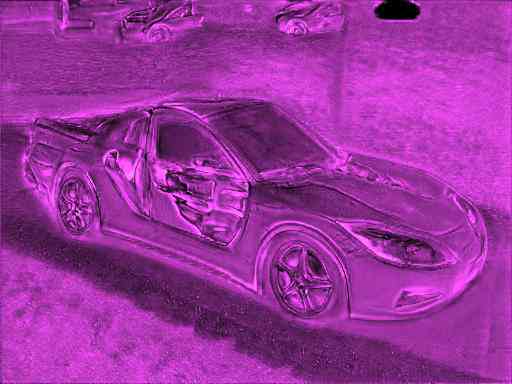}&
					\includegraphics[width=.19\textwidth,height=.15\textwidth]{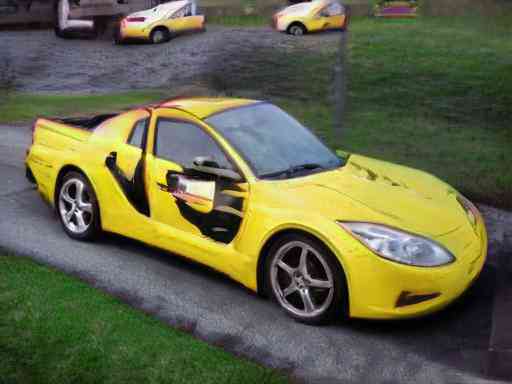}
					
					\hspace*{0pt}
				\end{tabular}
			\end{tabular}
		\end{center}
		\vspace*{-0.5cm}
	}
	\caption{\label{fig:layer} \textbf{\footnotesize Layer Visualization for Each Block}: Notice that the car contour starts to appear in blocks 4 and higher. This supports some of our findings that the early blocks control viewpoint (and other global properties), while shape, texture and background are controlled in the higher layers.}
\end{figure*}

As shown in Fig~\ref{fig:layer}, we find that blocks 1 and 2 do not exhibit interpretable structure while the car shape starts to appear in blocks 3-5. We observe that there is a rough car contour in block 4 which further becomes clear in block 5. From blocks 6 to 8, the car's shape becomes increasingly finer and background scene also  appears. This supports some of our findings, i.e., the viewpoint is controlled in block 1 and 2 (first 4 layers) while shape, texture, and background exist in the last 12 layers.

\section{Our ``StyleGAN" Dataset}
\label{sec:dataset}

\begin{figure*}[t]
	{
		\vspace*{0pt}
		\begin{center}
			\setlength{\tabcolsep}{1pt}
			\setlength{\fboxrule}{0pt}
			\hspace*{0pt}
			\begin{tabular}{c}
				\begin{tabular}{cccccccc}
					\includegraphics[width=.12\textwidth,height=.09\textwidth]{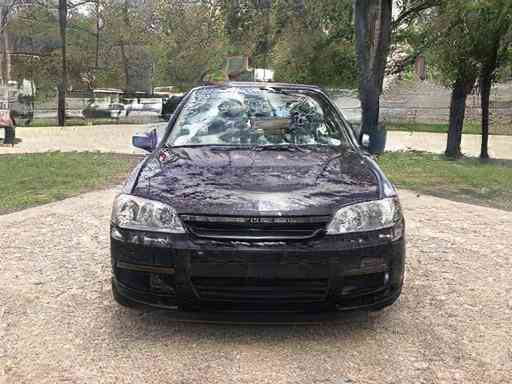}&
					\includegraphics[width=.12\textwidth,height=.09\textwidth]{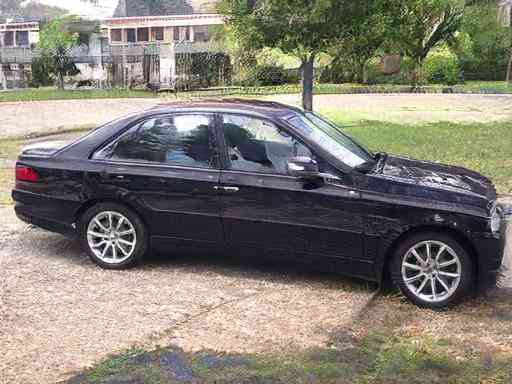}&
					\includegraphics[width=.12\textwidth,height=.09\textwidth]{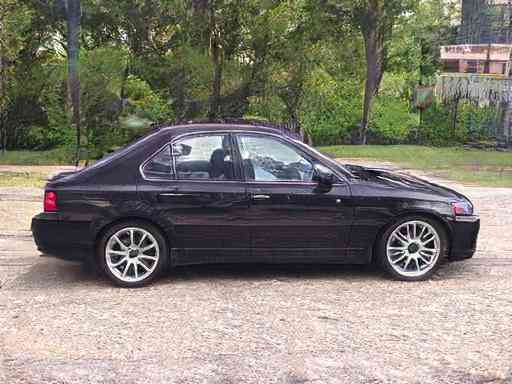}&
					\includegraphics[width=.12\textwidth,height=.09\textwidth]{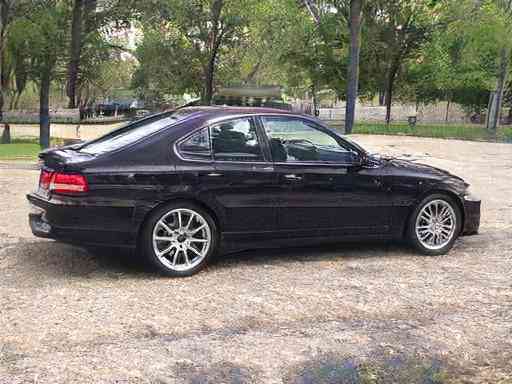}&
					\includegraphics[width=.12\textwidth,height=.09\textwidth]{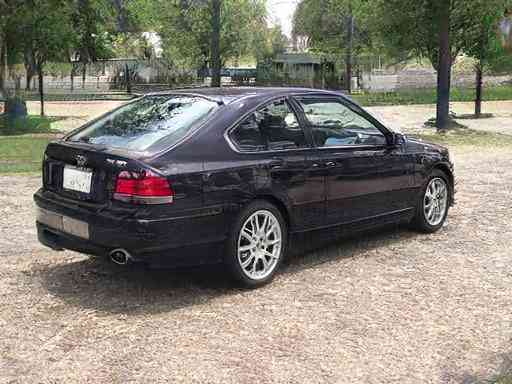}&
					\includegraphics[width=.12\textwidth,height=.09\textwidth]{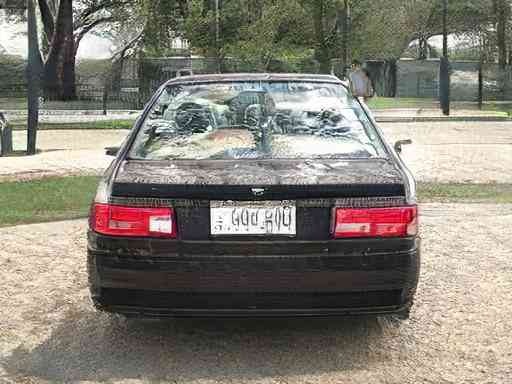}&
					\includegraphics[width=.12\textwidth,height=.09\textwidth]{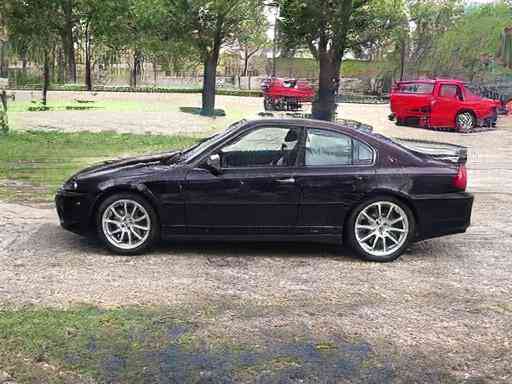}&
					\includegraphics[width=.12\textwidth,height=.09\textwidth]{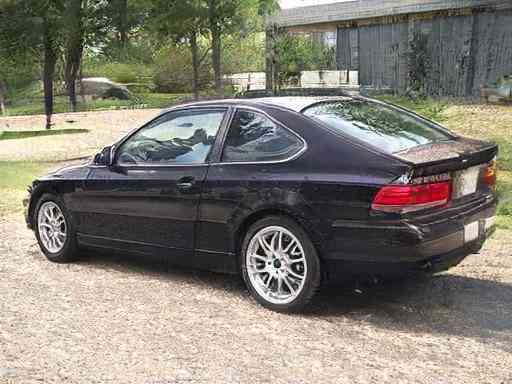}
					\\
					\includegraphics[width=.12\textwidth,height=.09\textwidth]{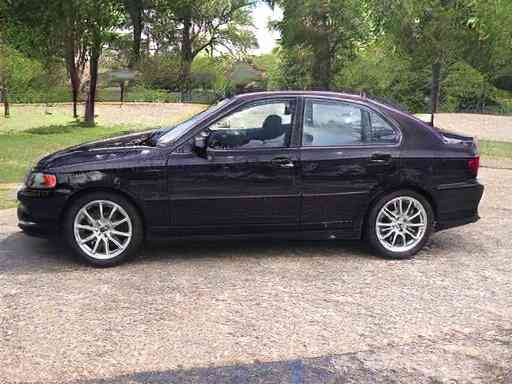}&
					\includegraphics[width=.12\textwidth,height=.09\textwidth]{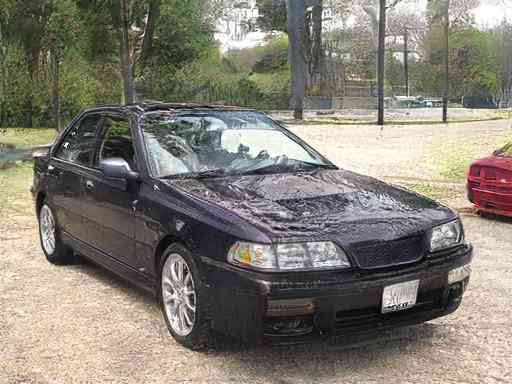}&
					\includegraphics[width=.12\textwidth,height=.09\textwidth]{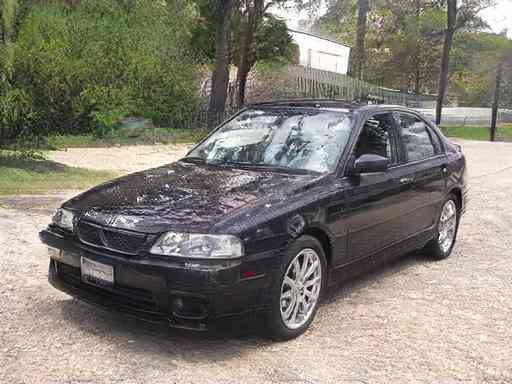}&
					\includegraphics[width=.12\textwidth,height=.09\textwidth]{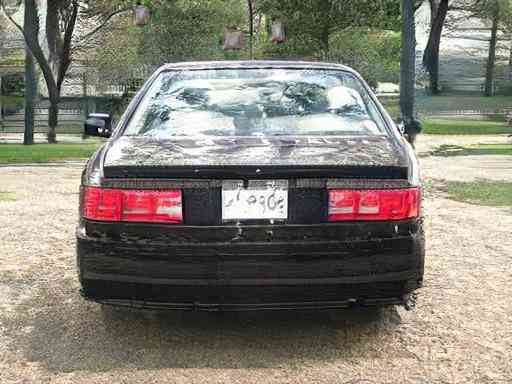}&
					\includegraphics[width=.12\textwidth,height=.09\textwidth]{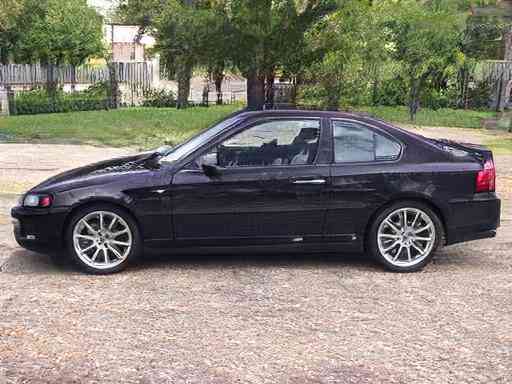}&
					\includegraphics[width=.12\textwidth,height=.09\textwidth]{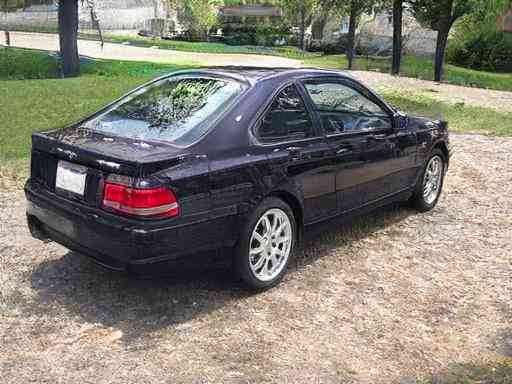}&
					\includegraphics[width=.12\textwidth,height=.09\textwidth]{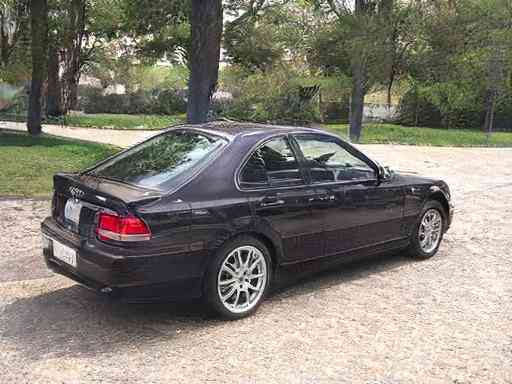}&
					\includegraphics[width=.12\textwidth,height=.09\textwidth]{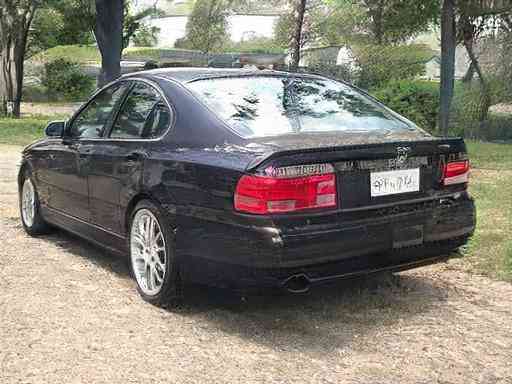}
					\\
					\includegraphics[width=.12\textwidth,height=.09\textwidth]{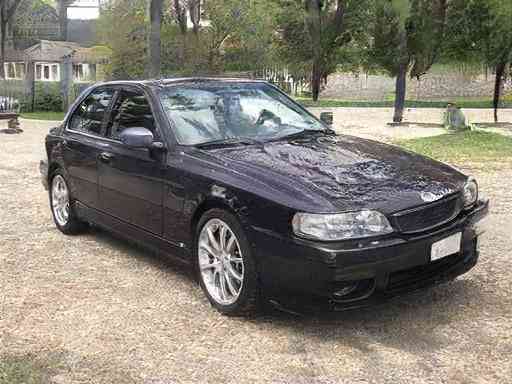}&
					\includegraphics[width=.12\textwidth,height=.09\textwidth]{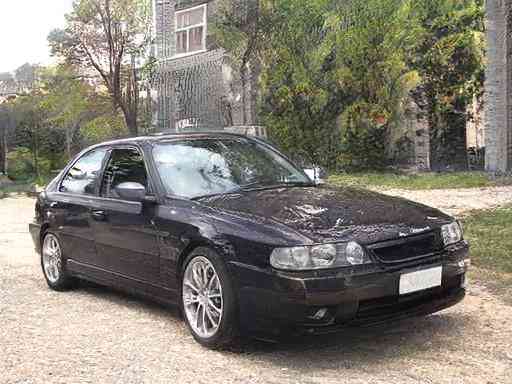}&
					\includegraphics[width=.12\textwidth,height=.09\textwidth]{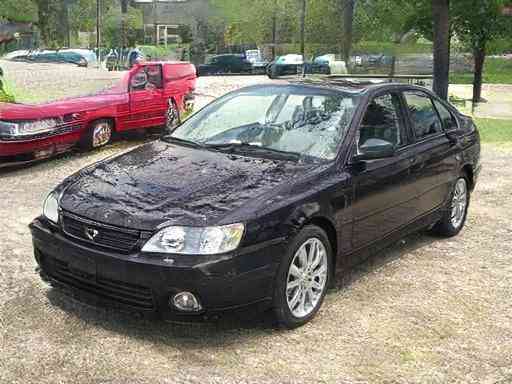}&
					\includegraphics[width=.12\textwidth,height=.09\textwidth]{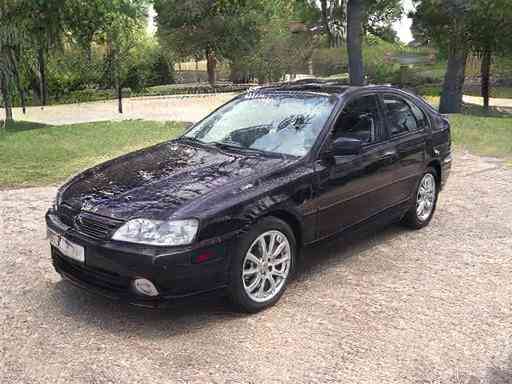}&
					\includegraphics[width=.12\textwidth,height=.09\textwidth]{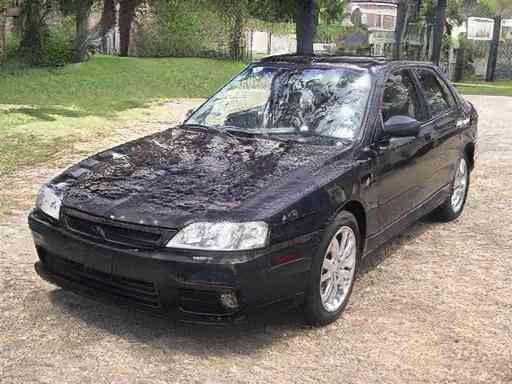}&
					\includegraphics[width=.12\textwidth,height=.09\textwidth]{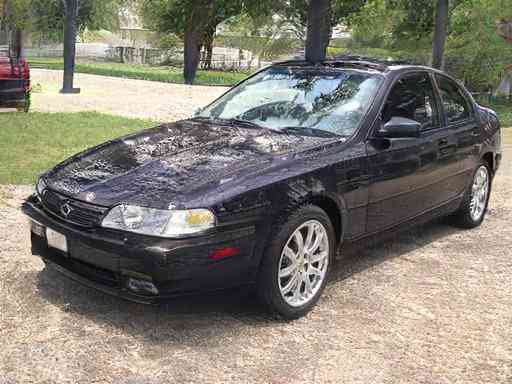}&
					\includegraphics[width=.12\textwidth,height=.09\textwidth]{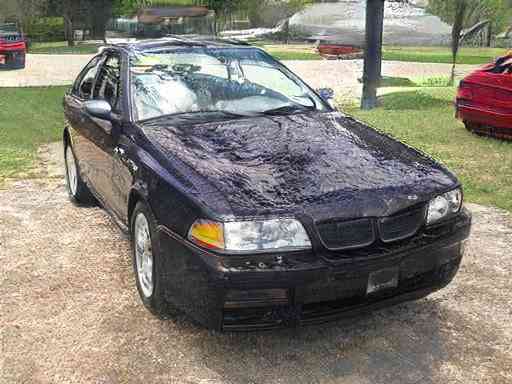}&
					\includegraphics[width=.12\textwidth,height=.09\textwidth]{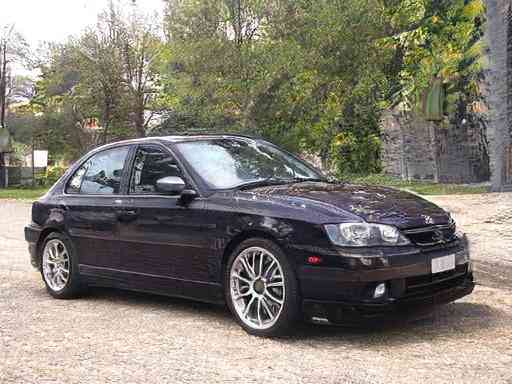}
					\\
					\includegraphics[width=.12\textwidth,height=.09\textwidth]{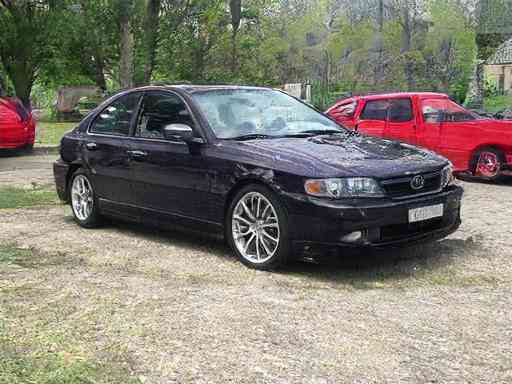}&
					\includegraphics[width=.12\textwidth,height=.09\textwidth]{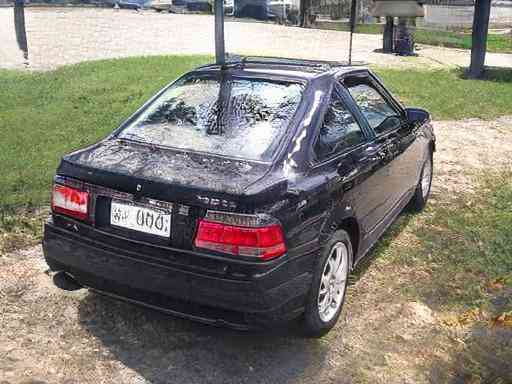}&
					\includegraphics[width=.12\textwidth,height=.09\textwidth]{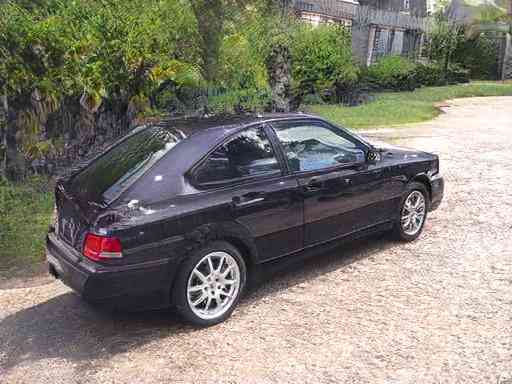}&
					\includegraphics[width=.12\textwidth,height=.09\textwidth]{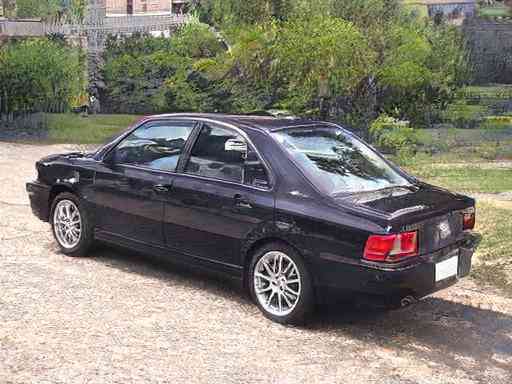}&
					\includegraphics[width=.12\textwidth,height=.09\textwidth]{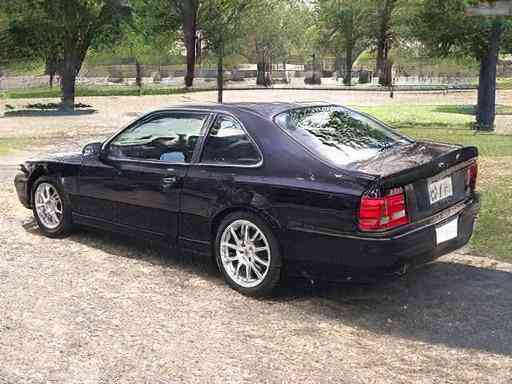}&
					\includegraphics[width=.12\textwidth,height=.09\textwidth]{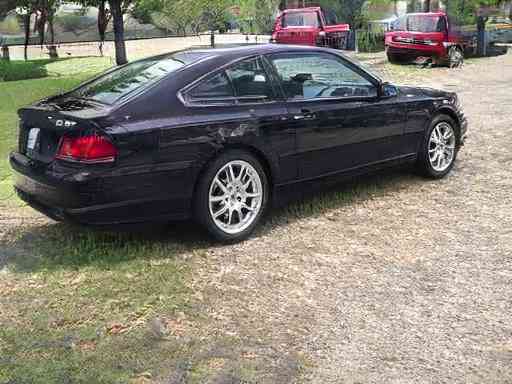}&
					\includegraphics[width=.12\textwidth,height=.09\textwidth]{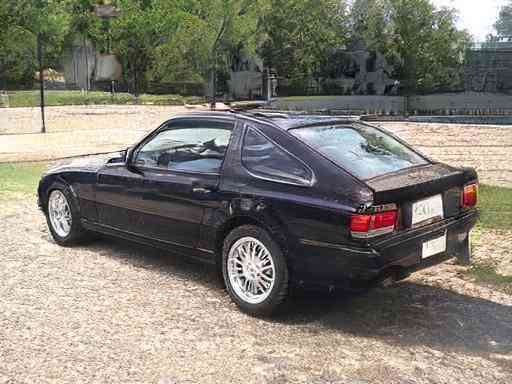}&
					\includegraphics[width=.12\textwidth,height=.09\textwidth]{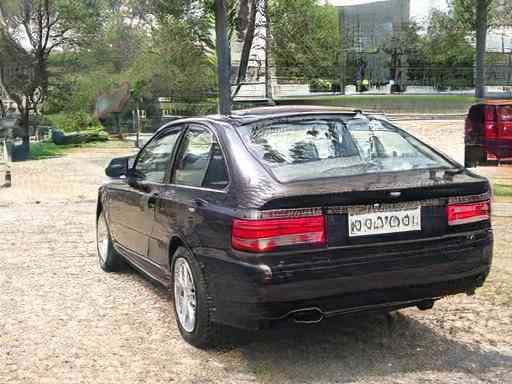}
					\\
					\includegraphics[width=.12\textwidth,height=.09\textwidth]{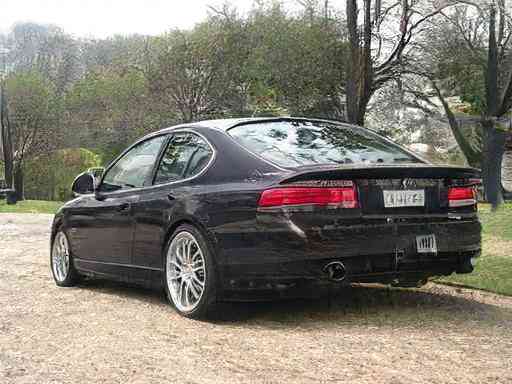}&
					\includegraphics[width=.12\textwidth,height=.09\textwidth]{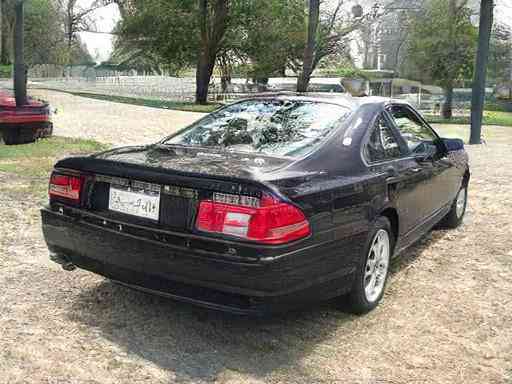}&
					\includegraphics[width=.12\textwidth,height=.09\textwidth]{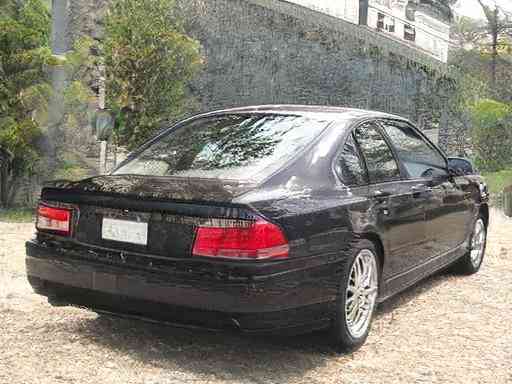}&
					\includegraphics[width=.12\textwidth,height=.09\textwidth]{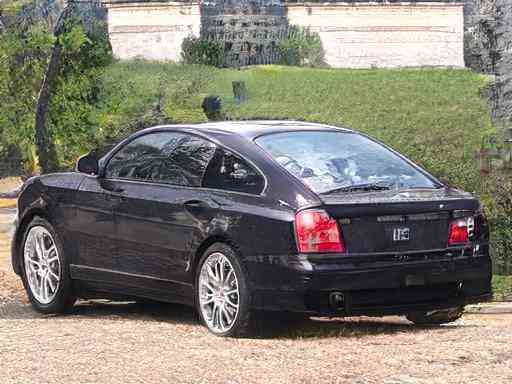}&
					\includegraphics[width=.12\textwidth,height=.09\textwidth]{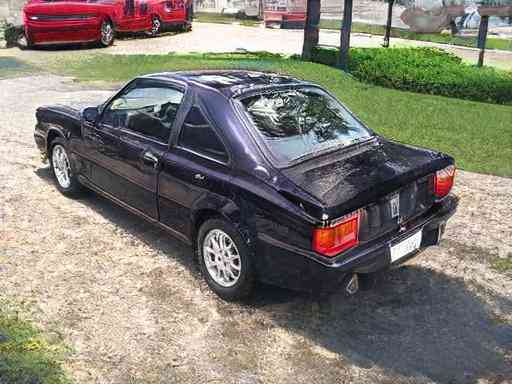}&
					\includegraphics[width=.12\textwidth,height=.09\textwidth]{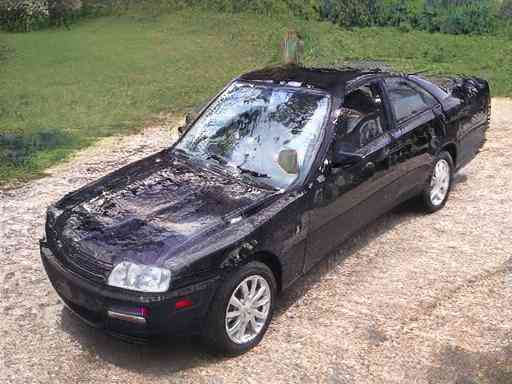}&
					\includegraphics[width=.12\textwidth,height=.09\textwidth]{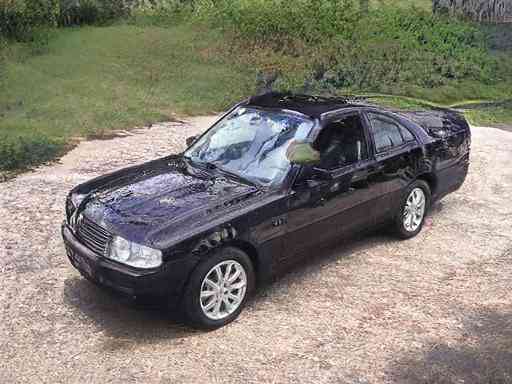}&
					\\	
					\multicolumn{8}{c}{\large Car Viewpoints} \\	
					& & & & & & &  \\
					\includegraphics[width=.12\textwidth,height=.09\textwidth]{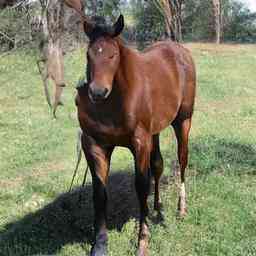}&
					\includegraphics[width=.12\textwidth,height=.09\textwidth]{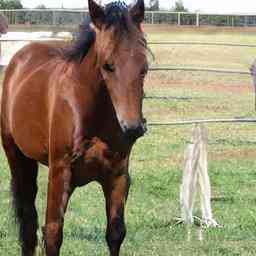}&
					\includegraphics[width=.12\textwidth,height=.09\textwidth]{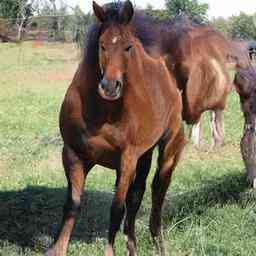}&
					\includegraphics[width=.12\textwidth,height=.09\textwidth]{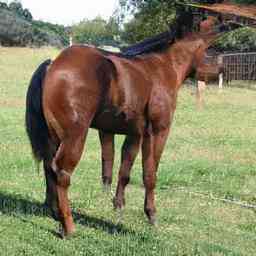}&
					\includegraphics[width=.12\textwidth,height=.09\textwidth]{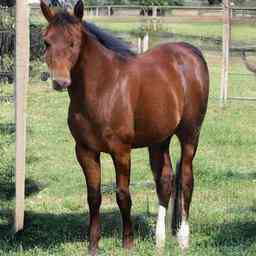}&
					\includegraphics[width=.12\textwidth,height=.09\textwidth]{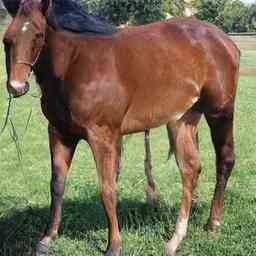}&
					\includegraphics[width=.12\textwidth,height=.09\textwidth]{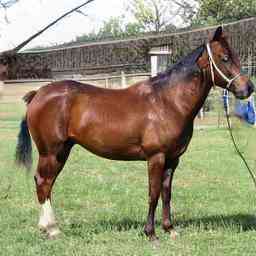}&
					\includegraphics[width=.12\textwidth,height=.09\textwidth]{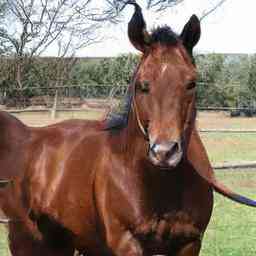}
					\\		
					
					\includegraphics[width=.12\textwidth,height=.09\textwidth]{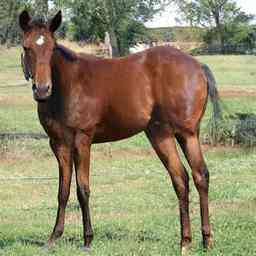}&
					\includegraphics[width=.12\textwidth,height=.09\textwidth]{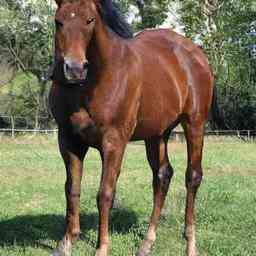}&
					\includegraphics[width=.12\textwidth,height=.09\textwidth]{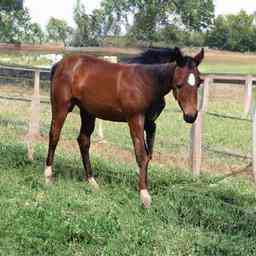}&
					\includegraphics[width=.12\textwidth,height=.09\textwidth]{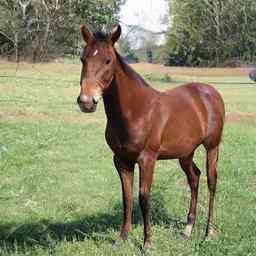}&
					\includegraphics[width=.12\textwidth,height=.09\textwidth]{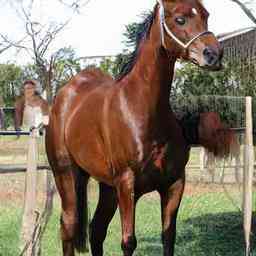}&
					\includegraphics[width=.12\textwidth,height=.09\textwidth]{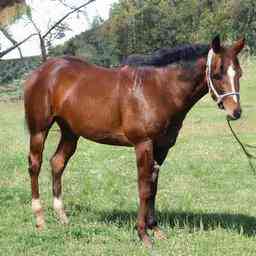}&
					\includegraphics[width=.12\textwidth,height=.09\textwidth]{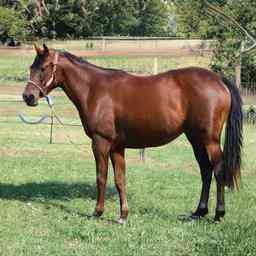}&
					\includegraphics[width=.12\textwidth,height=.09\textwidth]{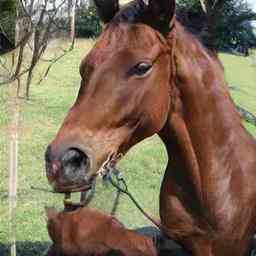}
					\\							
					\includegraphics[width=.12\textwidth,height=.09\textwidth]{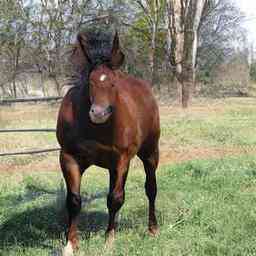}&
					\includegraphics[width=.12\textwidth,height=.09\textwidth]{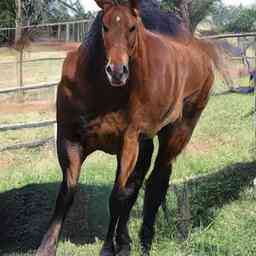}&
					\includegraphics[width=.12\textwidth,height=.09\textwidth]{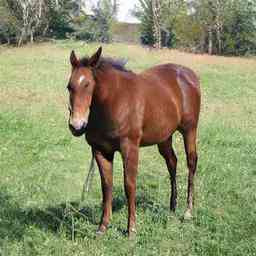}&
					\includegraphics[width=.12\textwidth,height=.09\textwidth]{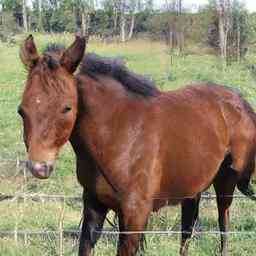}&
					\includegraphics[width=.12\textwidth,height=.09\textwidth]{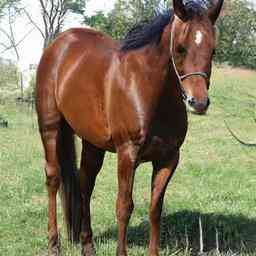}&
					\includegraphics[width=.12\textwidth,height=.09\textwidth]{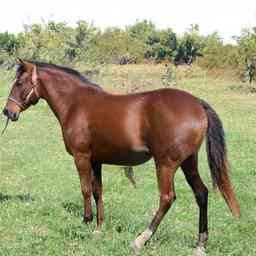}
					\\	
					\multicolumn{8}{c}{\large Horse Viewpoints} \\
					& & & & & & & \\				
					\includegraphics[width=.12\textwidth,height=.09\textwidth]{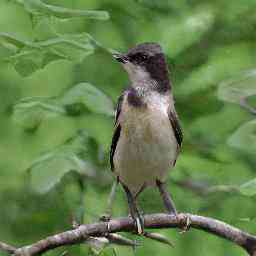}&
					\includegraphics[width=.12\textwidth,height=.09\textwidth]{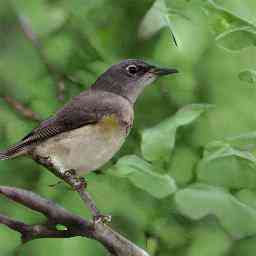}&
					\includegraphics[width=.12\textwidth,height=.09\textwidth]{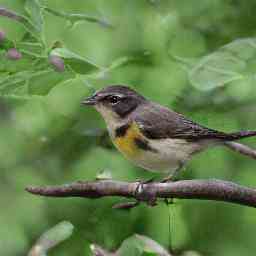}&
					\includegraphics[width=.12\textwidth,height=.09\textwidth]{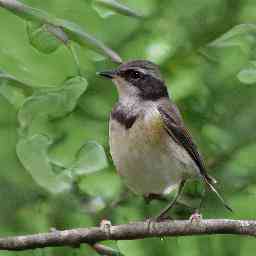}&
					\includegraphics[width=.12\textwidth,height=.09\textwidth]{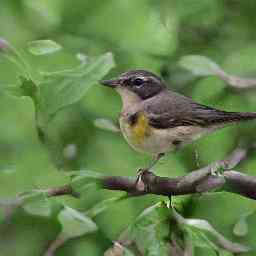}&
					\includegraphics[width=.12\textwidth,height=.09\textwidth]{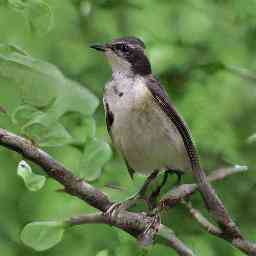}&
					\includegraphics[width=.12\textwidth,height=.09\textwidth]{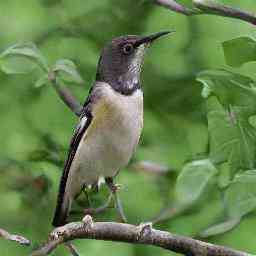}&
					\includegraphics[width=.12\textwidth,height=.09\textwidth]{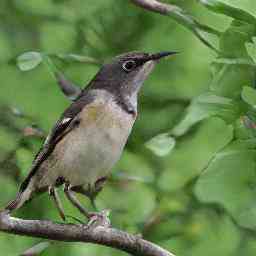}
					\\				
					\multicolumn{8}{c}{\large Bird Viewpoints} \\			

					\hspace*{0pt}
				\end{tabular}
			\end{tabular}
		\end{center}
		\vspace*{-0.4cm}
	}
	\caption{\label{fig:trainview:car} \textbf{\footnotesize All Viewpoints:} \footnotesize We show an example of a car, bird and a horse synthesized in all of our chosen viewpoints. While shape and texture are not perfectly consistent across views, they are sufficiently accurate to enable training accurate inverse graphics networks in our downstream tasks. Horses and birds are especially challenging due to articulation. One can notice small changes in articulation across viewpoints. Dealing with articulated objects is subject to future work. }
\end{figure*}

We visualize all of our selected viewpoints in our dataset in Fig.~\ref{fig:trainview:car}. 
Our car training dataset contains 39 viewpoints. For the horse and bird datasets, we choose 22 and 8 views, respectively. We find that these views are sufficient to learn accurate 3D inverse graphics networks. We could not find views that would depict the object from a higher up camera, i.e., a viewpoint from which the roof of the car or the back of the horse would be more clearly visible. This is mainly due to the original dataset on which StyleGAN was trained on, which lacked  such  views. This leads to challenges in training inverse graphics networks to accurately predict the top of the objects. 

Notice the high consistency of both the car shape and texture as well as the background scene across the different viewpoints. Note that for articulated objects such as the horse and bird classes, StyleGAN does not perfectly preserve object articulation in different viewpoints, which leads to challenges in training high accuracy models using multi-view consistency loss. We leave further investigation of articulated objects to future work. 

\begin{figure*}[t]
	{
		\vspace*{0pt}
		\begin{center}
			\setlength{\tabcolsep}{1pt}
			\setlength{\fboxrule}{0pt}
			\hspace*{0pt}
			\begin{tabular}{c}
				\begin{tabular}{ccccccc}
					\includegraphics[width=.14\textwidth,trim=0 0 0 30,clip]{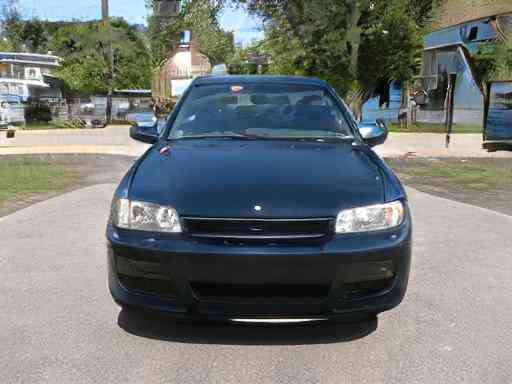}&
					\includegraphics[width=.14\textwidth,trim=0 0 0 30,clip]{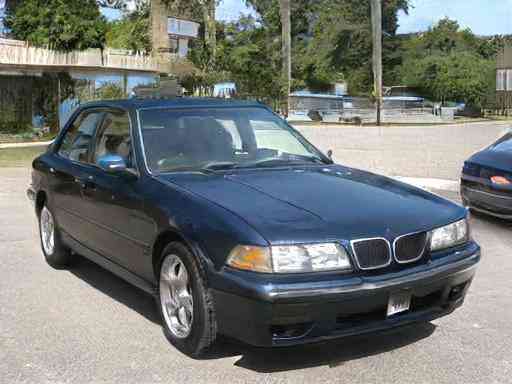}&
					\includegraphics[width=.14\textwidth,trim=0 0 0 30,clip]{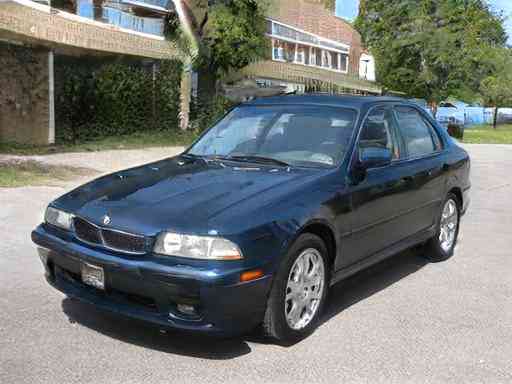}&
					\includegraphics[width=.14\textwidth,trim=0 0 0 30,clip]{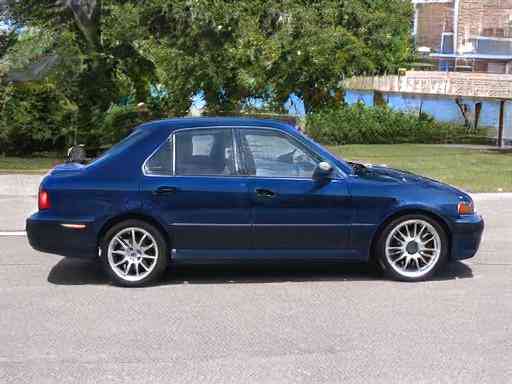}&
					\includegraphics[width=.14\textwidth,trim=0 0 0 30,clip]{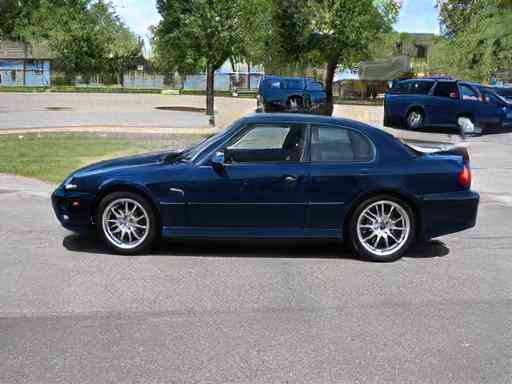}&
					\includegraphics[width=.14\textwidth,trim=0 0 0 30,clip]{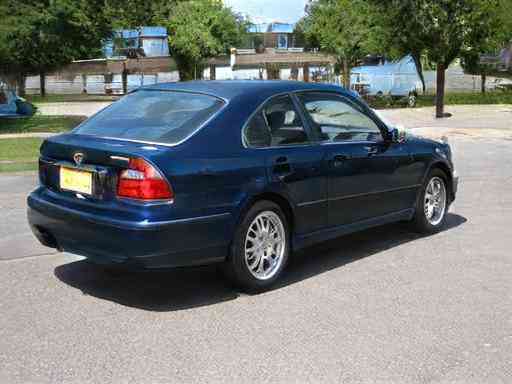}&
					\includegraphics[width=.14\textwidth,trim=0 0 0 30,clip]{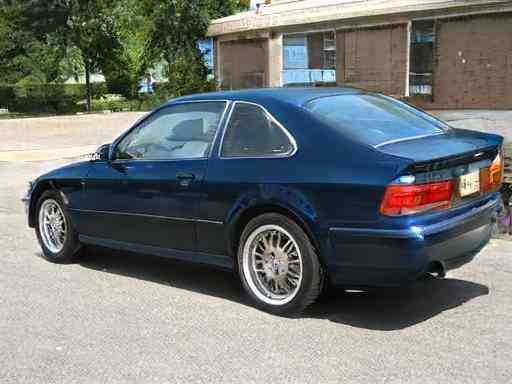}					
					\\
					\includegraphics[width=.14\textwidth,trim=0 0 0 30,clip]{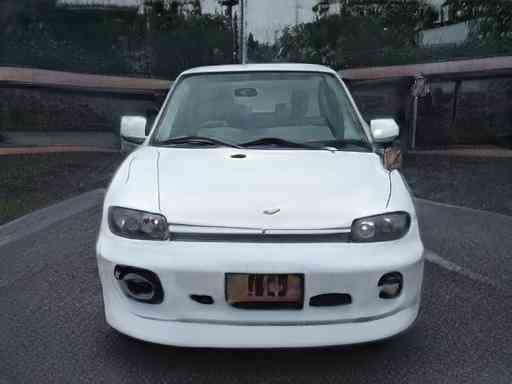}&
					\includegraphics[width=.14\textwidth,trim=0 0 0 30,clip]{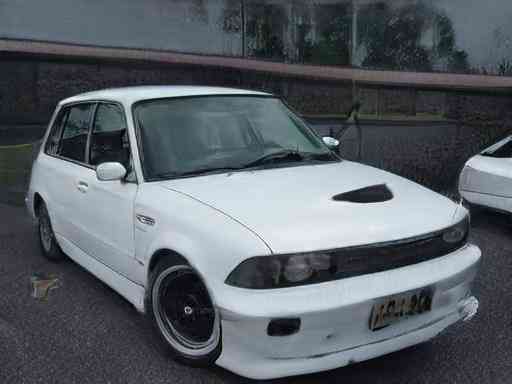}&
					\includegraphics[width=.14\textwidth,trim=0 0 0 30,clip]{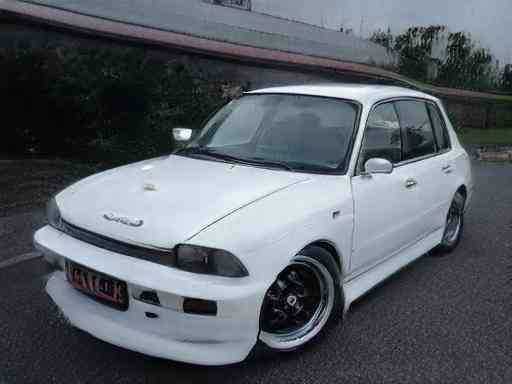}&
					\includegraphics[width=.14\textwidth,trim=0 0 0 30,clip]{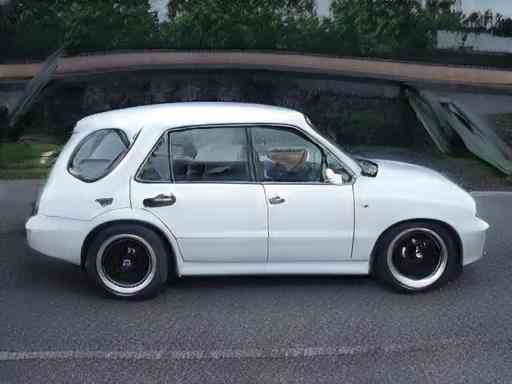}&
					\includegraphics[width=.14\textwidth,trim=0 0 0 30,clip]{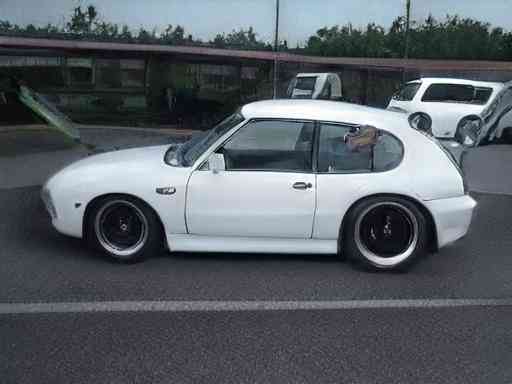}&
					\includegraphics[width=.14\textwidth,trim=0 0 0 30,clip]{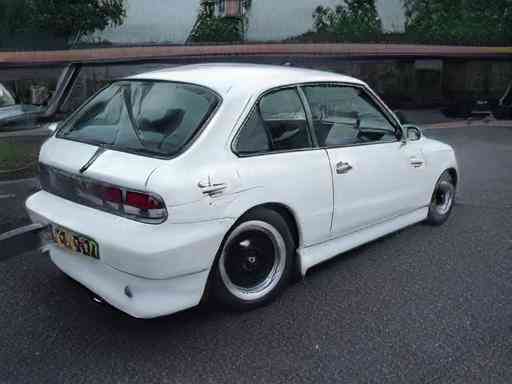}&
					\includegraphics[width=.14\textwidth,trim=0 0 0 30,clip]{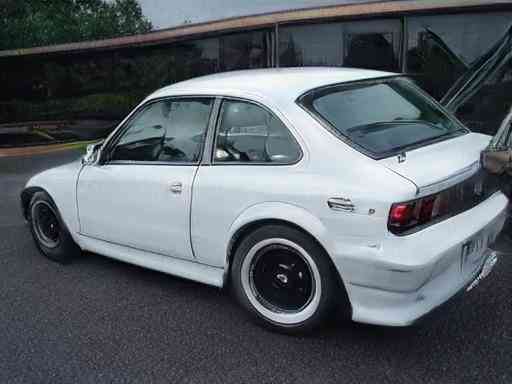}					
					\\

					\includegraphics[width=.14\textwidth,trim=0 0 0 30,clip]{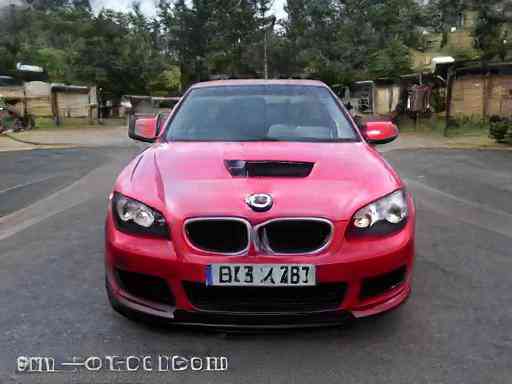}&
					\includegraphics[width=.14\textwidth,trim=0 0 0 30,clip]{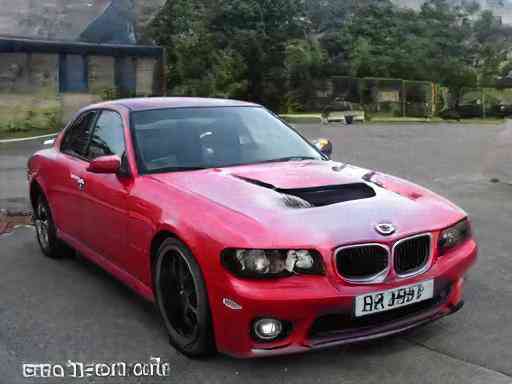}&
					\includegraphics[width=.14\textwidth,trim=0 0 0 30,clip]{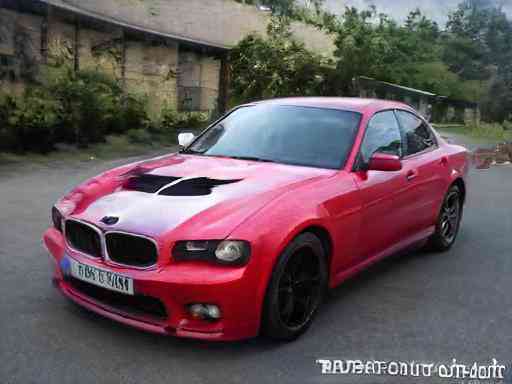}&
					\includegraphics[width=.14\textwidth,trim=0 0 0 30,clip]{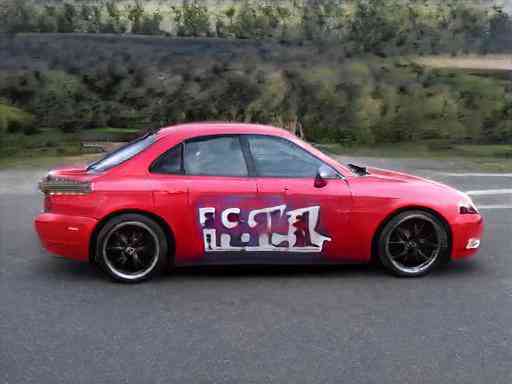}&
					\includegraphics[width=.14\textwidth,trim=0 0 0 30,clip]{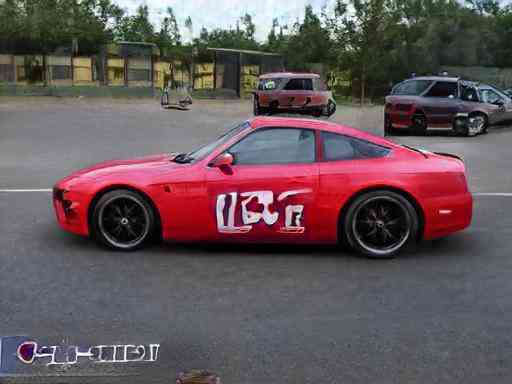}&
					\includegraphics[width=.14\textwidth,trim=0 0 0 30,clip]{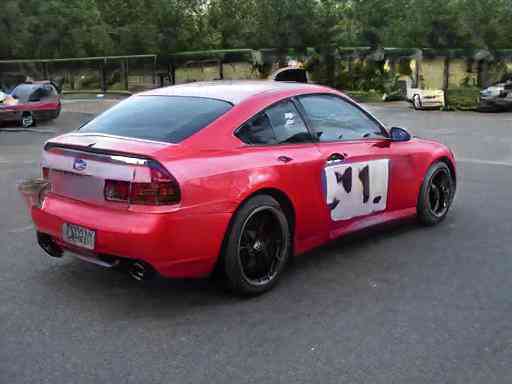}&
					\includegraphics[width=.14\textwidth,trim=0 0 0 30,clip]{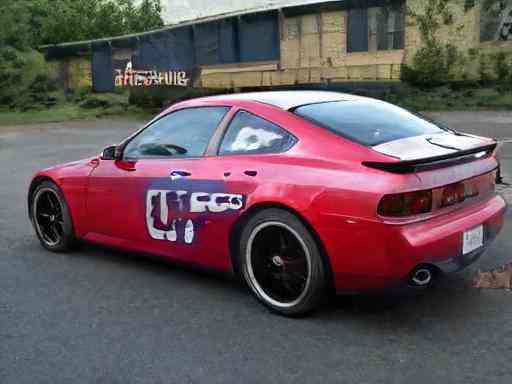}					
					\\
					
					\includegraphics[width=.14\textwidth,trim=0 0 0 30,clip]{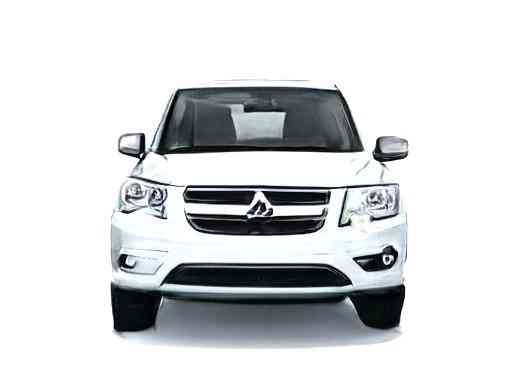}&
					\includegraphics[width=.14\textwidth,trim=0 0 0 30,clip]{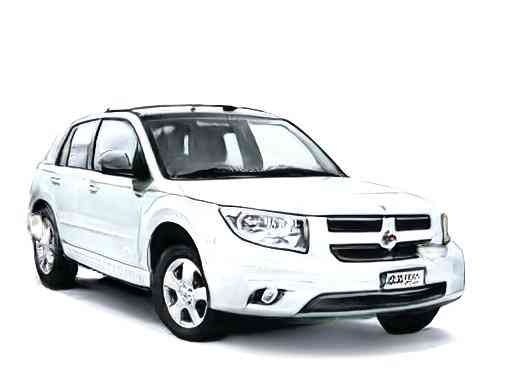}&
					\includegraphics[width=.14\textwidth,trim=0 0 0 30,clip]{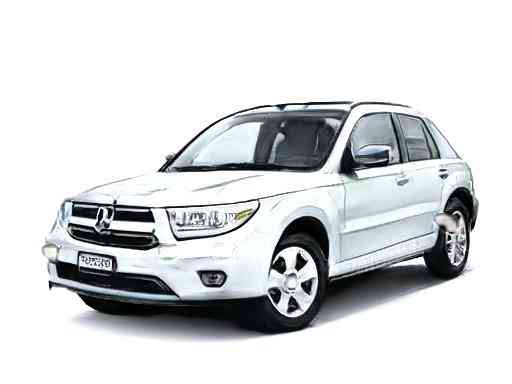}&
					\includegraphics[width=.14\textwidth,trim=0 0 0 30,clip]{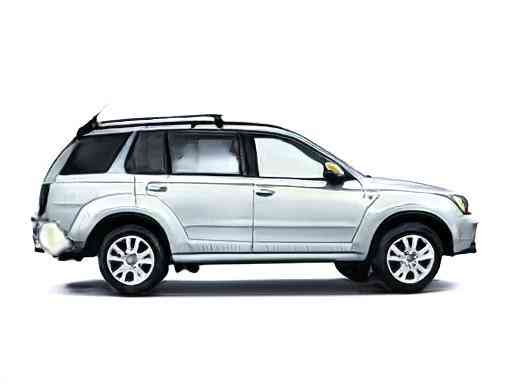}&
					\includegraphics[width=.14\textwidth,trim=0 0 0 30,clip]{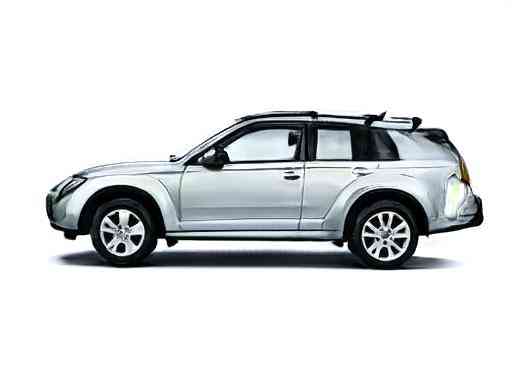}&
					\includegraphics[width=.14\textwidth,trim=0 0 0 30,clip]{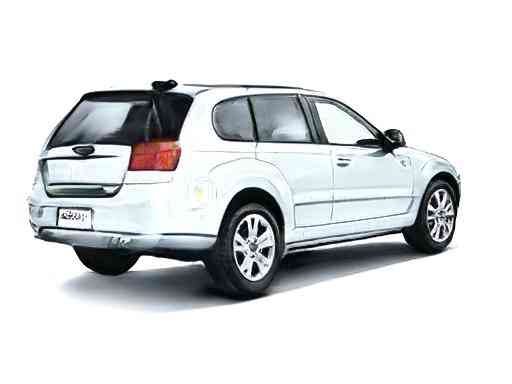}&
					\includegraphics[width=.14\textwidth,trim=0 0 0 30,clip]{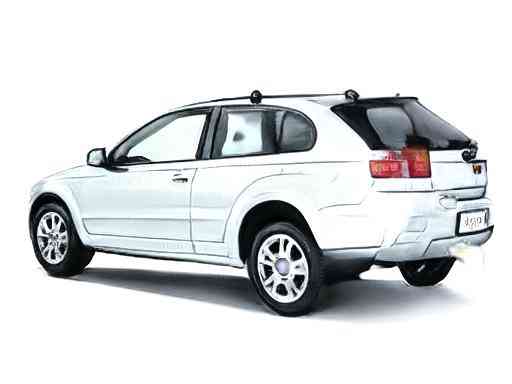}					
					\\

					\includegraphics[width=.14\textwidth,trim=0 0 0 30,clip]{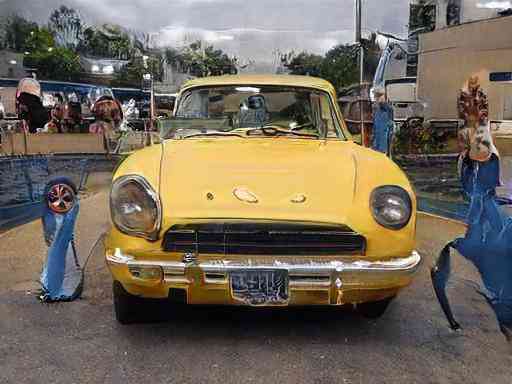}&
					\includegraphics[width=.14\textwidth,trim=0 0 0 30,clip]{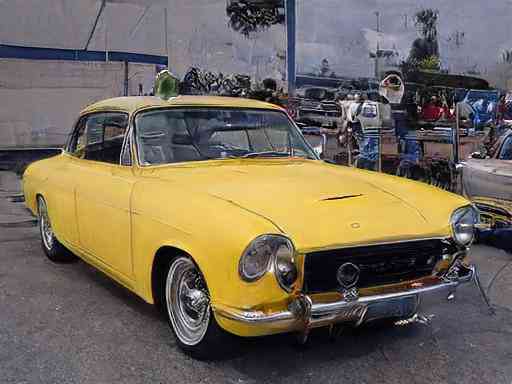}&
					\includegraphics[width=.14\textwidth,trim=0 0 0 30,clip]{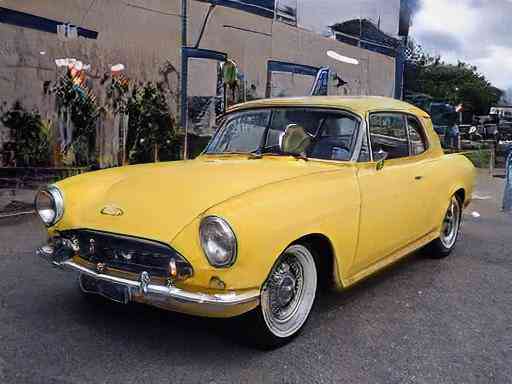}&
					\includegraphics[width=.14\textwidth,trim=0 0 0 30,clip]{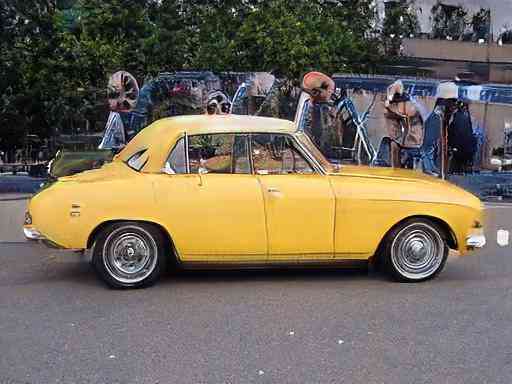}&
					\includegraphics[width=.14\textwidth,trim=0 0 0 30,clip]{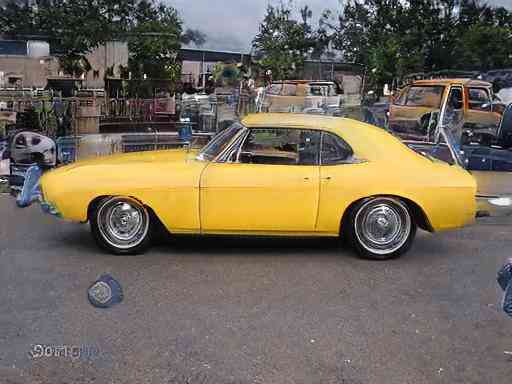}&
					\includegraphics[width=.14\textwidth,trim=0 0 0 30,clip]{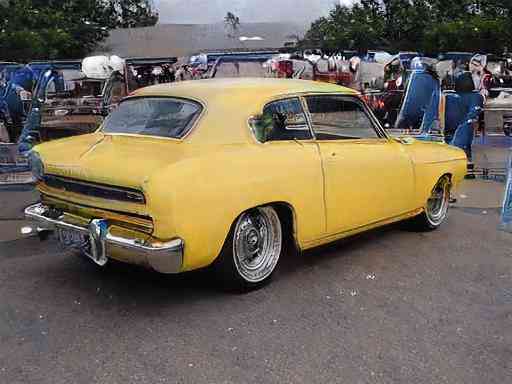}&
					\includegraphics[width=.14\textwidth,trim=0 0 0 30,clip]{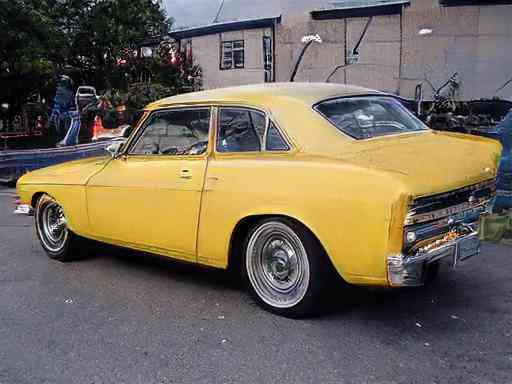}					
					\\
					
					\includegraphics[width=.14\textwidth,trim=0 0 0 30,clip]{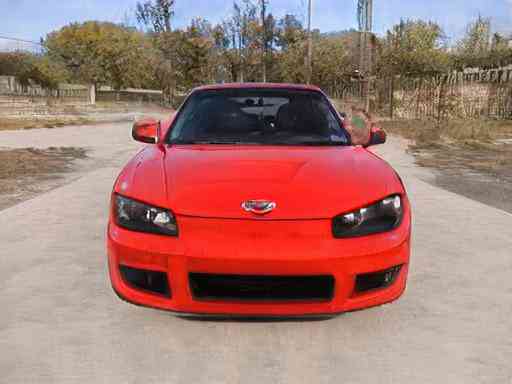}&
					\includegraphics[width=.14\textwidth,trim=0 0 0 30,clip]{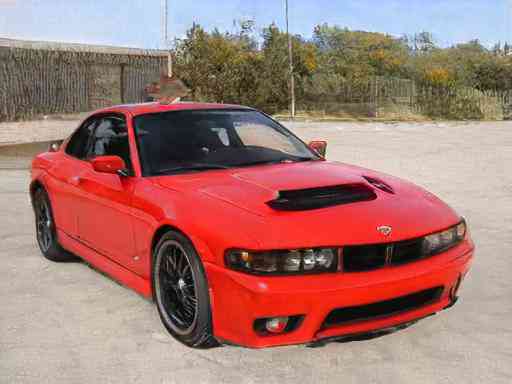}&
					\includegraphics[width=.14\textwidth,trim=0 0 0 30,clip]{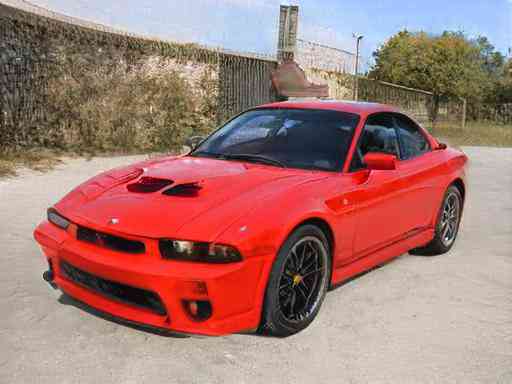}&
					\includegraphics[width=.14\textwidth,trim=0 0 0 30,clip]{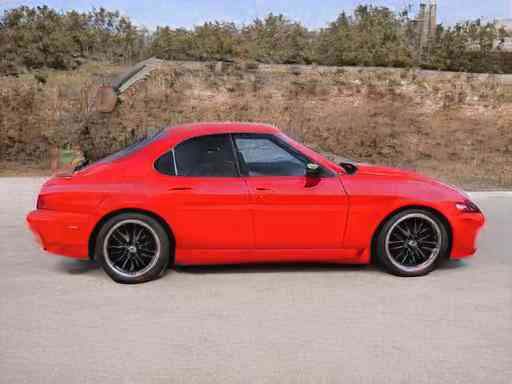}&
					\includegraphics[width=.14\textwidth,trim=0 0 0 30,clip]{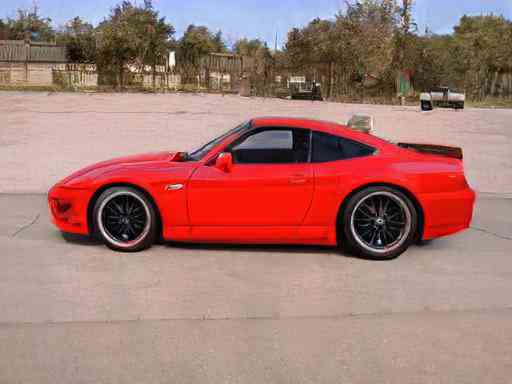}&
					\includegraphics[width=.14\textwidth,trim=0 0 0 30,clip]{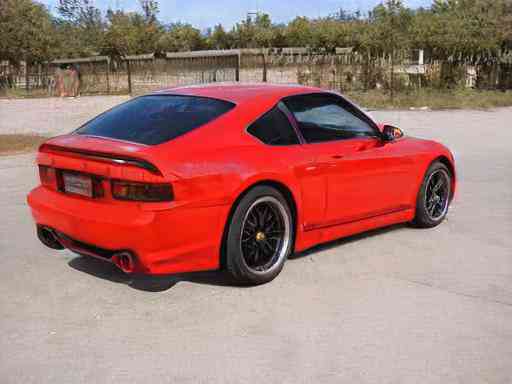}&
					\includegraphics[width=.14\textwidth,trim=0 0 0 30,clip]{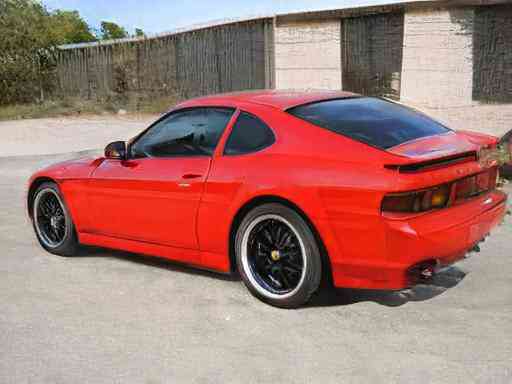}					
					\\

					\includegraphics[width=.14\textwidth,height=.105\textwidth, trim=0 0 0 30,clip]{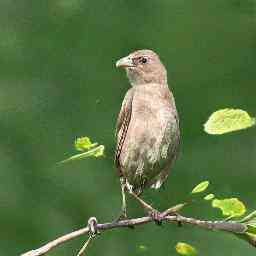}&
					\includegraphics[width=.14\textwidth,height=.105\textwidth, trim=0 0 0 30,clip]{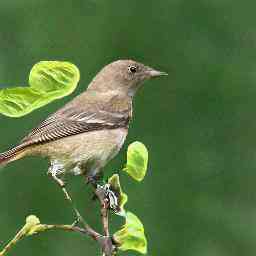}&
					\includegraphics[width=.14\textwidth,height=.105\textwidth, trim=0 0 0 30,clip]{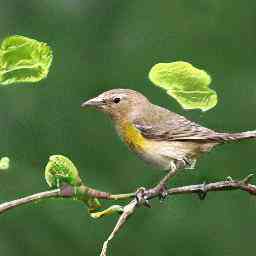}&
					\includegraphics[width=.14\textwidth,height=.105\textwidth, trim=0 0 0 30,clip]{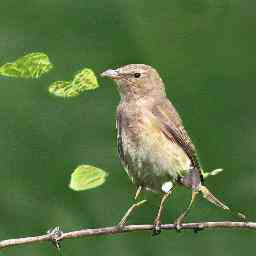}&
					\includegraphics[width=.14\textwidth,height=.105\textwidth, trim=0 0 0 30,clip]{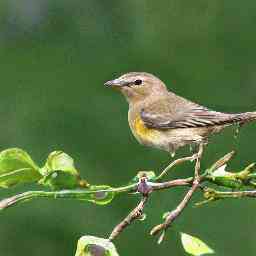}&
					\includegraphics[width=.14\textwidth,height=.105\textwidth, trim=0 0 0 30,clip]{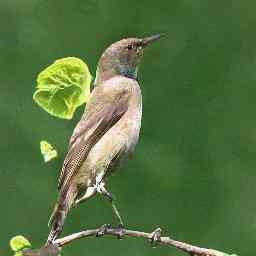}&
					\includegraphics[width=.14\textwidth,height=.105\textwidth, trim=0 0 0 30,clip]{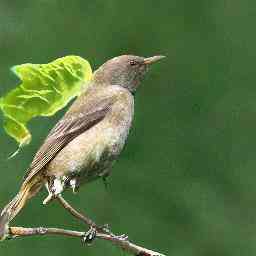}			
					\\	
					
					\includegraphics[width=.14\textwidth,height=.105\textwidth, trim=0 0 0 30,clip]{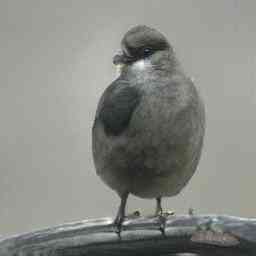}&
					\includegraphics[width=.14\textwidth,height=.105\textwidth, trim=0 0 0 30,clip]{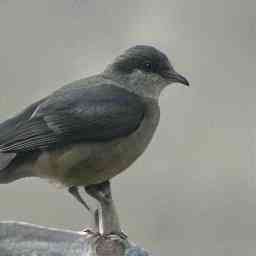}&
					\includegraphics[width=.14\textwidth,height=.105\textwidth, trim=0 0 0 30,clip]{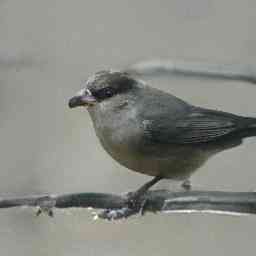}&
					\includegraphics[width=.14\textwidth,height=.105\textwidth, trim=0 0 0 30,clip]{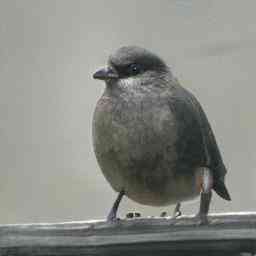}&
					\includegraphics[width=.14\textwidth,height=.105\textwidth, trim=0 0 0 30,clip]{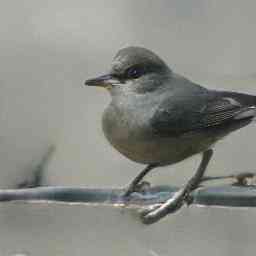}&
					\includegraphics[width=.14\textwidth,height=.105\textwidth, trim=0 0 0 30,clip]{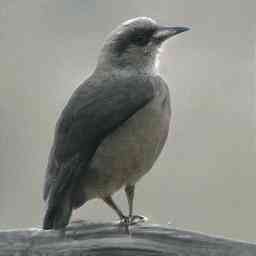}&
					\includegraphics[width=.14\textwidth,height=.105\textwidth, trim=0 0 0 30,clip]{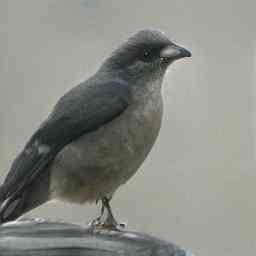}			
					\\
					\includegraphics[width=.14\textwidth,height=.105\textwidth, trim=0 0 0 30,clip]{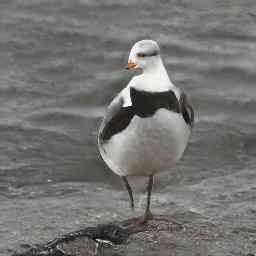}&
					\includegraphics[width=.14\textwidth,height=.105\textwidth, trim=0 0 0 30,clip]{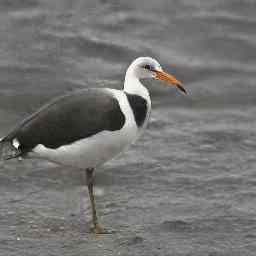}&
					\includegraphics[width=.14\textwidth,height=.105\textwidth, trim=0 0 0 30,clip]{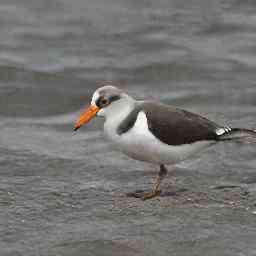}&
					\includegraphics[width=.14\textwidth,height=.105\textwidth, trim=0 0 0 30,clip]{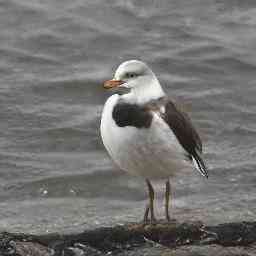}&
					\includegraphics[width=.14\textwidth,height=.105\textwidth, trim=0 0 0 30,clip]{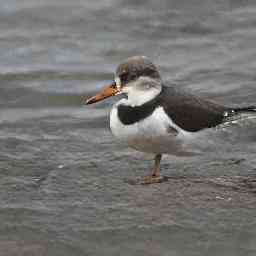}&
					\includegraphics[width=.14\textwidth,height=.105\textwidth, trim=0 0 0 30,clip]{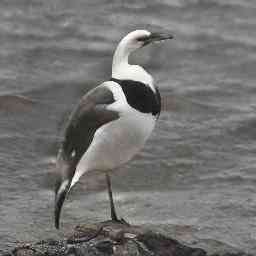}&
					\includegraphics[width=.14\textwidth,height=.105\textwidth, trim=0 0 0 30,clip]{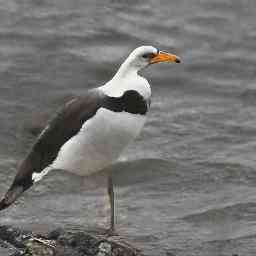}			
					\\									
					\includegraphics[width=.14\textwidth,height=.105\textwidth, trim=0 0 0 30,clip]{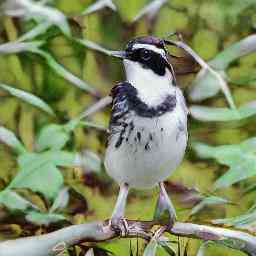}&
					\includegraphics[width=.14\textwidth,height=.105\textwidth, trim=0 0 0 30,clip]{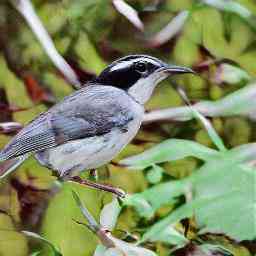}&
					\includegraphics[width=.14\textwidth,height=.105\textwidth, trim=0 0 0 30,clip]{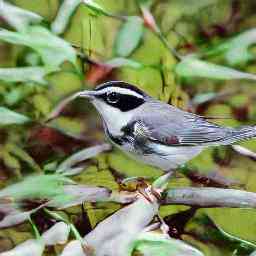}&
					\includegraphics[width=.14\textwidth,height=.105\textwidth, trim=0 0 0 30,clip]{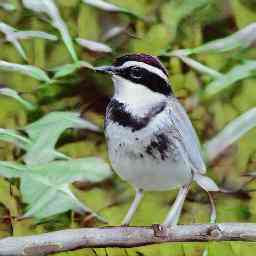}&
					\includegraphics[width=.14\textwidth,height=.105\textwidth, trim=0 0 0 30,clip]{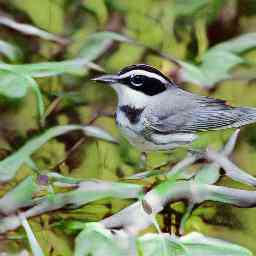}&
					\includegraphics[width=.14\textwidth,height=.105\textwidth, trim=0 0 0 30,clip]{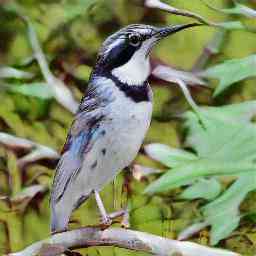}&
					\includegraphics[width=.14\textwidth,height=.105\textwidth, trim=0 0 0 30,clip]{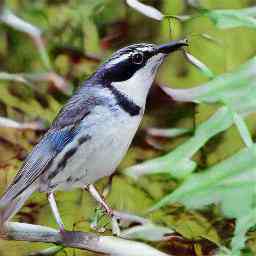}			
					\\

					\includegraphics[width=.14\textwidth,height=.105\textwidth, trim=0 0 0 30,clip]{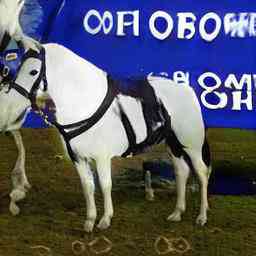}&
					\includegraphics[width=.14\textwidth,height=.105\textwidth, trim=0 0 0 30,clip]{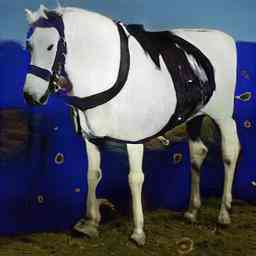}&
					\includegraphics[width=.14\textwidth,height=.105\textwidth, trim=0 0 0 30,clip]{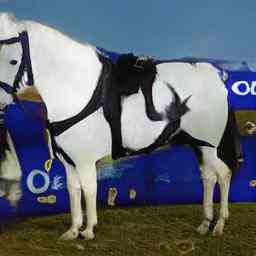}&
					\includegraphics[width=.14\textwidth,height=.105\textwidth, trim=0 0 0 30,clip]{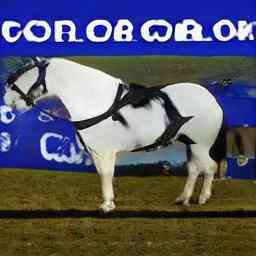}&
					\includegraphics[width=.14\textwidth,height=.105\textwidth, trim=0 0 0 30,clip]{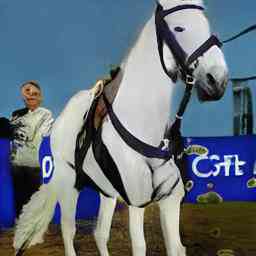}&
					\includegraphics[width=.14\textwidth,height=.105\textwidth, trim=0 0 0 30,clip]{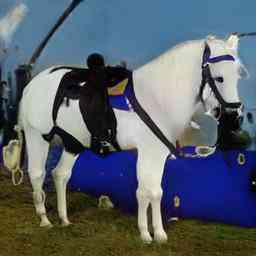}&
					\includegraphics[width=.14\textwidth,height=.105\textwidth, trim=0 0 0 30,clip]{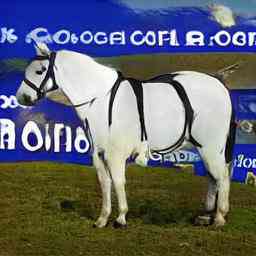}			
					\\		
					
					\includegraphics[width=.14\textwidth,height=.105\textwidth, trim=0 0 0 30,clip]{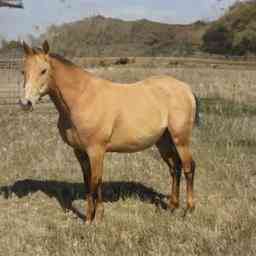}&
					\includegraphics[width=.14\textwidth,height=.105\textwidth, trim=0 0 0 30,clip]{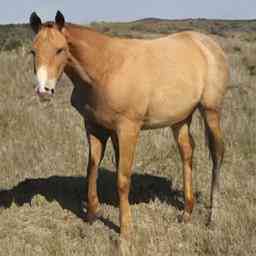}&
					\includegraphics[width=.14\textwidth,height=.105\textwidth, trim=0 0 0 30,clip]{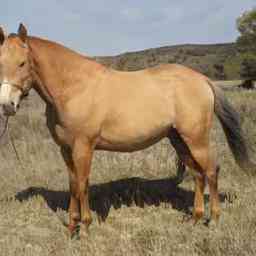}&
					\includegraphics[width=.14\textwidth,height=.105\textwidth, trim=0 0 0 30,clip]{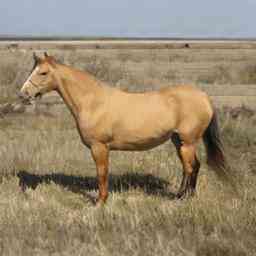}&
					\includegraphics[width=.14\textwidth,height=.105\textwidth, trim=0 0 0 30,clip]{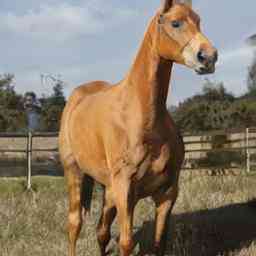}&
					\includegraphics[width=.14\textwidth,height=.105\textwidth, trim=0 0 0 30,clip]{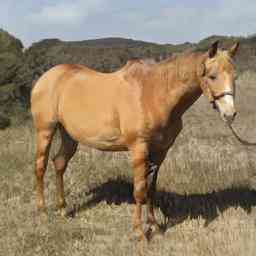}&
					\includegraphics[width=.14\textwidth,height=.105\textwidth, trim=0 0 0 30,clip]{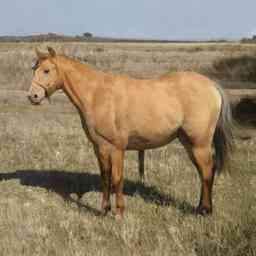}			
					\\		
					
					\includegraphics[width=.14\textwidth,height=.105\textwidth, trim=0 0 0 30,clip]{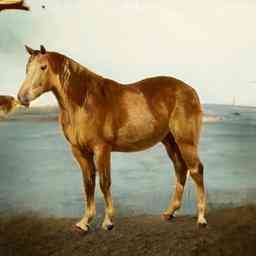}&
					\includegraphics[width=.14\textwidth,height=.105\textwidth, trim=0 0 0 30,clip]{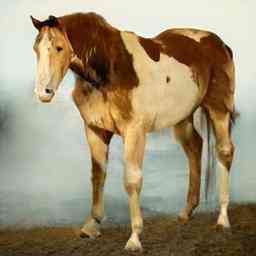}&
					\includegraphics[width=.14\textwidth,height=.105\textwidth, trim=0 0 0 30,clip]{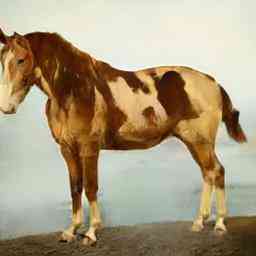}&
					\includegraphics[width=.14\textwidth,height=.105\textwidth, trim=0 0 0 30,clip]{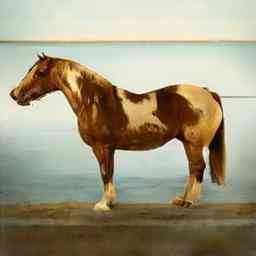}&
					\includegraphics[width=.14\textwidth,height=.105\textwidth, trim=0 0 0 30,clip]{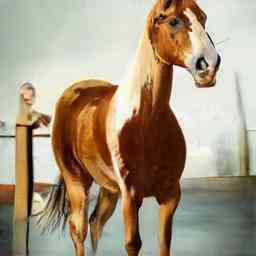}&
					\includegraphics[width=.14\textwidth,height=.105\textwidth, trim=0 0 0 30,clip]{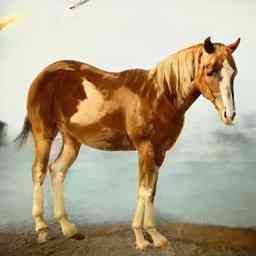}&
					\includegraphics[width=.14\textwidth,height=.105\textwidth, trim=0 0 0 30,clip]{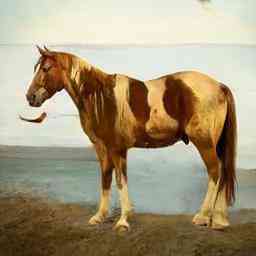}			
					\\		
					
					\includegraphics[width=.14\textwidth,height=.105\textwidth, trim=0 0 0 30,clip]{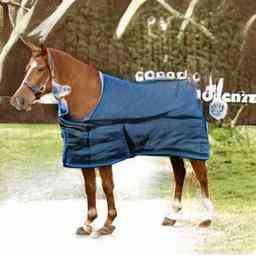}&
					\includegraphics[width=.14\textwidth,height=.105\textwidth, trim=0 0 0 30,clip]{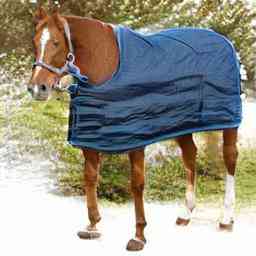}&
					\includegraphics[width=.14\textwidth,height=.105\textwidth, trim=0 0 0 30,clip]{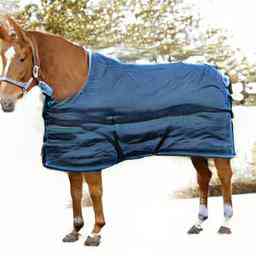}&
					\includegraphics[width=.14\textwidth,height=.105\textwidth, trim=0 0 0 30,clip]{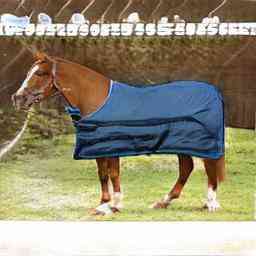}&
					\includegraphics[width=.14\textwidth,height=.105\textwidth, trim=0 0 0 30,clip]{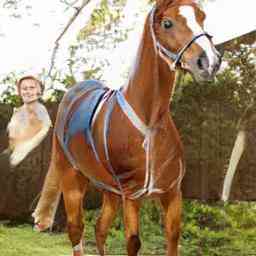}&
					\includegraphics[width=.14\textwidth,height=.105\textwidth, trim=0 0 0 30,clip]{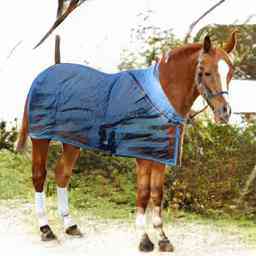}&
					\includegraphics[width=.14\textwidth,height=.105\textwidth, trim=0 0 0 30,clip]{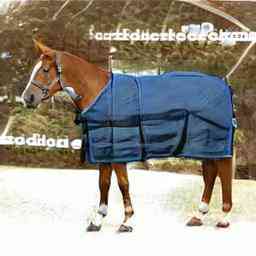}			
					\\
									
					\hspace*{0pt}
				\end{tabular}
			\end{tabular}
		\end{center}
		\vspace*{-7mm}
	}
	\caption{\label{fig:dataset} \textbf{\footnotesize Dataset Overview:} \footnotesize We synthesize multi-view datasets for three classes: \emph{car}, \emph{horse}, and \emph{bird}. Our datasets contain objects with various shapes, textures and viewpoints. Notice the consistency of pose of object in each column (for each class). Challenges include the fact that for all of these objects StyleGAN has not learned to synthesize views that overlook the object from above due to the photographer bias in the original dataset that StyleGAN was trained on.}	
\end{figure*}

We further show examples from our StyleGAN-generated dataset in Fig.~\ref{fig:dataset}. Our dataset contains objects with various shapes, textures and viewpoints. In particular, in the first six rows, one can notice a diverse variants of car types (Standard Car, SUV, Sports car, Antique Car, etc) . We find that StyleGAN can also produce rare car shapes like trucks, but with a lower probability.

\clearpage

\section{Camera Initialization}
\label{sec:camera_init}

Inverse graphics tasks require camera pose information during training, which is challenging to acquire for real imagery. Pose is generally obtained by annotating keypoints for each object and running structure-from-motion (SFM) techniques~\citep{cub,xiang_wacv14} to compute camera parameters. However, keypoint annotation is quite time consuming -- requiring roughly 3-5minutes per object which we verify in practice using the LabelMe interface~\citep{labelme}. In our work, we utilize StyleGAN to significantly reduce annotation effort since samples with the same $w_v^*$ share the same viewpoint. Therefore, we only need to assign a few selected $w_v^*$ into camera poses. In particular, we assign poses into several bins which we show is sufficient for training inverse graphics networks where, along with the network parameters, cameras get jointly optimized during training using these bins as initialization. 

Specifically, we assign poses into 39, 22 and 8 bins for the car, horse and bird classes, respectively. 
This allows us to annotate all the views (and thus the entire dataset) in only \emph{1 minute}. We do acknowledge additional time in selecting good views out of several candidates. 

We annotate each view with a rough absolute camera pose (which we further optimize during training). To be specific, we first select 12 azimuth angles: [0$^\circ$, 30$^\circ$, 60$^\circ$, 90$^\circ$, 120$^\circ$, 150$^\circ$, 180$^\circ$, 210$^\circ$, 240$^\circ$, 270$^\circ$, 300$^\circ$, 330$^\circ$]. Given a StyleGAN viewpoint, we manually classify which azimuth angle it is close to and assign it to the corresponding label with fixed elevation (0$^\circ$) and camera distance.

To demonstrate the effectiveness of our camera initialization, we make a comparison with another inverse graphics network trained with a more \emph{accurate} camera initialization. Such an initialization is done by manually annotating object keypoints in each of the selected views ($w_v^*$) of a single car example, which takes about 3-4 hours (around 200 minutes, 39 views). Note that this is still a significantly lower annotation effort compared to 200-350 hours required to annotate keypoints for every single object in the Pascal3D dataset. We then compute the camera parameters using SfM. We refer to the two inverse graphics networks trained with different camera initializations as $view$-model and $keypoint$ -model, respectively.

\begin{figure*}[t!]
	{
		\vspace*{-10mm}
		\begin{center}
			\setlength{\tabcolsep}{1pt}
			\setlength{\fboxrule}{0pt}
			\hspace*{0pt}
			\begin{tabular}{c}
				\begin{tabular}{cccccc}
										\footnotesize{Azimuth=0$^\circ$} & \footnotesize{Azimuth=30$^\circ$} &\footnotesize{Azimuth=30$^\circ$} & \footnotesize{Azimuth=180$^\circ$} & \footnotesize{Azimuth=210$^\circ$} & \footnotesize{Azimuth=270$^\circ$}
										\\
										\includegraphics[width=.161\textwidth]{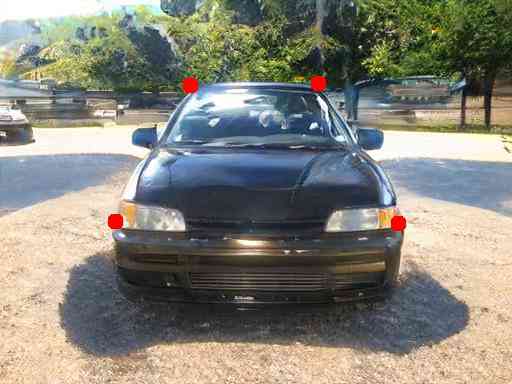}&
										\includegraphics[width=.161\textwidth]{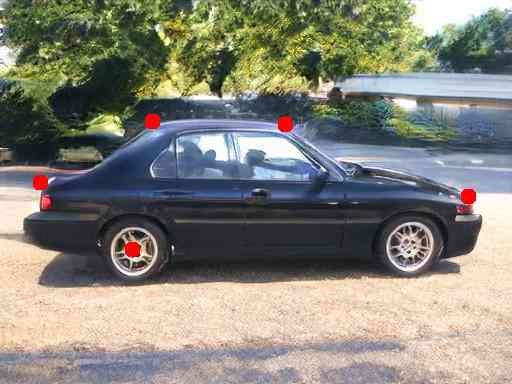}&
										\includegraphics[width=.161\textwidth]{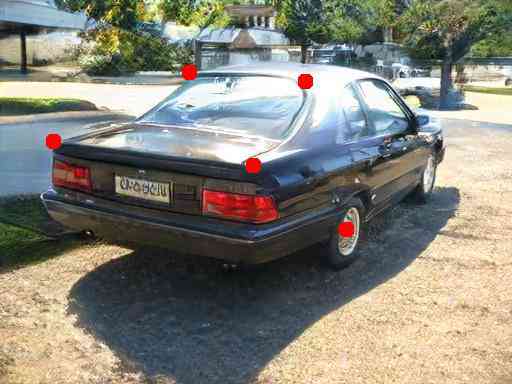}&
										\includegraphics[width=.161\textwidth]{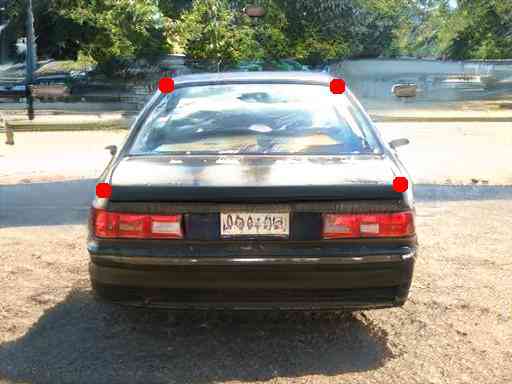}&
										\includegraphics[width=.161\textwidth]{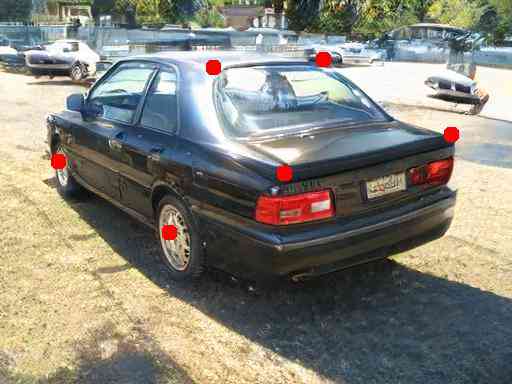}
										&
										\includegraphics[width=.161\textwidth]{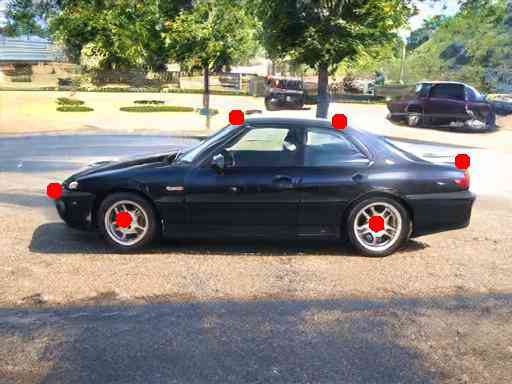}
										\\

										\includegraphics[width=.161\textwidth]{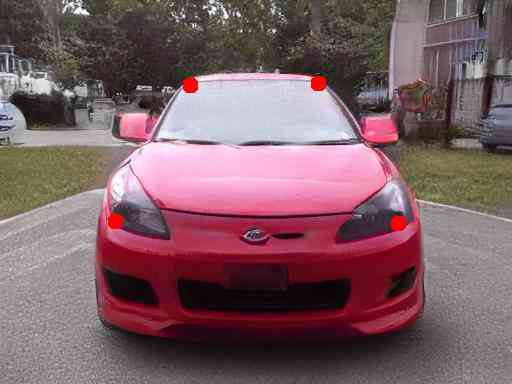}&
										\includegraphics[width=.161\textwidth]{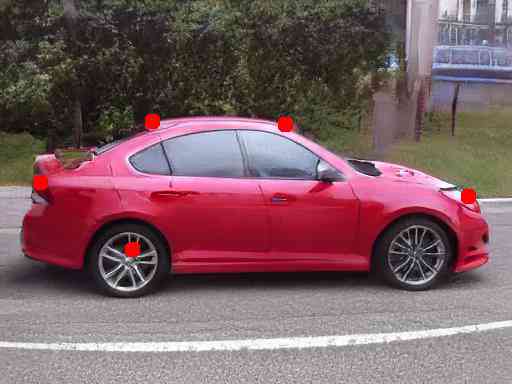}&
										\includegraphics[width=.161\textwidth]{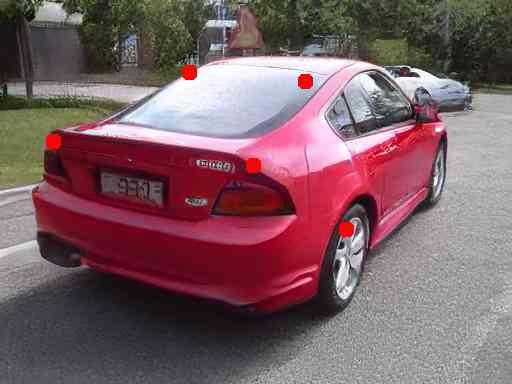}&
										\includegraphics[width=.161\textwidth]{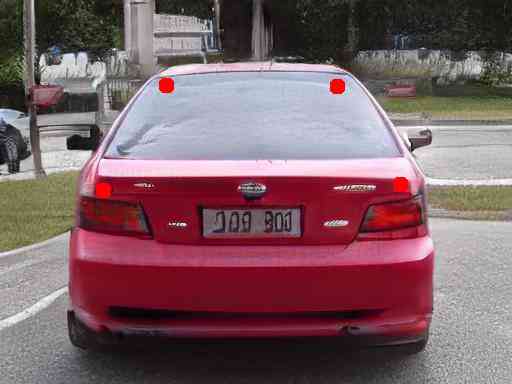}&
										\includegraphics[width=.161\textwidth]{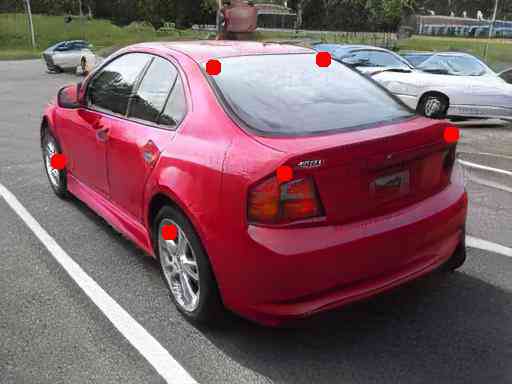}
										&
										\includegraphics[width=.161\textwidth]{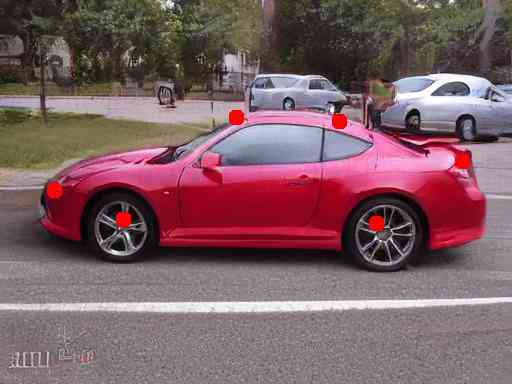}
										\\
																		
										\includegraphics[width=.161\textwidth]{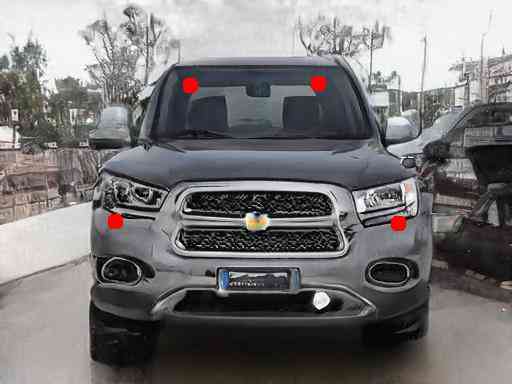}&
										\includegraphics[width=.161\textwidth]{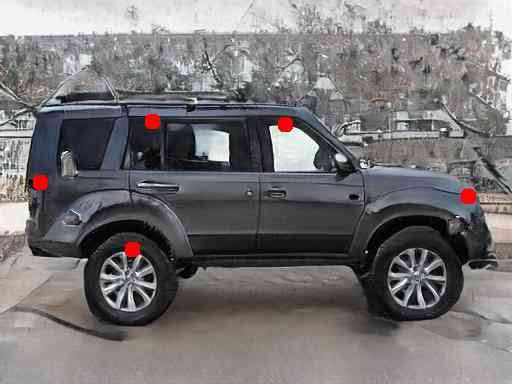}&
										\includegraphics[width=.161\textwidth]{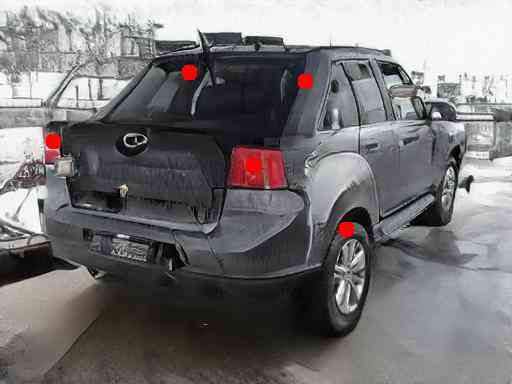}&
										\includegraphics[width=.161\textwidth]{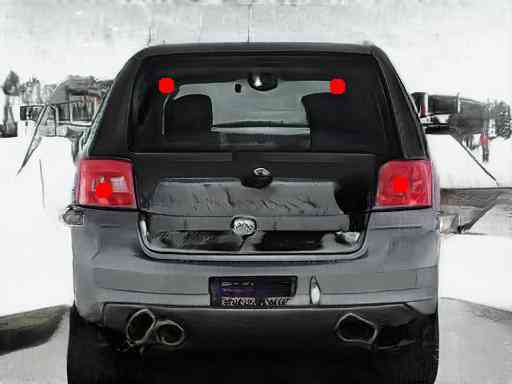}&
										\includegraphics[width=.161\textwidth]{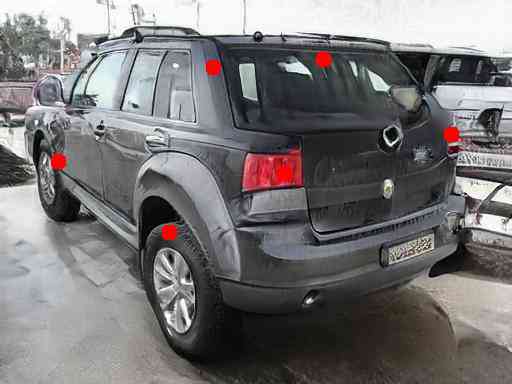}&
										\includegraphics[width=.161\textwidth]{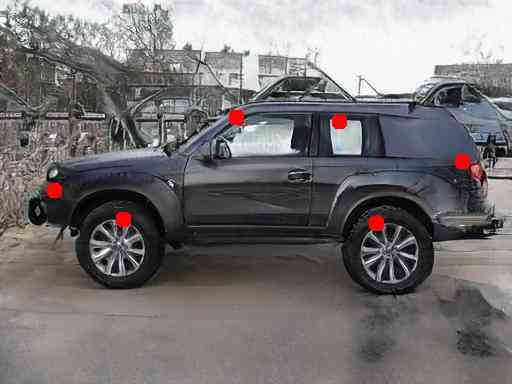}
										\\	
										
										\includegraphics[width=.161\textwidth]{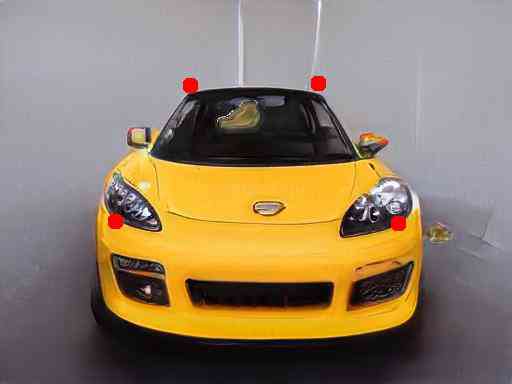}&
										\includegraphics[width=.161\textwidth]{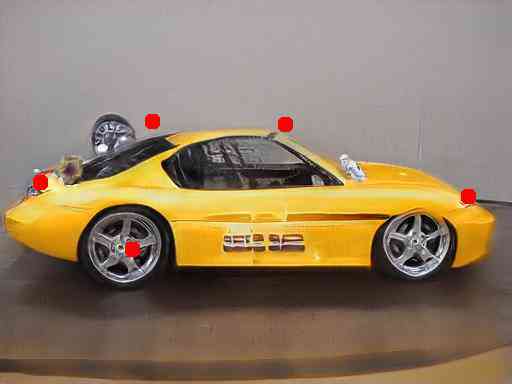}&
										\includegraphics[width=.161\textwidth]{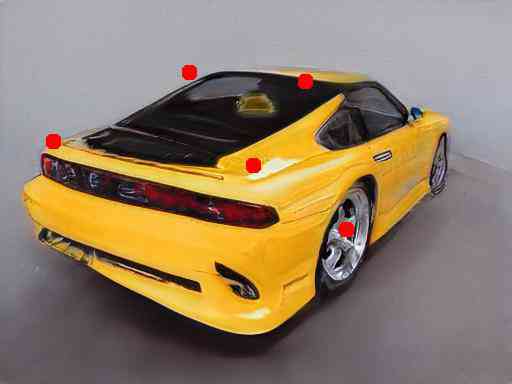}&
										\includegraphics[width=.161\textwidth]{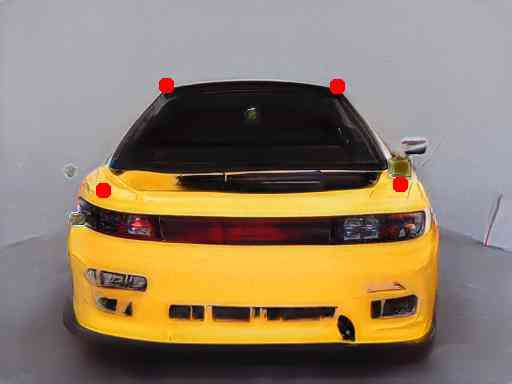}&
										\includegraphics[width=.161\textwidth]{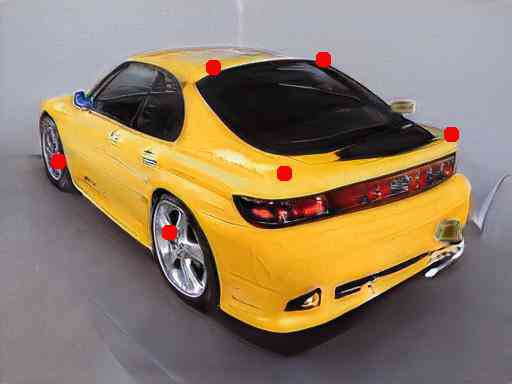}&
										\includegraphics[width=.161\textwidth]{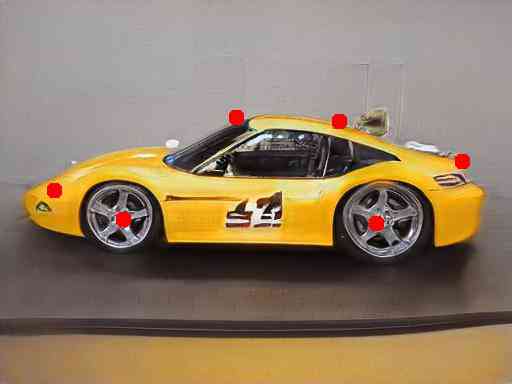}
										\\											


										\includegraphics[width=.161\textwidth]{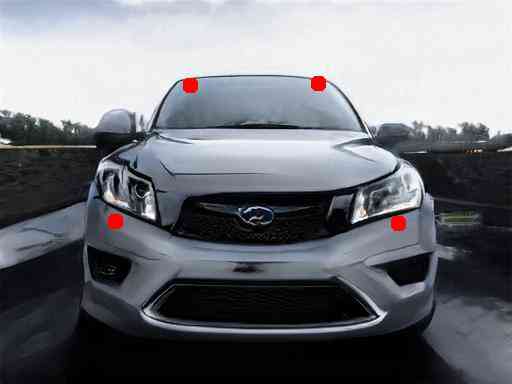}&
										\includegraphics[width=.161\textwidth]{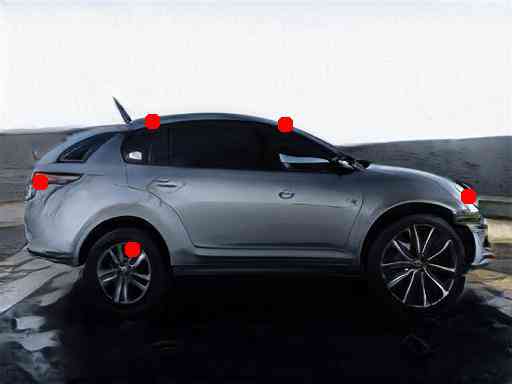}&
										\includegraphics[width=.161\textwidth]{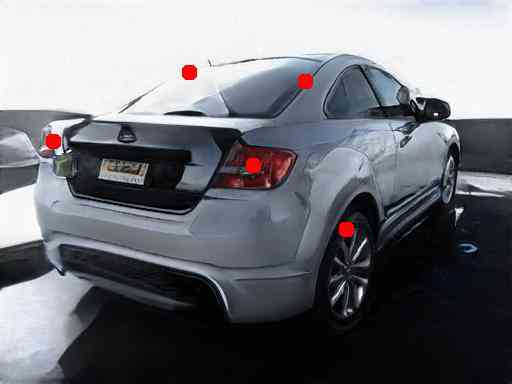}&
										\includegraphics[width=.161\textwidth]{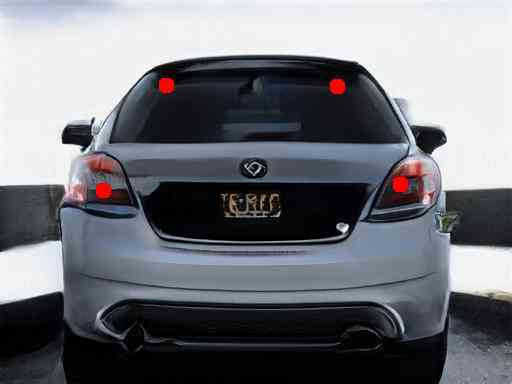}&
										\includegraphics[width=.161\textwidth]{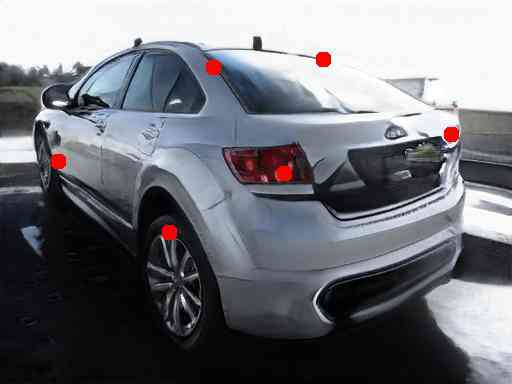}&
										\includegraphics[width=.161\textwidth]{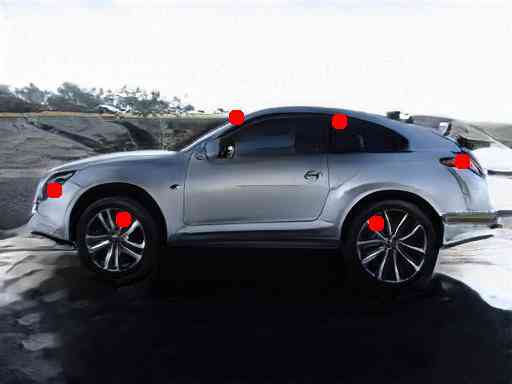}
										\\

					\end{tabular}
					\end{tabular}
		\end{center}
		\vspace*{-4mm}
	}
	\caption{\label{fig:viewpointannot}\footnotesize We show examples of cars  synthesized in chosen viewpoints (columns) along with annotations. Top row shows the pose bin annotation, while the images show the annotated keypoints. We annotated keypoints for the car example in the first image-row based on which we compute the accurate camera parameters using SfM. To showcase how well aligned the objects are for the same viewpoint latent code, we visualize the annotated keypoints on all other synthesized car examples. Note that we do not assume that these keypoints are accurate for these cars (only the implied viewpoint).  Annotating pose bins took 1 min for the car class, while keypoint annotation took 3-4 hours, both types of annotations thus being quite efficient. 
	We empirically find that pose bin annotation is sufficient in training accurate inverse graphics networks (when optimizing camera parameters during training in addition to optimizing the network parameters).
	}
	\vspace{-2mm}
\end{figure*}

We visualize our two different annotation types in Fig~\ref{fig:viewpointannot}. We show annotated bins in the top. We annotated keypoints for the (synthesized) car example in the first image row based on which we compute the accurate viewpoint using SfM. To showcase how well aligned the objects are for the same viewpoint code, we visualize the annotated keypoints on all other synthesized car examples. Note that we do not assume that these keypoints are accurate for these cars (only the implied viewpoint). 


We quantitatively evaluate two initialization methods in Table.~\ref{tab:ablation}. We first compare the annotation and training times. While it takes the same amount of time to train, $view$-model saves on annotation time. The performance of  $view$-model and $keypoint$ -model are comparable with almost the same 2D IOU re-projection score on the StyleGAN test set. Moreover, during training the two camera systems converge to the same position. We evaluate this by converting all the views into quaternions and compare the difference between the rotation axes and rotation angles. Among all views, the average difference of the rotation axes is only 1.43$^\circ$ and the rotation angle is 0.42$^\circ$. The maximum difference of the rotation axes  is only 2.95$^\circ$ and the rotation angle is 1.11$^\circ$. 

\begin{figure*}[h]
	{
		\vspace*{0pt}
		\begin{center}
			\setlength{\tabcolsep}{1pt}
			\setlength{\fboxrule}{0pt}
			\hspace*{-0.25cm}
			\begin{tabular}{c}
				\begin{tabular}{cccc}
					\rotatebox{90}{\,\,{\color{black}{\scriptsize $view$-Init.}}}~~~& 
					\hspace*{-0.3cm}
					\includegraphics[height=0.115\linewidth]{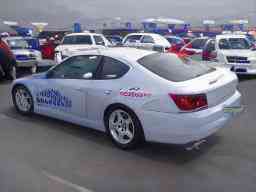}&
					\includegraphics[height=0.115\linewidth]{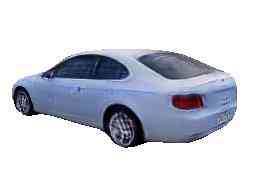}&
					\includegraphics[height=0.115\linewidth]{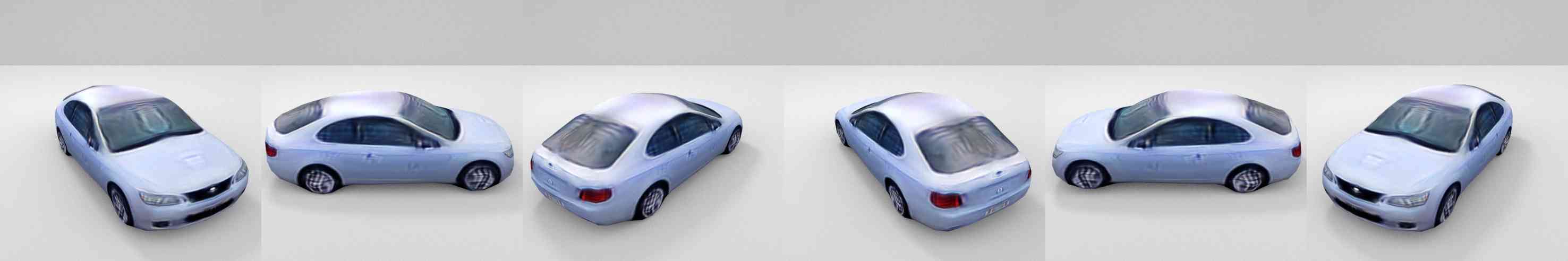}
					\\
					\rotatebox{90}{{\color{black}{\scriptsize $keypoint$-Init.}}}~~~&
					\hspace*{-0.3cm}
					\includegraphics[height=0.115\linewidth]{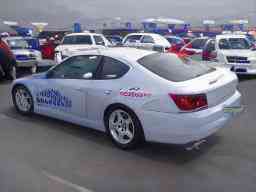}&
					\includegraphics[height=0.115\linewidth]{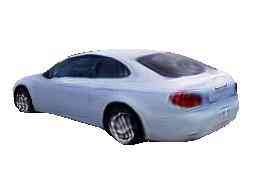}&
					\includegraphics[height=0.115\linewidth]{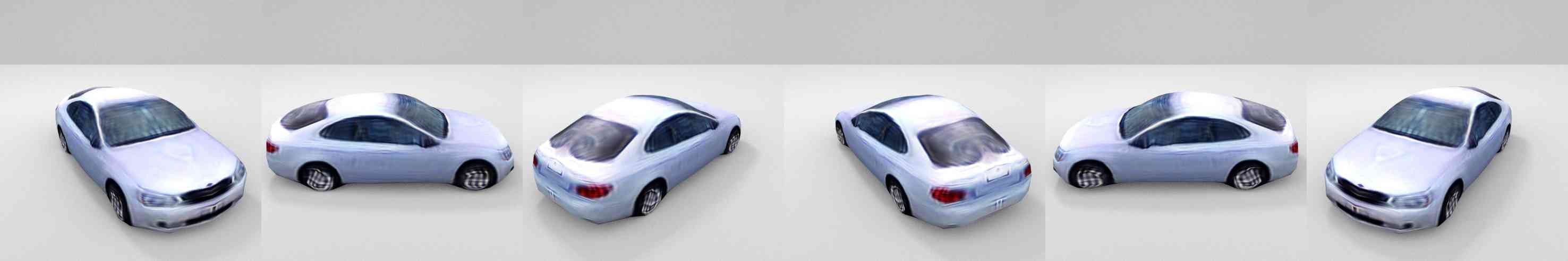}
					\\
					& {\footnotesize Input} & {\footnotesize Pred.} & {\footnotesize Multiple Views for the Predicted Shape and Texture}
				\end{tabular}
			\end{tabular}
		\end{center}
		\vspace*{-0.3cm}
	}
	\caption{\label{fig:camera_comp} \footnotesize \textbf{Comparison of Different Camera Initializations}: The first row shows predictions from $keypoint$-Initialization (cameras computed by running SFM on annotated keypoints) and the second row show results obtained by training with $view$-Initialization (cameras are coarsely annotated into 12 view bins). Notice how close the two predictions are, indicating that coarse viewpoint annotation is sufficient for training accurate inverse graphics networks. Coarse viewpoint annotation can be done in 1 minute. }
	\vspace*{2mm}
\end{figure*}

We further qualitatively compare the two methods in Fig.~\ref{fig:camera_comp}, showing that they perform very similarly. 
Both, qualitative and quantitative comparisons, demonstrated that $view$-camera initialization is sufficient for training accurate inverse graphics networks and no additional annotation is required. This demonstrates a scaleable way for creating multi-view datasets with StyleGAN, with roughy a minute of annotation time per class. 

\begin{table*}   
	\vspace*{4pt}
	\begin{center}
		\addtolength{\tabcolsep}{-1pt}
		\footnotesize
		\centering
			\begin{tabular}{cc}
		\begin{tabular}{lccc}
			\toprule
			Annotation Type & Annotation Time  & Training Time  & 2D IOU \\
			\midrule
			 $keypoint$  & 3-4h  & 60h & 0.953 \\
			$view$& 1min  & 60h & 0.952  \\
			\bottomrule
		\end{tabular}
	&\hspace{1.5mm}
	\begin{tabular}{cccc}
		\toprule
		& Quaternion & Mean & Max  \\
			\midrule
		  & $q_{xyz}$ &  1.43$^\circ$ & 2.95$^\circ$ \\
		 & $q_{w}$  & 0.42$^\circ$  & 1.11$^\circ$\\
		\bottomrule
	\end{tabular}
\\
\\[-2mm]
(a)  Time \& Performance & (b) Camera Difference after Training\\
	\end{tabular}
\end{center}
	\vspace{-7pt}
	\label{tab:ablation}
		\caption{\footnotesize \textbf{Comparison of Different Camera Initializations}: First table shows annotation time required for the StyleGAN dataset, and training times of  the $view$-model and  $keypoint$-model on the dataset with respective annotations (binned viewpoints or cameras computed with SFM from annotated keypoints). The $view$-model requires significantly less annotation time, and its final performance is comparable to the $keypoint$-model. Second table shows the difference of the camera parameters after training both methods (which optimize cameras during training). They converge to very similar camera positions. This shows that coarse view annotation along with camera optimization during training is sufficient in training high accuracy inverse graphics networks.}
	\vspace{-1mm}
\end{table*}


\section{3D Inference}
\label{sec:3d_inference}

\begin{figure*}[t]
	{
		\vspace*{0pt}
		\begin{center}
			\setlength{\tabcolsep}{1pt}
			\setlength{\fboxrule}{0pt}
			\hspace*{-0.25cm}
			\begin{tabular}{c}
				\begin{tabular}{cccc}
					\rotatebox{90}{\,\,\,\,{\color{black}{\tiny Pascal3D}}}&
					\includegraphics[width=2cm,height=1.5cm]{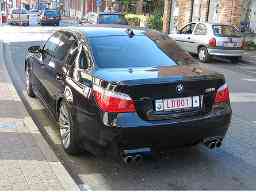}&
					\includegraphics[width=2cm,height=1.5cm]{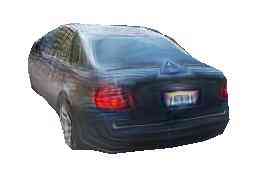}&
					\includegraphics[height=1.5cm]{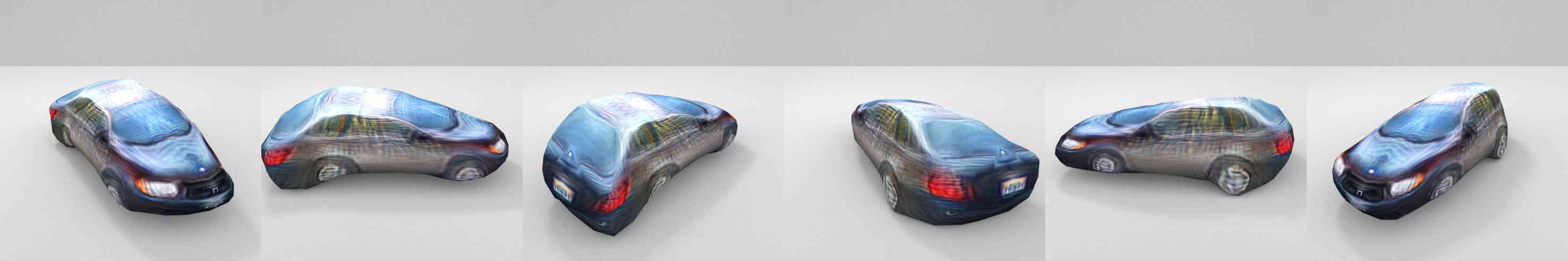}
					\\
					\rotatebox{90}{\,\,\,\,\,\,\,\,{\color{black}{\tiny Ours}}}&
					\includegraphics[width=2cm,height=1.5cm]{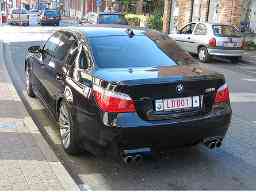}&
					\includegraphics[width=2cm,height=1.5cm]{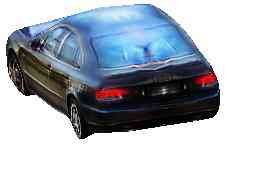}&
					\includegraphics[height=1.5cm]{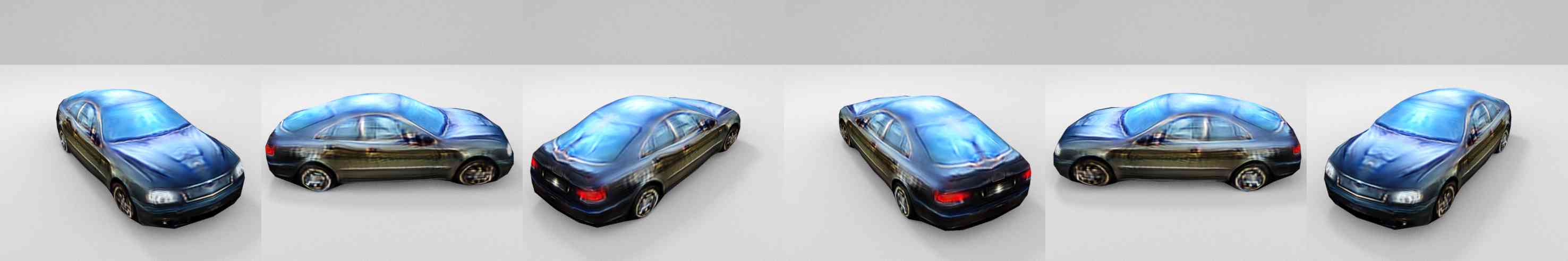}
					\\
					\rotatebox{90}{\,\,\,\,{\color{black}{\tiny Pascal3D}}}&
					\includegraphics[width=2cm,height=1.5cm]{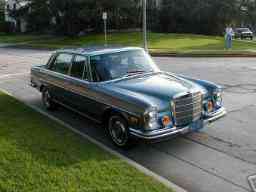}&
					\includegraphics[width=2cm,height=1.5cm]{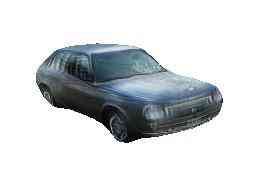}&
					\includegraphics[height=1.5cm]{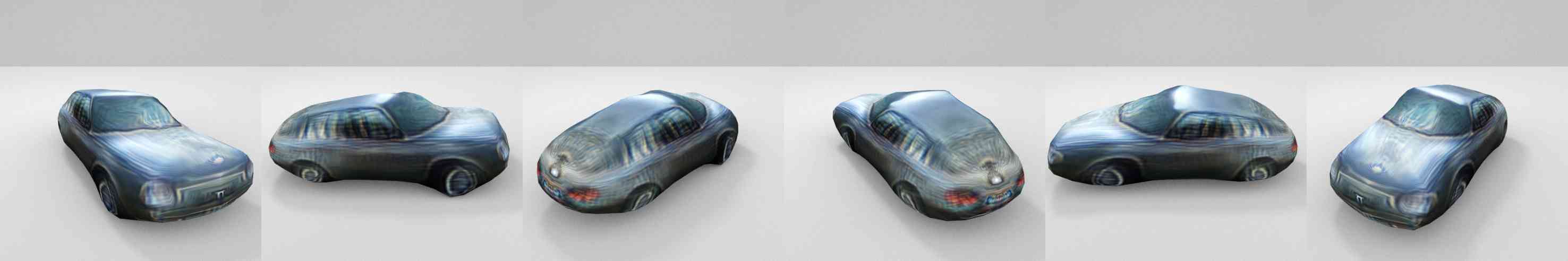}
					\\
					\rotatebox{90}{\,\,\,\,\,\,\,\,{\color{black}{\tiny Ours}}}&
					\includegraphics[width=2cm,height=1.5cm]{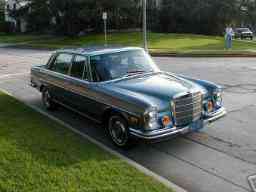}&
					\includegraphics[width=2cm,height=1.5cm]{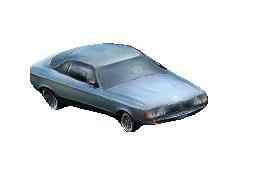}&
					\includegraphics[height=1.5cm]{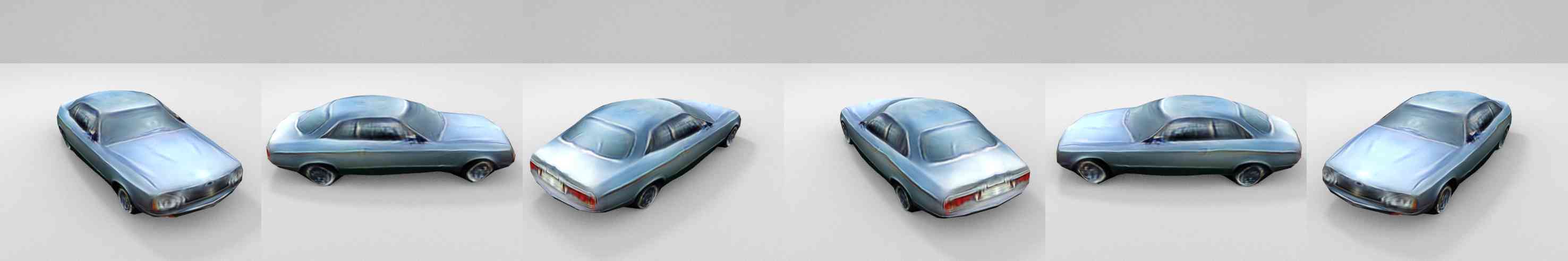}
					\\
					\rotatebox{90}{\,\,\,\, {\color{black}{\tiny Pascal3D}}}&
					\includegraphics[width=2cm,height=1.5cm]{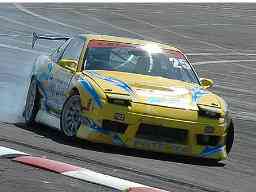}&
					\includegraphics[width=2cm,height=1.5cm]{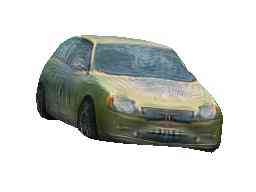}&
					\includegraphics[height=1.5cm]{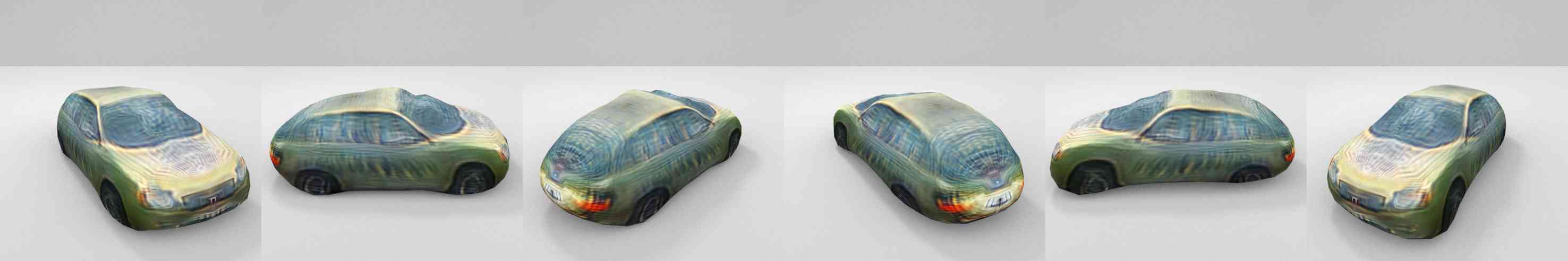}
					\\
					\rotatebox{90}{\,\,\,\,\,\,\,\,{\color{black}{\tiny Ours}}}&
					\includegraphics[width=2cm,height=1.5cm]{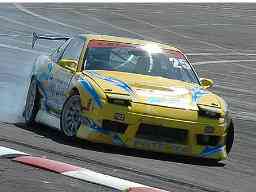}&
					\includegraphics[width=2cm,height=1.5cm]{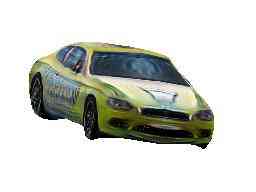}&
					\includegraphics[height=1.5cm]{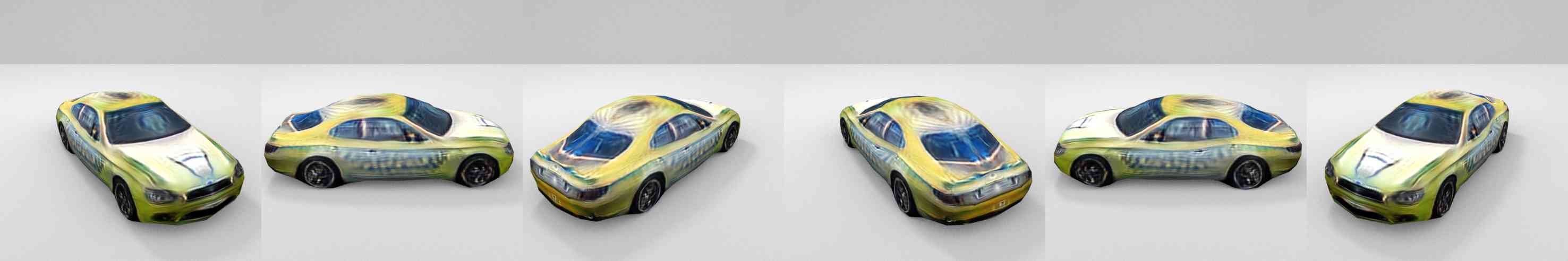}
					\\
					\rotatebox{90}{\,\,\,\, {\color{black}{\tiny Pascal3D}}}&
					\includegraphics[width=2cm,height=1.5cm]{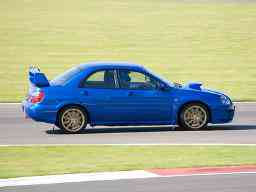}&
					\includegraphics[width=2cm,height=1.5cm]{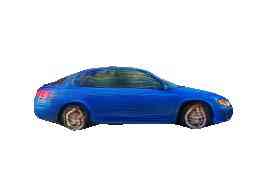}&
					\includegraphics[height=1.5cm]{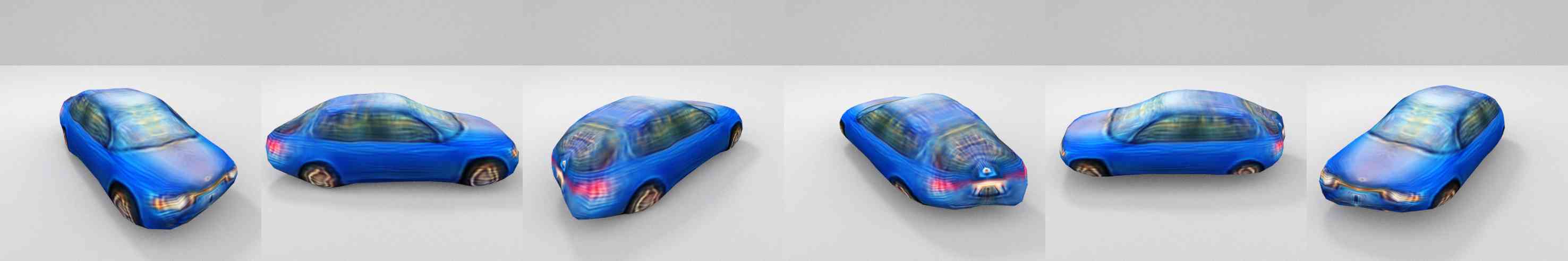}
					\\
					\rotatebox{90}{\,\,\,\,\,\,\,\,{\color{black}{\tiny Ours}}}&
					\includegraphics[width=2cm,height=1.5cm]{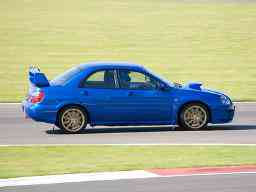}&
					\includegraphics[width=2cm,height=1.5cm]{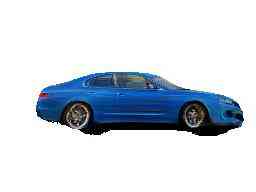}&
					\includegraphics[height=1.5cm]{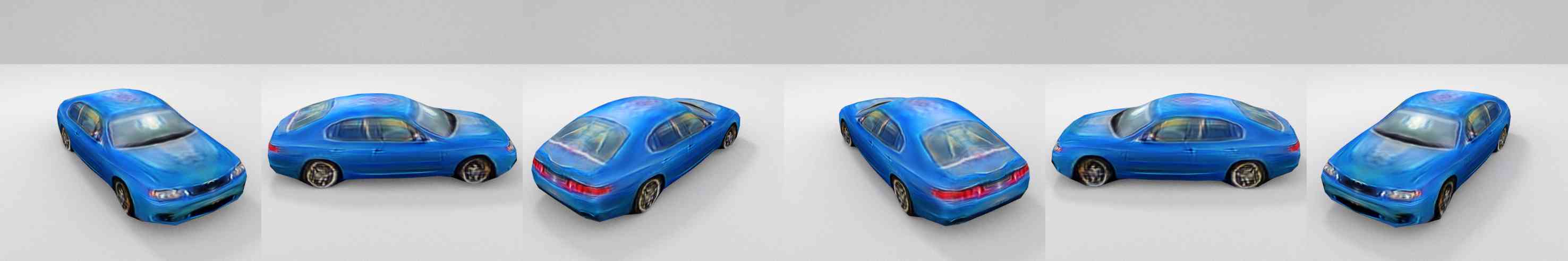}
					\\
					
					\rotatebox{90}{\,\,\,\, {\color{black}{\tiny Pascal3D}}}&
					\includegraphics[width=2cm,height=1.5cm]{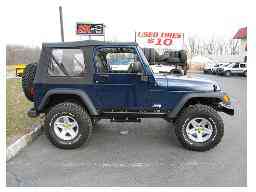}&
					\includegraphics[width=2cm,height=1.5cm]{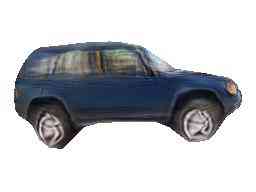}&
					\includegraphics[height=1.5cm]{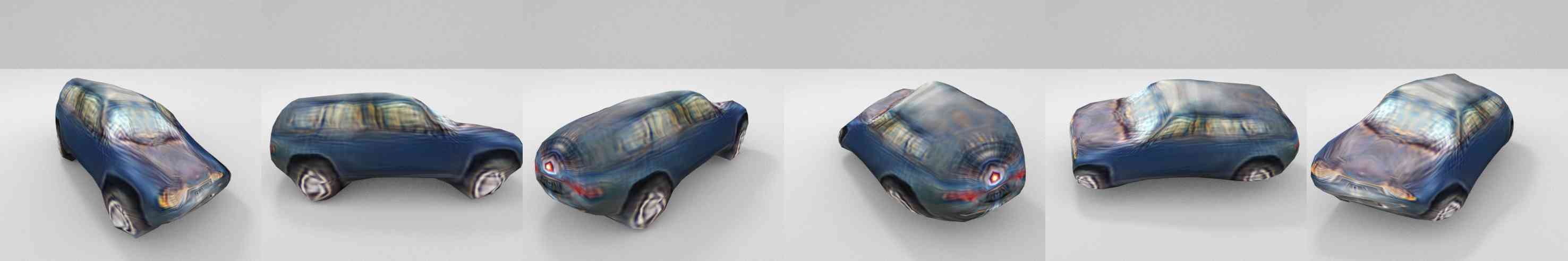}
					\\
					\rotatebox{90}{\,\,\,\,\,\,\,\,{\color{black}{\tiny Ours}}}&
					\includegraphics[width=2cm,height=1.5cm]{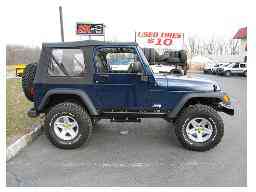}&
					\includegraphics[width=2cm,height=1.5cm]{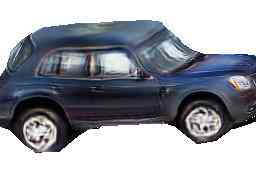}&
					\includegraphics[height=1.5cm]{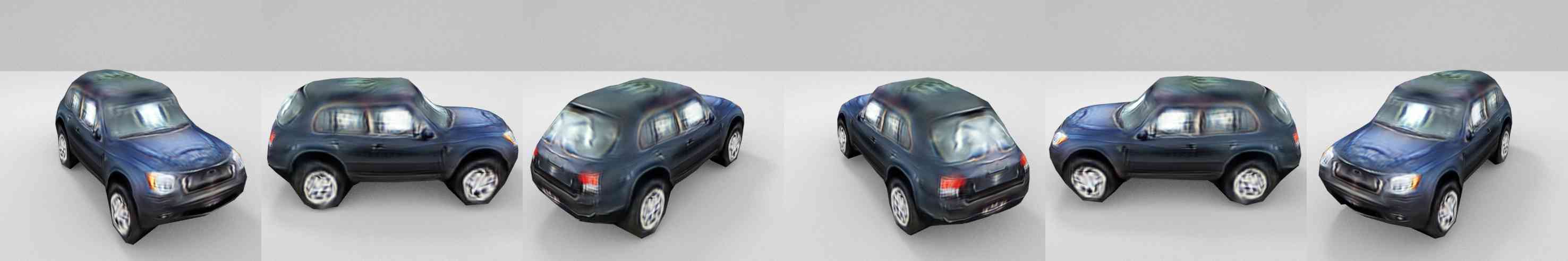}
					\\					
					
					\rotatebox{90}{\,\,\,\, {\color{black}{\tiny Pascal3D}}}&
					\includegraphics[width=2cm,height=1.5cm]{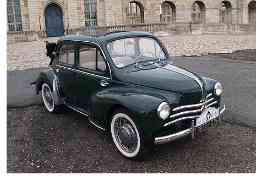}&
					\includegraphics[width=2cm,height=1.5cm]{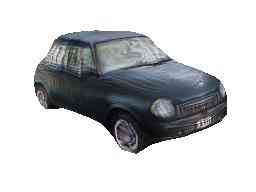}&
					\includegraphics[height=1.5cm]{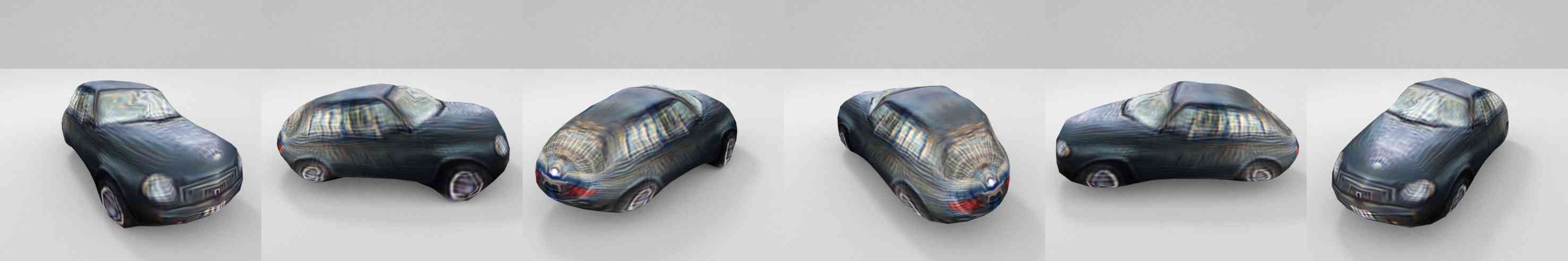}
					\\
					\rotatebox{90}{\,\,\,\,\,\,\,\,{\color{black}{\tiny Ours}}}&
					\includegraphics[width=2cm,height=1.5cm]{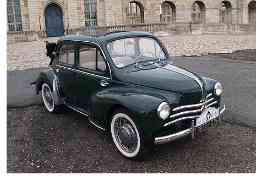}&
					\includegraphics[width=2cm,height=1.5cm]{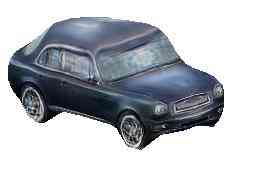}&
					\includegraphics[height=1.5cm]{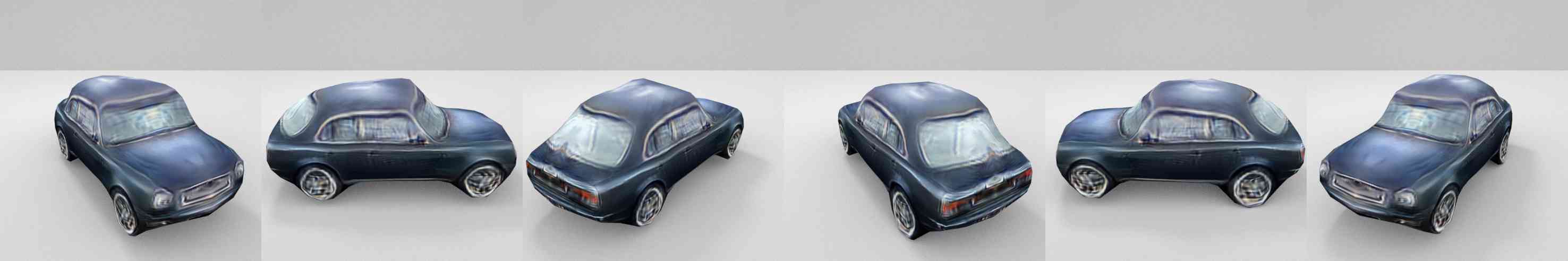}
					\\
					& {\footnotesize Input }& {\footnotesize Pred.}  & {\footnotesize Multiple Views}
					\hspace*{0pt}
				\end{tabular}
			\end{tabular}
		\end{center}
	}
	\caption{\footnotesize \textbf{Comparison on PASCAL3D imagery:} We compare PASCAL-model  with StyleGAN-model on PASCAL3D test set. While the predictions from both models are visually good in the corresponding image view, the prediction from StyleGAN-model have much better shapes and textures as observed in other views.}
	\label{fig:p3d}
\end{figure*}

\begin{figure*}[t]
	{
		\vspace*{0pt}
		\begin{center}
			\setlength{\tabcolsep}{1pt}
			\setlength{\fboxrule}{0pt}
			\hspace*{-0.25cm}
			\begin{tabular}{c}
				\begin{tabular}{ccc}
					\rotatebox{90}{\,\,\,\,\,\, {\color{black}{\tiny Pascal3D}}}&
					\includegraphics[height=0.14\linewidth]{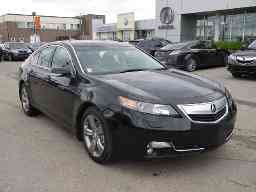}&
					\includegraphics[height=0.14\linewidth]{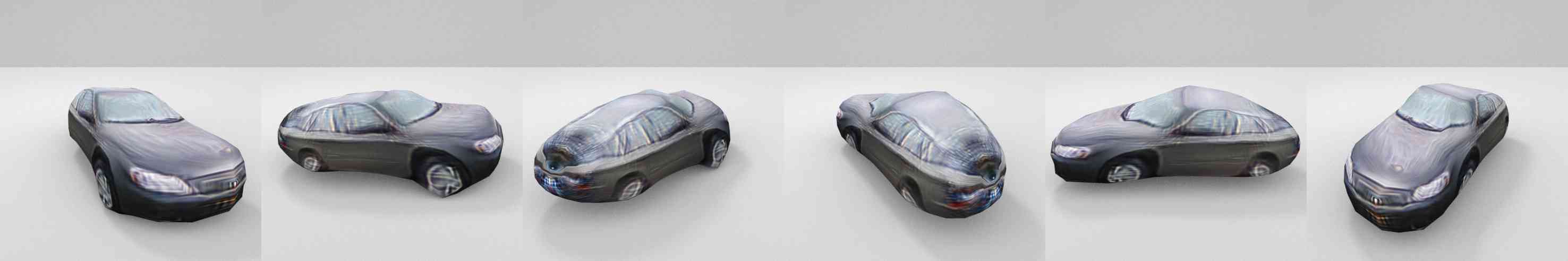}
					\\
					\rotatebox{90}{\,\,\,\,\,\,\,\,\,\, {\color{black}{\tiny Ours}}}&
					\includegraphics[height=0.14\linewidth]{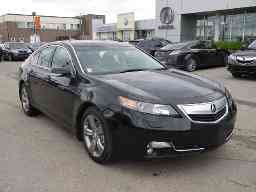}&
					\includegraphics[height=0.14\linewidth]{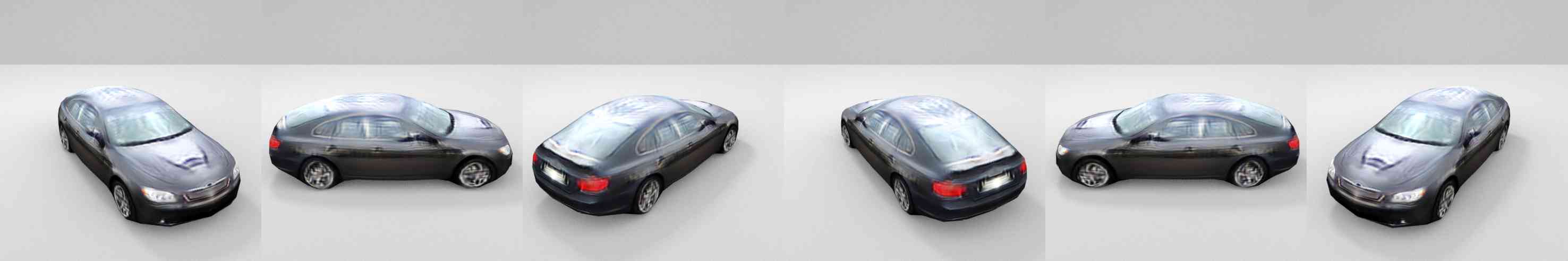}
					\\
					
					\rotatebox{90}{\,\,\,\,\,\, {\color{black}{\tiny Pascal3D}}}&
					\includegraphics[height=0.14\linewidth]{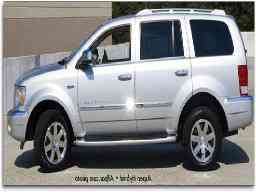}&
					\includegraphics[height=0.14\linewidth]{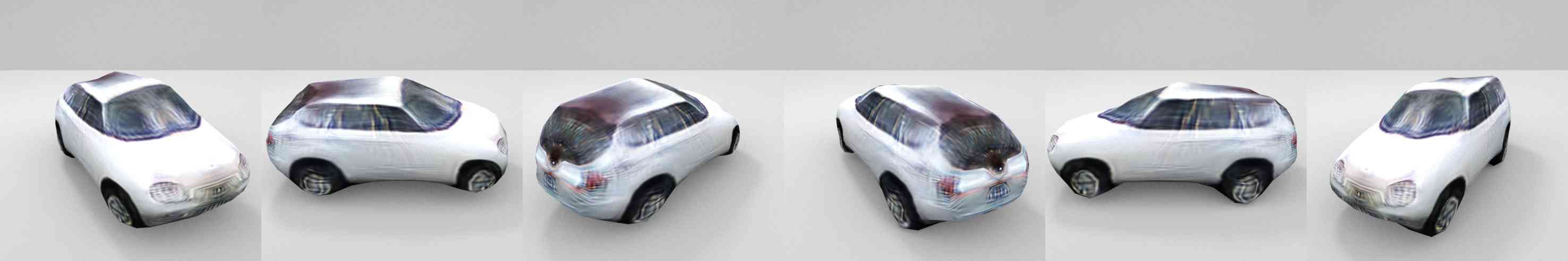}
					\\
					\rotatebox{90}{\,\,\,\,\,\,\,\,\,\, {\color{black}{\tiny Ours}}}&
					\includegraphics[height=0.14\linewidth]{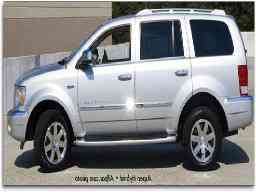}&
					\includegraphics[height=0.14\linewidth]{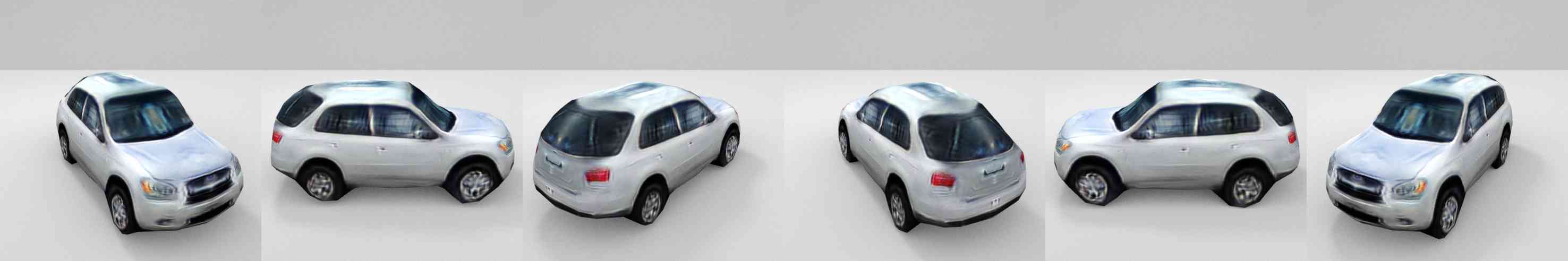}
					\\
					
					\rotatebox{90}{\,\,\,\,\,\, {\color{black}{\tiny Pascal3D}}}&
					\includegraphics[height=0.14\linewidth]{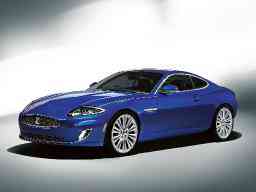}&
					\includegraphics[height=0.14\linewidth]{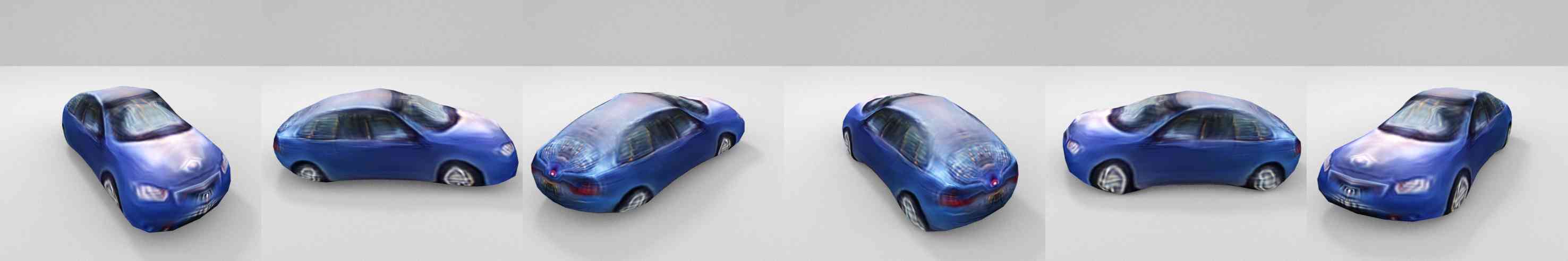}
					\\
					\rotatebox{90}{\,\,\,\,\,\,\,\,\,\, {\color{black}{\tiny Ours}}}&
					\includegraphics[height=0.14\linewidth]{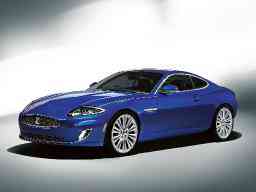}&
					\includegraphics[height=0.14\linewidth]{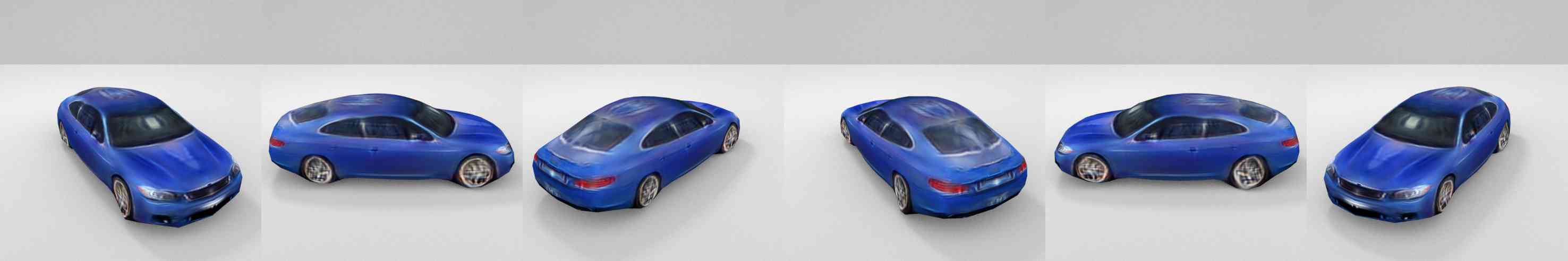}
					\\
					
					\rotatebox{90}{\,\,\,\,\,\, {\color{black}{\tiny Pascal3D}}}&
					\includegraphics[height=0.14\linewidth]{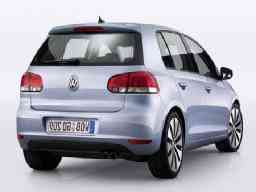}&
					\includegraphics[height=0.14\linewidth]{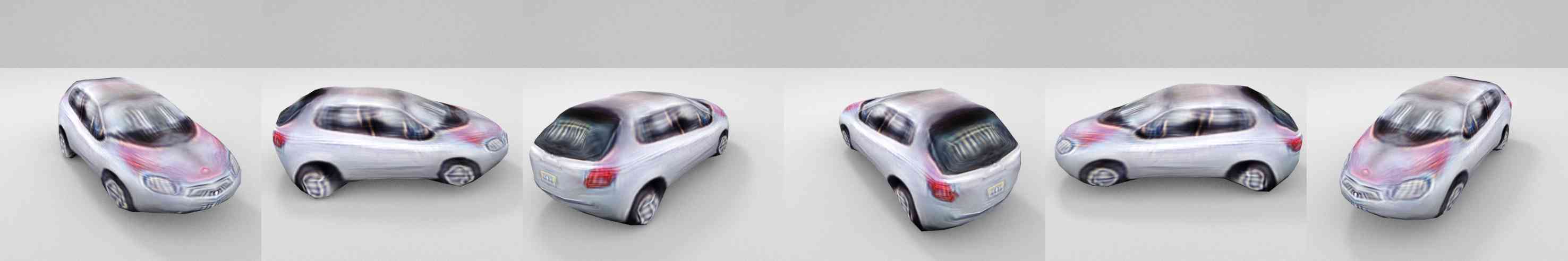}
					\\
					\rotatebox{90}{\,\,\,\,\,\,\,\,\,\, {\color{black}{\tiny Ours}}}&
					\includegraphics[height=0.14\linewidth]{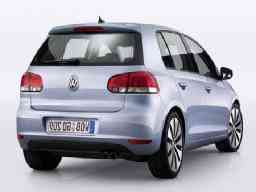}&
					\includegraphics[height=0.14\linewidth]{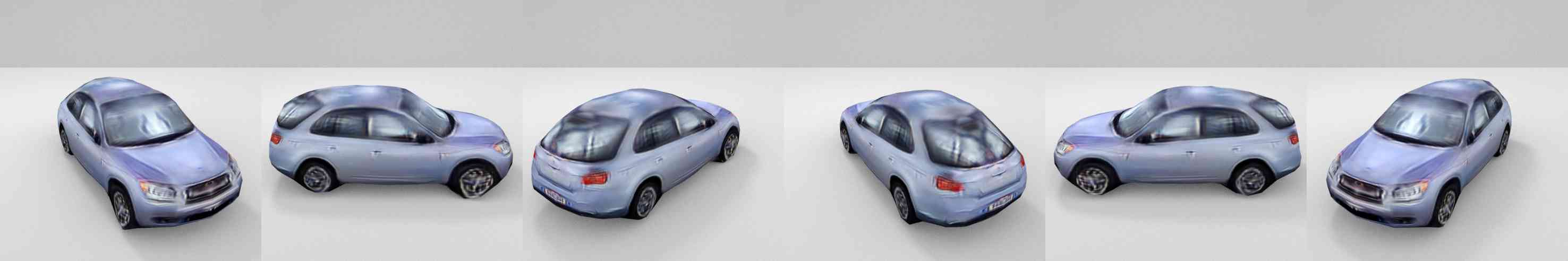}
					\\
					
					\rotatebox{90}{\,\,\,\,\,\, {\color{black}{\tiny Pascal3D}}}&
					\includegraphics[height=0.14\linewidth]{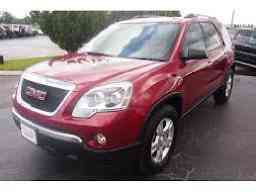}&
					\includegraphics[height=0.14\linewidth]{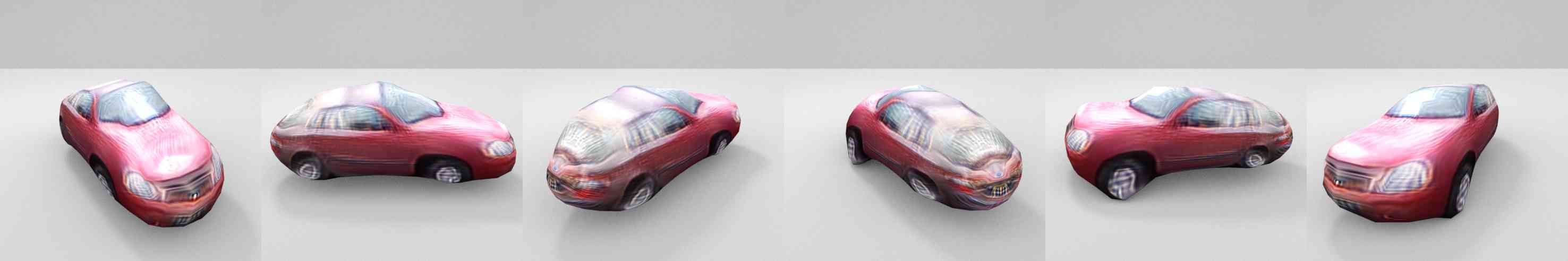}
					\\
					\\
					\rotatebox{90}{\,\,\,\,\,\,\,\,\,\, {\color{black}{\tiny Ours}}}&
					\includegraphics[height=0.14\linewidth]{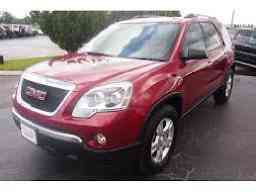}&
					\includegraphics[height=0.14\linewidth]{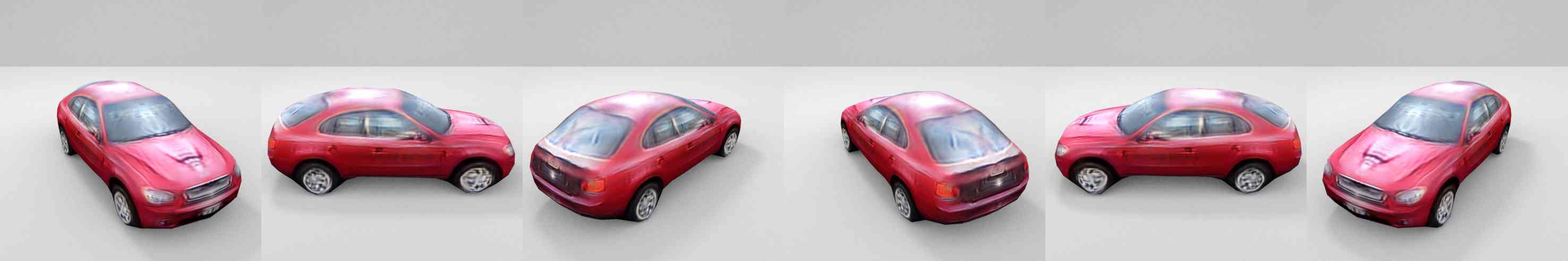}
					\\
					& {\scriptsize Input } & {\scriptsize Pred. Multiple Views}
					\hspace*{0pt}
				\end{tabular}
			\end{tabular}
		\end{center}
	}
	\caption{\footnotesize \textbf{Comparison on Images from the Web:} We compare the PASCAL-model with our StyleGAN-model on images downloaded from the web. While the predictions from both models are visually good in the corresponding image view, the prediction from StyleGAN-model have much better shapes and textures as observed in other views.}
	\label{fig:web}
\end{figure*}

\begin{figure*}[h]
	{
		\vspace*{0pt}
		\begin{center}
			\setlength{\tabcolsep}{1pt}
			\setlength{\fboxrule}{0pt}
			\hspace*{-0.2cm}
			\begin{tabular}{c}
				\begin{tabular}{ccc}
					\includegraphics[height=0.12\linewidth]{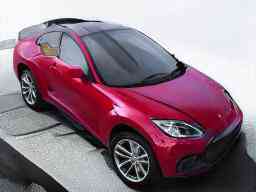}&
					\includegraphics[height=0.12\linewidth]{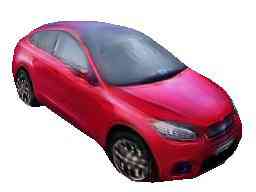}&
					\includegraphics[height=0.12\linewidth]{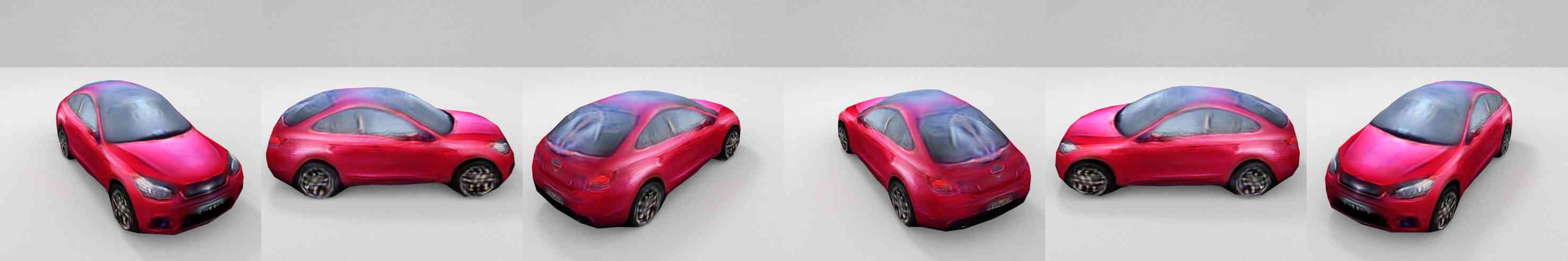}
					
					\\
					\includegraphics[height=0.12\linewidth]{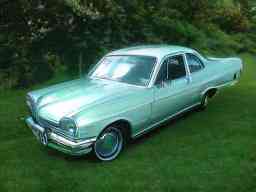}&
					\includegraphics[height=0.12\linewidth]{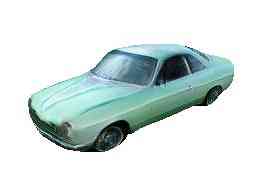}&
					\includegraphics[height=0.12\linewidth]{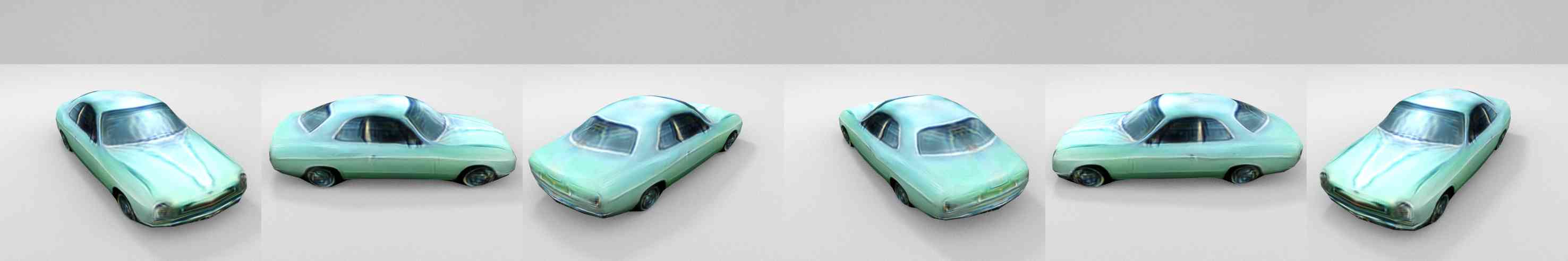}
					
					\\
					\includegraphics[height=0.12\linewidth]{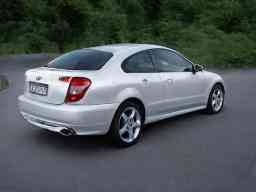}&
					\includegraphics[height=0.12\linewidth]{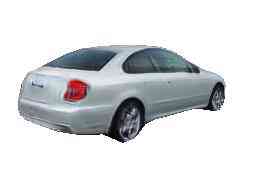}&
					\includegraphics[height=0.12\linewidth]{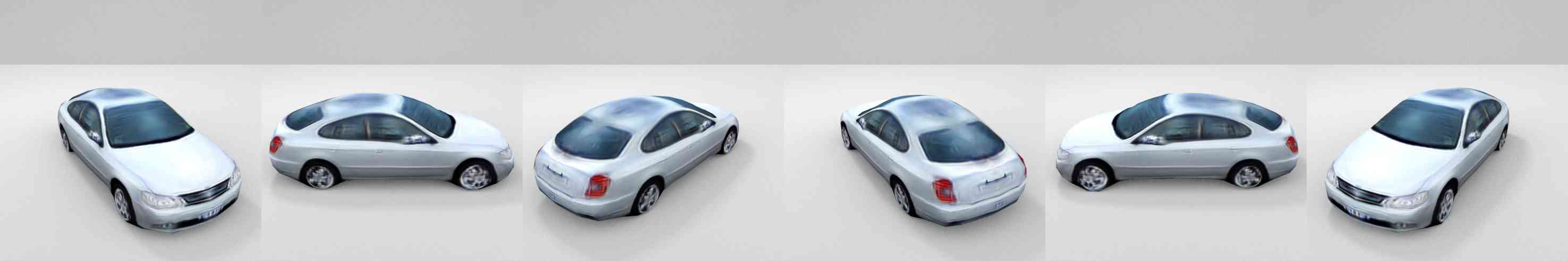}
					
					\\
					\includegraphics[height=0.12\linewidth]{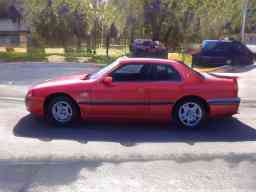}&
					\includegraphics[height=0.12\linewidth]{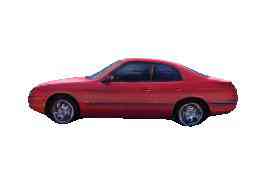}&
					\includegraphics[height=0.12\linewidth]{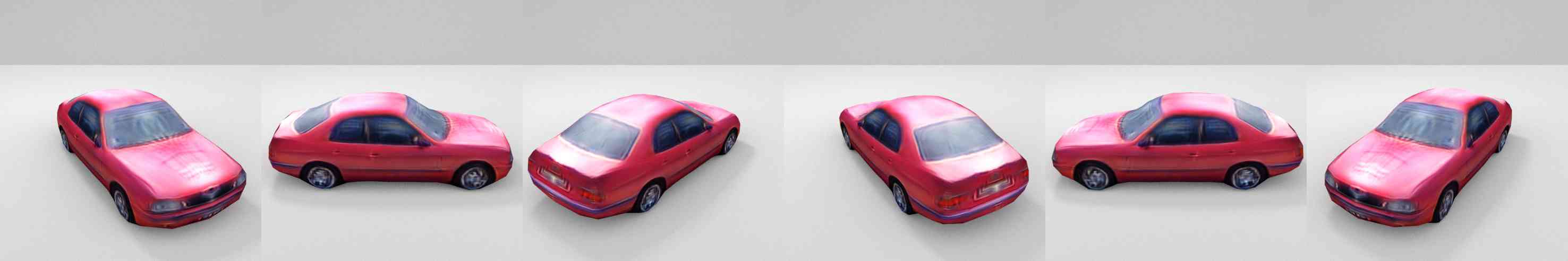}
					\\		
					
					\includegraphics[height=0.12\linewidth]{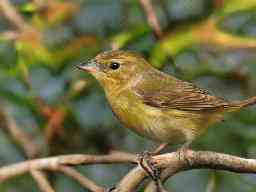}&
					\includegraphics[height=0.12\linewidth]{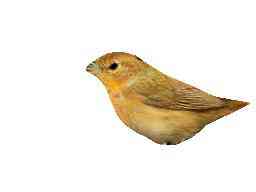}&
					\includegraphics[height=0.12\linewidth]{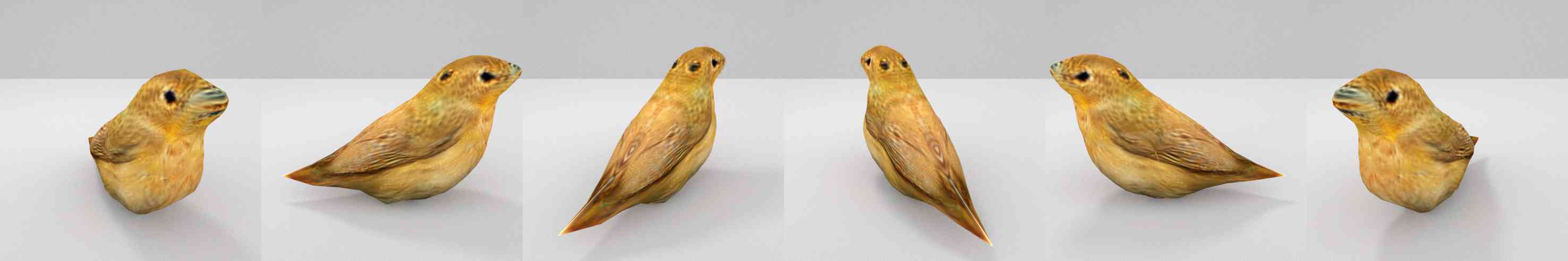}
					\\								
					
					\includegraphics[height=0.12\linewidth]{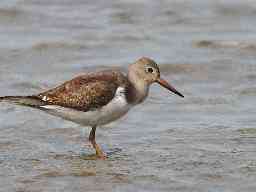}&
					\includegraphics[height=0.12\linewidth]{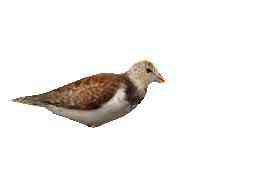}&
					\includegraphics[height=0.12\linewidth]{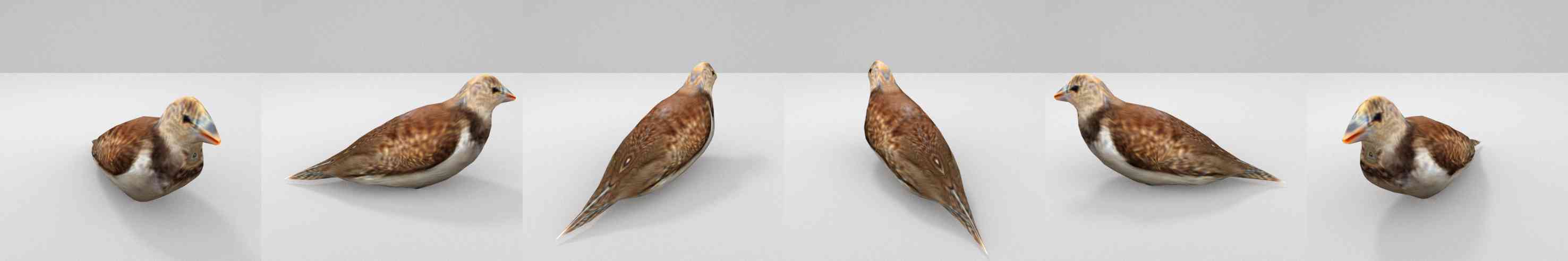}
					\\		
											
					\includegraphics[height=0.12\linewidth]{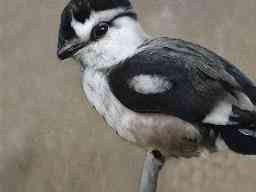}&
					\includegraphics[height=0.12\linewidth]{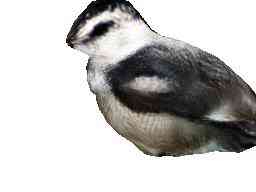}&
					\includegraphics[height=0.12\linewidth]{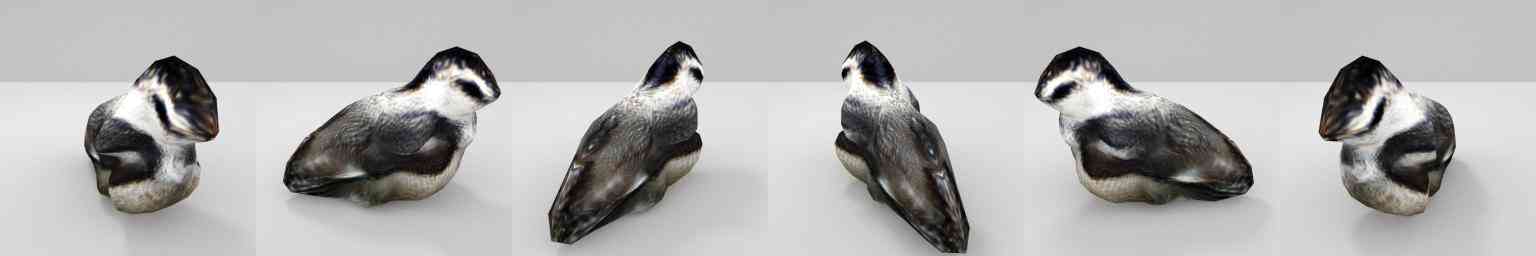}
					\\								
					
					\includegraphics[height=0.12\linewidth]{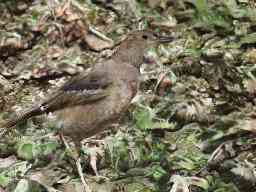}&
					\includegraphics[height=0.12\linewidth]{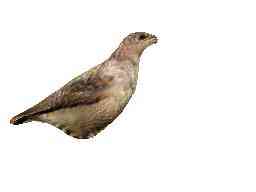}&
					\includegraphics[height=0.12\linewidth]{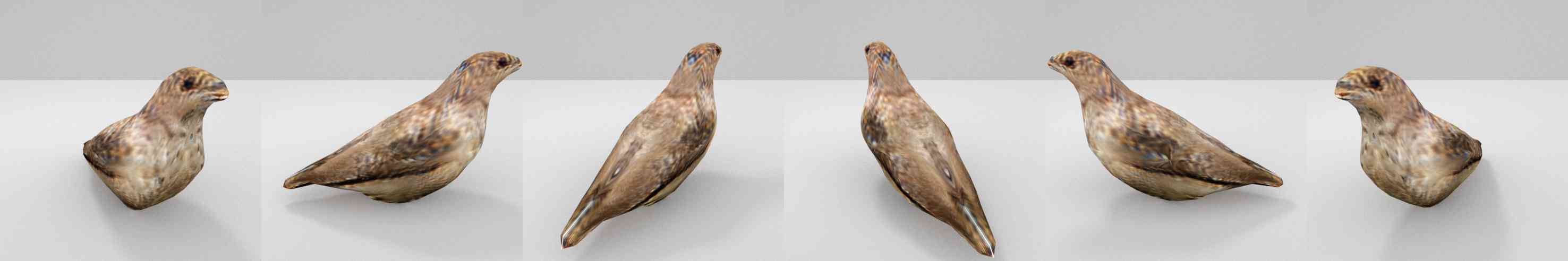}
					\\			
								
					\includegraphics[height=0.12\linewidth]{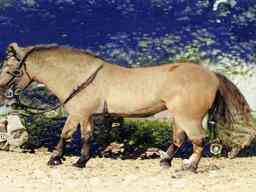}&
					\includegraphics[height=0.12\linewidth]{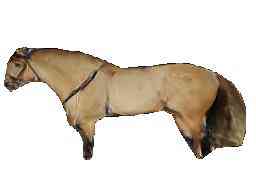}&
					\includegraphics[height=0.12\linewidth]{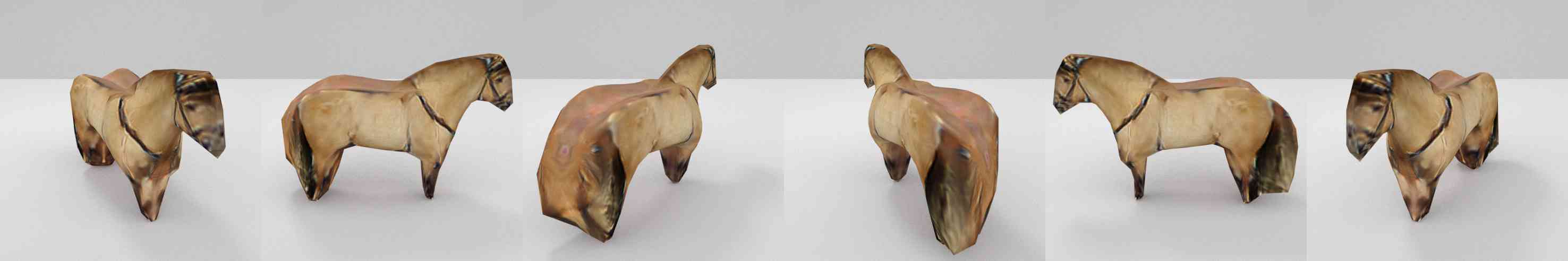}
					\\			
					
					\includegraphics[height=0.12\linewidth]{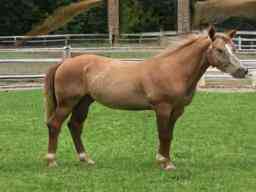}&
					\includegraphics[height=0.12\linewidth]{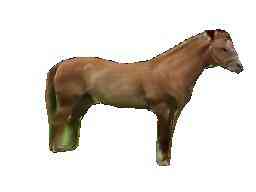}&
					\includegraphics[height=0.12\linewidth]{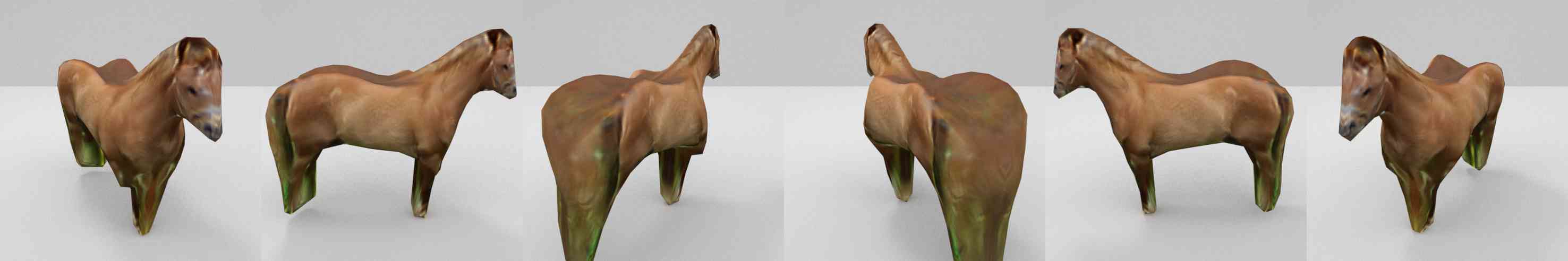}
					\\									
					
					\includegraphics[height=0.12\linewidth]{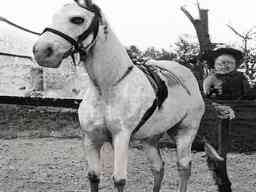}&
					\includegraphics[height=0.12\linewidth]{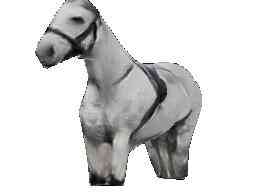}&
					\includegraphics[height=0.12\linewidth]{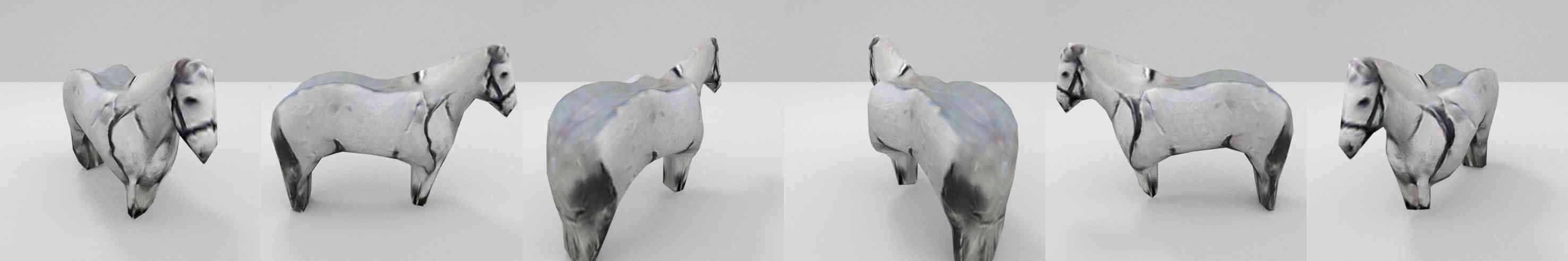}
					\\

					\includegraphics[height=0.12\linewidth]{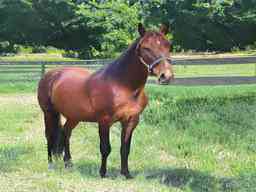}&
					\includegraphics[height=0.12\linewidth]{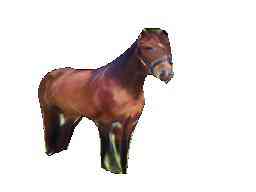}&
					\includegraphics[height=0.12\linewidth]{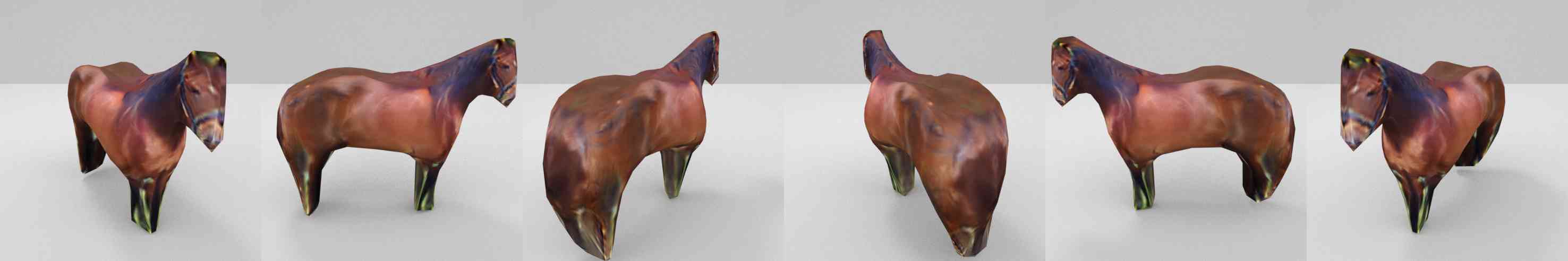}
					\\

					{\footnotesize Input} & {\footnotesize Pred.} 	& {\footnotesize Multiple Views for the predicted shape and texture} 
				\end{tabular}
			\end{tabular}
		\end{center}
		\vspace*{-4mm}
	}
	\caption{\label{fig:3d_for_otherclass} \footnotesize \textbf{3D Reconstruction Results for Car, Horse and Bird Classes}: We show car, horse and bird examples tested on the images from the StyleGAN dataset test sets. Notice that the model struggles a little in reconstructing the top of the back of the horse, since such views are lacking in training. } 
	\vspace*{-5mm}
\end{figure*}

We here present additional 3D prediction results and compare our model, which is trained on our StyleGAN generated dataset (StyleGAN-model),  with the one trained on the Pascal 3D dataset~\citep{xiang_wacv14} (PASCAL-model). We qualitatively compare two models on the Pascal3D test set in Fig.~\ref{fig:p3d} and web imagery in Fig.~\ref{fig:web}. Our StyleGAN-model produces better shape and texture predictions in all  the testing datasets, which is particularly noticeable when looking at  different rendered views of the prediction. We also present additional 3D prediction results on horses and birds in Fig.~\ref{fig:3d_for_otherclass}.

\clearpage

\clearpage

\clearpage

\vspace{-2mm}
\section{User Study}
\label{sec:user_study}

\begin{figure*}[t]
 \vspace{-5mm}
	{
		\begin{minipage}{0.64\linewidth}
			\includegraphics[width=1\textwidth,trim=10 0 100 0,clip]{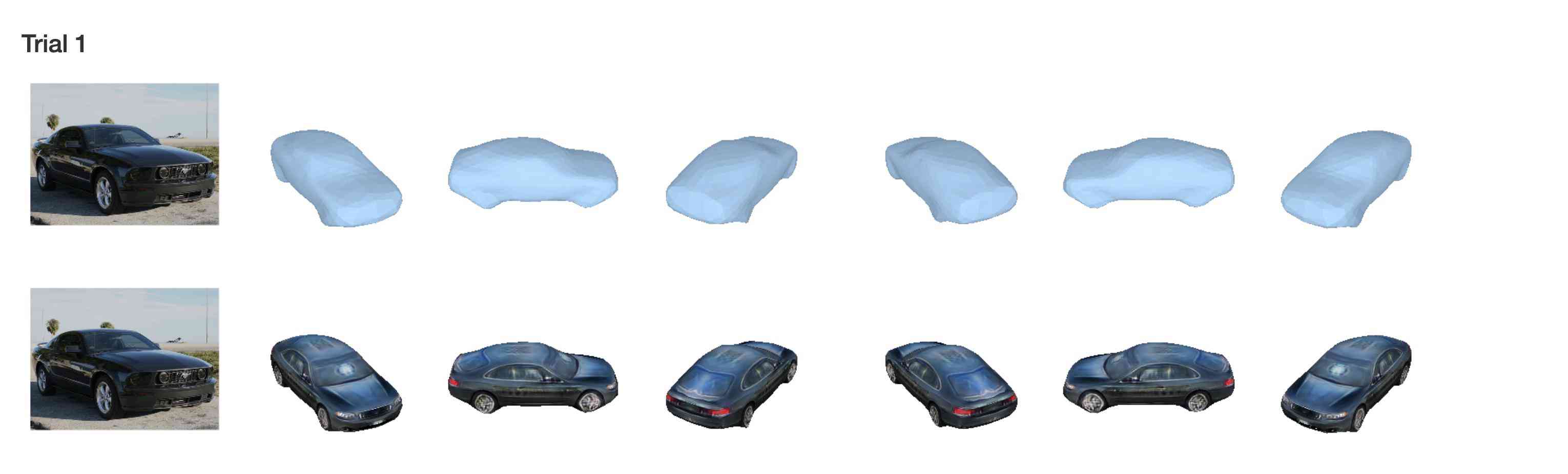}
			\includegraphics[width=1\textwidth,trim=10 0 100 0,clip]{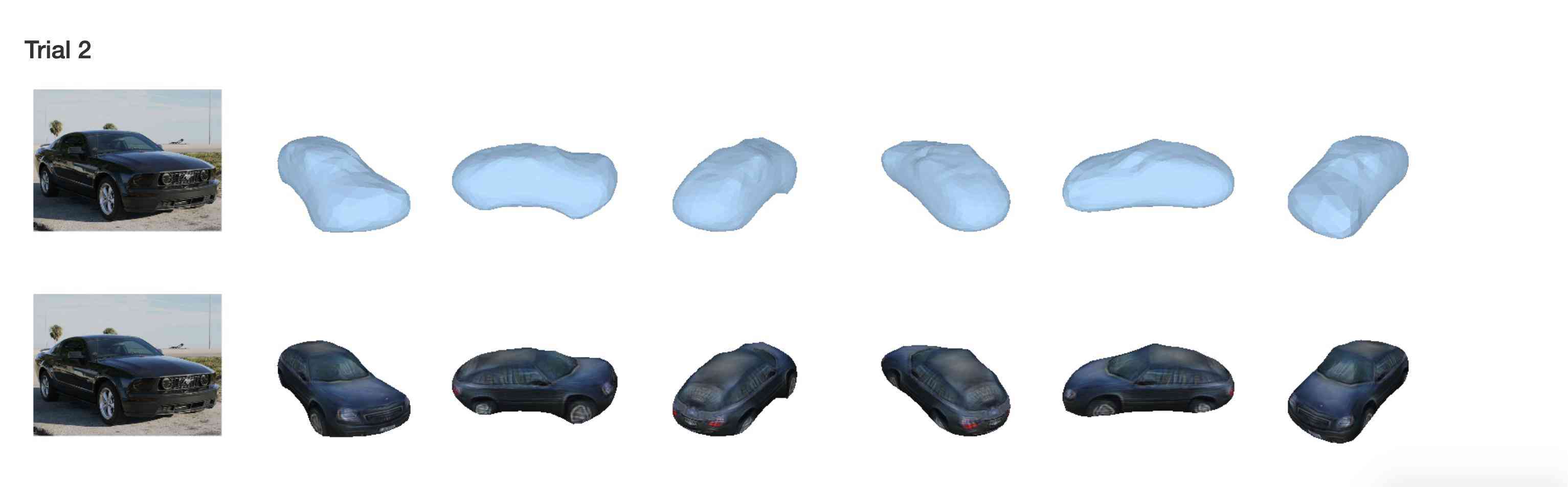}
		\end{minipage}
		\begin{minipage}{0.355\linewidth}	
			\includegraphics[width=1\textwidth,trim=15 0 10 0,clip]{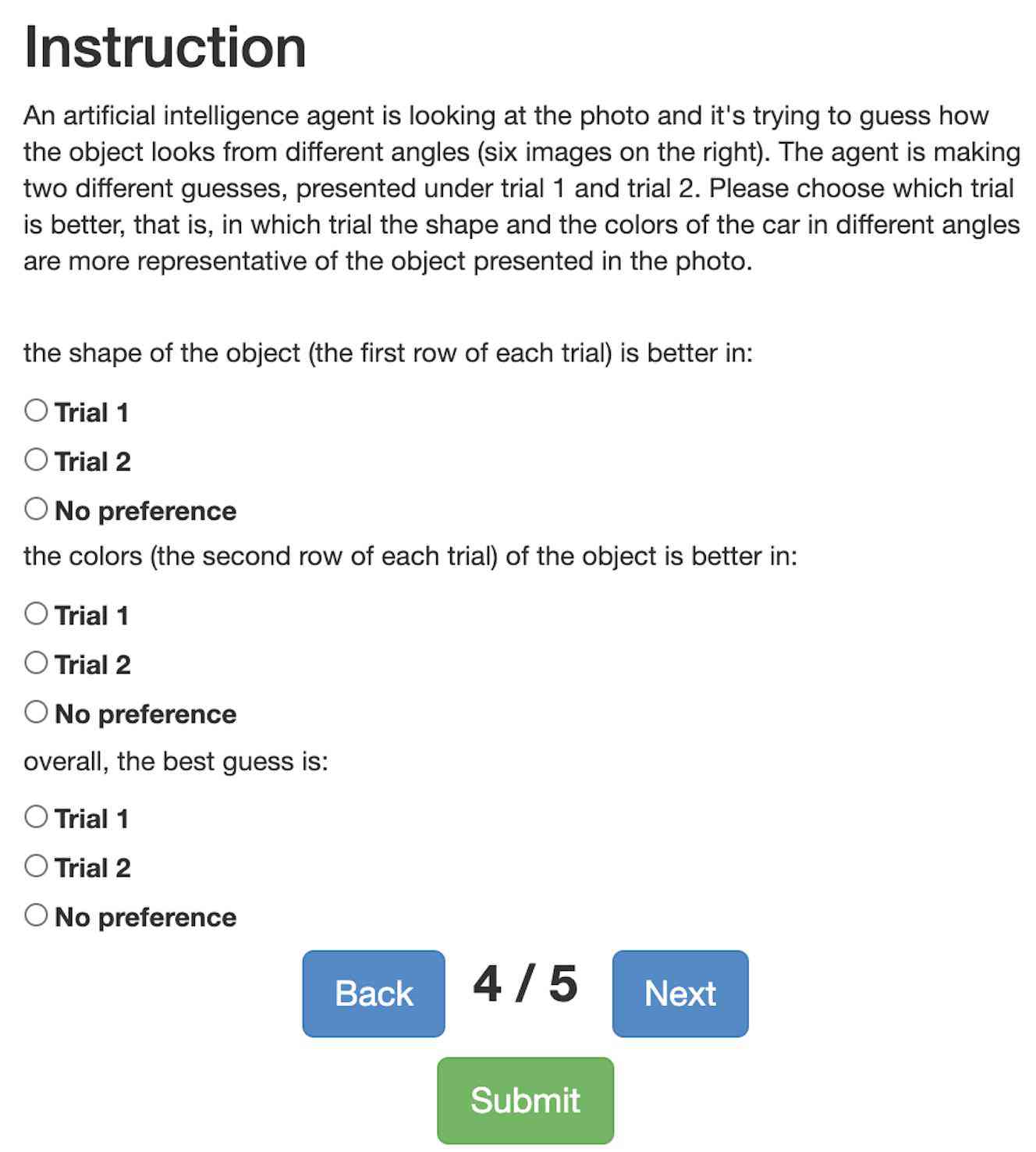}
		\end{minipage}
	}
	\vspace{-4mm}
	\caption{\footnotesize \textbf{User Study Interface (AMT)}:  Predictions are rendered in 6 views and we ask users to choose the result with a more realistic shape and texture that is relevant to the input object. We compare both the baseline (trained on Pascal3D dataset) and ours (trained on StyleGAN dataset). We randomize their order in each HIT.}
	\label{fig:interface}
\end{figure*}

We provide user study details in this section. We implement our user interface, visualized in in Fig.~\ref{fig:interface},  on Amazon Mechanical Turk. We show the input image and predictions rendered in 6 views such that users can better judge the quality of 3D reconstruction. We show results for both, our inverse graphics network (trained on the StyleGAN dataset) and the one trained on the Pascal3D dataset. We show shape reconstruction and textured models separately, such that users can judge the quality of both, shape and texture,  more easily. 
We randomize the order of ours vs baseline in each HIT to avoid any bias. We ask users to choose results that produce more realistic and representative shape, texture and overall quality with respect to the input image. We separate judgement of quality into these three categories to disentangle effects of 3D reconstruction from texture prediction. We also provide ``no preference" options in case of ties.  Our instructions emphasize that more ``representative" results of the input should be selected, to avoid users being biased by good looking predictions that are not consistent with the input (e.g., such as in the case of overfit networks). 

We evaluate the two networks on all 220 images from the Pascal3D test set (which are ``in-domain" for the Pascal3D-trained network).  For each image we ask three users to perform evaluation, which results in 660 votes in total. We report the average of all votes as our final metric. We further report annotator agreement analysis in Table~\ref{tab:agreement}. For shape, texture, and overall evaluation, there are 88.2\%, 89.2\%, and 87.2\% cases where at least two out of three users choose the same option.

{}
\begin{table*} [htb!]  
	\vspace{0mm}
	\begin{center}
		\addtolength{\tabcolsep}{-1.8pt}
		\small
		\centering

			\begin{tabular}{cc}

\begin{tabular}{lccc}
	\toprule
	 & Overall & Shape & Texture \\
	\midrule
	Ours  & \textbf{57.5\%}  & \textbf{61.6\%}  & \textbf{56.3\%}  \\
	Pascal3D-model & 25.9\% &26.4\% & 32.8\% \\
	No Preference & 16.6\% & 11.9\% & 10.8\%  \\
	\bottomrule
\end{tabular}
&\hspace{5mm}
\begin{tabular}{lccc}
	\toprule
	 & Overall & Shape & Texture \\
	\midrule
	All Agree  & \textbf{26.1\%}  & \textbf{29.6\%}  & \textbf{27.1\%}  \\
	Two Agree & 61.1\% &58.6\% & 62.1\% \\
	No Agreement & 12.8\% & 11.8\% & 10.8\%  \\
	\bottomrule
\end{tabular}

\\
\\[-2mm]
 (a) 3D Quality Study	& (b) Annotator Agreement
\end{tabular}	

	\end{center}
	\vspace{-4mm}
\caption{\footnotesize User study results: \textbf{(a):} Quality of 3D estimation (shape, texture and overall).  
	\textbf{(b):} Annotators agreement analysis. ``No agreement" stands for the case where all three annotators choose different options.
	}
	\vspace{-4pt}
	\label{tab:agreement}
\end{table*}

\section{{\ours} Disentanglement}
\label{sec:stylegan_disentangle}

\begin{figure*}[t]
\vspace{-7mm}
	{
		\vspace*{0pt}
		\begin{center}
			\setlength{\tabcolsep}{1pt}
			\setlength{\fboxrule}{0pt}
			\hspace*{-0.25cm}
			\begin{tabular}{c}
				\begin{tabular}{cc}
					\\
					\rotatebox{90}{\ \ \ \ \ \  {\color{black}\tiny{DIB-R}}}&	
					\includegraphics[height=1.7cm]{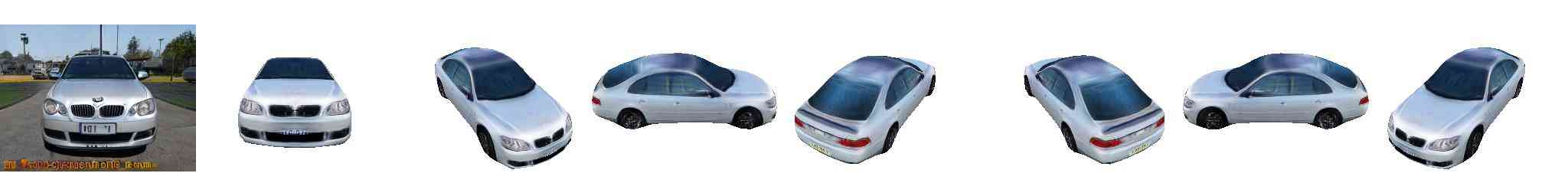}
					\\
					\rotatebox{90}{\ \ \  {\color{black}\tiny{StyleGAN-R}}}&
					\includegraphics[height=1.7cm]{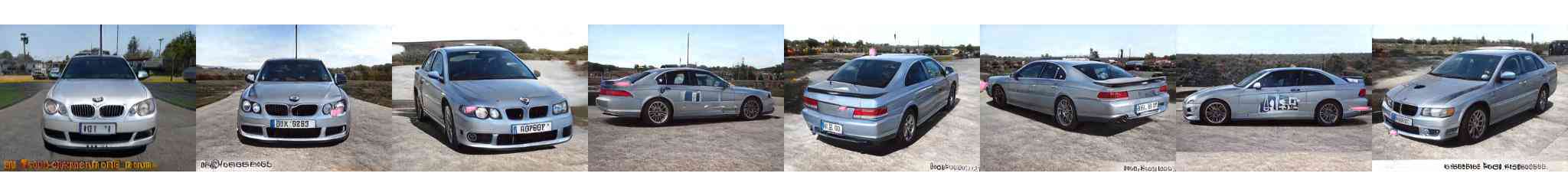}
					\\
					\rotatebox{90}{\ \ \ \ \  {\color{black}\tiny{StyleGAN}}}&
					\includegraphics[height=1.7cm]{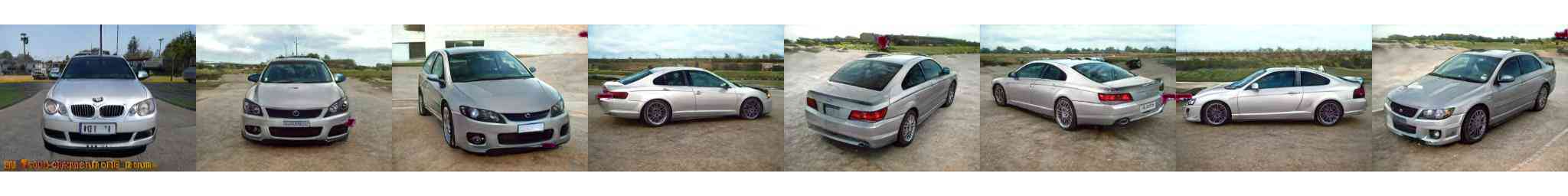}
					\\
					\\
					\rotatebox{90}{\ \ \ \ \ \ {\color{black}\tiny{DIB-R}}}&	
					\includegraphics[height=1.7cm]{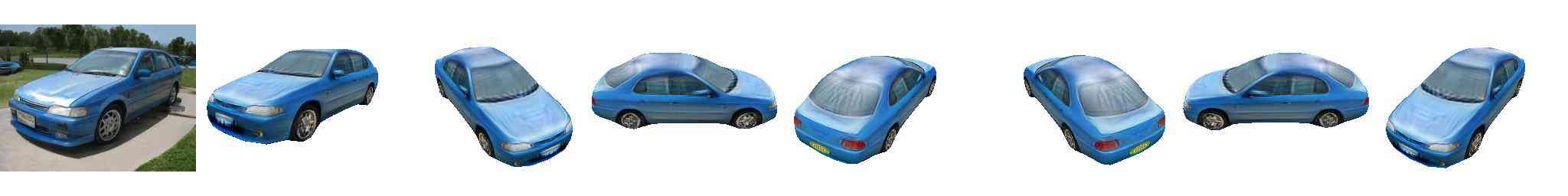}
					\\
					\rotatebox{90}{\ \ \ {\color{black}\tiny{StyleGAN-R}}}&
					\includegraphics[height=1.7cm]{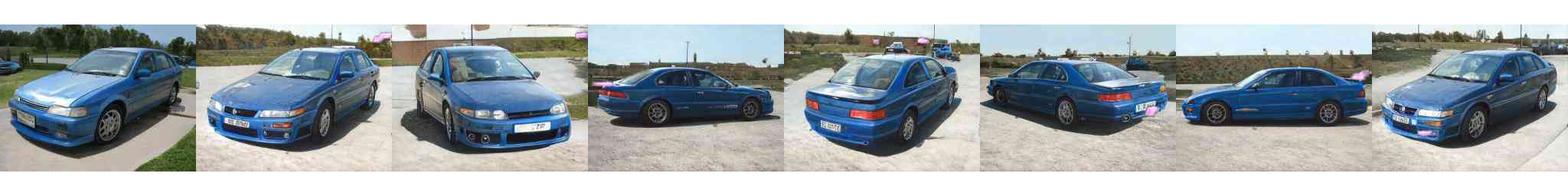}
					\\
					\rotatebox{90}{\ \ \ \ \ {\color{black}\tiny{StyleGAN}}}&
					\includegraphics[height=1.7cm]{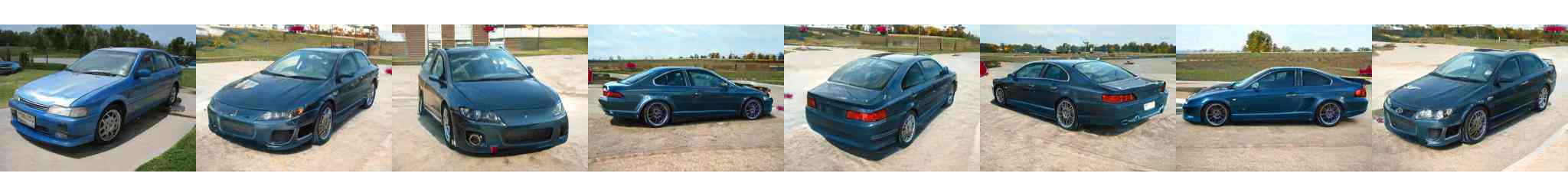}
					\\[-2.5mm]
					& \,\,\,\,\,\,\,\, \footnotesize Input \,\,\,\,\,\,\,\,\,\,\,\,\,\,\,\,\,\, \footnotesize Pred. \,\,\,\,\,\,\,\,\,\,\,\,\,\,\,\,\,\,\,\,\,\,\,\,\,\,\,\,\,\,\,\,\,\,\,\,\,\,\,\,\,\,\,\,\,\,\,\,\,\,\,\,\,\,\,\,\,\,\,\,\,\,\,\,\,\,\,\,\,\,\,\,\,\,\,\,\,\,\,\,\,\,\,\,\,\,\,\,\,\,\,\, \footnotesize Multiple Views \,\,\,\,\,\,\,\,\,\,\,\,\,\,\,\,\,\,\,\,\,\,\,\,\,\,\,\,\,\,\,\,\,\,\,\,\,\,\,\,\,\,\,\,\,\,\,\,\,\,\,\,\,\,\,\,\,\,\,\,\,\,\,\,\,\,\,\,\
					\hspace*{0pt}
				\end{tabular}
			\end{tabular}
		\end{center}
		\vspace{-4mm}
	}
	\caption{\footnotesize \textbf{Dual Rendering:} Given the input image, we show the DIB-R-rendered predictions in rows (1, 4) and {\ours}'s results in rows (2, 5). We further shows the neural rendering results from the original StyleGAN model, where we only learn the mapping network but keep the StyleGAN weights fixed. Clearly, after fine-tuning, {\ours} produces more consistent results.}
	\label{fig:compstylegan}
	\vspace*{-0.0cm}
\end{figure*}

Given an input image, we infer 3D properties of an object (shape, texture, background) using our inverse graphics network, but can also map these properties back to the latent code and use our {\ours} to synthesize a new image. We show the results in Fig.~\ref{fig:compstylegan}. Similar to Fig.~\ref{fig:camera_controlll} in the main paper, we show DIB-R-rendered predictions and neural rendering {\ours}'s predictions, and manipulate their viewpoints in rows (1, 4) and (2, 5). We further show ``neural rendering" results from the original StyleGAN in row (3, 6), where we only learn the mapping network but keep the StyleGAN's weights fixed. We find that fine-tuning is necessary and {\ours} produces more consistent shape, texture and background.

\begin{figure*}[t]
	{
		\vspace*{0pt}
		\begin{center}
			\setlength{\tabcolsep}{1pt}
			\setlength{\fboxrule}{0pt}
			\hspace*{-0.25cm}
			\begin{tabular}{c}
				\begin{tabular}{cccccc}
					\includegraphics[width=0.16\textwidth]{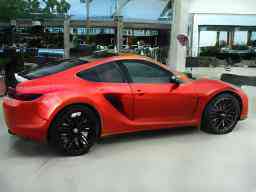}&
					\includegraphics[width=0.16\textwidth]{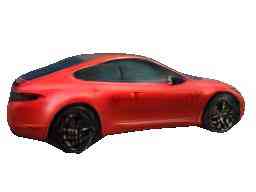}&
					\includegraphics[width=0.16\textwidth]{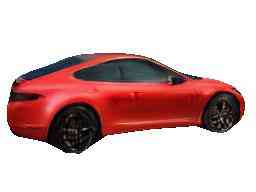}&
					\includegraphics[width=0.16\textwidth]{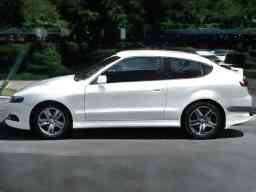}&
					\includegraphics[width=0.16\textwidth]{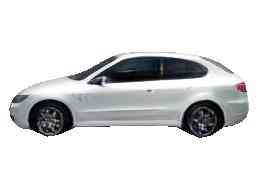}&
					\includegraphics[width=0.16\textwidth]{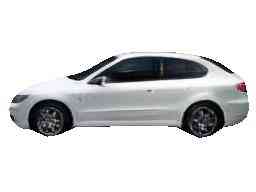}
					\\
					\includegraphics[width=0.16\textwidth]{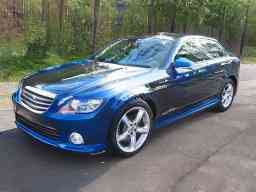}&
					\includegraphics[width=0.16\textwidth]{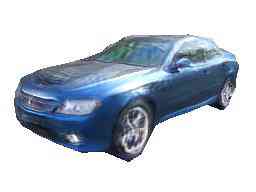}&
					\includegraphics[width=0.16\textwidth]{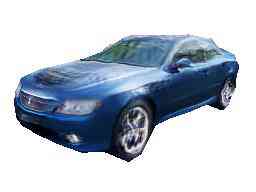}&
					\includegraphics[width=0.16\textwidth]{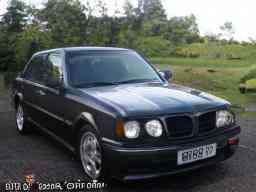}&
					\includegraphics[width=0.16\textwidth]{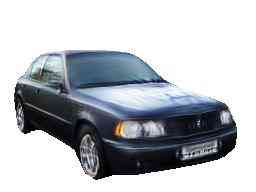}&
					\includegraphics[width=0.16\textwidth]{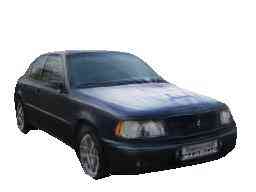}
					\\
					{\scriptsize Input } & {\scriptsize w. Light } & {\scriptsize w.o. Light} & {\scriptsize Input } & {\scriptsize w. Light } & {\scriptsize w.o. Light}
					\hspace*{0pt}
				\end{tabular}
			\end{tabular}
		\end{center}
		\vspace{-4mm}
	}
	\caption{\footnotesize \textbf{Light Prediction:} Given the input image, we show rendering (using the OpenGL renderer used in DIB-R) results with light (columns 2, 5) and results with just textures (columns 3, 6). We find that the two results are quite similar, which indicates that we did not learn a good predictor for lighting. Moreover, we find that higher order lighting, such as reflection, high-specular light are merged into texture, as shown in the second row. We aim to resolve this limitation in future work.}
	\label{fig:light}
	\vspace*{-0.0cm}
\end{figure*}

\begin{figure*}[t]
	{
		\vspace*{-0mm}
		\begin{center}
			\setlength{\tabcolsep}{1pt}
			\setlength{\fboxrule}{0pt}
			\hspace*{0pt}
			\begin{tabular}{c}
				\begin{tabular}{ccccccc}
					\footnotesize Input & \multicolumn{2}{c}{\footnotesize \ours} & \multicolumn{4}{c}{\footnotesize Manipulate Scales} 
					\\
					\includegraphics[width=.14\textwidth,height=.10\textwidth]{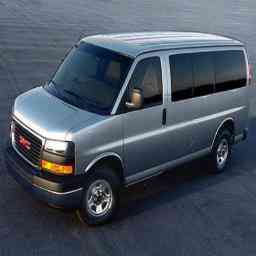}&
					\includegraphics[width=.14\textwidth,height=.10\textwidth]{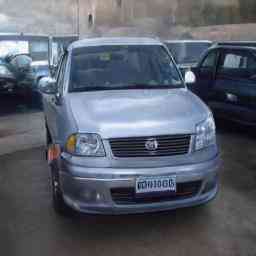}&
					\includegraphics[width=.14\textwidth,height=.10\textwidth]{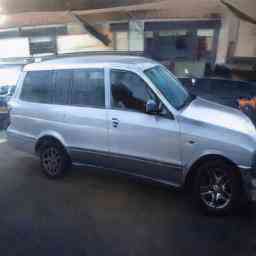}&
					\includegraphics[width=.14\textwidth,height=.10\textwidth]{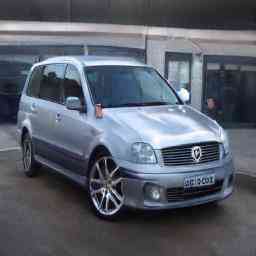}&
					\includegraphics[width=.14\textwidth,height=.10\textwidth]{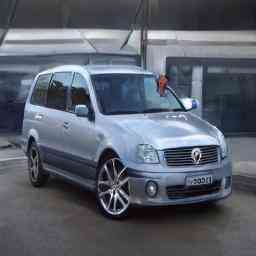}&
					\includegraphics[width=.14\textwidth,height=.10\textwidth]{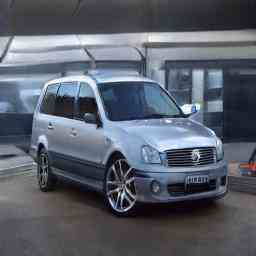}&
					\includegraphics[width=.14\textwidth,height=.10\textwidth]{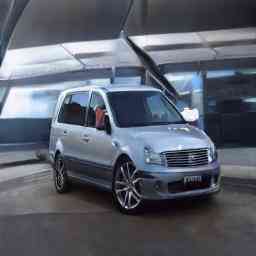}
					\\
					\includegraphics[width=.14\textwidth,height=.10\textwidth]{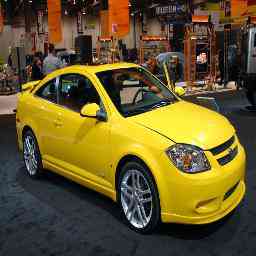}&
					\includegraphics[width=.14\textwidth,height=.10\textwidth]{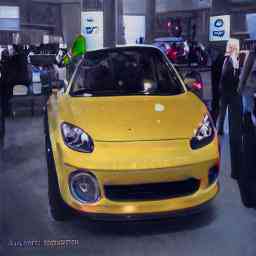}&
					\includegraphics[width=.14\textwidth,height=.10\textwidth]{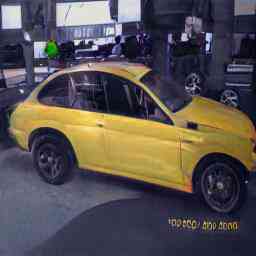}&
					\includegraphics[width=.14\textwidth,height=.10\textwidth]{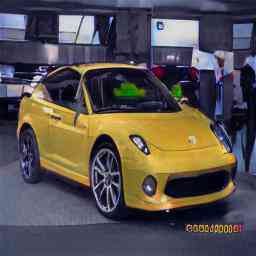}&
					\includegraphics[width=.14\textwidth,height=.10\textwidth]{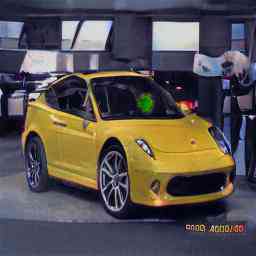}&
					\includegraphics[width=.14\textwidth,height=.10\textwidth]{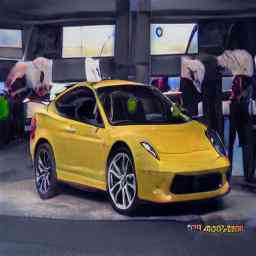}&
					\includegraphics[width=.14\textwidth,height=.10\textwidth]{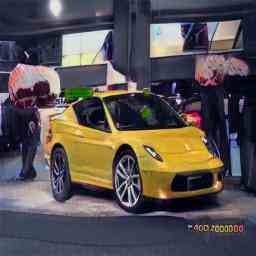}
					\\
					\includegraphics[width=.14\textwidth,height=.10\textwidth]{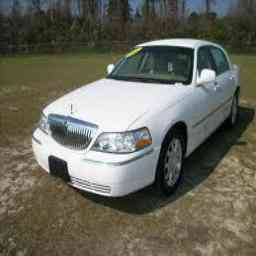}&
					\includegraphics[width=.14\textwidth,height=.10\textwidth]{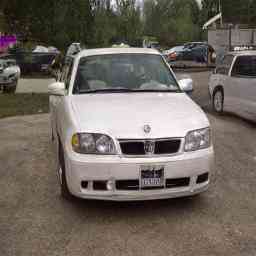}&
					\includegraphics[width=.14\textwidth,height=.10\textwidth]{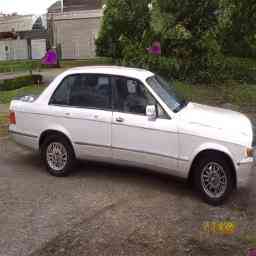}&
					\includegraphics[width=.14\textwidth,height=.10\textwidth]{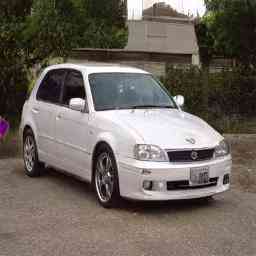}&
					\includegraphics[width=.14\textwidth,height=.10\textwidth]{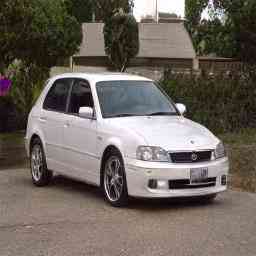}&
					\includegraphics[width=.14\textwidth,height=.10\textwidth]{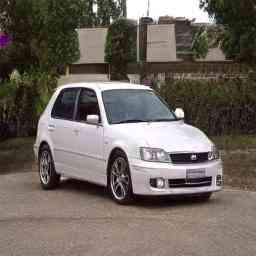}&
					\includegraphics[width=.14\textwidth,height=.10\textwidth]{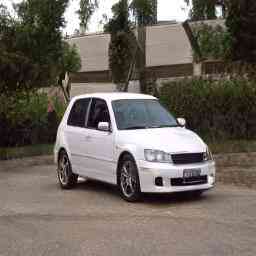}
					\\
					\\
					\footnotesize Input & \multicolumn{2}{c}{\footnotesize \ours} & \multicolumn{4}{c}{\footnotesize Manipulate Azimuths} 
					\\
					\includegraphics[width=.14\textwidth,height=.10\textwidth]{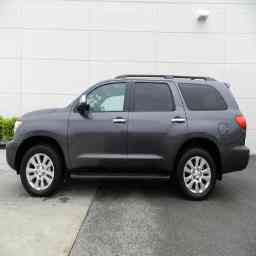}&
					\includegraphics[width=.14\textwidth,height=.10\textwidth]{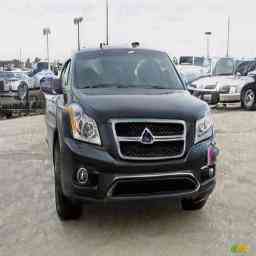}&
					\includegraphics[width=.14\textwidth,height=.10\textwidth]{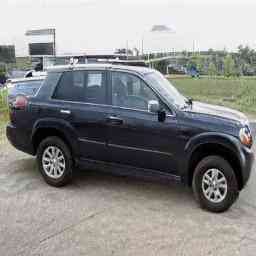}&
					\includegraphics[width=.14\textwidth,height=.10\textwidth]{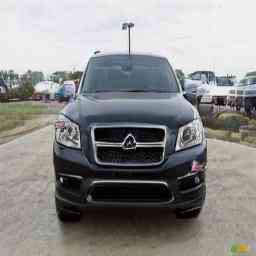}&
					\includegraphics[width=.14\textwidth,height=.10\textwidth]{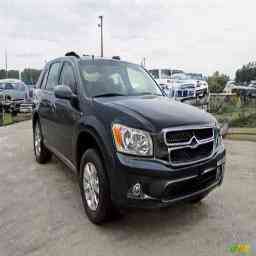}&
					\includegraphics[width=.14\textwidth,height=.10\textwidth]{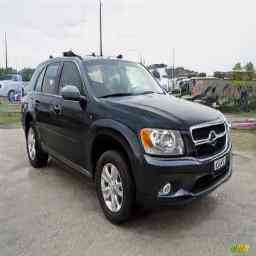}&
					\includegraphics[width=.14\textwidth,height=.10\textwidth]{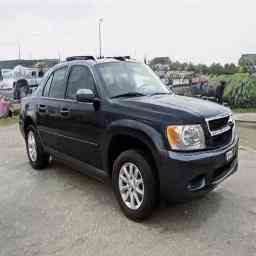}
					\\
					\includegraphics[width=.14\textwidth,height=.10\textwidth]{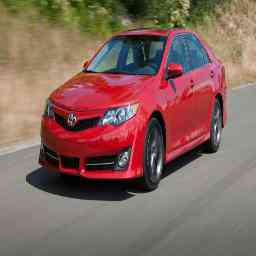}&
					\includegraphics[width=.14\textwidth,height=.10\textwidth]{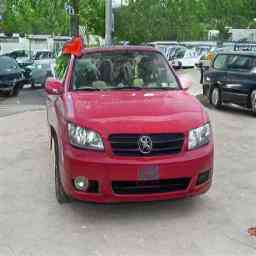}&
					\includegraphics[width=.14\textwidth,height=.10\textwidth]{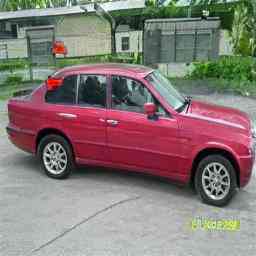}&
					\includegraphics[width=.14\textwidth,height=.10\textwidth]{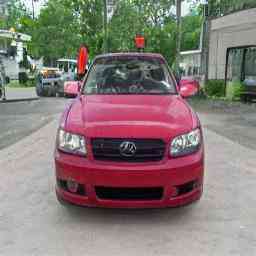}&
					\includegraphics[width=.14\textwidth,height=.10\textwidth]{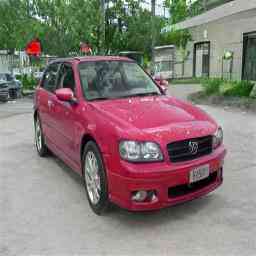}&
					\includegraphics[width=.14\textwidth,height=.10\textwidth]{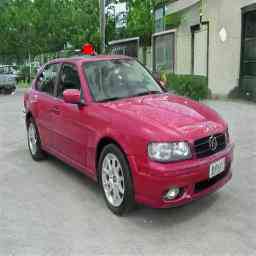}&
					\includegraphics[width=.14\textwidth,height=.10\textwidth]{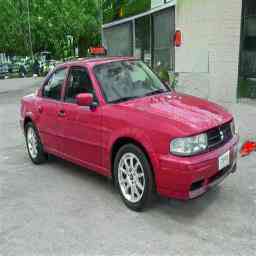}
					\\
					\includegraphics[width=.14\textwidth,height=.10\textwidth]{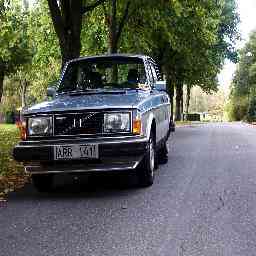}&
					\includegraphics[width=.14\textwidth,height=.10\textwidth]{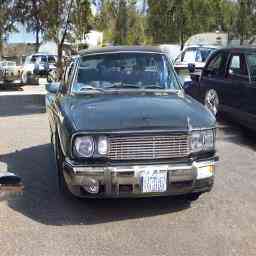}&
					\includegraphics[width=.14\textwidth,height=.10\textwidth]{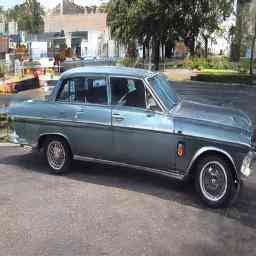}&
					\includegraphics[width=.14\textwidth,height=.10\textwidth]{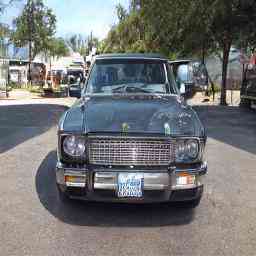}&
					\includegraphics[width=.14\textwidth,height=.10\textwidth]{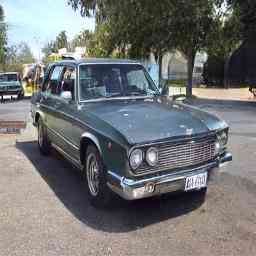}&
					\includegraphics[width=.14\textwidth,height=.10\textwidth]{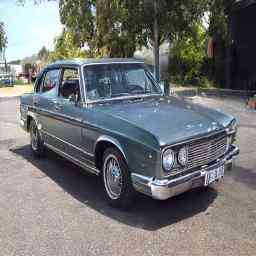}&
					\includegraphics[width=.14\textwidth,height=.10\textwidth]{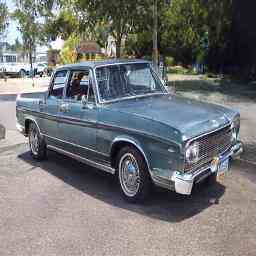}
					\\
					\\
					\footnotesize Input & \multicolumn{2}{c}{\footnotesize \ours} & \multicolumn{4}{c}{\footnotesize Manipulate Elevations} 
					\\
					\includegraphics[width=.14\textwidth,height=.10\textwidth]{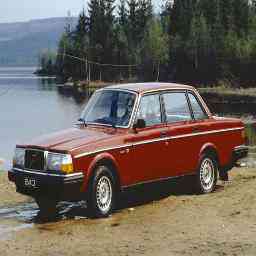}&
					\includegraphics[width=.14\textwidth,height=.10\textwidth]{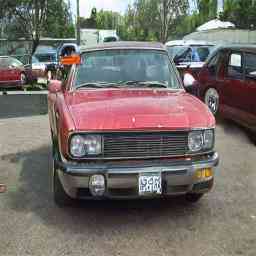}&
					\includegraphics[width=.14\textwidth,height=.10\textwidth]{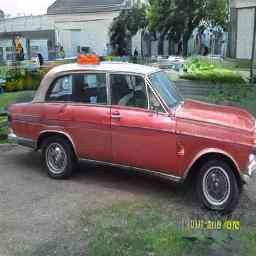}&
					\includegraphics[width=.14\textwidth,height=.10\textwidth]{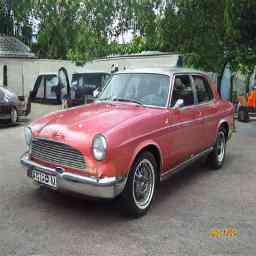}&
					\includegraphics[width=.14\textwidth,height=.10\textwidth]{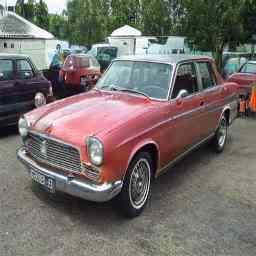}&
					\includegraphics[width=.14\textwidth,height=.10\textwidth]{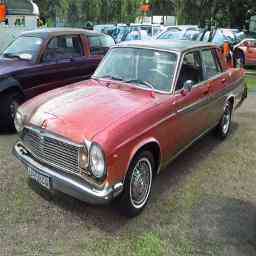}&
					\includegraphics[width=.14\textwidth,height=.10\textwidth]{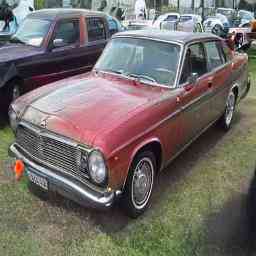}
					\\
					\includegraphics[width=.14\textwidth,height=.10\textwidth]{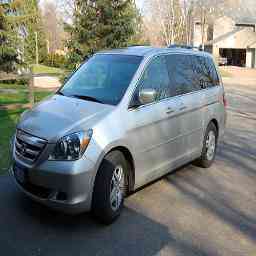}&
					\includegraphics[width=.14\textwidth,height=.10\textwidth]{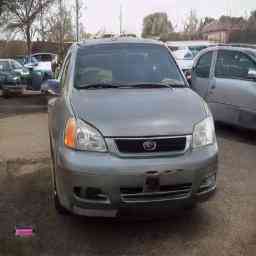}&
					\includegraphics[width=.14\textwidth,height=.10\textwidth]{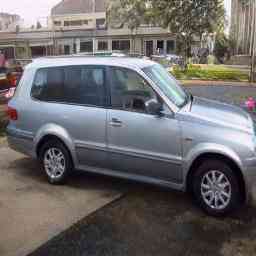}&
					\includegraphics[width=.14\textwidth,height=.10\textwidth]{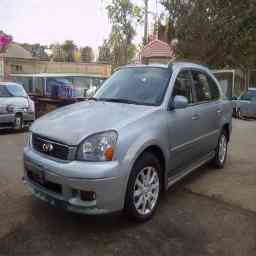}&
					\includegraphics[width=.14\textwidth,height=.10\textwidth]{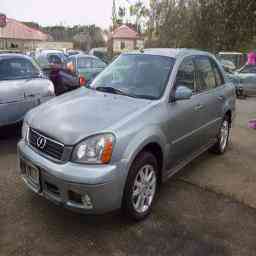}&
					\includegraphics[width=.14\textwidth,height=.10\textwidth]{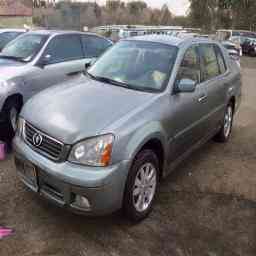}&
					\includegraphics[width=.14\textwidth,height=.10\textwidth]{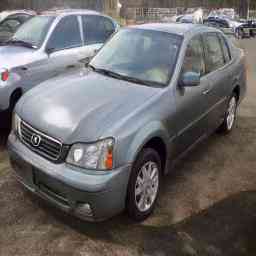}
					\\
					\includegraphics[width=.14\textwidth,height=.10\textwidth]{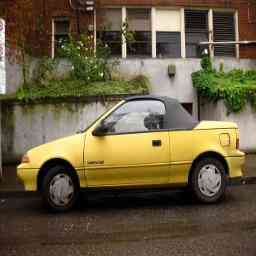}&
					\includegraphics[width=.14\textwidth,height=.10\textwidth]{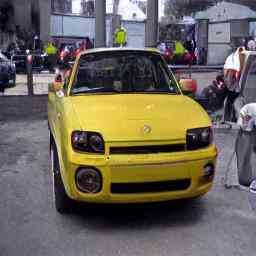}&
					\includegraphics[width=.14\textwidth,height=.10\textwidth]{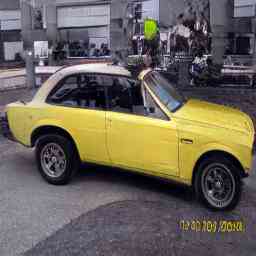}&
					\includegraphics[width=.14\textwidth,height=.10\textwidth]{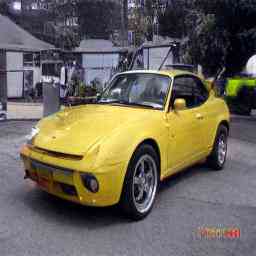}&
					\includegraphics[width=.14\textwidth,height=.10\textwidth]{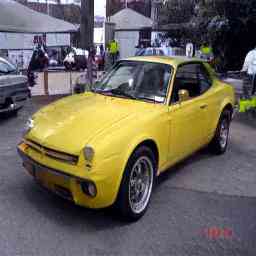}&
					\includegraphics[width=.14\textwidth,height=.10\textwidth]{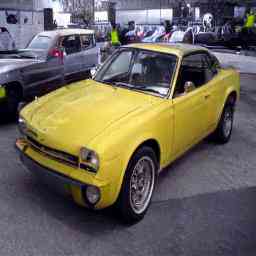}&
					\includegraphics[width=.14\textwidth,height=.10\textwidth]{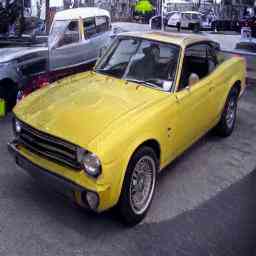}
					\hspace*{0pt}
				\end{tabular}
			\end{tabular}
		\end{center}
		\vspace*{-0.5cm}
	}
	\caption{\label{fig:realimgedit} \textbf{\footnotesize Real Image Editing.} \footnotesize Given an input image (column 1), we use our inverse graphics network to predict the 3D properties and apply {\ours} to re-render these (column 2, 3). We manipulate the car size/scale (row 1-3), azimuth (row 4-6) and elevation (Row 7-9).}
	\vspace*{-0.0mm}	
\end{figure*}

\paragraph{Real Image Editing: } We show additional real-image editing examples in Fig.~\ref{fig:realimgedit}. With our {\ours}, we can easily change the car's size, azimuth and elevation and synthesize a new image while preserving the shape and texture of the car with a consistent background.

\vspace{-2mm}
\section{Ablation Studies}
\label{sec:ablation}
\vspace{-1.7mm}

\begin{figure*}[t]
	{
		\vspace*{0pt}
		\begin{center}
			\setlength{\tabcolsep}{1pt}
			\setlength{\fboxrule}{0pt}
			\hspace*{-0.25cm}
			\begin{tabular}{c}
				\begin{tabular}{ccccc}
					\rotatebox{90}{\,\,\,\,\,\, {\color{black}{\tiny Full}}}&
					\includegraphics[width=0.13\textwidth]{figures_sup/abl2/orig/0_img.jpg}&
					\includegraphics[width=0.13\textwidth]{figures_sup/abl2/orig/0_rendered.jpg}&
					\includegraphics[width=0.1\textwidth]{figures_sup/abl2/orig/imview_102_102_front.jpg}&
					\includegraphics[width=0.6\textwidth]{figures_sup/abl2/orig/0_multiview.jpg}
					\\
					\rotatebox{90}{\,\,\, {\color{black}{\tiny w.o M. V.}}}&
					\includegraphics[width=0.13\textwidth]{figures_sup/abl2/nomv/0_img.jpg}&
					\includegraphics[width=0.13\textwidth]{figures_sup/abl2/nomv/0_rendered.jpg}&
					\includegraphics[width=0.1\textwidth]{figures_sup/abl2/nomv/imview_102_102_front.jpg}&
					\includegraphics[width=0.6\textwidth]{figures_sup/abl2/nomv/0_multiview.jpg}
					\\
					\rotatebox{90}{\,\,\,\,\, {\color{black}{\tiny w.o P.}}}&
					\includegraphics[width=0.13\textwidth]{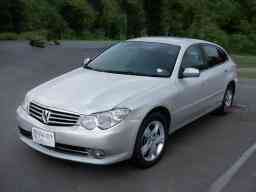}&
					\includegraphics[width=0.13\textwidth]{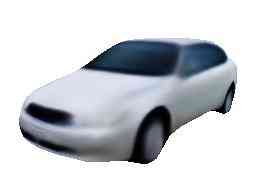}&
					\includegraphics[width=0.1\textwidth]{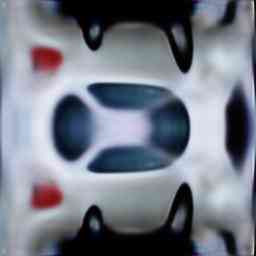}&
					\includegraphics[width=0.6\textwidth]{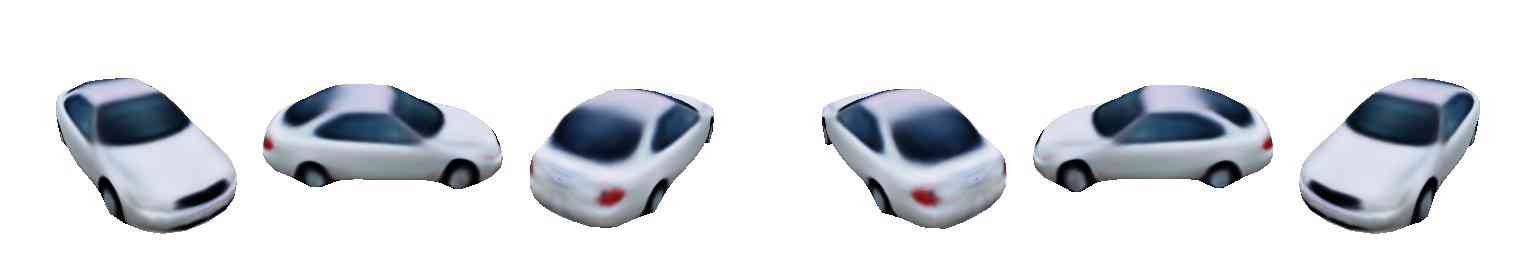}
					\\
					\rotatebox{90}{\,\,\,\,\,\,  {\color{black}{\tiny Full}}}&
					\includegraphics[width=0.13\textwidth]{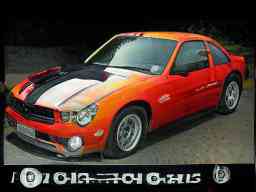}&
					\includegraphics[width=0.13\textwidth]{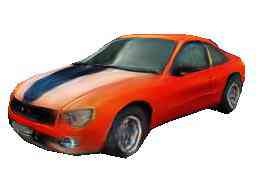}&
					\includegraphics[width=0.1\textwidth]{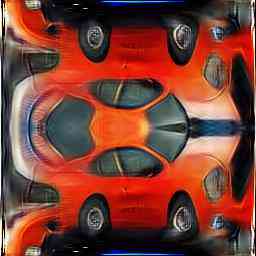}&
					\includegraphics[width=0.6\textwidth]{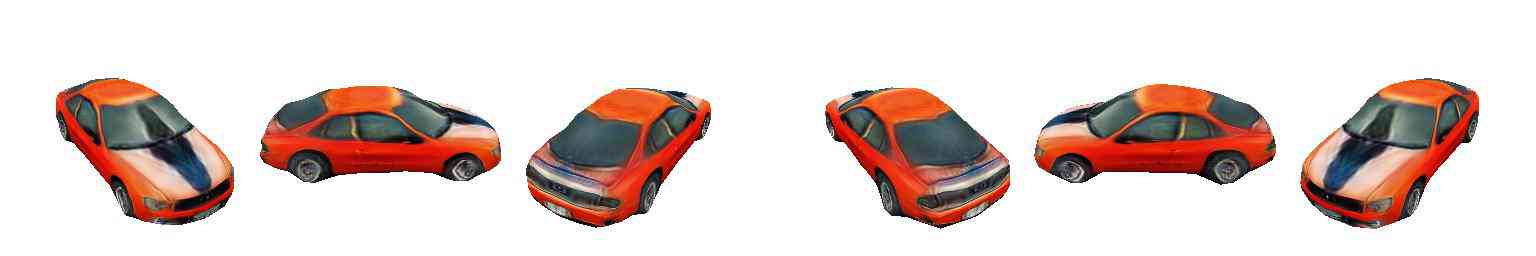}
					\\
					\rotatebox{90}{\,\,\, {\color{black}{\tiny w.o M. V.}}}&
					\includegraphics[width=0.13\textwidth]{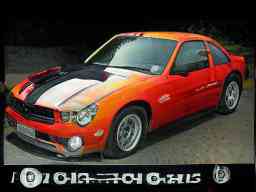}&
					\includegraphics[width=0.13\textwidth]{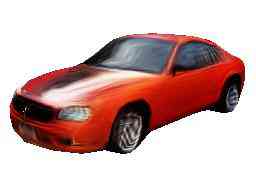}&
					\includegraphics[width=0.1\textwidth]{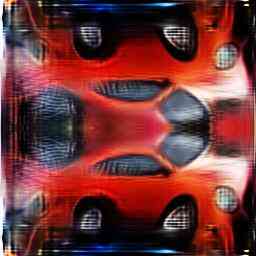}&
					\includegraphics[width=0.6\textwidth]{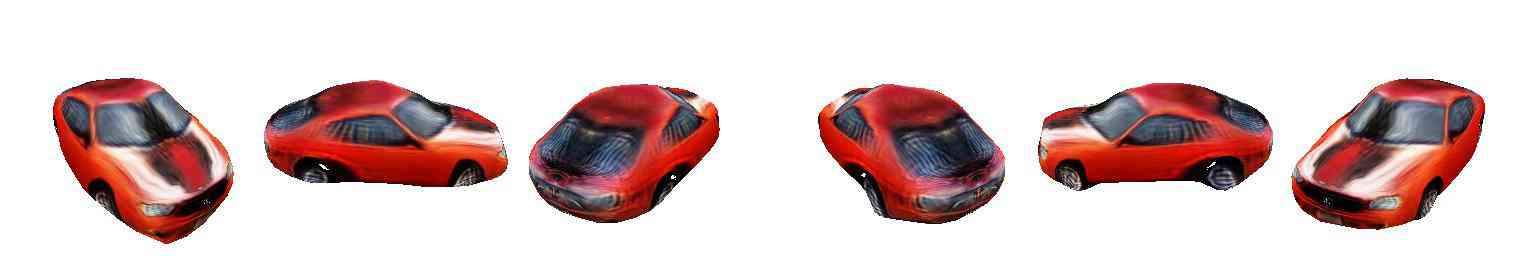}
					\\
					\rotatebox{90}{\,\,\,\,\,  {\color{black}{\tiny w.o P.}}}&
					\includegraphics[width=0.13\textwidth]{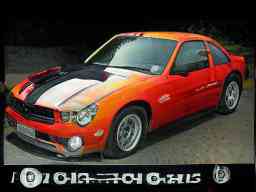}&
					\includegraphics[width=0.13\textwidth]{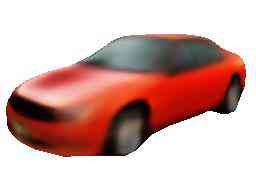}&
					\includegraphics[width=0.1\textwidth]{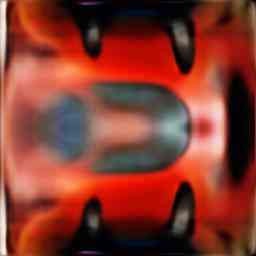}&
					\includegraphics[width=0.6\textwidth]{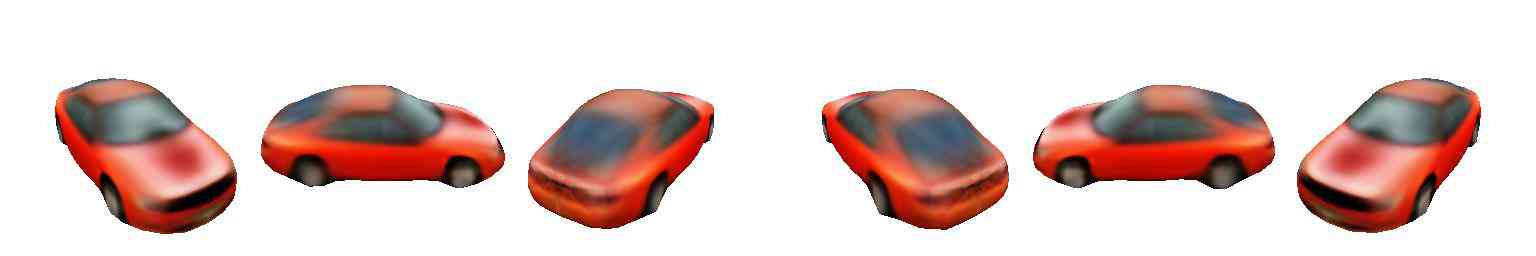}
					\\
					& {\scriptsize Input } & {\scriptsize Pred. } & {\scriptsize Texture} & {\scriptsize Pred. Multiple Views}
					\hspace*{0pt}
				\end{tabular}
			\end{tabular}
		\end{center}
		\vspace*{-0.3cm}
	}
	\caption{\footnotesize \textbf{Ablation Study:} We ablate the use of multi-view consistency and perceptual losses by showing results of 3D predictions. Clearly, the texture becomes worse in the invisible part if we remove the multi-view consistency loss (rows 2, 5, denoted by ``w.o M. V.'', which denotes that no multi-view consistency was used during training), showcasing the importance of our StyleGAN-multivew dataset. Moreover, the textures become quite smooth and lose details if we do not use the perceptual loss (rows 3, 6, noted by ``w.o P.'', which denotes that no perceptual loss was used during training).}
	\label{fig:abl}
\end{figure*}

We find that the multi-view consistency and perceptual losses play an import role in training, as shown in Fig.~\ref{fig:abl}. Multi-view consistency loss helps in training a more accurate inverse graphics network in terms of shape, while the perceptual loss helps to keep  texture more realistic.

\section{StyleGAN Manipulation}
\label{sec:rebuttal}
\begin{figure*}[t!]
	{
		\vspace{-3mm}
		\hspace{-1mm}
		\begin{minipage}{0.70\linewidth}
			\setlength{\tabcolsep}{1pt}
			\setlength{\fboxrule}{0pt}
			\begin{tabular}{c}
				\begin{tabular}{ccccccc}
					\\
					\rotatebox{90}{$\qquad\ \ ${\color{black}{\scriptsize Scale}}}&
					\includegraphics[width=.23\textwidth]{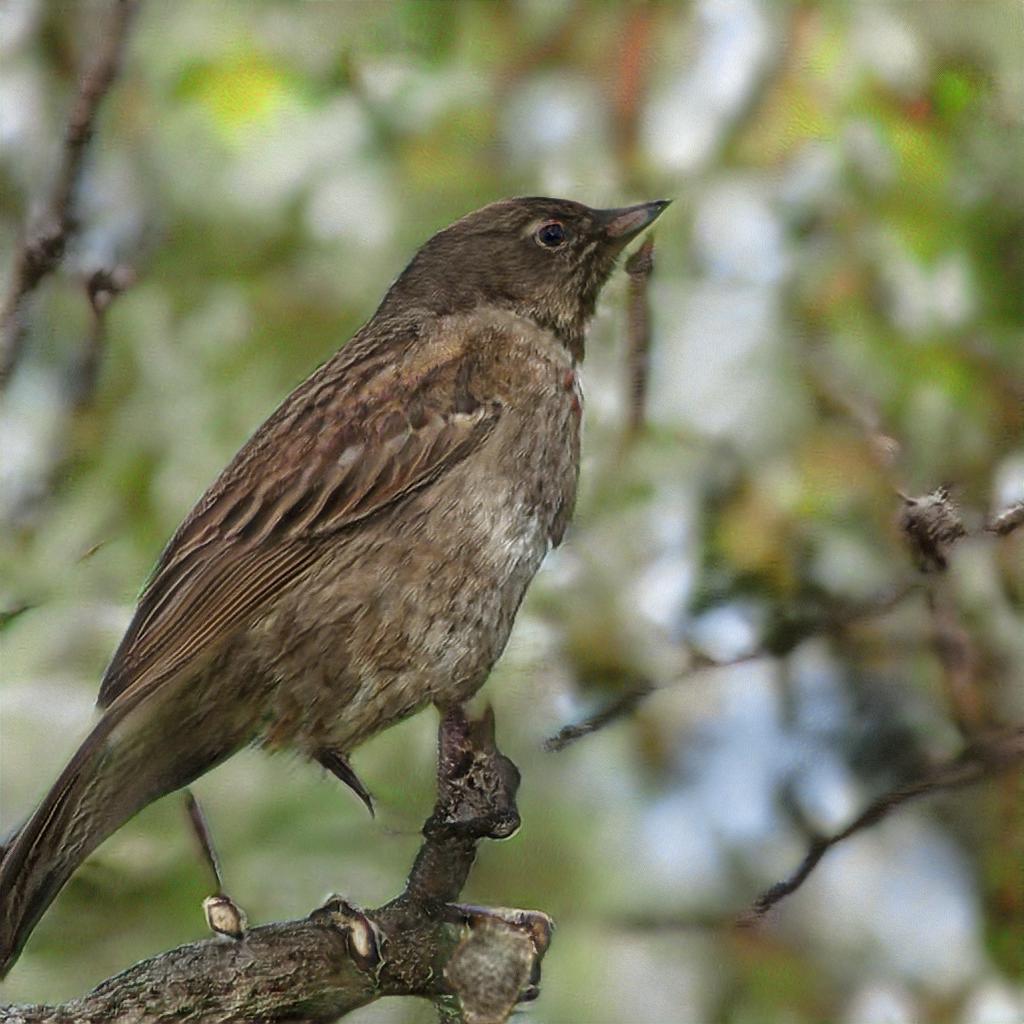}&
					\includegraphics[width=.23\textwidth]{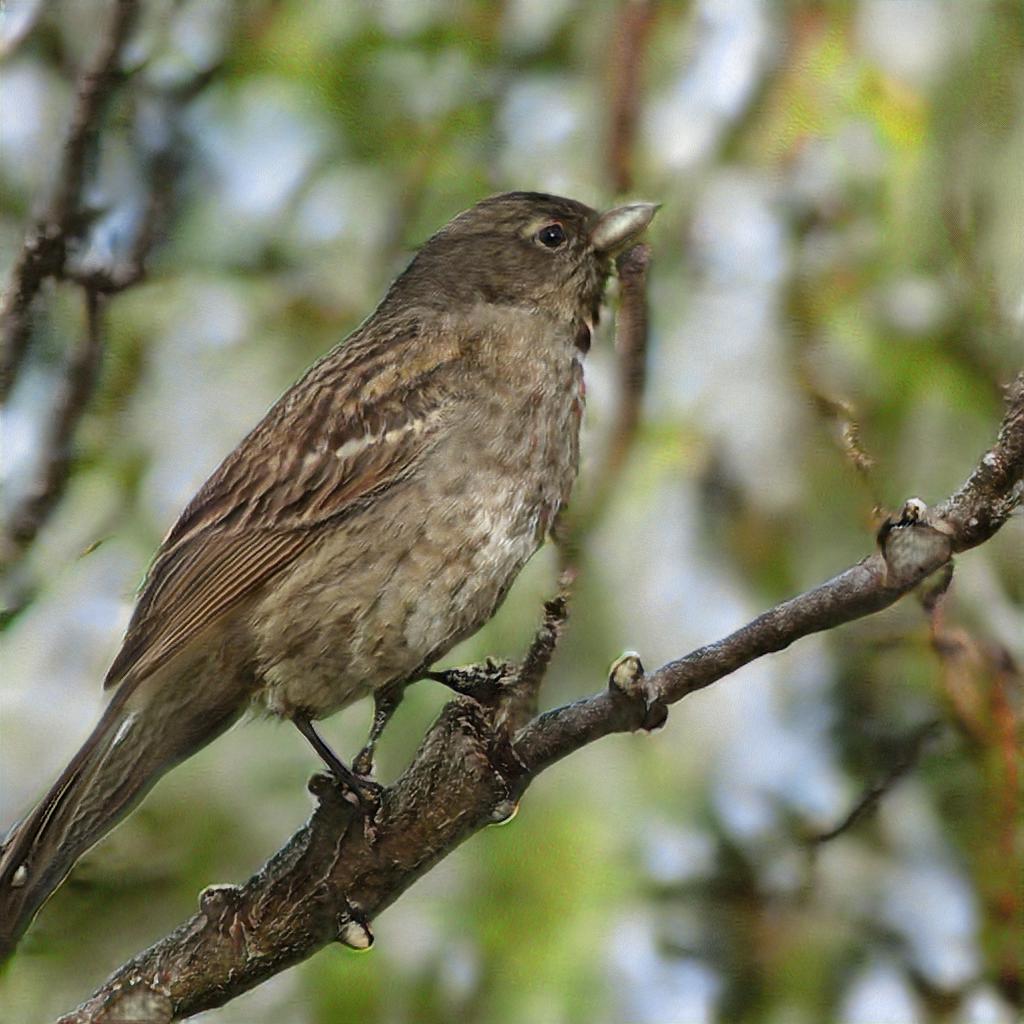}&
					\includegraphics[width=.23\textwidth]{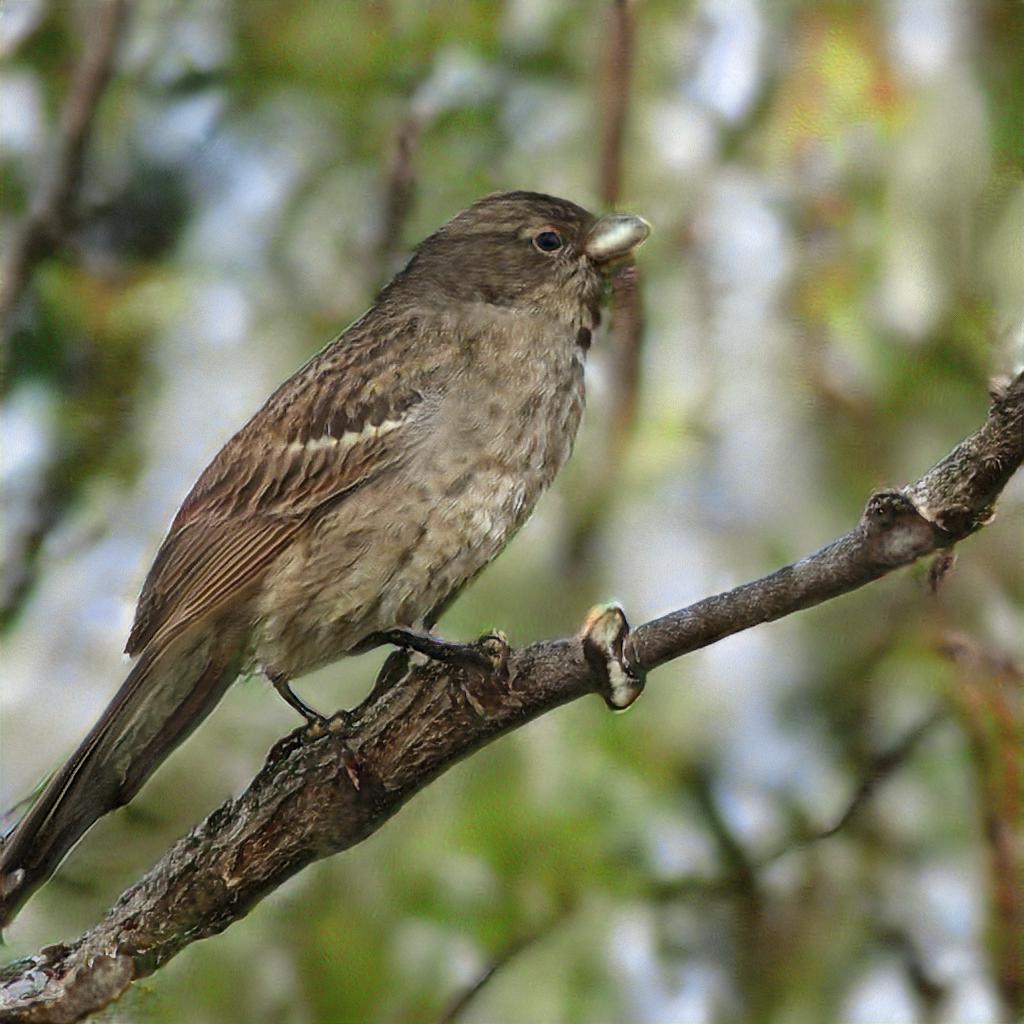}&
					\includegraphics[width=.23\textwidth]{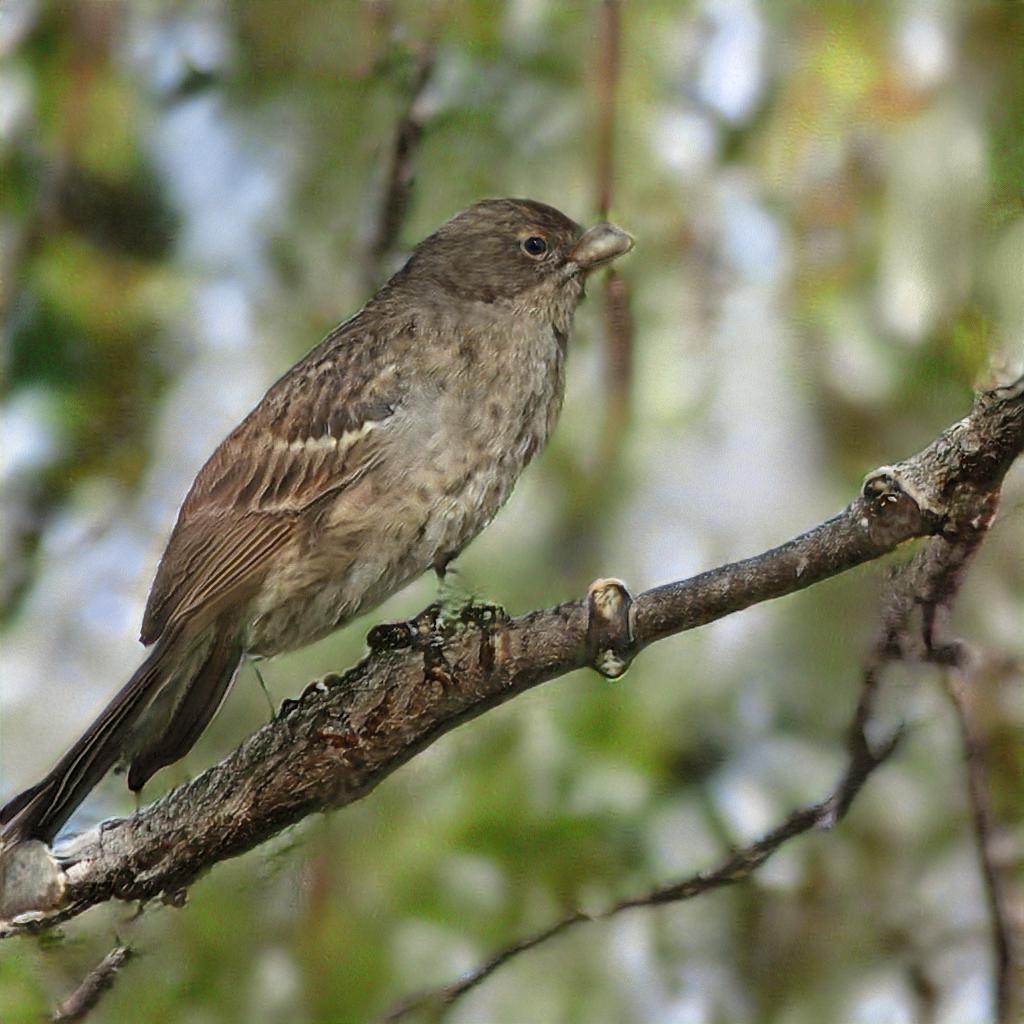}&
					\includegraphics[width=.23\textwidth]{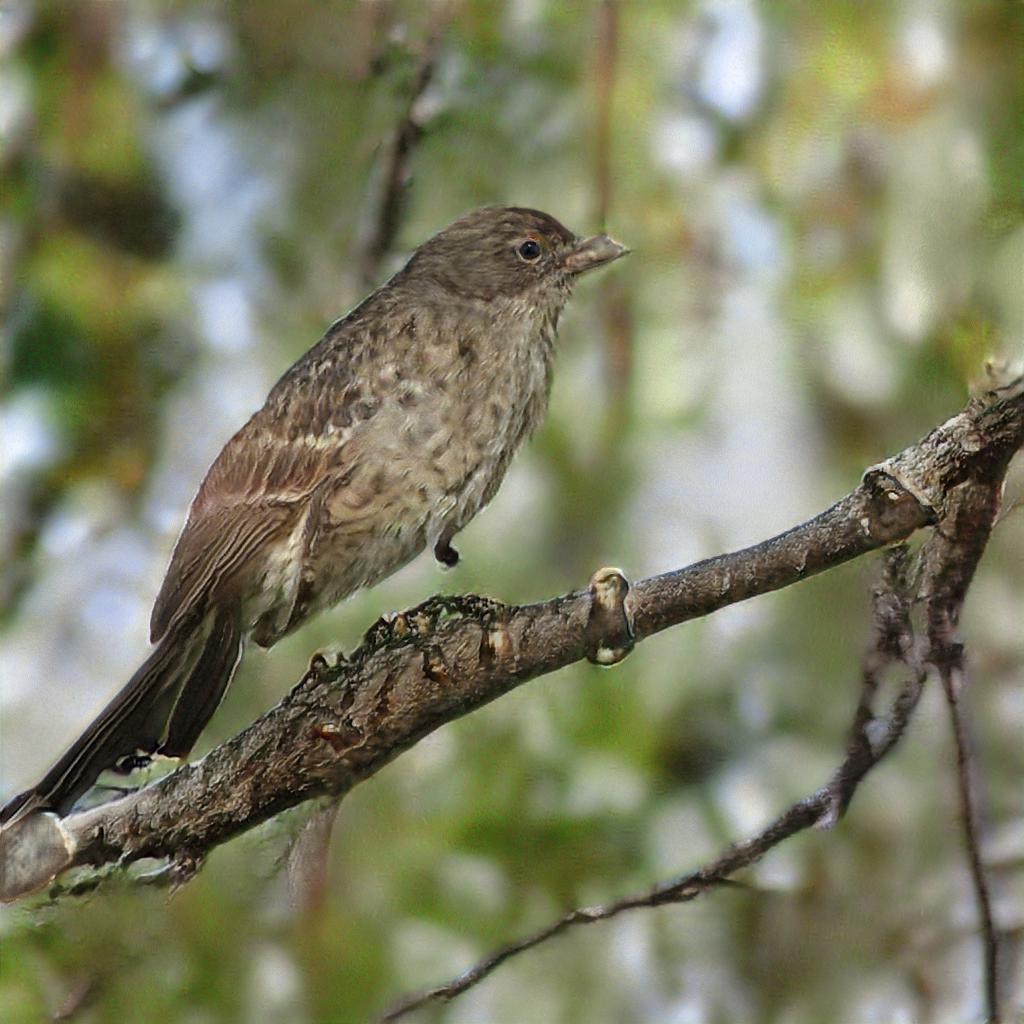}&
					\includegraphics[width=.23\textwidth]{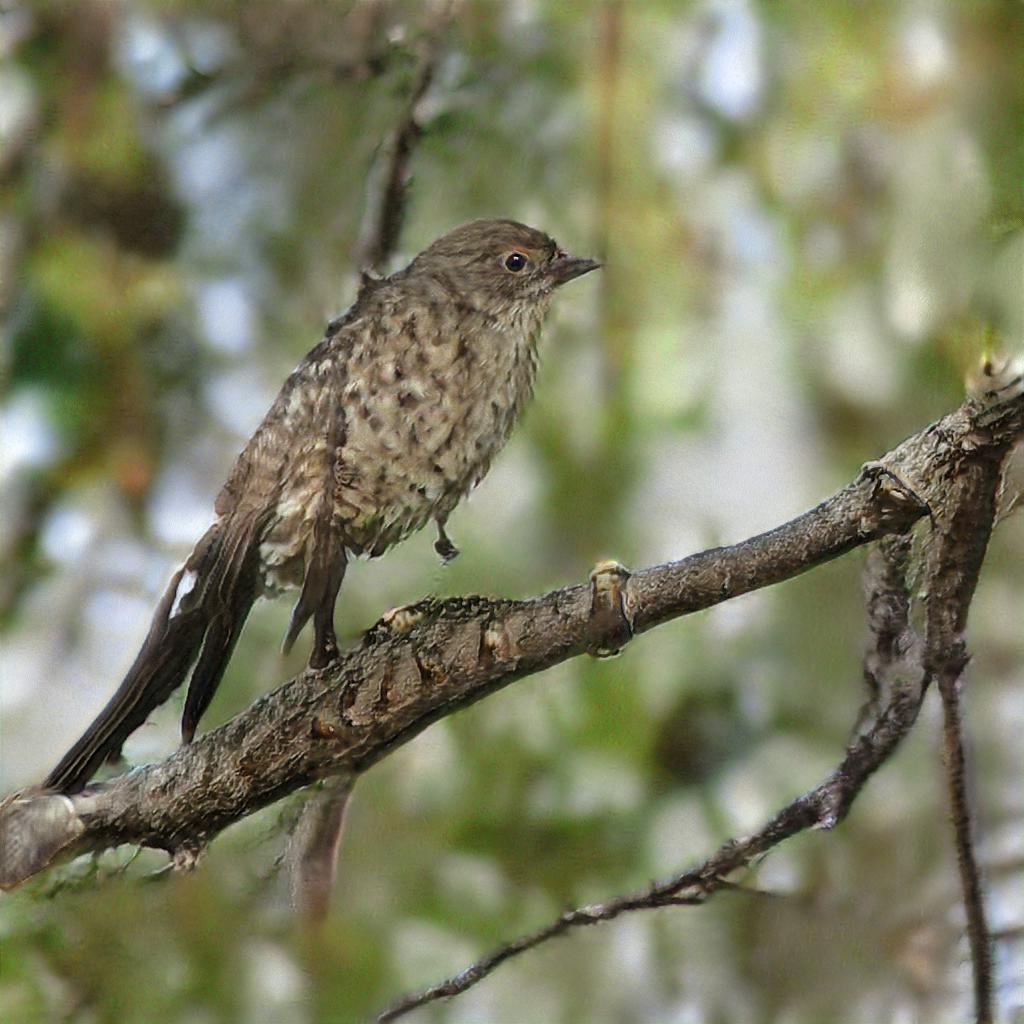}
					\\
					\rotatebox{90}{$\qquad${\color{black}{\scriptsize Azimuth}}}&
					\includegraphics[width=.23\textwidth]{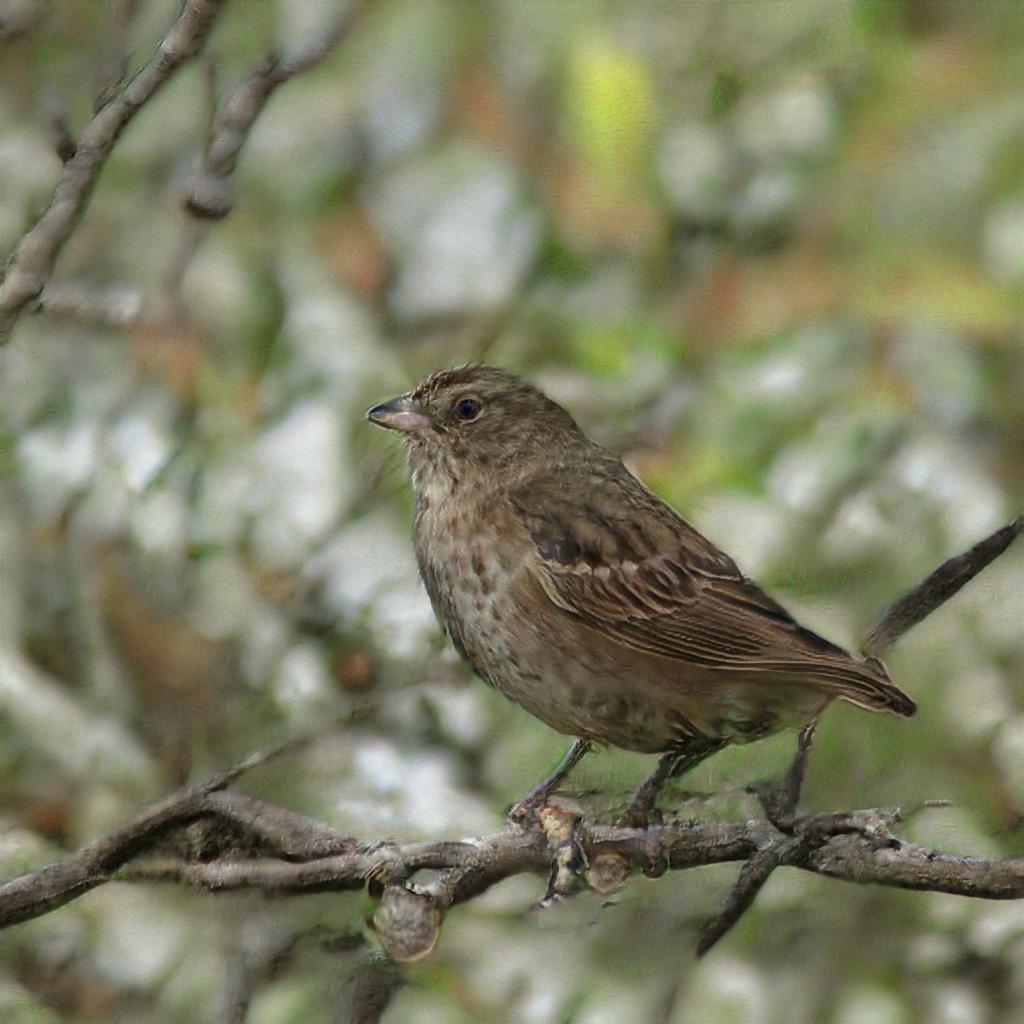}&
					\includegraphics[width=.23\textwidth]{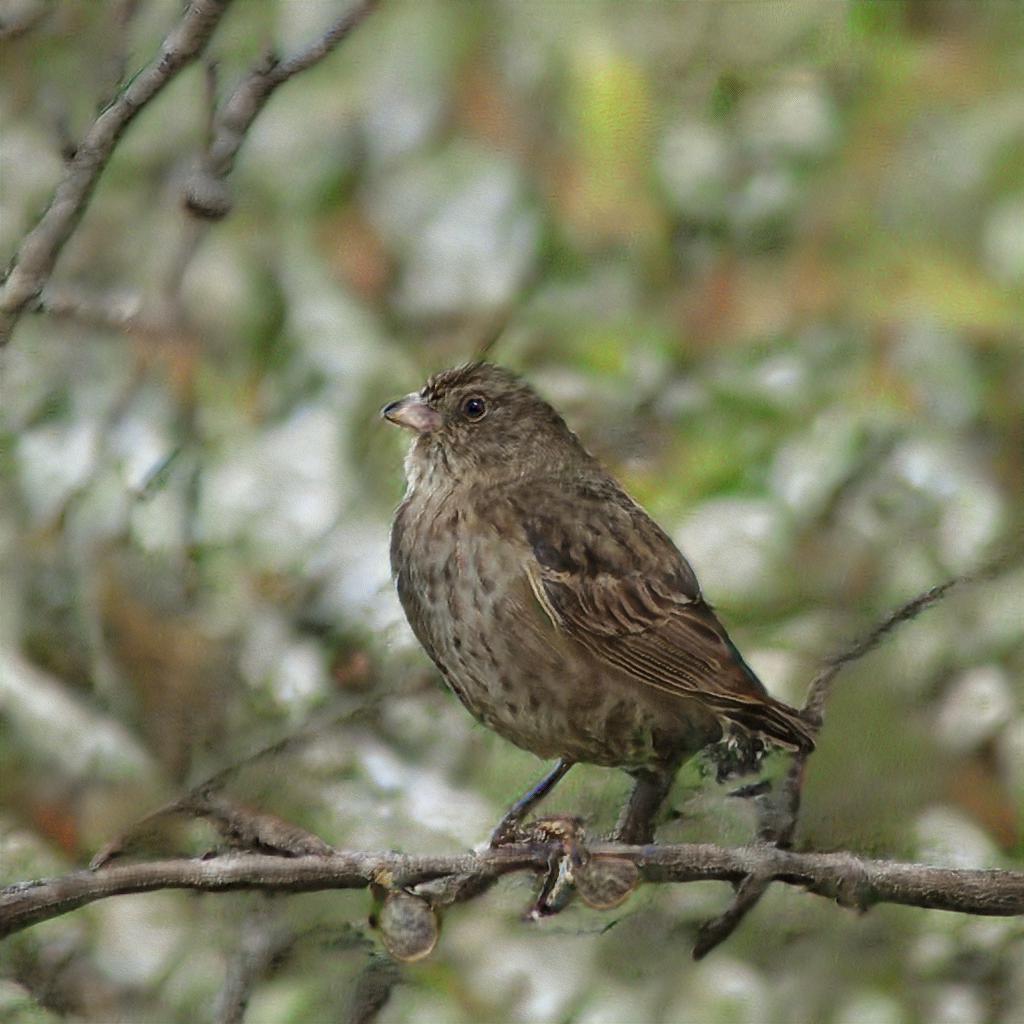}&
					\includegraphics[width=.23\textwidth]{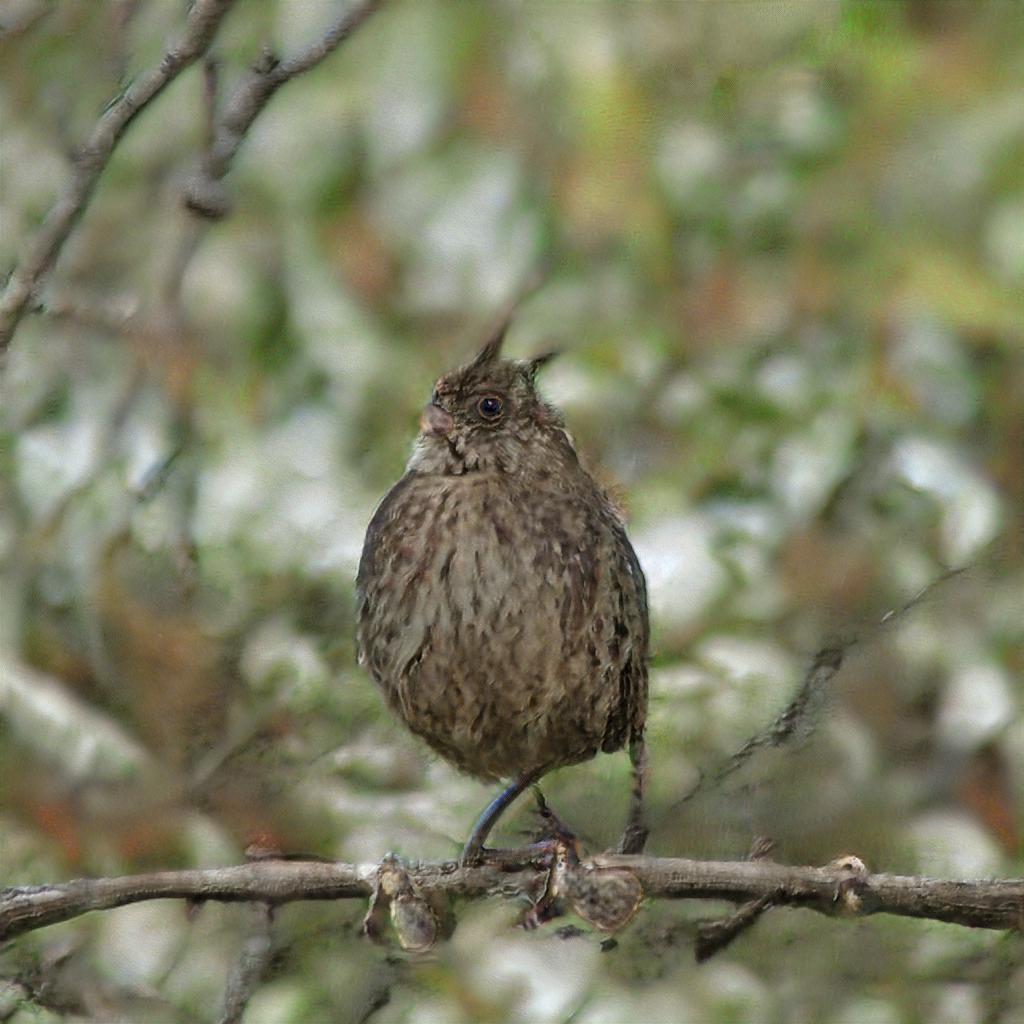}&
					\includegraphics[width=.23\textwidth]{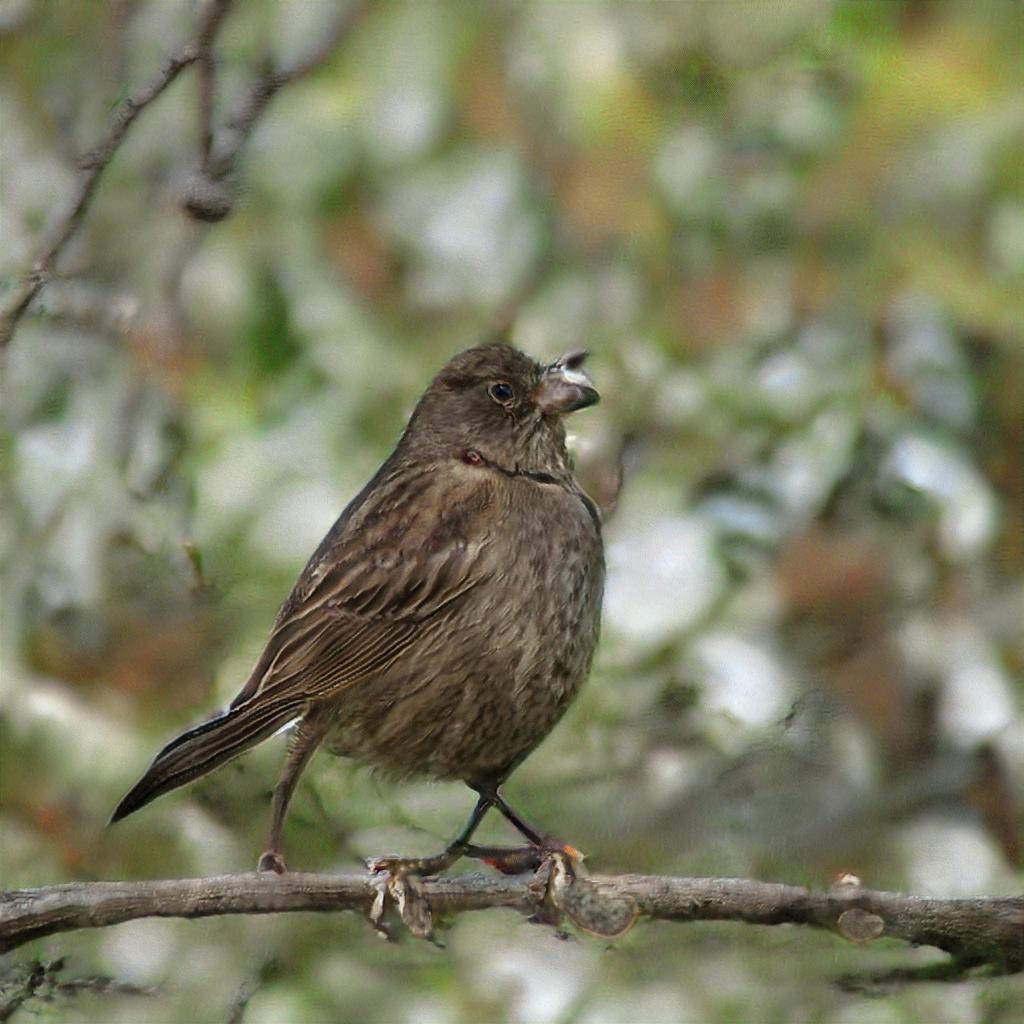}&
					\includegraphics[width=.23\textwidth]{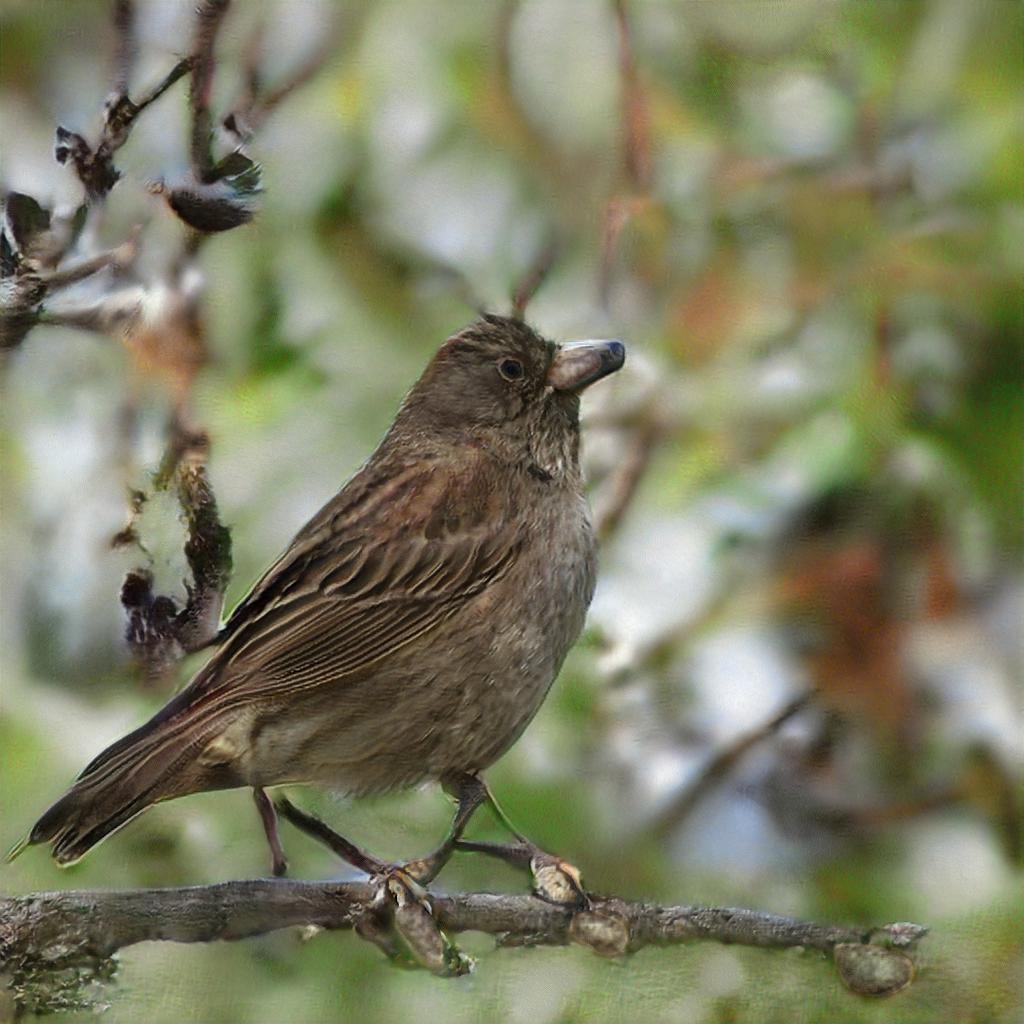}&
					\includegraphics[width=.23\textwidth]{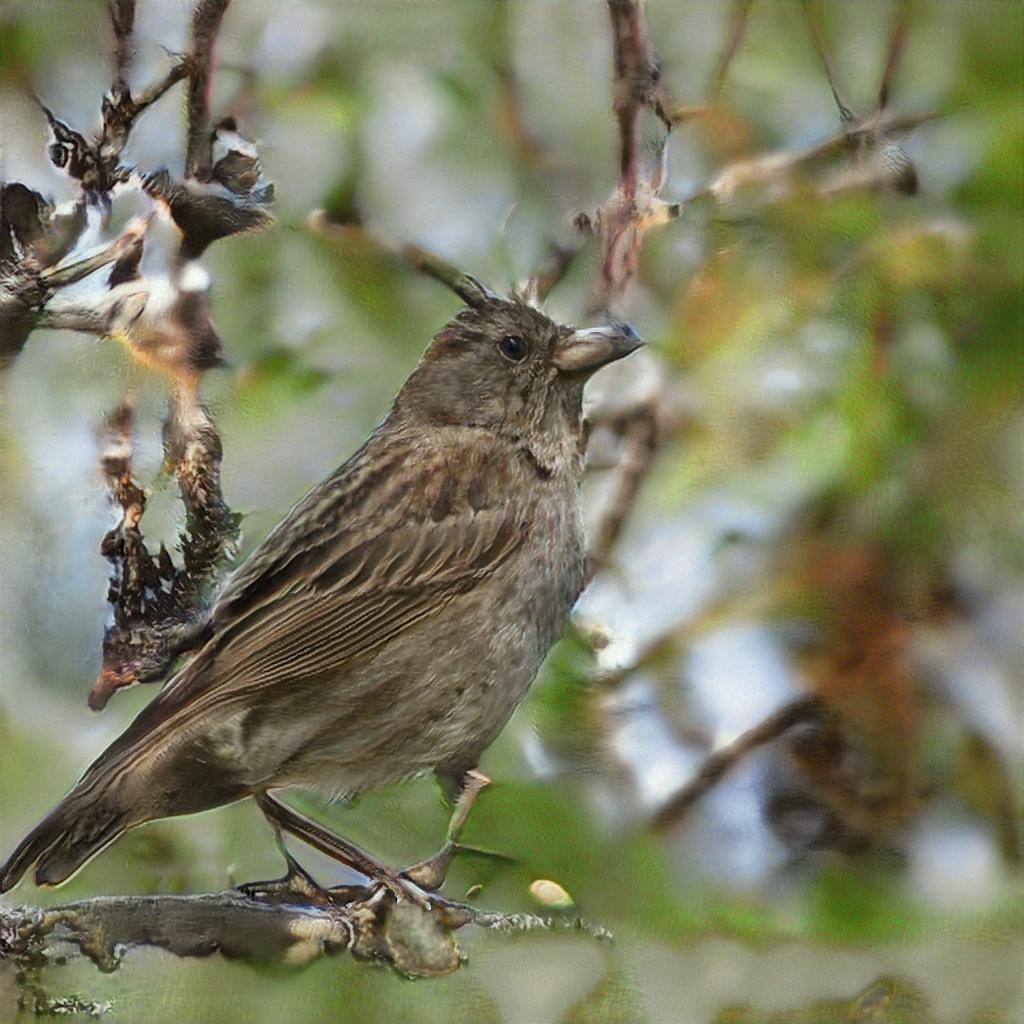}
					\\
					\rotatebox{90}{$\qquad${\color{black}{\scriptsize Elevation}}}&
					\includegraphics[width=.23\textwidth]{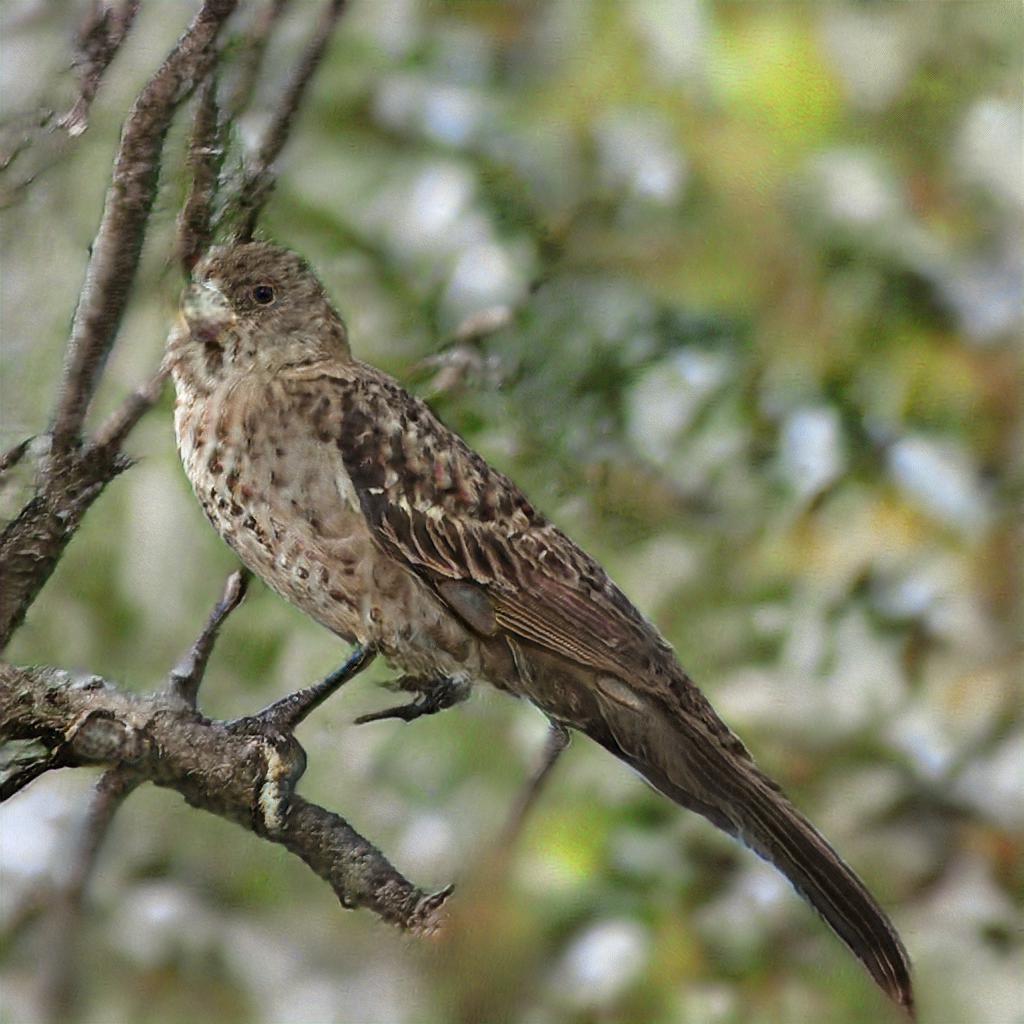}&
					\includegraphics[width=.23\textwidth]{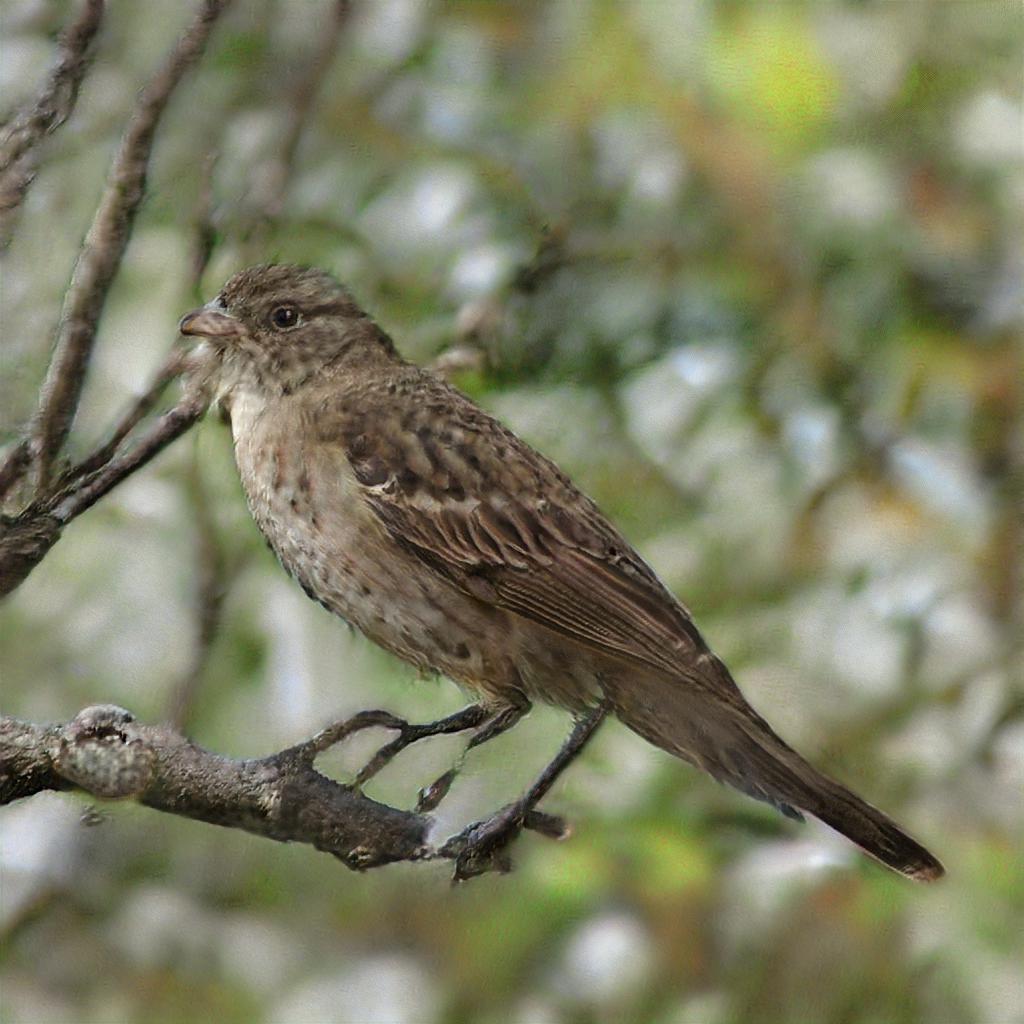}&
					\includegraphics[width=.23\textwidth]{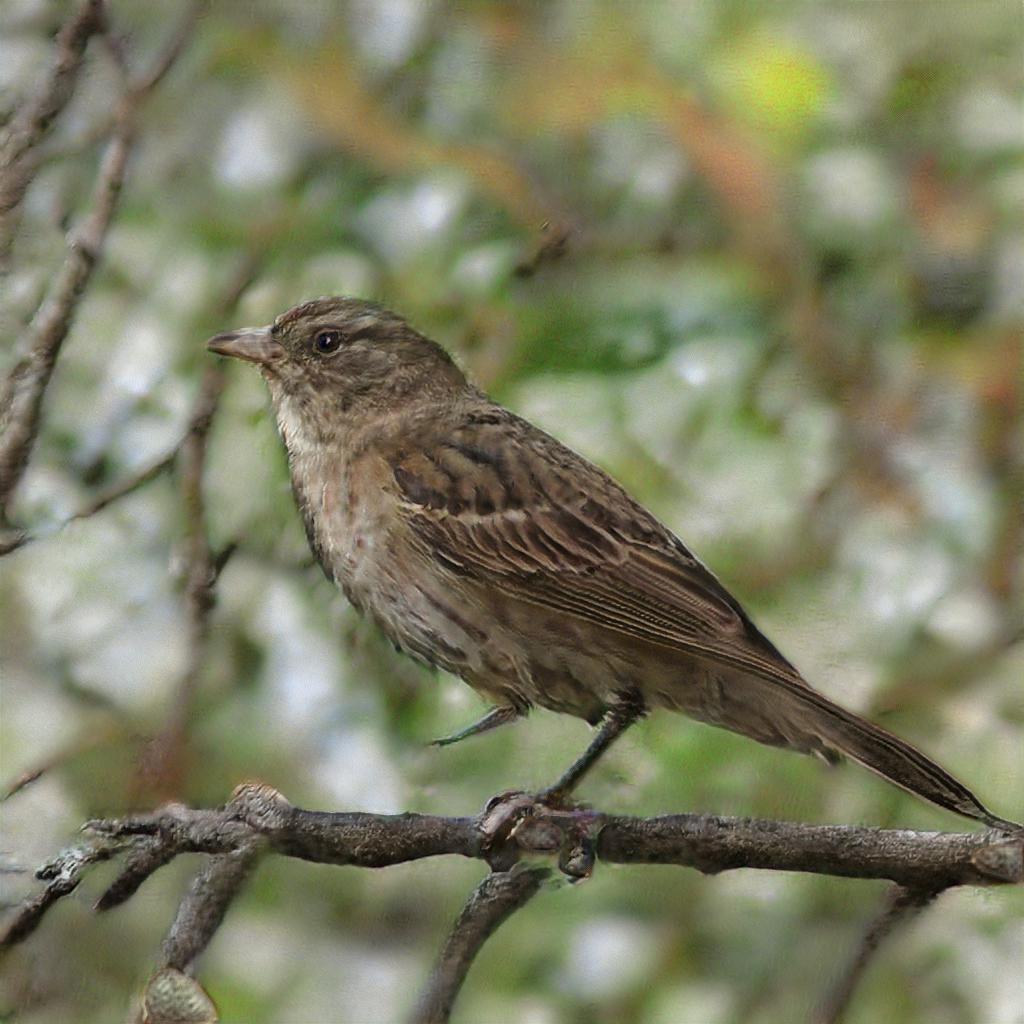}&
					\includegraphics[width=.23\textwidth]{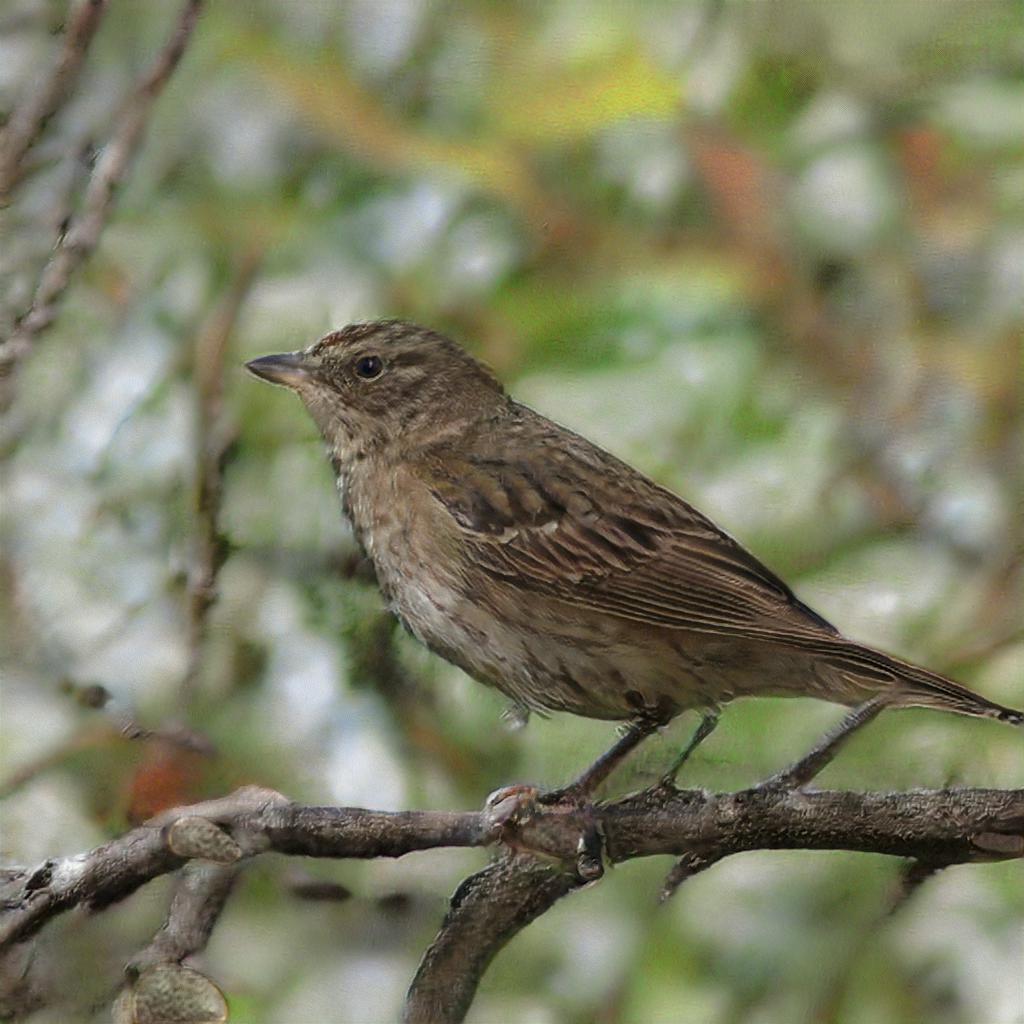}&
					\includegraphics[width=.23\textwidth]{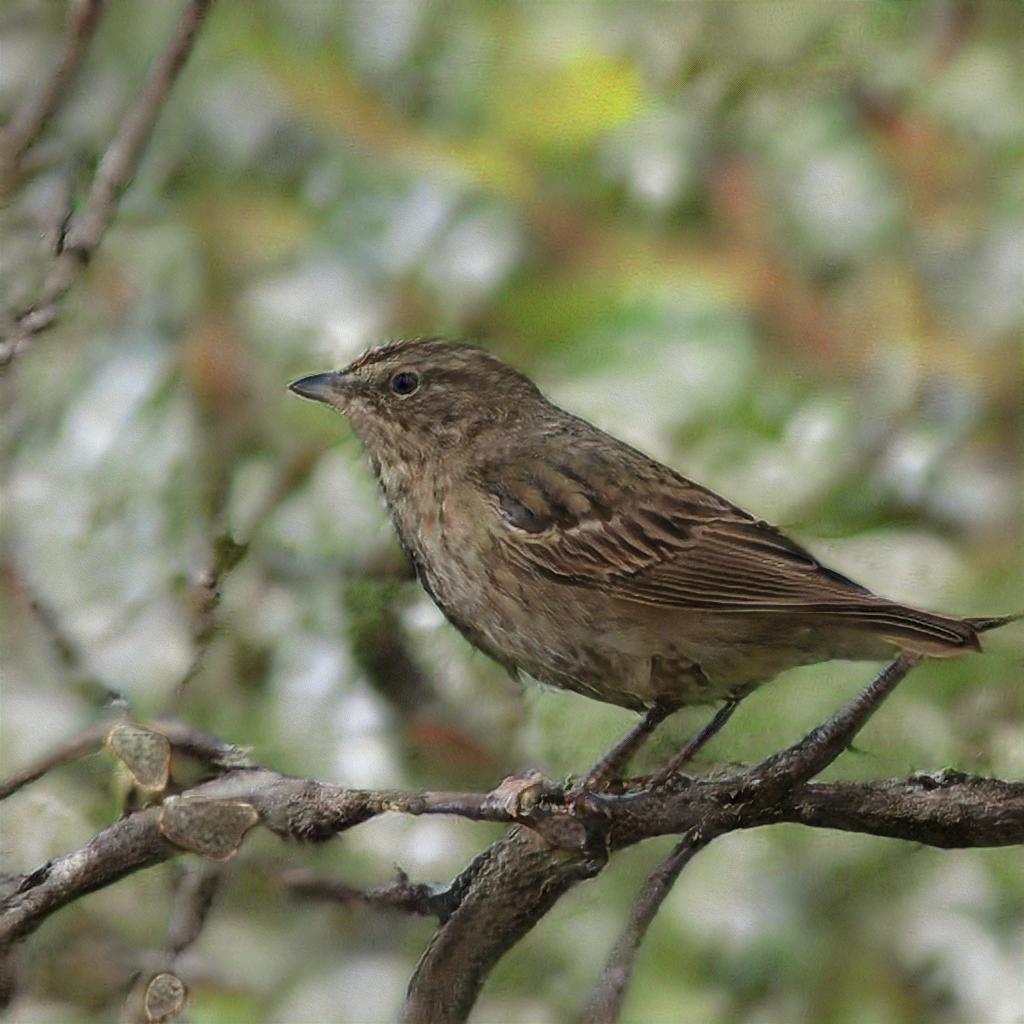}&
					\includegraphics[width=.23\textwidth]{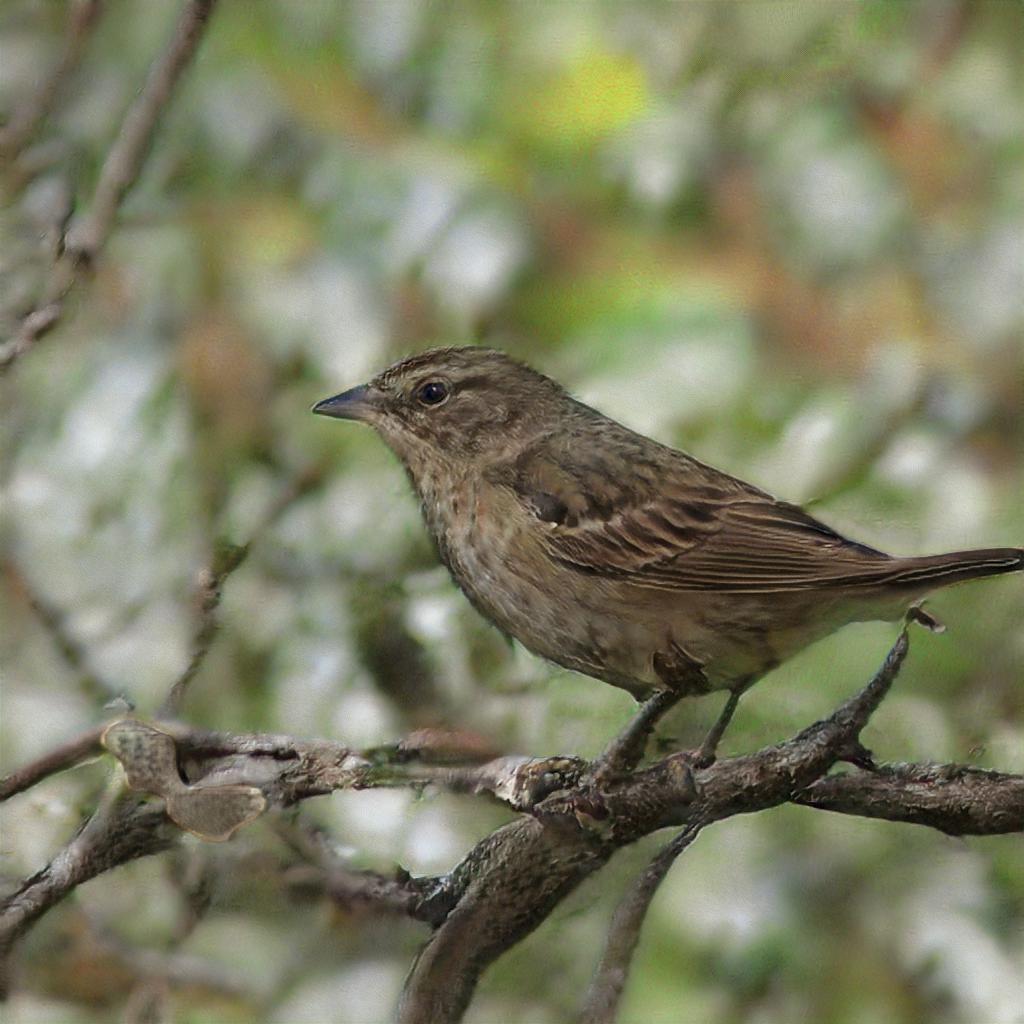}\\
				\end{tabular}
			\end{tabular}
		\end{minipage}
		\vspace{-4mm}
		\caption{\label{fig:camera_controll_bird}\footnotesize \textbf{Bird Camera Controller:} We manipulate azimuth, scale, elevation parameters with {\ours} to synthesize images in new viewpoints while keeping content code fixed.}
	}
	\vspace*{1mm}
\end{figure*}

\begin{figure*}[t!]
	{
		\vspace*{-2mm}
		\begin{minipage}{0.69\linewidth}
			\setlength{\tabcolsep}{0.0pt}
			\setlength{\fboxrule}{0pt}
			\hspace*{-0.25cm}
			\begin{tabular}{c}
				\begin{tabular}{ccccc}
					& {\scriptsize Sampled Cars} & \cellcolor{red!40}{\scriptsize Shape Swap} & {\cellcolor{lime!50}{\scriptsize Texture Swap}} & {\cellcolor{cyan!40}{\scriptsize Background Swap}}
					\\
					\rotatebox{90}{\,\,\,\,\,{\color{black}{\scriptsize Car 1}}}&
					{\setlength{\fboxrule}{0mm}\fcolorbox{red}{red}{\includegraphics[width=.35\textwidth]{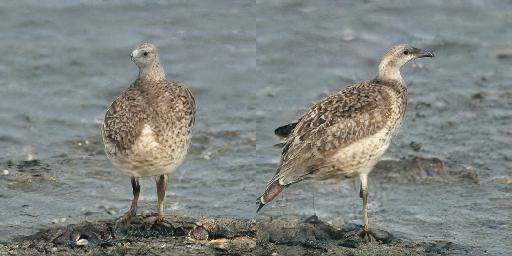}}}&
					{\setlength{\fboxrule}{0mm}\fcolorbox{white}{red}{\includegraphics[width=.35\textwidth]{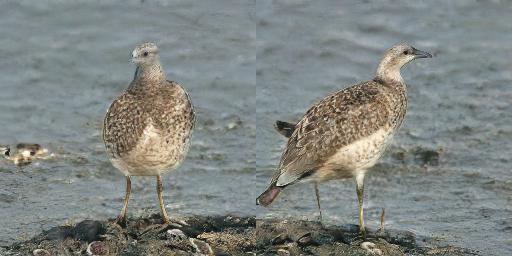}}}&
					\includegraphics[width=.35\textwidth]{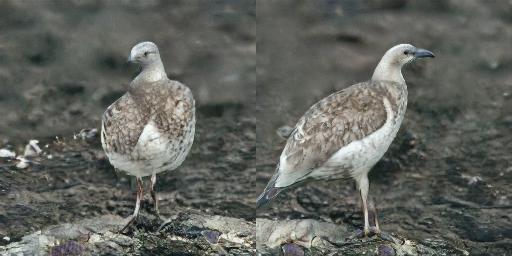}&
					\includegraphics[width=.35\textwidth]{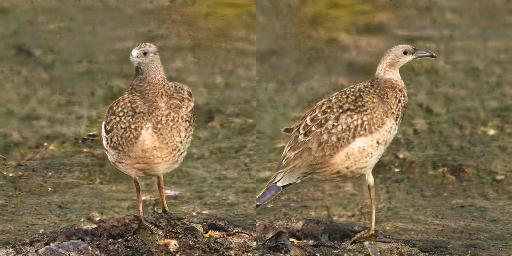}
					\\
					\rotatebox{90}{\,\,\,\,\, {\color{black}{\scriptsize Car 2}}}&
					{\setlength{\fboxrule}{0mm}\fcolorbox{lime}{lime}{\includegraphics[width=.35\textwidth]{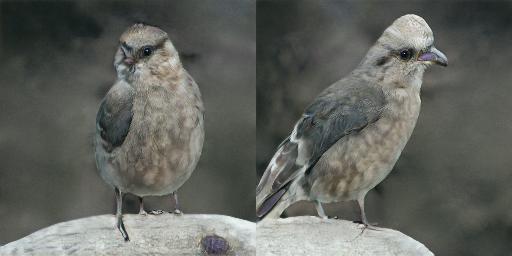}}}&
					\includegraphics[width=.35\textwidth]{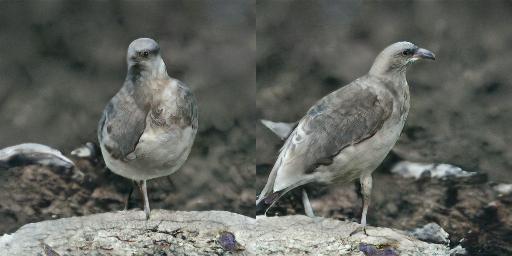}&
					\fcolorbox{white}{lime}{\includegraphics[width=.35\textwidth]{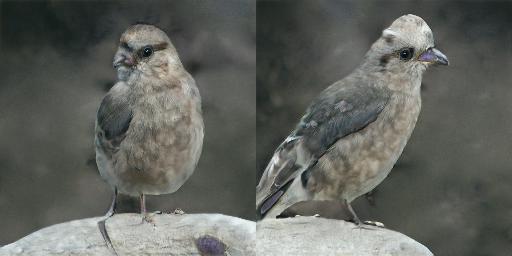}}&
					\includegraphics[width=.35\textwidth]{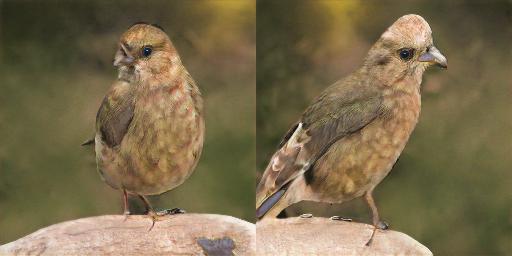}
					\\
					\rotatebox{90}{\,\,\,\,\,{\color{black}{\scriptsize Car 3}}}&
					{\setlength{\fboxrule}{0mm}\fcolorbox{white}{cyan}{\includegraphics[width=.35\textwidth]{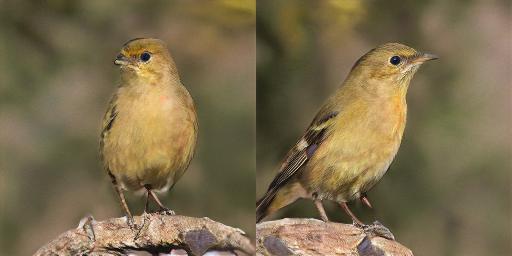}}}&
					\includegraphics[width=.35\textwidth]{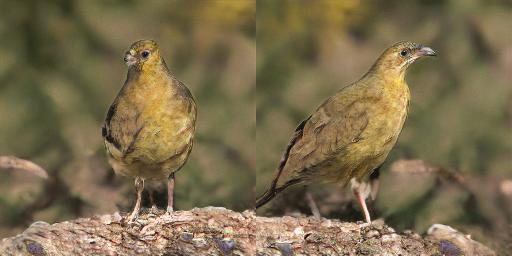}&
					\includegraphics[width=.35\textwidth]{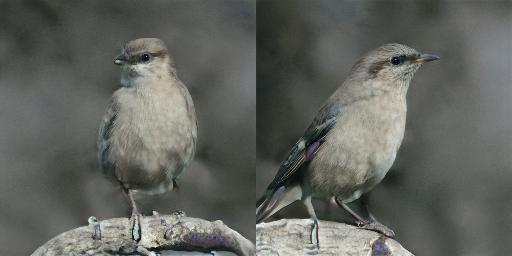}&
					{\setlength{\fboxrule}{0mm}\fcolorbox{white}{cyan}{\includegraphics[width=.35\textwidth]{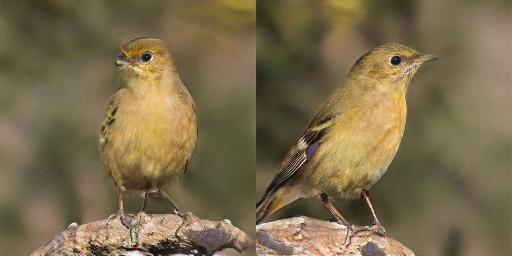}}}
					\hspace*{0pt}
				\end{tabular}
			\end{tabular}
		\end{minipage}
		\hspace{-1mm}
		\caption{\footnotesize\textbf{Bird 3D Manipulation:} We  sample 3 birds in column 1. We replace the shape of all birds with the shape of Bird 1 ({\color{red}red} box) in 2nd column. We transfer texture of Bird 2 ({\color{green}green} box)  to other birds (3rd col). In last column, we paste background of Bird 3 ({\color{cyan}cyan} box) to the other birds. Examples indicated with boxes are unchanged.}	
		\label{fig:styleganmanip_bird}
	}
	\vspace*{-4mm}
\end{figure*}

We show that our method for manipulating StyleGAN is generalizable and can be generalized to other class, as illustrated in the StyleGAN-R manipulation results for the bird in Fig~\ref{fig:camera_controll_bird} and Fig~\ref{fig:styleganmanip_bird}. 

\vspace{-0mm}
\section{Failure Cases}
\label{sec:failure}
\vspace{-0mm}

\begin{figure*}[h]
	{
		\vspace*{0pt}
		\begin{center}
			\setlength{\tabcolsep}{1pt}
			\setlength{\fboxrule}{0pt}
			\hspace*{-0.2cm}
			\begin{tabular}{c}
				\begin{tabular}{cc}
					\includegraphics[height=0.142\linewidth]{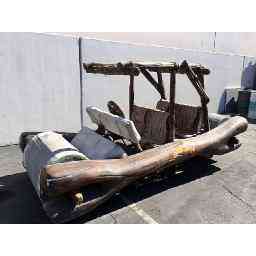}&
					\includegraphics[height=0.142\linewidth]{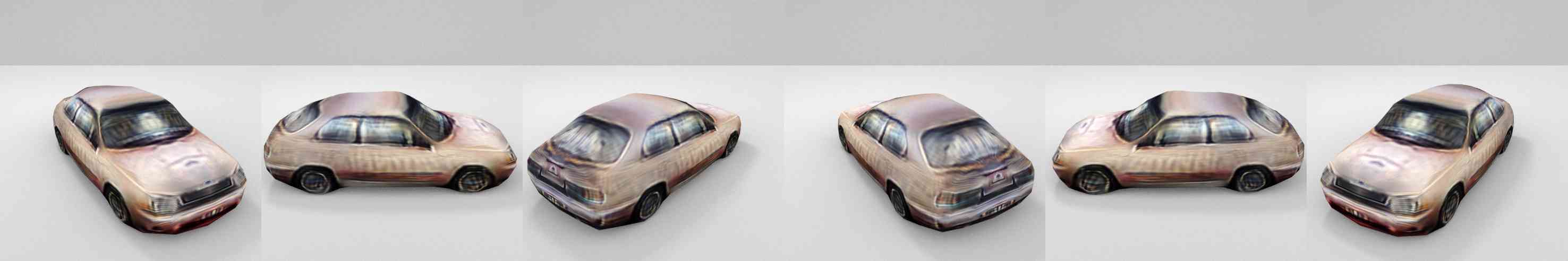}
					
					\\
					\includegraphics[height=0.142\linewidth]{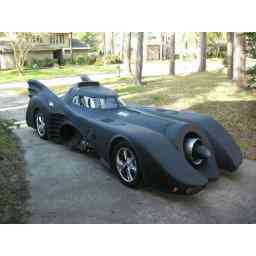}&
					\includegraphics[height=0.142\linewidth]{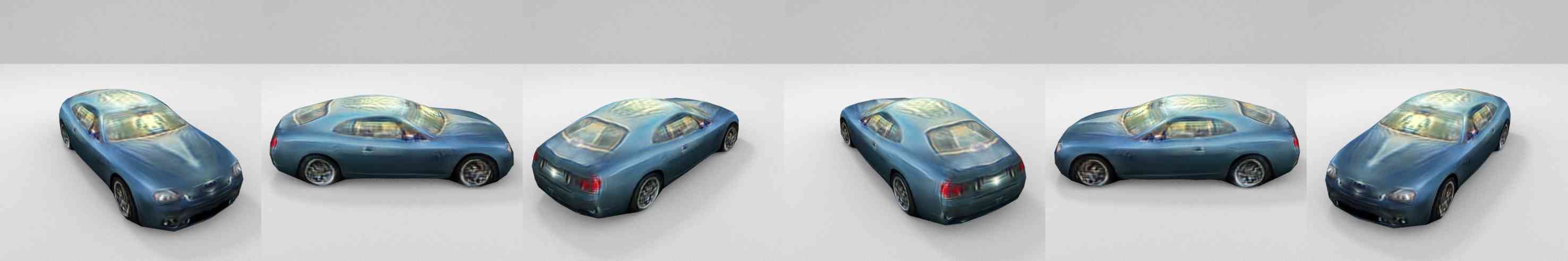}
					
					\\
					\includegraphics[height=0.142\linewidth]{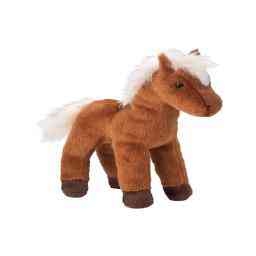}&
					\includegraphics[height=0.142\linewidth]{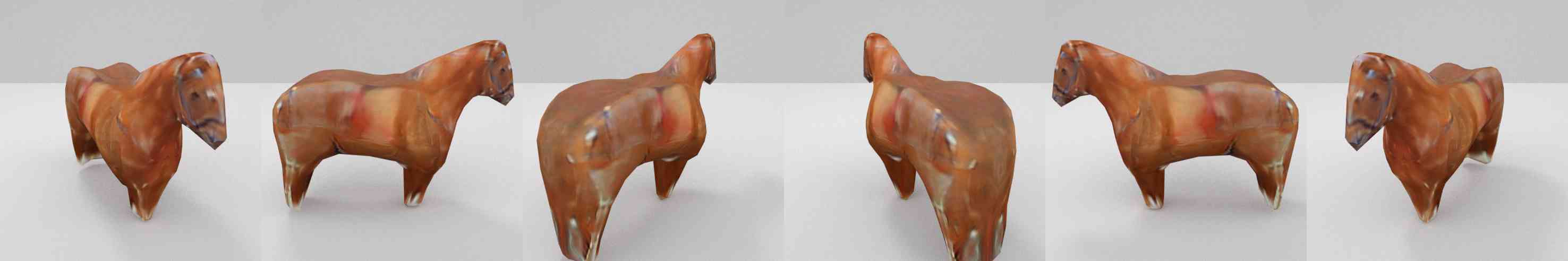}
					
					\\
					\includegraphics[height=0.142\linewidth]{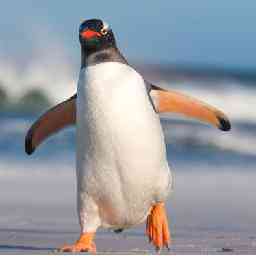}&
					\includegraphics[height=0.142\linewidth]{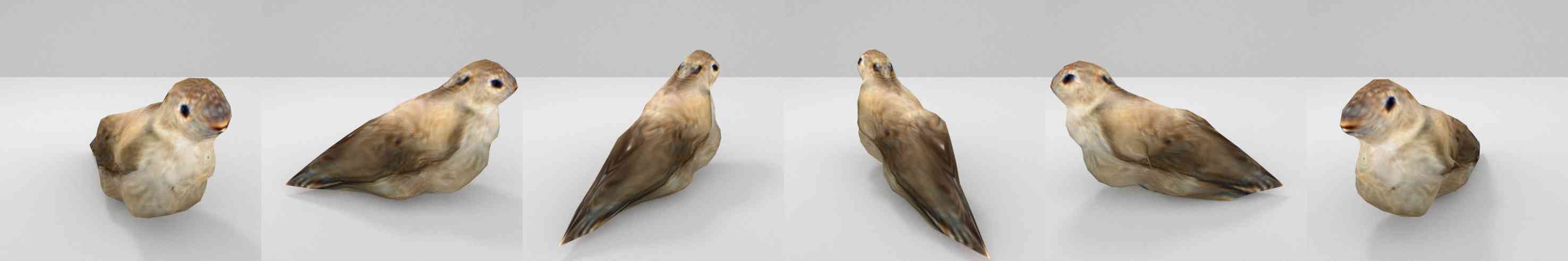}	
					
					\\

					{\footnotesize Input} & {\footnotesize Multiple Views for the predicted shape and texture} 
				\end{tabular}
			\end{tabular}
		\end{center}
		\vspace*{-2mm}
	}
	\caption{\label{fig:failure_case} \footnotesize \textbf{3D Reconstruction Failure Cases}: We show examples of failure cases for car, bird and horse. Our method tends to fail to produce relevant shapes for objects with out-of-distribution shapes (or textures).} 
	\vspace*{1mm}
\end{figure*}

We find that our inverse graphics network fails on out-of-distribution images/shapes, as shown in Fig.~\ref{fig:failure_case}. For example, the reconstruction results for Batmobile and Flinstone cars are not representative of the input cars. We anticipate that this issue can be addressed by augmenting the dataset on which StyleGAN is trained with more diverse objects. Part of the issue is also caused by GANs not capturing the tails of the distribution well, which is an active area of research. 

\vspace{-0mm}
\section{Limitations}
\label{sec:supp_limit}
\vspace{-0mm}

Our simple spherical harmonics model fails to separate light from textures. We show several examples in Fig.\ref{fig:light}. We leave this issue for future work.

\end{document}